\PassOptionsToPackage{dvipsnames}{xcolor}  
\documentclass{article}
\usepackage{iclr2027_conference,times}
\iclrfinalcopy

\usepackage{amsmath,amsfonts,bm}

\def\eqref#1{equation~\ref{#1}}

\def\1{\bm{1}}

\DeclareMathAlphabet{\mathsfit}{\encodingdefault}{\sfdefault}{m}{sl}
\SetMathAlphabet{\mathsfit}{bold}{\encodingdefault}{\sfdefault}{bx}{n}

\usepackage[dvipsnames]{xcolor}
\definecolor{MidnightBlue}{RGB}{25, 25, 112}
\definecolor{RoyalBlue}{RGB}{65, 105, 225} 
\usepackage{graphicx}
\usepackage{listings}
\usepackage{booktabs}
\usepackage{multirow}
\usepackage{enumitem}
\usepackage{subcaption}
\usepackage{caption}
\usepackage{framed}
\usepackage{fancybox}
\usepackage{soul}
\usepackage{tcolorbox}
\usepackage{tabularx}
\usepackage{amssymb}
\usepackage{pifont}
\colorlet{punct}{red!60!black}
\definecolor{background}{HTML}{EEEEEE}
\definecolor{delim}{RGB}{20,105,176}
\colorlet{numb}{magenta!60!black}

\usepackage{titlesec}

\lstdefinelanguage{json}{
    basicstyle=\scriptsize\ttfamily,
    showstringspaces=false,
    breaklines=true,
    frame=single,
    backgroundcolor=\color{white},
    literate=
     *{0}{{{\color{numb}0}}}{1}
      {1}{{{\color{numb}1}}}{1}
      {2}{{{\color{numb}2}}}{1}
      {3}{{{\color{numb}3}}}{1}
      {4}{{{\color{numb}4}}}{1}
      {5}{{{\color{numb}5}}}{1}
      {6}{{{\color{numb}6}}}{1}
      {7}{{{\color{numb}7}}}{1}
      {8}{{{\color{numb}8}}}{1}
      {9}{{{\color{numb}9}}}{1}
      {:}{{{\color{punct}{:}}}}{1}
      {,}{{{\color{punct}{,}}}}{1}
      {\{}{{{\color{delim}{\{}}}}{1}
      {\}}{{{\color{delim}{\}}}}}{1}
      {[}{{{\color{delim}{[}}}}{1}
      {]}{{{\color{delim}{]}}}}{1},
}

\usepackage[utf8]{inputenc} 
\usepackage[T1]{fontenc}    
\usepackage{hyperref}       
\usepackage{url}            
\usepackage{booktabs}       
\usepackage{amsfonts}       
\usepackage{nicefrac}       
\usepackage{microtype}      
\usepackage{xcolor}         

\usepackage{multirow}
\usepackage{makecell}
\usepackage{adjustbox}
\usepackage{threeparttable}
\usepackage{pifont}

\usepackage{amsmath}
\usepackage{mathtools}
\usepackage{mdframed}
\usepackage{soul}
\usepackage{xcolor}
\usepackage{ragged2e}
\usepackage[ruled,vlined]{algorithm2e}

\tcbuselibrary{breakable}

\usepackage{comment}

\sethlcolor{lightgray}
\definecolor{lightgreen}{rgb}{0.56, 0.93, 0.56}
\definecolor{deepblue}{HTML}{003D79}

\title{\capsysname: Evaluating Interactive World Models Grounded on Physical Simulations}

\author{
    Lanxiang Hu\thanks{Main contributors.}\textsuperscript{\hspace{1.5mm}}$^{1}$
    \hspace{1pt}
    Yixin Huang\footnotemark[1]\textsuperscript{\hspace{1.5mm}}$^{1}$
    \hspace{1pt}
    Abhilash Shankarampeta\footnotemark[1]\textsuperscript{\hspace{1.5mm}}$^{1}$
    \hspace{1pt}
    Zilin Dai\footnotemark[1]\textsuperscript{\hspace{1.5mm}}$^{1}$
    \hspace{1pt}
    Yuyuan Wu\footnotemark[1]\textsuperscript{\hspace{1.5mm}}$^{1}$
    \hspace{1pt}
    \And
    \hspace{1pt}
    Kun Qian\footnotemark[1]\textsuperscript{\hspace{1.5mm}}$^{2}$
    \hspace{1pt}
    Junze Liu\footnotemark[1]\textsuperscript{\hspace{1.5mm}}$^{2}$
    \hspace{1pt}
    Arvind Srinivasan\footnotemark[1]\textsuperscript{\hspace{1.5mm}}$^{2}$
    \hspace{1pt}
    Mingjia Huo$^{1}$
    \hspace{1pt}
    Will Lin$^{1}$
    \hspace{1pt}
    Kai Zhong$^{2}$
    \And
    \hspace{1pt}
    Weiqi Zhang$^{2}$
    \hspace{1pt}
    Tian Wang$^{2}$
    \hspace{1pt}
    Qingjun Cui$^{2}$
    \hspace{1pt}
    Tajana Rosing$^{1}$
    \hspace{1pt}
    Hao Zhang$^{1}$ \\
    $^1$UC San Diego
    \hspace{1pt}
    $^2$Unity
}

\definecolor{darkmagenta}{HTML}{8B008B}

\newcommand{\capsysname}{\textsc{WMEval}}
\newcommand{\capsysnamespace}{\textsc{WMEval }}

\definecolor{lxagentgreen}{HTML}{1A9E5C}

\newif\ifcomments
\commentstrue
\ifcomments
    \providecommand{\ion}[1]{{\color{blue}{[ion: #1]}}}
\else
    \providecommand{\ion}[1]{}
\fi

\begin{document}

\maketitle

\begin{abstract}

Interactive world models hold immense potential for high-stakes applications in robotics, game engines, and immersive digital twins, yet existing benchmarks primarily focus on perceptual similarity rather than the physical-causal coherence essential for real-world utility. We introduce \capsysname, a comprehensive benchmark that evaluates interactive world models against Unity-based physical simulations. By leveraging Unity as a high-fidelity ground-truth oracle, we shift evaluation beyond perceptual visual assessment toward the objective state-based verification required for downstream applications. Spanning 9 diverse scenes, \capsysnamespace assesses action controllability, scene and character consistency, prompt adherence, kinematics, fluid mechanics, physical causality, and lighting. By executing matched action sequences in both the simulator and the world model, \capsysnamespace enables evaluation at both the perceptual level and the physical-causal level, using simulator-derived states as reference.  Experiments on state-of-the-art models reveal significant room for improvement in dimensions
that remain hidden by standard perceptual metrics but are vital for real-world deployments. Our code is available at: \textcolor{blue}{https://github.com/hao-ai-lab/WMEval}.

\end{abstract}

\section{Introduction}

Video generative models are increasingly positioned as interactive world-modeling engines~\citep{bruce2024genie,deepmind2025genie3,sun2026worldplay,wang2026matrix}: given observations and actions, they are expected to generate futures that are visually plausible and physically coherent~\citep{valevski2025diffusion,10.1609/aaai.v40i6.42465}. This shift moves video models beyond open visual prediction toward action-conditioned simulation~\citep{qin2025worldsimbench,zhang2026worldinworld,liu2026world}. However, evaluation remains challenging: current models struggle with controllability, scene consistency, and physical state evolution. 

Existing evaluations only partially address this need. Video-generation benchmarks primarily measure visual quality, spatio-temporal consistency, or prompt alignment~\citep{huang2024vbench,zheng2025vbench,Han_2025_CVPR}. Recent interactive world-model benchmarks evaluate scene consistency and action control through perceptual similarity to references~\citep{ye2026mind,qin2025worldsimbench}. Broader world-modeling video benchmarks further test scientific-law adherence or physics-related generation violations~\citep{10.1007/978-3-032-37624-4_9,li2025worldmodelbench}. However, for interactive applications such as robotics, autonomous driving, and gaming, an interactive world model must preserve consistency, controllability, and causal outcomes under interaction. Whether action-conditioned generations obey the same physical constraints as a reference engine therefore remains largely untested. Table~\ref{tab:benchmark_comparison} summarizes this gap. This calls for a standardized evaluation protocol specifying how action space should be configured, what references should be used, and how generations should be compared.

We introduce \capsysname, a comprehensive benchmark for evaluating interactive world models against physical simulation references. We build on Unity because it is a widely used real-time 3D development platform with mature tooling and a large asset ecosystem, making it practical to construct diverse scenes. Across 9 Unity scenes, \capsysnamespace pairs each scene with a set of deliberately designed action trajectories to stress test 6 different world modeling dimensions including \textit{action controllability}, \textit{scene and character consistency}, \textit{kinematics and mechanics}, \textit{fluid mechanics}, \textit{physical causality}, as well as \textit{lighting and shading}. 

By grounding evaluation in a Unity-based physical simulator, \capsysnamespace executes matched action trajectories in both the simulator and the interactive world model. We use Unity not just as a visual target, but as a controllable reference engine that provides simulator-derived states, and causal event traces. This enables \capsysnamespace to evaluate generations beyond perceptual quality through reference quantities such as velocity, momentum, energy, motion trajectories, and material properties.


\newcommand{\wmcmark}{\ding{51}}
\newcommand{\wmxmark}{\ding{55}}

\begin{table*}[t]
\centering
\small
\caption{Coverage comparison across video and world-model benchmarks. Unlike prior benchmarks that focus on video quality, prompt alignment, or physical-law adherence, \capsysnamespace evaluates interactive world models against physical-simulator references under matched actions.}
\label{tab:benchmark_comparison}
\begin{adjustbox}{max width=\textwidth}
\begin{tabular}{ccccccc}
\toprule
\makecell{Benchmark \\Category}
& \makecell{Representative \\Works}
& \makecell{Perceptual\\Quality}
& \makecell{Prompt\\Consistency}
& \makecell{Physical Law\\Adherence}
& \makecell{Action\\Controllability}
& \makecell{Causality} \\
\midrule

Video model benchmarks
& \makecell[c]{VBench~\citep{huang2024vbench}\\VBench-2.0~\citep{zheng2025vbench}\\Video-Bench~\citep{Han_2025_CVPR}}
& \wmcmark
& \wmcmark
& \wmxmark
& \wmxmark
& \wmxmark \\

\midrule

World-modeling video model benchmarks
& \makecell[c]{VideoScience-Bench~\citep{10.1007/978-3-032-37624-4_9}\\WorldModelBench~\citep{li2025worldmodelbench}}
& \wmcmark
& \wmcmark
& \wmcmark
& \wmxmark
& \wmxmark \\

\midrule

\multirow{2}{*}{\makecell[c]{Interactive world-model benchmarks}}
& \makecell[c]{MIND~\citep{ye2026mind}}
& \wmcmark
& \wmxmark
& \wmxmark
& \wmcmark
& \wmxmark \\

& \textbf{\capsysname} \textbf{(Ours)}
& \wmcmark
& \wmcmark
& \wmcmark
& \wmcmark
& \wmcmark \\

\bottomrule
\end{tabular}
\end{adjustbox}

\vspace{2pt}
\footnotesize
\textit{Legend.} \wmcmark: explicitly evaluated; \wmxmark: not a primary evaluation target.
\end{table*}

\capsysnamespace provides a metric suite that combines dimension-specific CV-based metrics for comparing generations with simulator-derived states, and holistic metrics for video-level similarity and prompt-scene consistency. We evaluate six state-of-the-art systems that are sufficiently capable and compatible with our standardized evaluation protocol. Results on \capsysnamespace reveal a substantial gap between current interactive world model systems and physical-simulator grounded references: models often preserve visual fidelity and action-responsive motion, but fail on simulator-consistent spatial relations, mechanical quantities, fluid invariants, lighting coherence, and multi-stage causal chains.



To summarize, in this work we make the following contributions:
\begin{itemize}
    \item We propose \capsysname, an interactive world-model benchmark that evaluates interactive world models through matched comparisons against physical-simulator references, using deliberately designed action trajectories across 9 diverse Unity scenes.

    \item We develop an evaluation suite that combines dimension-specific CV-based metrics, with holistic video-level and VLM-as-a-judge evaluations, covering video quality, prompt consistency, physical-law adherence, action controllability, and action-driven causality.

    \item We benchmark six state-of-the-art interactive world models and reveal world modeling failure cases that are not captured by video-quality or perceptual-similarity metrics alone. 
\end{itemize}

\begin{figure*}[t]
\centering
\includegraphics[width=\textwidth]{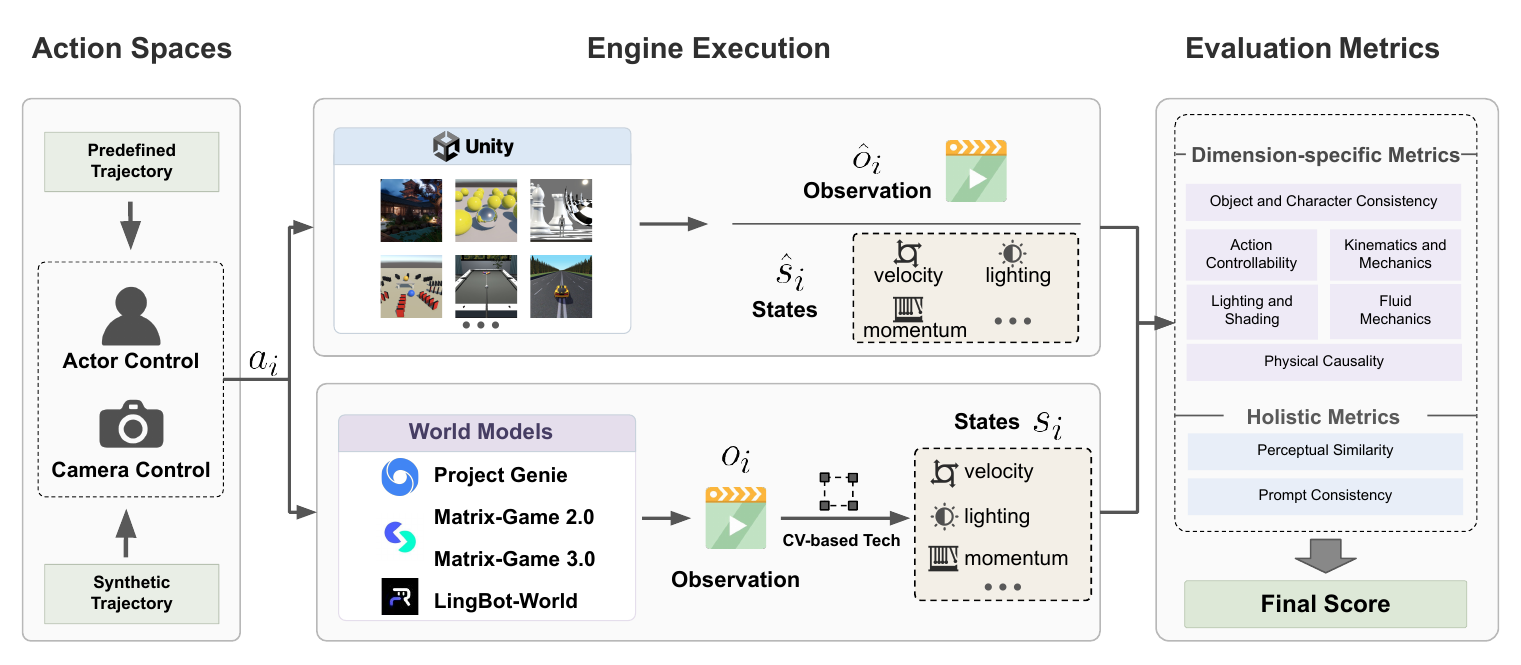}
\caption{\textbf{Overview of WMEval.} 
A shared action sequence $\{a_i\}_{i=1}^T$, sampled from predefined or rule-based  synthetic trajectories over actor and camera controls, is dispatched  to two parallel flows: a Unity physical simulator yielding reference observations $\hat{o}_i$ and simulator-derived states $\hat{s}_i$, and a world model under evaluation producing generated observations $o_i$, from which state-like quantities $s_i$ are recovered via CV-based estimators. 
}
    \label{fig:overview}
\end{figure*}

\section{Related Work}

\textbf{3D World Construction Models.}
One representative line of world-modeling work constructs explicit, navigable 3D world representations from text, images, or videos. Models such as HY-World 2.0 and Marble generate high-fidelity 3D environments by combining generative modeling with geometric representations such as 3D Gaussian Splatting~\citep{hyworld2026hyworld2,worldlabs2025marble}. These methods prioritize 3D scene reconstruction, view-consistent rendering, and interactive exploration, rather than explicitly modeling action-conditioned environment dynamics.


\textbf{Interactive World Models.}
The other line of work uses action-conditioned video diffusion models to simulate interactive environments that evolve over time. 
In \capsysname, we focus on image-to-video interactive world models
that generate rollouts autoregressively under either action-control or camera-control interactions. We evaluate Project Genie~\citep{bruce2024genie}, LingBot-World~\citep{gao2026lingbotworld}, Matrix-Game 2.0 and 3.0~\citep{he2025matrixgame2, wang2026matrix}, HY-WorldPlay~\citep{sun2026worldplay}, and Waypoint-1.5~\citep{overworld2026waypoint} because they represent a diverse set of latest interactive world-model systems with publicly accessible or experimentally usable interfaces.


\textbf{Interactive World-Model Benchmarks.}
Recent benchmarks have begun to evaluate video generators as world models. MIND evaluates memory consistency and action control in open-domain closed-loop generations~\citep{ye2026mind}, while WorldSimBench evaluates world simulators through perceptual assessment and action-level embodied tasks~\citep{qin2025worldsimbench}. More recently, WBench provides broad multi-turn evaluation of interactive world models across interaction adherence, consistency, and physics-related dimensions~\citep{ying2026wbench}. Broader video-world-model benchmarks such as WorldModelBench and VideoScience-Bench test instruction following, physics adherence, or scientific-law reasoning in generated videos~\citep{li2025worldmodelbench,10.1007/978-3-032-37624-4_9}. However, existing benchmarks remain limited in standardizing world-model inputs, interaction protocols, and evaluation protocols across models. \capsysnamespace fills this gap by providing simulator-based physical references, controlled action trajectories, and a comprehensive metrics suite. 

\newenvironment{compactscenes}{%
  \begingroup
  \titlespacing*{\paragraph}{0pt}{0.25em}{0.45em}%
  \titleformat{\paragraph}[runin]{\normalfont\normalsize\bfseries}{}{0pt}{}%
  \setlength{\parskip}{0.15em}%
}{%
  \endgroup
}

\section{\capsysname}
\label{sec:benchmark}


\subsection{Benchmark Design}

Figure~\ref{fig:overview} summarizes the end-to-end \capsysnamespace
evaluation protocol. The benchmark consists of three stages.
First, for each scene and evaluation dimension, we define a
\emph{semantic action trajectory} over actor and camera controls.
Second, this trajectory is executed in parallel by a Unity reference
environment and the interactive world model under evaluation. Unity
provides both rendered observations and simulator-derived physical quantities,
while the world model exposes generated observations only.
Third, \capsysnamespace compares the two rollouts using dimension-specific
metrics and holistic video-level metrics. When simulator-grounded quantities
are required, CV- or VLM-based evaluators recover the corresponding
physical quantities from the world-model rollouts. We next introduce the Unity scenes, formalize how matched action-conditioned rollouts are generated, and describe the dimension-specific and holistic metrics used to evaluate them.


\subsubsection{Scenes}
\label{sec:scenes}

We construct nine controlled Unity environments as reference testbeds for
interactive world models. Each scene exposes controls, observations, and simulator-derived states through Unity ML-Agents~\citep{unitymlagents}, enabling paired execution of the same semantic action trajectory in the simulator and the world model. Together, the scenes target six evaluation dimensions: \textit{action controllability}, \textit{scene and character consistency}, \textit{kinematics and mechanics}, \textit{fluid mechanics}, \textit{physical causality}, and \textit{lighting and shading}. Detailed scene configurations, action spaces, and simulator states are
provided in Appendix~\ref{app:scenes}.

For action-conditioned dynamics, \texttt{BallRoller} uses a controllable
sphere moving through ramps, obstacles, and passive objects to test action
responsiveness and motion consistency. \texttt{BeamDomino} triggers a
multi-stage \textsc{rolling}$\rightarrow$\textsc{impact}$\rightarrow$
\textsc{topple}$\rightarrow$\textsc{cascade} sequence, providing a controlled
test of action-driven physical causality~\citep{li2026causalstep}.
\texttt{BilliardGame} uses three-ball carom billiards with deterministic
physics to test collision dynamics and post-collision motion, while
\texttt{RacingCar} evaluates vehicle motion under steering, throttle, and braking.

For scene and appearance consistency, \texttt{ChessBoard} places a humanoid
character among large reflective objects, providing a testbed for character
and lighting consistency. \texttt{MirrorRollingBall} increases scene
complexity with a reflective ball and up to 48 surrounding objects, stressing
multi-object spatial consistency and tracking. \texttt{SunRisesAndSets}
isolates lighting and shading through controlled camera motion and directional
illumination, while \texttt{WanderingGarden} provides the most visually
complex scene, with foliage, particles, NPC motion, and designated anchor
objects for evaluating spatial and object consistency.

Finally, \texttt{LiquidContainer} manipulates a sealed bottle through
six-degree-of-freedom motion to induce controlled tilt and sloshing. Its
stylized liquid dynamics provide a reproducible reference for sealed-container
motion rather than a general-purpose fluid simulation.

\subsubsection{Evaluation Settings}

\paragraph{Action Space and Action Trajectories.}
Each scene $e$ defines a semantic action space $\mathcal{A}^{(e)}$ with $d_c^{(e)}$ continuous controls and, when needed, $d_d^{(e)}$ discrete branches:
\begin{equation}
    \mathcal{A}^{(e)}
    =
    [-1,1]^{d_c^{(e)}}
    \times
    \prod_{k=1}^{d_d^{(e)}} \{0,\ldots,K_k^{(e)}-1\}.
\end{equation}
These controls represent scene-level semantics such as actor motion, steering, braking, camera motion, or object manipulation. For each evaluation dimension $d$, we define scene-specific trajectories
\begin{equation}
    \Gamma_d
    =
    \{(e_i,\gamma_i^{(d)})\}_{i=1}^{N_d},
    \qquad
    \gamma_i^{(d)}
    =
    (a^{(i)}_1,\ldots,a^{(i)}_{T_i}),
    \quad
    a^{(i)}_t \in \mathcal{A}^{(e_i)} .
\end{equation}

Each trajectory is designed to elicit a measurable target behavior. For example, controllability trajectories induce directional motion, mechanics trajectories trigger collisions, and causality trajectories trigger multi-stage event chains.

\paragraph{Action Adaptation.}
Interactive world models expose heterogeneous native control interfaces. We therefore match actions at the level of semantic controls and physical time rather than low-level action tokens. For each model, a lightweight adapter maps the semantic trajectory to the closest supported native controls and aligns their duration according to the model's frame rate or action stride. A model is evaluated only on dimensions for which its interface supports the required control semantics.

\paragraph{Matched Rollouts.}
Given semantic action $a_{i+1}$, the Unity reference produces
\begin{equation}
    (\hat{o}_{i+1}, \hat{s}_{i+1}, \hat{m}_{i+1})
    =
    \mathcal{E}^{(e)}(\hat{s}_i,a_{i+1}),
\end{equation}
where $\hat{o}_{i+1}$ is the rendered observation, $\hat{s}_{i+1}$ the simulator state, and $\hat{m}_{i+1}$ evaluation metadata such as object velocities, trajectories, lighting states, and event labels. The world model receives the controls and produces observations,
\begin{equation}
    o_{i+1}
    \sim
    p_\theta(o_{i+1}\mid o_{\le i},a^{(m)}_{\le i+1}),
\end{equation}
while its underlying state remains latent. This yields paired rollouts indexed by the same semantic action trajectory and physical timing. Simulator states $\{\hat{s}_i,\hat{m}_i\}_{i=1}^{T}$ then provide the reference quantities for the simulator-grounded metrics.

\paragraph{Simulator Implementation.} \capsysnamespace works on any simulator exposing the required reference rollouts and physical states can define benchmark cases. Once packaged, evaluation only requires generating from the seed frame under the prescribed actions. We provide a Unity-based reference package in the supplementary material.

\subsection{Metrics Design}

\capsysnamespace evaluates generated rollouts using dimension-specific and
holistic metrics. Dimension-specific metrics compare candidate-side
measurements against Unity-derived states and event traces to diagnose
consistency, controllability, physical dynamics, lighting, and causal
transitions. Holistic metrics evaluate video-level perceptual similarity and
prompt-scene consistency. 
Detailed metric definitions, calibration, and implementation details are provided in Appendix~\ref{app:metrics}.



\paragraph{Action Controllability.}
This dimension evaluates whether generated motion follows the intended action
or camera-control input. Using RAFT optical flow~\citep{teed2020raft}, we
separate state-level agreement with Unity-reference flow
(\textit{Ground Truth RAFT}) from alignment with the commanded motion
(\textit{Synthetic RAFT}).

\paragraph{Physical Causality.}
This dimension evaluates whether matched actions preserve multi-stage causal
state transitions. Following CausalStep~\citep{li2026causalstep}, we measure
full- and partial-chain preservation and causal reasoning, applying the same
judging pipeline to Unity references and world-model rollouts to quantify
reference-side judge error. Detailed metrics are provided in
Appendix~\ref{app:physical_causality_metrics}.

\paragraph{Kinematics and Mechanics.}
This dimension evaluates whether matched controls produce simulator-consistent
motion and mechanical outcomes. We report Final Displacement Error (\textit{FDE}) and Dynamic Time Warping (\textit{DTW}) for trajectory agreement, together with \textit{ExactMatch} and \textit{ErrorRate} for quantities such as terminal momentum and kinetic
energy.

\paragraph{Fluid Mechanics.}
This dimension evaluates sealed-container slosh against a controlled
parametric reference in which surface tilt and wobble depend on container
acceleration. We measure motion-field agreement, tracked-region stability,
orientation consistency, slosh-spectrum agreement, and LPIPS. The
candidate-side measurements depend on SAM~2 tracking; its Unity-side
calibration and the resulting scope of these metrics are detailed in
Appendix~\ref{app:fluid_validity}.

\paragraph{Lighting and Shading.}
This dimension evaluates whether illumination and cast shadows remain
geometrically and temporally coherent under controlled camera motion. We
report scene retention (Scene Retention Rate, \textit{SRR}), shadow presence
(Shadow Detection Rate, \textit{SDR}), light-direction coherence
(Multi-Object Sun Disagreement, \textit{MOSD}), temporal smoothness
(Shadow Direction Jitter, \textit{SDJ}), footprint stability
(Shadow Area Stability, \textit{SAS}), and shape stability
(Shadow Shape Drift, \textit{SSD}), via GroundingDINO, SAM~2, Hu moments
~\citep{10.1007/978-3-031-72970-6_3,
ravi2025sam, hu1962visual}.

\paragraph{Environment Object and Character Consistency.}
This dimension evaluates whether objects, spatial relations, and character
identity persist under controlled camera motion. We report \textit{Depth
AbsRel} and \textit{DistMAE} for geometry, \textit{ObjPerm} for anchor-object
persistence, and \textit{CharConsis} for character identity, using
GroundingDINO, SAM 2, and DINOv2~\citep{oquab2024dinov}.

\section{Experiments}



\subsection{Experiment Settings}
We evaluate six state-of-the-art interactive world models: Google Project Genie~\citep{bruce2024genie,deepmind2025genie3}, LingBot-World~\citep{gao2026lingbotworld}, Matrix-Game 2.0~\citep{he2025matrixgame2}, Matrix-Game 3.0~\citep{wang2026matrix}, HY-WorldPlay~\citep{sun2026worldplay}, and Waypoint-1.5~\citep{overworld2026waypoint}. Since these systems expose different interfaces, we evaluate each dimension on the subset of models that satisfies its interface requirements, e.g.,~third-person camera control for object consistency and action control for mechanics and causality. 
The open-source model runs were executed on a single host with 8 NVIDIA A100 GPUs. Unless otherwise noted, we report mean $\pm$ standard deviation over five generations. 
For the model comparisons that support our main conclusions, we additionally test whether the conclusion remains unchanged over all $\binom{5}{3}=10$ three-run subsets and apply metric-appropriate statistical tests.



\subsubsection{Measurement Reliability}

\paragraph{Evaluator and Generation Size Calibration.} We calibrate learned evaluators against simulator-derived physical quantities and audit their validity on generated videos. Across evaluation dimensions, the resulting calibration errors are sufficiently small for the reported comparisons within each metric's stated scope (Table~\ref{tab:evaluator_validity}). Full calibration and validity analyses are in Appendix~\ref{app:evaluator_validity}. 
Moreover, we ensure the main findings do not depend on a particular choice of five generations. The controllability ordering is unchanged across all $\binom{5}{3}=10$ three-run subsets. When the evidence does not separate two models for example, Project Genie and LingBot-World on prompt consistency -- we report them as comparable. Full analyses are in Appendix~\ref{app:statistical_robustness}.



\subsection{Results and Insights}

Across evaluation dimensions, three recurring gaps emerge in current world models. First, following an immediate control does not guarantee correct causal state evolution. Second, perceptually plausible generations can still violate physical quantities or multi-stage causal structure. Third, consistency along one axis, such as temporal smoothness or foreground appearance, does not imply consistency along others, such as spatial geometry or lighting. We analyze these failures below.


\subsubsection{Action and Causal State Evolution}


\begin{table*}[t]
\centering
\small
\caption{Controllability evaluation via optical-flow analysis. 
Cells report mean $\pm$ std over 5 generations per scene (Appendix~\ref{app:raft_pipeline}). Waypoint-1.5 exposes first-person camera control only; it is scored with Ground Truth RAFT on the two scenes.
}
\label{tab:worldmodel_controllability}
\begin{adjustbox}{max width=\textwidth}
\begin{tabular}{lllccc}
\toprule
Model & Method & Metric & BallRoller & BeamDomino & MirrorRollingBall \\
\midrule
\multirow{4}{*}{Project Genie}
& \multirow{2}{*}{Ground Truth RAFT}
& Raw Relative EPE Dynamic Mean $\downarrow$
& 0.55 $\pm$ 0.11 & 0.44 $\pm$ 0.07 & 0.77 $\pm$ 0.05 \\
&
& Raw Cosine Similarity Dynamic Mean $\uparrow$
& 0.71 $\pm$ 0.19 & $\mathbf{0.98 \pm 0.00}$ & 0.84 $\pm$ 0.10 \\
& \multirow{2}{*}{Synthetic RAFT}
& Mean Flow Cosine $\uparrow$
& -0.33 $\pm$ 0.35 & $\mathbf{0.99 \pm 0.00}$ & 0.98 $\pm$ 0.01 \\
&
& Pixel EPE Mean $\downarrow$
& 665.46 $\pm$ 55.24 & $\mathbf{841.07 \pm 145.12}$ & $\mathbf{270.86 \pm 73.49}$ \\
\midrule
\multirow{4}{*}{LingBot-World (Act)}
& \multirow{2}{*}{Ground Truth RAFT}
& Raw Relative EPE Dynamic Mean $\downarrow$
& $\mathbf{0.16 \pm 0.01}$ & $\mathbf{0.34 \pm 0.03}$ & $\mathbf{0.49 \pm 0.10}$ \\
&
& Raw Cosine Similarity Dynamic Mean $\uparrow$
& $\mathbf{0.99 \pm 0.00}$ & $\mathbf{0.98 \pm 0.00}$ & $\mathbf{0.94 \pm 0.06}$ \\
& \multirow{2}{*}{Synthetic RAFT}
& Mean Flow Cosine $\uparrow$
& $\mathbf{1.00 \pm 0.00}$ & $\mathbf{0.99 \pm 0.00}$ & $\mathbf{1.00 \pm 0.00}$ \\
&
& Pixel EPE Mean $\downarrow$
& $\mathbf{620.90 \pm 33.32}$ & 970.21 $\pm$ 73.09 & 678.48 $\pm$ 214.26 \\
\midrule
\multirow{2}{*}{Waypoint-1.5}
& \multirow{2}{*}{Ground Truth RAFT}
& Raw Relative EPE Dynamic Mean $\downarrow$
& 0.50 $\pm$ 0.15 & 0.51 $\pm$ 0.03 & N/A \\
&
& Raw Cosine Similarity Dynamic Mean $\uparrow$
& 0.94 $\pm$ 0.04 & 0.97 $\pm$ 0.01 & N/A \\
\bottomrule
\end{tabular}
\end{adjustbox}
\end{table*}

\paragraph{Controllability is necessary but insufficient for faithful state evolution.} Physical causality first requires the model to respond to the intended control: if an action fails to trigger the correct motion, its downstream physical consequences cannot be correct. We therefore ask whether successful immediate action following is also sufficient for faithful state evolution. Table~\ref{tab:worldmodel_controllability} shows that it is not. On \texttt{MirrorRollingBall}, Project Genie and LingBot-World nearly perfectly follow the general commanded direction (Synthetic-RAFT cosine $0.98$ and $1.00$), yet their subsequent state evolution still deviates substantially from the Unity reference (relative flow errors of $0.77$ and $0.49$, respectively). The same pattern appears on \texttt{BeamDomino}: both achieve a Synthetic-RAFT cosine of $0.99$, while retaining state-level errors of $0.44$ and $0.34$. 
By contrast, Project Genie already fails at the control stage on \texttt{BallRoller}, with a Synthetic-RAFT cosine of $-0.33$ versus $1.00$ for LingBot-World. Failures thus arise either when the action is not executed or when a correct response later evolves toward the wrong physical state.
Per-scene analyses are in Appendix~\ref{app:raft_per_scene}.


\begin{table}[t]
  \centering
  \small
  \caption{
  Physical Causality on \texttt{BeamDomino} ($n=15$: $5$ rollouts $\times$ $3$ trajectory shapes). CSR reports VLM judgment on whether the full causal chain is answered correctly for a trajectory, so we report success counts with $95\%$ confidence intervals. All remaining CausalStep columns report degradation relative to Unity, $\overline{\Delta}=m_{\textsc{unity}}-m_{\textsc{wm}}$. For example, $\overline{\Delta}\mathrm{ICRA}=10$ means isolated causal-question accuracy is $10$ percentage points below Unity. Details on per-trajectory results are provided in Table~\ref{tab:causality_per_traj}.
  }
  
  \label{tab:causal-chain}
  \label{tab:causal-physical-causality}
  \begin{adjustbox}{max width=\textwidth}
  \begin{tabular}{lcccccccc}
    \toprule
    & \multicolumn{3}{c}{Causal-chain preservation}
    & \multicolumn{5}{c}{CausalStep deltas} \\
    \cmidrule(lr){2-4}\cmidrule(lr){5-9}
    Model
      & CSR $\uparrow$ & $95\%$ CI & Fisher $p$ vs.\ ref.
      & $\overline{\Delta}$AMCL $\downarrow$
      & $\overline{\Delta}$MCL $\downarrow$
      & $\overline{\Delta}$WS $\downarrow$
      & $\overline{\Delta}$DUA $\downarrow$
      & $\overline{\Delta}$ICRA $\downarrow$ \\
    \midrule
    Project Genie
      & $3/15$ & $[4.3,\ 48.1]$ & $0.0007$
      & $0.80{\pm}0.75$
      & $0.80{\pm}0.75$
      & $2.60{\pm}2.41$
      & $13.9{\pm}27.6$
      & $10.0{\pm}18.4$ \\
    \bottomrule
  \end{tabular}
  \end{adjustbox}
  \\[2pt]
\end{table}

\paragraph{Short-horizon causal understanding does not compose into full causal chains.} Finally, \texttt{BeamDomino} tests whether an action propagates through the four-stage \textsc{rolling}$\rightarrow$\textsc{impact}$\rightarrow$ \textsc{topple}$\rightarrow$\textsc{cascade} chain, rather than merely producing recognizable individual events. For example, Project Genie generations remain close to Unity references on Isolated Causal Reasoning Accuracy (ICRA; $90\%$ vs.\ $100\%$), meaning that it can usually answer causal questions about individual stages for instance, \textit{Why does the red ball begin to move in this video segment?} Yet it preserves the complete causal chain in only $3/15$ rollouts, compared with $13/15$ for Unity.
This failure is systematic across trajectory types: none of its ten \texttt{glancing} and \texttt{direct\_strike} rollouts preserves the full chain. 
These results reveal a compositional gap: plausible individual causal transitions do not guarantee that their physical consequences propagate correctly through the full event chain. Further causal trajectory analyses are provided in Appendix~\ref{appx:physics_causality}.

\begin{table*}[t]
\centering
\small
\caption{Mechanical physical law evaluation on \texttt{RacingCar} (third person, 3P) and \texttt{BilliardGame} (first person, 1P). We report trajectory consistency using DTW and final displacement error (FDE, m); \texttt{RacingCar} trajectories are measured in road coordinates, from the road's motion relative to the camera. Mechanical-law fidelity is the robust error: the mean momentum and kinetic-energy error over the motion span, normalized by the reference peak. Unity self-match references are in Table~\ref{tab:physical_laws_unity}. Cells show mean $\pm$ sample standard deviation over 5 generations. 
$^{\ast}$the car stays parked and leaves the view as the camera passes it.}
\label{tab:worldmodel_physical_laws}
\begin{adjustbox}{max width=\textwidth}
\begin{tabular}{lllcc}
\toprule
Model
& Category
& Metric
& Racing Car (3P)
& Billiard Game (1P) \\
\midrule



\multirow{3}{*}{Project Genie}
& \multirow{2}{*}{Kinematics}
& DTW $\downarrow$
& $\mathbf{2.28 \pm 0.54}$
& $0.81 \pm 0.19$ \\
&
& FDE $\downarrow$
& $\mathbf{9.39 \pm 2.25}$
& $0.83 \pm 0.52$ \\
& Mechanical Laws
& Robust Error (\%) $\downarrow$
& $25.7 \pm 2.5$
& $26.8 \pm 0.4$ \\

\midrule

\multirow{3}{*}{Matrix-Game 2.0}
& \multirow{2}{*}{Kinematics}
& DTW $\downarrow$
& $17.10 \pm 0.27$
& $\mathbf{0.64 \pm 0.00}$ \\
&
& FDE $\downarrow$
& $29.65 \pm 0.20$
& $\mathbf{0.45 \pm 0.01}$ \\
& Mechanical Laws
& Robust Error (\%) $\downarrow$
& N/A$^{\ast}$
& $\mathbf{23.3 \pm 0.2}$ \\

\bottomrule
\end{tabular}
\end{adjustbox}
\end{table*}

\paragraph{Contact-dependent mechanics remain hard.} We next ask whether the resulting trajectories preserve mechanical consequences under matched controls. Table~\ref{tab:worldmodel_physical_laws} compares the relatively smooth vehicle motion in \texttt{RacingCar} with the collision-dependent dynamics of \texttt{BilliardGame}.
On \texttt{BilliardGame}, both models track the ball closely (FDE $0.83$ and $0.45$\,m), yet their robust mechanics error ($26.8\%$ and $23.3\%$) is about five times Unity's $4.9\%$ (Table~\ref{tab:physical_laws_unity}). On \texttt{RacingCar}, Project Genie executes the commanded turn (DTW $2.28$\,m) but overshoots the endpoint by $9.39$\,m, with a similar $25.7\%$ mechanics error; Matrix-Game 2.0 instead advances the camera past a car that stays parked, so its FDE ($29.65$\,m) equals that of a stationary car. Generating action-responsive motion is thus easier than reproducing its momentum and energy, with or without contact.


\subsubsection{Perceptual Quality and Physical Adherence}

\begin{table*}[t]
\centering
\small
\caption{Causal LPIPS evaluation (Eq.~\ref{eq:clpips}). Causal contact events are matched chronologically between the Unity reference and world model generations, and LPIPS is computed on frame pairs at offsets $\{-2,-1,0,+1,+2\}$ around each match (Appendix~\ref{app:event_detection}). Generations with no matched events are excluded. Cells report mean $\pm$ std over 5 generations per scene.
}
\label{tab:worldmodel_causal_lpips}
\begin{adjustbox}{max width=\textwidth}
\begin{tabular}{llccc}
\toprule
Model & Metric & BallRoller & BeamDomino & MirrorRollingBall \\
\midrule
Project Genie         & Causal LPIPS $\downarrow$ & 0.56 $\pm$ 0.02 & $\mathbf{0.60 \pm 0.02}$ & 0.72 $\pm$ 0.02 \\
LingBot-World (Act)   & Causal LPIPS $\downarrow$ & $\mathbf{0.38 \pm 0.02}$ & 0.61 $\pm$ 0.02 & $\mathbf{0.62 \pm 0.05}$ \\
\bottomrule
\end{tabular}
\end{adjustbox}
\end{table*}

\paragraph{Perceptual similarity can hide distinct interaction failures.}
Perceptual metrics are widely used as proxies for rollout fidelity: MIND reports LPIPS, SSIM, and PSNR, while iWorld-Bench uses LPIPS, SSIM, and MSE for motion smoothness~\citep{ye2026mind,fang2026iworldbench}. Yet visual similarity does not reveal whether the interaction evolves correctly. On \texttt{BeamDomino}, Project Genie's cLPIPS is nearly identical to LingBot-World's ($0.60$ vs.\ $0.61$), yet it preserves the full causal chain in only $3/15$ rollouts versus $13/15$ for Unity. On \texttt{BallRoller}, Project Genie has cLPIPS $0.56$ versus $0.38$ for LingBot-World, while its generated motion is misaligned with the commanded direction (Synthetic-RAFT cosine $-0.33$ vs.\ $1.00$). Thus, perceptual similarity captures appearance-level divergence but can obscure failures in control execution and causal state propagation. 


\begin{table}[t]
\centering
\caption{Sealed-container slosh evaluation on \texttt{LiquidContainer}.
Mean $\pm$ std over $4$ action profiles $\times$ $5$ generations per model.
Flow cosine is a diagnostic for motion; the remaining columns compare reference-grounded area stability, surface orientation, slosh spectrum, and whole-frame perceptual similarity. $L_1$ does not separate the two models, so no winner is highlighted. 
Detailed metric definitions and estimator validity are provided in Appendix~\ref{app:fluid_metrics} and Table~\ref{tab:fluid-validity}.}
\label{tab:fluid-headline}
\small
\begin{adjustbox}{max width=\textwidth}
\begin{tabular}{lrrrrrr}
\toprule
& \multicolumn{1}{c}{\emph{Diagnostic}} & \multicolumn{4}{c}{\emph{Reference-grounded dynamics}}
& \multicolumn{1}{c}{\emph{Perceptual}} \\
\cmidrule(lr){2-2}\cmidrule(lr){3-6}\cmidrule(lr){7-7}
Model & cos $\uparrow$ & $\Delta\sigma_A\!\downarrow$ &
$|\Delta\dot A|\!\downarrow$ & $|\Delta\theta_{\mathrm{rms}}|\!\downarrow$ &
$L_1\!\downarrow$ & LPIPS $\downarrow$ \\
\midrule
Project Genie    & $0.818{\pm}0.167$         & $0.171{\pm}0.314$         & $0.017{\pm}0.036$         & $\mathbf{8.87{\pm}7.70}$ & $1.243{\pm}0.280$ & $0.624{\pm}0.021$ \\
HY-WorldPlay     & $\mathbf{0.985{\pm}0.008}$ & $\mathbf{0.015{\pm}0.023}$ & $\mathbf{0.001{\pm}0.001}$ & $9.62{\pm}9.80$          & $\mathbf{1.203{\pm}0.305}$ & $\mathbf{0.300{\pm}0.040}$ \\
\bottomrule
\end{tabular}
\end{adjustbox}
\end{table}

On \texttt{LiquidContainer}, Table~\ref{tab:fluid-headline} evaluates whether the liquid moves like the reference, maintains a stable apparent size and surface orientation, matches its oscillation pattern, and remains perceptually similar. Compared to Project Genie, HY-WorldPlay achieves a much lower LPIPS ($0.300$ vs.\ $0.624$), yet it exhibits more surface orientation deviations. It also renders the bottle at about half of the reference's apparent size (median scale ratio $0.51$, Table~\ref{tab:fluid-validity}), allowing the well-matched background to dominate LPIPS. Thus, strong perceptual similarity does not imply physical adherence.

\paragraph{Physical axes expose distinct failure modes across models.} 
HY-WorldPlay keeps the liquid's apparent size stable but tilts its surface unlike Unity, whereas Project Genie keeps the tilt closer but lets the liquid area and flow drift; on third-person \texttt{RacingCar}, Project Genie turns the car but overshoots, whereas Matrix-Game~2.0 moves the camera instead of the car.
Additional results and analyses for fluid mechanics are provided in Appendix~\ref{appx:fluid_mechanics}. 

\subsubsection{Local and Long-Horizon Consistency}

\newcommand{\sdev}[1]{{\scriptsize $\pm$#1}}

\begin{table*}[t]
\centering
\scriptsize
\setlength{\tabcolsep}{2.4pt}
\renewcommand{\arraystretch}{0.88}

\caption{Environmental object and character consistency evaluation.
Outward-facing views; character consistency pools both orientations. Cells report mean $\pm$ sample standard deviation across 5 runs. Geometry errors require anchor reference objects. N/A denotes non-applicable metrics in the setting.}
\label{tab:worldmodel_consistency}

\begin{adjustbox}{max width=\textwidth}
\begin{tabular}{llcccccc}
\toprule
\multirow{2}{*}{Model}
& \multirow{2}{*}{Metric}
& \multicolumn{2}{c}{Garden}
& \multicolumn{2}{c}{Billiard Game}
& \multicolumn{2}{c}{MirrorRollingBall} \\
\cmidrule(lr){3-4}
\cmidrule(lr){5-6}
\cmidrule(lr){7-8}

&
& {Perturb.}
& {360$^\circ$}
& {Perturb.}
& {360$^\circ$}
& {Perturb.}
& {360$^\circ$} \\

\midrule

\multirow{3}{*}{Project Genie}
& Depth $\downarrow$
& 169.9\sdev{48.1}
& 117.5\sdev{18.6}
& \textbf{14.9\sdev{6.3}}
& \textbf{11.5\sdev{6.4}}
& N/A & N/A \\

& Dist. $\downarrow$
& 78.5\sdev{41.1}
& 52.5\sdev{6.9}
& 76.9\sdev{35.4}
& 38.4\sdev{35.4}
& 36.9\sdev{4.4}
& \textbf{29.1\sdev{13.0}} \\


& Char. Cons. $\uparrow$
& N/A
& N/A
& 76.2\sdev{18.7}
& 29.8\sdev{14.9}
& \textbf{99.9\sdev{0.2}}
& \textbf{84.9\sdev{20.9}} \\

\midrule

\multirow{3}{*}{LingBot-World (Cam)}
& Depth $\downarrow$
& N/A
& 55.1\sdev{7.0}
& 25.4\sdev{5.8}
& 16.6\sdev{6.5}
& N/A & N/A \\

& Dist. $\downarrow$
& N/A
& \textbf{46.0\sdev{11.5}}
& 65.0\sdev{42.9}
& 69.7\sdev{36.3}
& \textbf{29.7\sdev{1.0}}
& 44.4\sdev{83.0} \\


& Char. Cons. $\uparrow$
& N/A
& N/A
& 66.5\sdev{36.2}
& \textbf{88.1\sdev{11.2}}
& 51.3\sdev{21.9}
& 65.0\sdev{12.7} \\

\midrule

\multirow{3}{*}{Matrix-Game 3.0}
& Depth $\downarrow$
& N/A
& 37.6\sdev{15.6}
& 32.8\sdev{16.0}
& 34.6\sdev{17.2}
& N/A & N/A \\

& Dist. $\downarrow$
& N/A
& 61.4\sdev{18.5}
& \textbf{26.8\sdev{5.4}}
& \textbf{20.9\sdev{12.2}}
& 88.5\sdev{26.6}
& 72.6\sdev{47.6} \\


& Char. Cons. $\uparrow$
& N/A
& N/A
& 91.2\sdev{19.7}
& 16.0\sdev{2.1}
& 14.8\sdev{19.7}
& 4.7\sdev{2.5} \\

\midrule

\multirow{3}{*}{Matrix-Game 2.0}
& Depth $\downarrow$
& N/A
& \textbf{33.1\sdev{9.2}}
& 25.2\sdev{13.4}
& 23.9\sdev{14.6}
& N/A & N/A \\

& Dist. $\downarrow$
& N/A
& N/A
& 44.9\sdev{9.1}
& 56.4\sdev{8.0}
& 65.3\sdev{13.0}
& 78.3\sdev{35.8} \\


& Char. Cons. $\uparrow$
& N/A
& N/A
& \textbf{95.0\sdev{6.7}}
& 14.7\sdev{0.6}
& 48.3\sdev{3.5}
& 6.7\sdev{2.8} \\

\bottomrule
\end{tabular}
\end{adjustbox}

\vspace{-2mm}
\end{table*}

\paragraph{Large viewpoint changes expose spatial inconsistency.} Table~\ref{tab:worldmodel_consistency} evaluates whether environment background, objects, and characters remain stable as the camera departs from the initial view. The results show substantial spatial inconsistency across models. On \texttt{Garden}, Project Genie exhibits large depth error, ranging from $117.5\%$ to $169.9\%$. On \texttt{BilliardGame}, its character consistency drops from $76.2\%$ under perturbation to $29.8\%$ under the full $360^\circ$ trajectory. 
\texttt{MirrorRollingBall} is more stable, retaining $99.9\%$ character consistency under perturbation and $84.9\%$ under $360^\circ$ rotation. The degradation is more pronounced for weaker models: Matrix-Game 3.0 drops from $91.2\%$ to $16.0\%$ character consistency on \texttt{BilliardGame} and from $14.8\%$ to $4.7\%$ on \texttt{MirrorRollingBall}, while Matrix-Game 2.0 falls from $95.0\%$ to $14.7\%$ and from $48.3\%$ to $6.7\%$, respectively. Thus, maintaining persistent scene geometry and character identity over large camera motion remains challenging even when local views appear visually consistent.


\begin{table*}[t]
\centering
\small
\caption{Lighting and shading evaluation on the \texttt{SunRisesAndSets} scene. Cells report mean $\pm$ std over $5$ rollouts per model. Metric definitions in Appendix~\ref{app:lighting_metrics}.}
\label{tab:worldmodel_lighting_shading}
\begin{adjustbox}{max width=\textwidth}
\begin{tabular}{lcccccc}
\toprule
\multirow{2}{*}{Model} 
& \multicolumn{2}{c}{Coverage} 
& Shadow Geometry 
& \multicolumn{3}{c}{Shadow Dynamics} \\
\cmidrule(lr){2-3} \cmidrule(lr){4-4} \cmidrule(lr){5-7}
& \shortstack{Scene\\Retention (\%)$\uparrow$} 
& \shortstack{Shadow\\Detection (\%)$\uparrow$} 
& \shortstack{Multi-Object Sun\\Disagreement ($^\circ$)$\downarrow$} 
& \shortstack{Shadow Direction\\Jitter ($^\circ$/frame)$\downarrow$} 
& \shortstack{Shadow Area\\Stability (\%)$\downarrow$} 
& \shortstack{Shadow Shape\\Drift (Hu L2)$\downarrow$} \\
\midrule
Project Genie 
& $96.48 \pm 7.86$ & $97.37 \pm 2.87$ & $65.78 \pm 8.51$ & $2.42 \pm 2.05$ & $13.95 \pm 4.84$ & $6.78 \pm 1.03$ \\
LingBot-World 
& $\mathbf{100.00 \pm 0.00}$ & $\mathbf{100.00 \pm 0.00}$ & $60.06 \pm 14.20$ & $7.61 \pm 11.19$ & $17.97 \pm 10.23$ & $6.66 \pm 1.44$ \\
Matrix-Game 2.0 
& $22.92 \pm 16.23$ & $\mathbf{100.00 \pm 0.00}$ & $65.26 \pm 9.58$ & $4.12 \pm 3.78$ & $14.22 \pm 2.42$ & $7.37 \pm 1.18$ \\
Matrix-Game 3.0 
& $14.41 \pm 7.06$ & $91.50 \pm 9.34$ & $\mathbf{59.07 \pm 13.81}$ & $23.18 \pm 19.09$ & $62.43 \pm 30.57$ & $11.43 \pm 1.90$ \\
HY-WorldPlay
& $73.92 \pm 2.44$ & $99.78 \pm 0.50$ & $77.29 \pm 5.61$ & $6.60 \pm 1.94$ & $\mathbf{10.30 \pm 1.17}$ & $5.75 \pm 0.63$ \\
Waypoint-1.5
& $\mathbf{100.00 \pm 0.00}$ & $\mathbf{100.00 \pm 0.00}$ & $98.29 \pm 2.49$ & $\mathbf{2.34 \pm 0.75}$ & $12.29 \pm 2.45$ & $\mathbf{4.55 \pm 0.57}$ \\
\bottomrule
\end{tabular}
\end{adjustbox}
\end{table*}


\paragraph{Stable-looking shadows do not imply a coherent global light source.}
Table~\ref{tab:worldmodel_lighting_shading} asks whether models keep the anchor scene in view, maintain stable individual shadows, and make shadows across objects agree on a common light direction. Project Genie and LingBot-World retain the anchor scene in $96.5\%$ and $100\%$ of frames, yet their inferred sun directions from different objects disagree by $65.8^\circ$ and $60.1^\circ$. Waypoint-1.5 makes the gap clearest: it retains the scene throughout the rollout and produces the smoothest, most shape-stable shadows, yet has the largest sun-direction disagreement ($98.3^\circ$). Thus, locally plausible shadows need not correspond to a coherent scene-level light source.


\begin{table*}[!ht]
\centering
\scriptsize
\setlength{\tabcolsep}{4pt}
\renewcommand{\arraystretch}{0.88}
\caption{Prompt-scene consistency results measured by the rubric-based VLM judge. Scores combine character (object) and environment rubrics (first-person RacingCar: environment only).
Mean $\pm$ std over 5 generations. Waypoint-1.5 has no text channel.}
\label{tab:vlm_eval}
\begin{adjustbox}{max width=\textwidth}
\begin{tabular}{lcccccc}
\toprule
\multirow{2}{*}{Model} & \multicolumn{3}{c}{First Person} & \multicolumn{3}{c}{Third Person} \\
\cmidrule(lr){2-4} \cmidrule(lr){5-7}
& ChessBoard $\uparrow$ & MirrorRollingBall $\uparrow$ & RacingCar $\uparrow$ & ChessBoard $\uparrow$ & MirrorRollingBall $\uparrow$ & RacingCar $\uparrow$ \\
\midrule
Project Genie & $0.912\!\pm\!0.074$ & $\mathbf{0.980\!\pm\!0.030}$ & $0.720\!\pm\!0.073$ & $0.936\!\pm\!0.057$ & $\mathbf{0.936\!\pm\!0.054}$ & $\mathbf{0.856\!\pm\!0.054}$ \\
LingBot-World (Cam) & $\mathbf{0.936\!\pm\!0.073}$ & $0.920\!\pm\!0.022$ & $\mathbf{0.760\!\pm\!0.060}$ & $\mathbf{0.992\!\pm\!0.018}$ & $0.899\!\pm\!0.036$ & $0.832\!\pm\!0.044$ \\
Matrix-Game 3.0 & $0.840\!\pm\!0.049$ & $0.663\!\pm\!0.057$ & $0.627\!\pm\!0.076$ & $0.893\!\pm\!0.053$ & $0.752\!\pm\!0.111$ & $0.616\!\pm\!0.064$ \\
HY-WorldPlay & $0.789\!\pm\!0.082$ & $0.740\!\pm\!0.070$ & $0.573\!\pm\!0.101$ & $0.837\!\pm\!0.059$ & $0.904\!\pm\!0.067$ & $0.824\!\pm\!0.083$ \\
Waypoint-1.5 & $0.933\!\pm\!0.037$ & $0.750\!\pm\!0.031$ & $0.667\!\pm\!0.082$ & $0.891\!\pm\!0.047$ & $0.763\!\pm\!0.061$ & $0.808\!\pm\!0.018$ \\
\bottomrule
\end{tabular}
\end{adjustbox}
\vspace{-2mm}
\end{table*}


\paragraph{Foreground objects are easier to preserve than the surroundings.} Table~\ref{tab:vlm_eval} shows the same local--global gap in prompt consistency: with the car in view, third-person \texttt{RacingCar} reaches $0.81$--$0.86$ for four of five models, while first person, scored on the environment alone, falls to $0.57$--$0.76$. In its breakdown (Table~\ref{tab:vlm_eval_breakdown}), on third-person \texttt{RacingCar}, Project Genie, LingBot-World, and HY-WorldPlay score $1.00$ on foreground objects but only $0.76$, $0.72$, and $0.71$ on the environment. Thus, salient objects are preserved more reliably than the broader scene. Human evaluation shows a strong alignment with the VLM results with a Spearman coefficient $\rho=0.90$ (Appendix~\ref{app:prompt_scene_validity}), Project Genie and LingBot-World performances are statistically in-distinguishable ($p=0.99$). 
Across geometry, lighting, and prompts, local evidence outlasts global state.

\section{Conclusion}

We introduced \capsysname, a comprehensive world model benchmark grounded in a physical simulator to evaluate interactive world models under matched action trajectories. Across nine Unity scenes and six state-of-the-art interactive world models, our results show that current models can often preserve coarse visual fidelity, foreground entities like the main character in third-person view, and action-responsive motion. However, current world models still fall short in many aspects: models struggle to preserve spatial relations under camera motion, maintain object and character identity over long horizons, reproduce mechanically valid trajectories and physical quantities, conserve fluid properties, render coherent lighting and shadows, and propagate multi-stage causal chains. These failures are often hidden by perceptual metrics, demonstrating that \capsysnamespace can effectively serve its intended role: exposing physical-causal gaps through standardized and simulator-grounded evaluation.



\newpage
\section*{AI Use Statement}

In this work, we used generative AI tools to provide feedback on technical formulations and statistical data analysis, we also used AI to help with simulation environment implementations. We have not used generative AI tools to generate synthetic datasets, develop theoretical models, assist in writing proofs, propose hypotheses, translate content, clean or reformat datasets, or interpret experimental results. Additionally, we used generative AI tools to edit and refine manuscript text for clarity and readability, suggest improvements to presentation and structure, and assist with literature search and verification. We reviewed all AI-assisted work and manually verified technical claims, citations, and suggested revisions against the original sources and our own experiments. We take responsibility for the final content of this work, including text, claims, and artifacts produced with the aid of generative AI.

\section*{Reproducibility Statement}
We provide detailed descriptions of the benchmark construction, evaluation metrics, and implementation procedures to facilitate reproducibility. Appendix~\ref{app:scenes} specifies the nine Unity environments, action spaces, simulator states, and scene configurations; Appendix~\ref{app:metrics} defines the dimension-specific and holistic evaluation metrics; Appendix~\ref{app:measurement_reliability} reports evaluator calibration and sample size robustness analyses; and Appendix~\ref{app:implementation} describes the evaluation pipelines and implementation details for the evaluation pipeline. We also provide the Unity-based reference package, including evaluation code, reference simulator states, and action scripts, in the supplementary material.

\section*{Ethics Statement}

All authors have read and adhere to the \href{https://iclr.cc/public/CodeOfEthics}{ICLR Code of Ethics}. This work does not involve human subjects, sensitive personal data, or experiments with the potential to cause harm. No confidential or proprietary data was used. The methods and experiments were conducted in accordance with principles of research integrity, fairness, and transparency. Potential societal impacts, including limitations and biases of large language models, are explicitly discussed in the paper. All conclusions are the sole responsibility of the authors.

\bibliography{wmbench}
\bibliographystyle{iclr2027_conference}

\newpage
\appendix
\section{Unity Scenes}
\label{app:scenes}

\begin{figure*}[ht]
    \centering
    
    \begin{subfigure}[b]{0.32\textwidth}
        \centering
        \includegraphics[width=\textwidth]{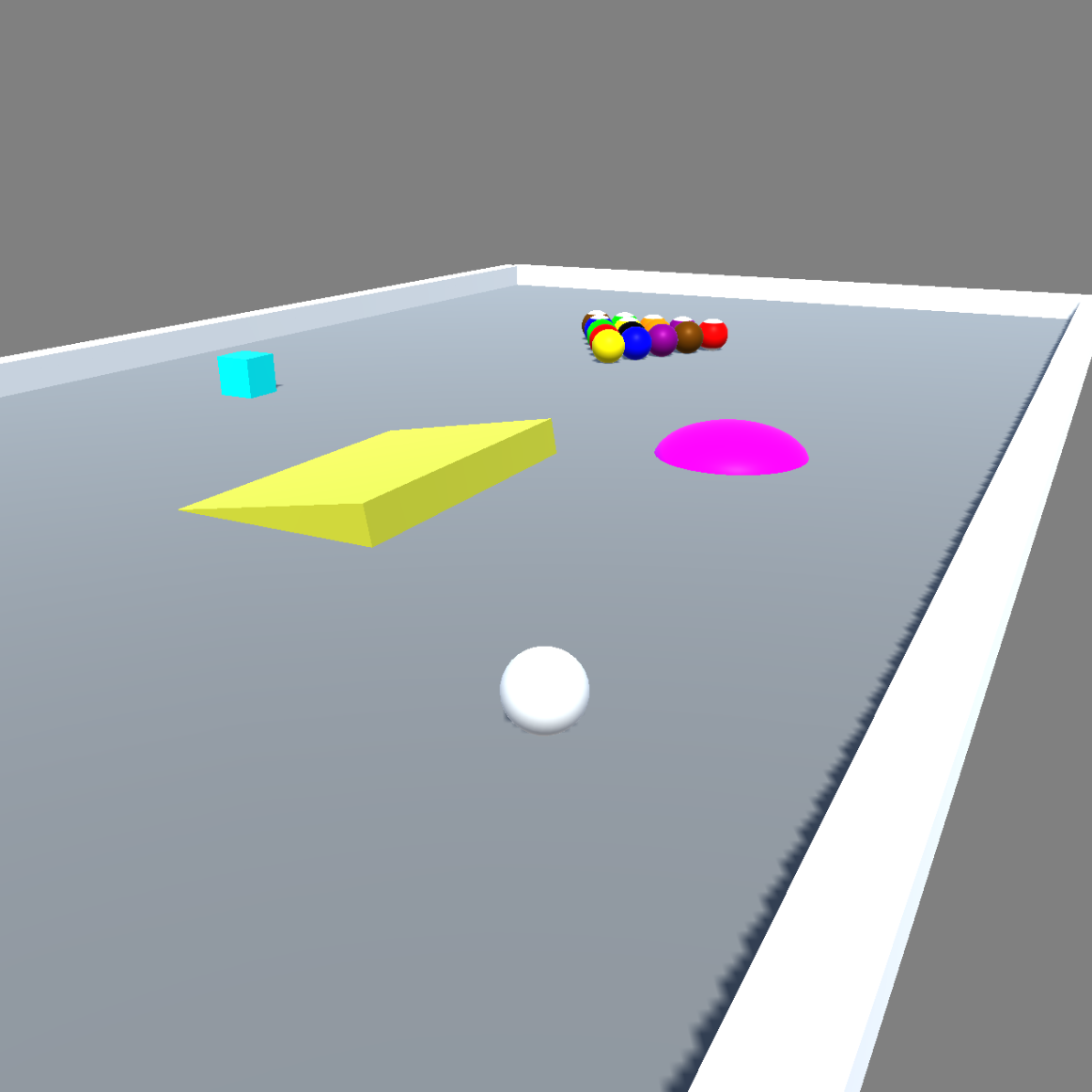}
        \caption{BallRoller}
        \label{fig:scenes:ballroller}
    \end{subfigure}
    \hfill
    \begin{subfigure}[b]{0.32\textwidth}
        \centering
        \includegraphics[width=\textwidth]{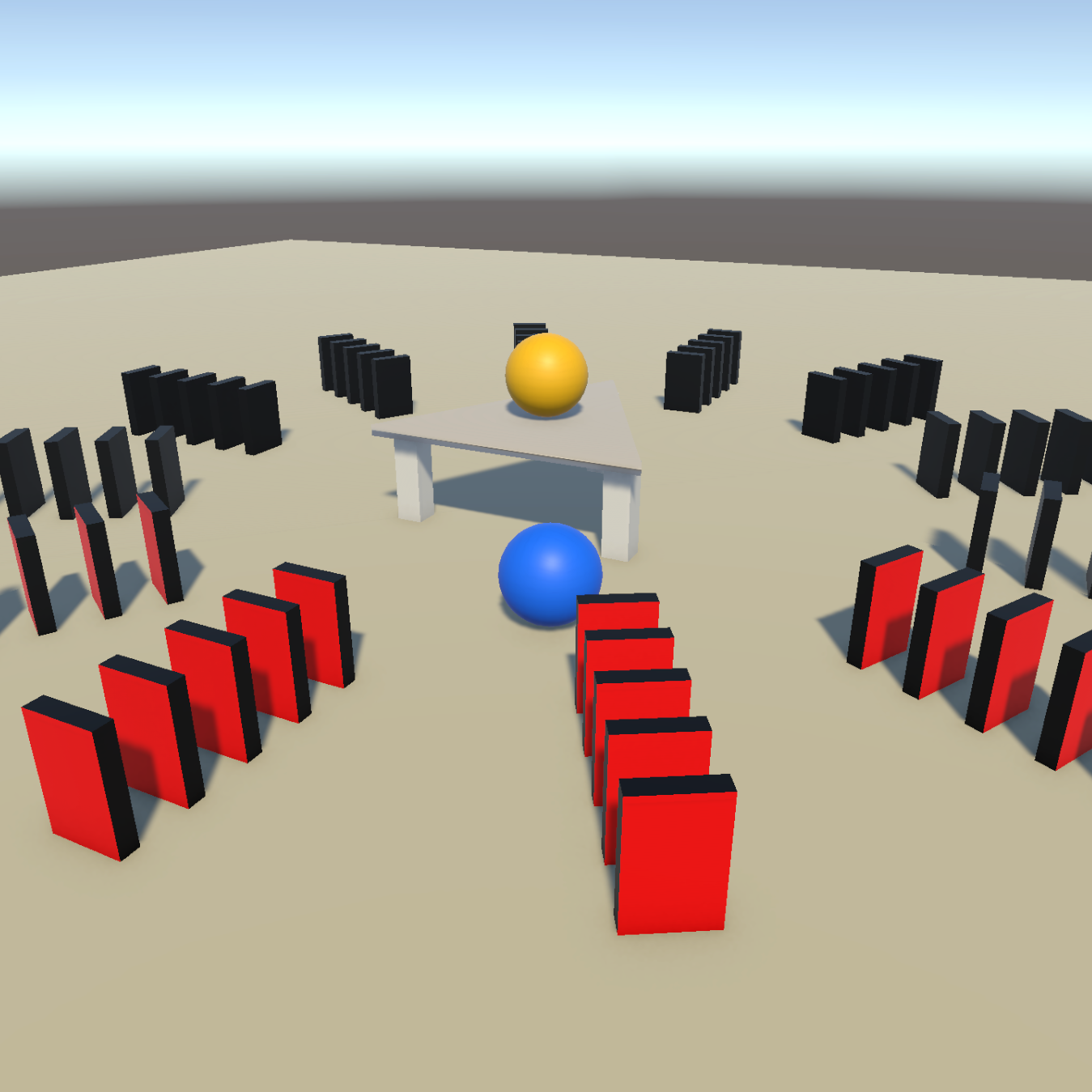}
        \caption{BeamDomino}
        \label{fig:scenes:beamdomino}
    \end{subfigure}
    \hfill
    \begin{subfigure}[b]{0.32\textwidth}
        \centering
        \includegraphics[width=\textwidth]{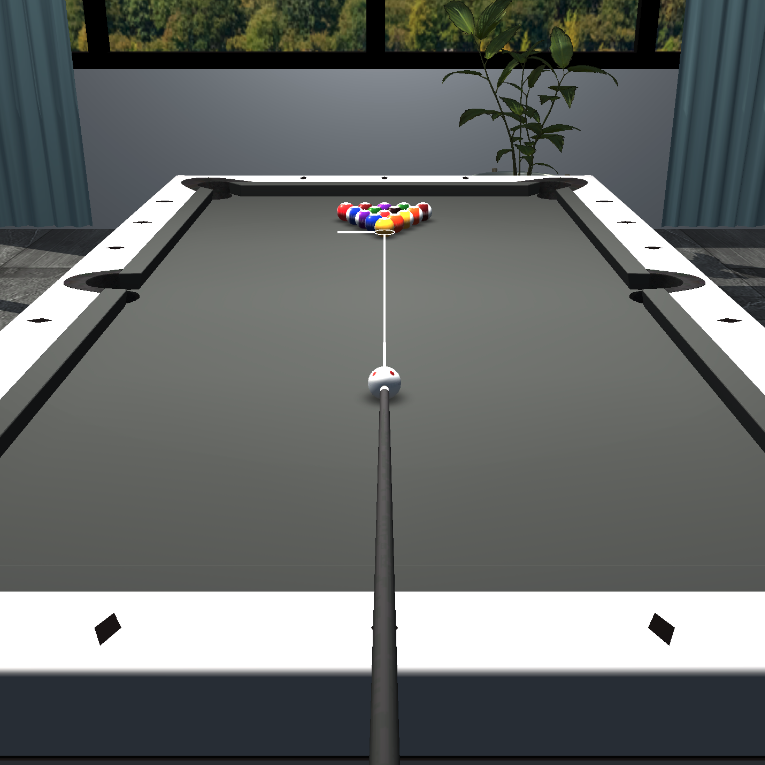}
        \caption{BilliardGame}
        \label{fig:scenes:billiard}
    \end{subfigure}

    \vspace{8pt} 

    \begin{subfigure}[b]{0.32\textwidth}
        \centering
        \includegraphics[width=\textwidth]{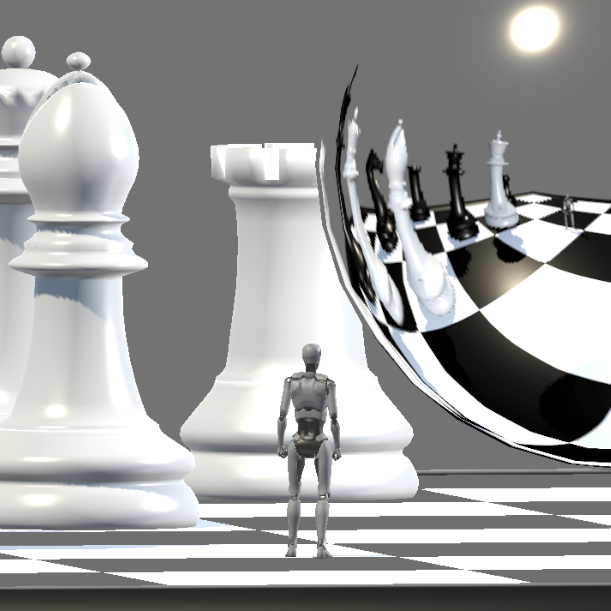}
        \caption{ChessBoard}
        \label{fig:scenes:chess}
    \end{subfigure}
    \hfill
    \begin{subfigure}[b]{0.32\textwidth}
        \centering
        \includegraphics[width=\textwidth]{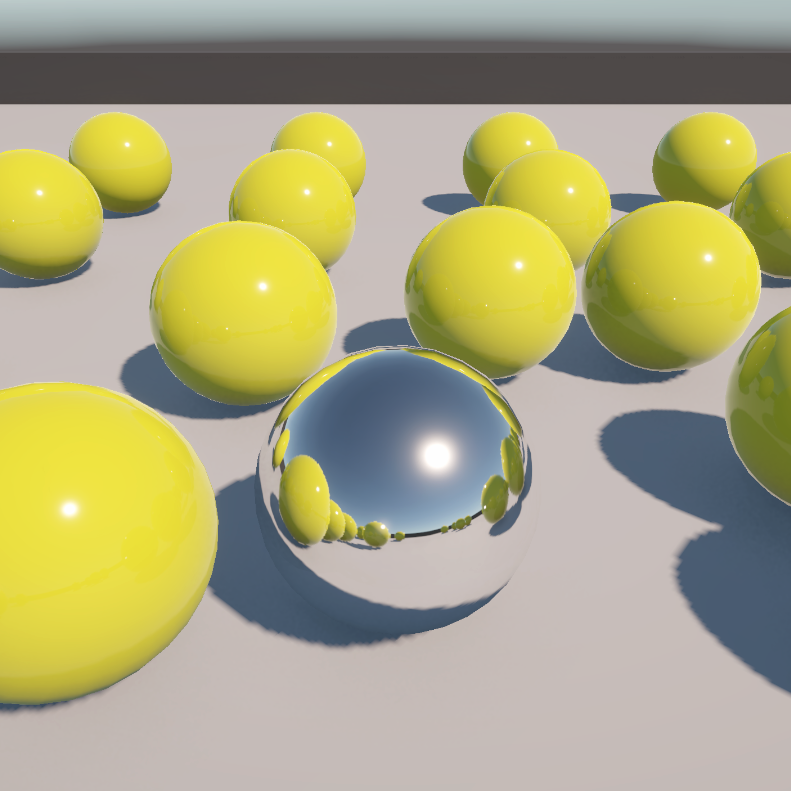}
        \caption{MirrorRollingBall}
        \label{fig:scenes:mirror}
    \end{subfigure}
    \hfill
    \begin{subfigure}[b]{0.32\textwidth}
        \centering
        \includegraphics[width=\textwidth]{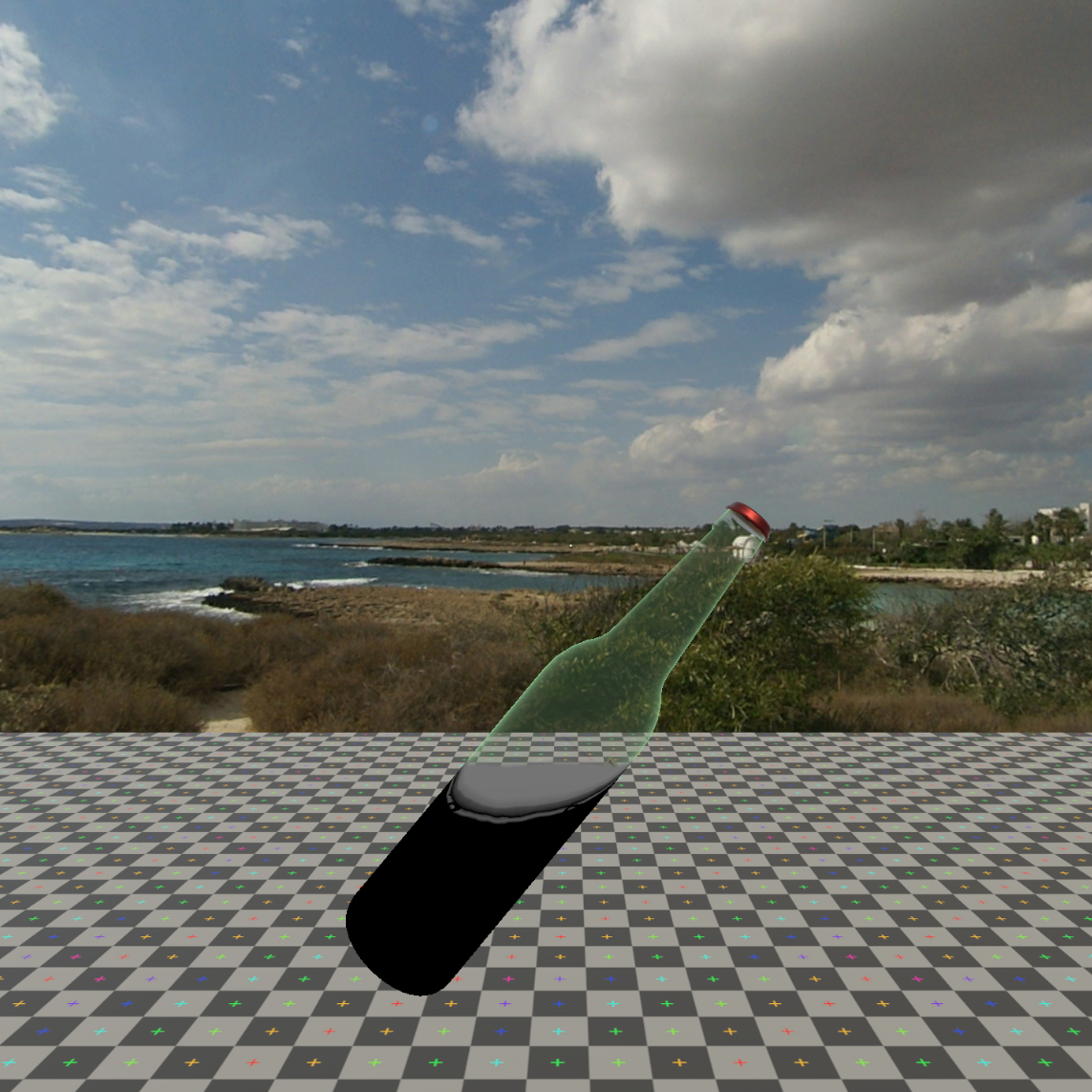}
        \caption{LiquidContainer}
        \label{fig:scenes:liquid}
    \end{subfigure}

    \vspace{8pt}

    \begin{subfigure}[b]{0.32\textwidth}
        \centering
        \includegraphics[width=\textwidth]{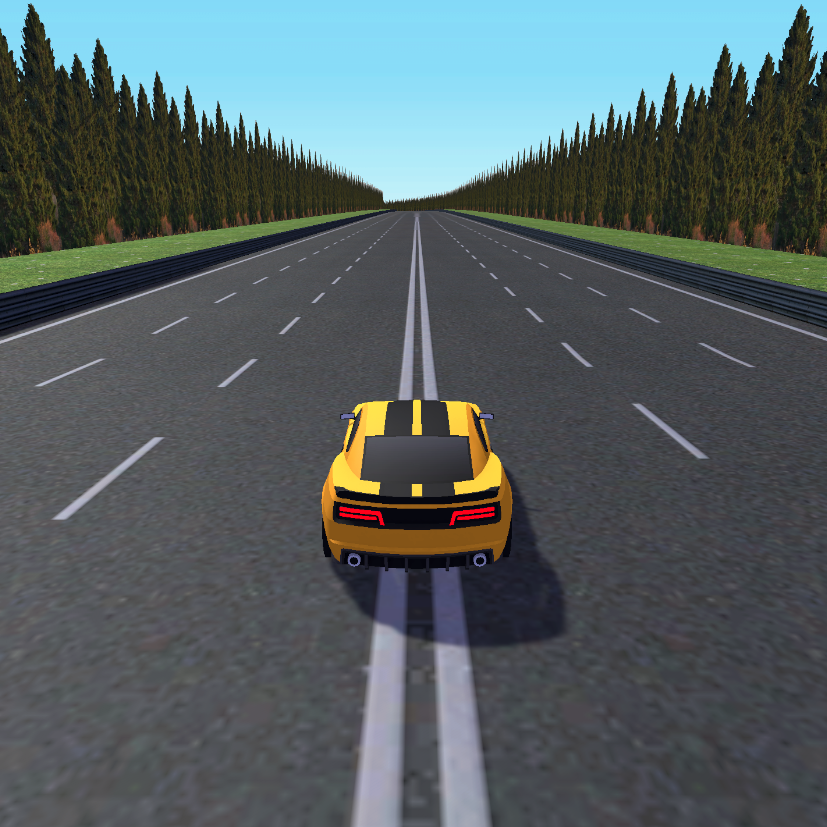}
        \caption{RacingCar}
        \label{fig:scenes:racing}
    \end{subfigure}
    \hfill
    \begin{subfigure}[b]{0.32\textwidth}
        \centering
        \includegraphics[width=\textwidth]{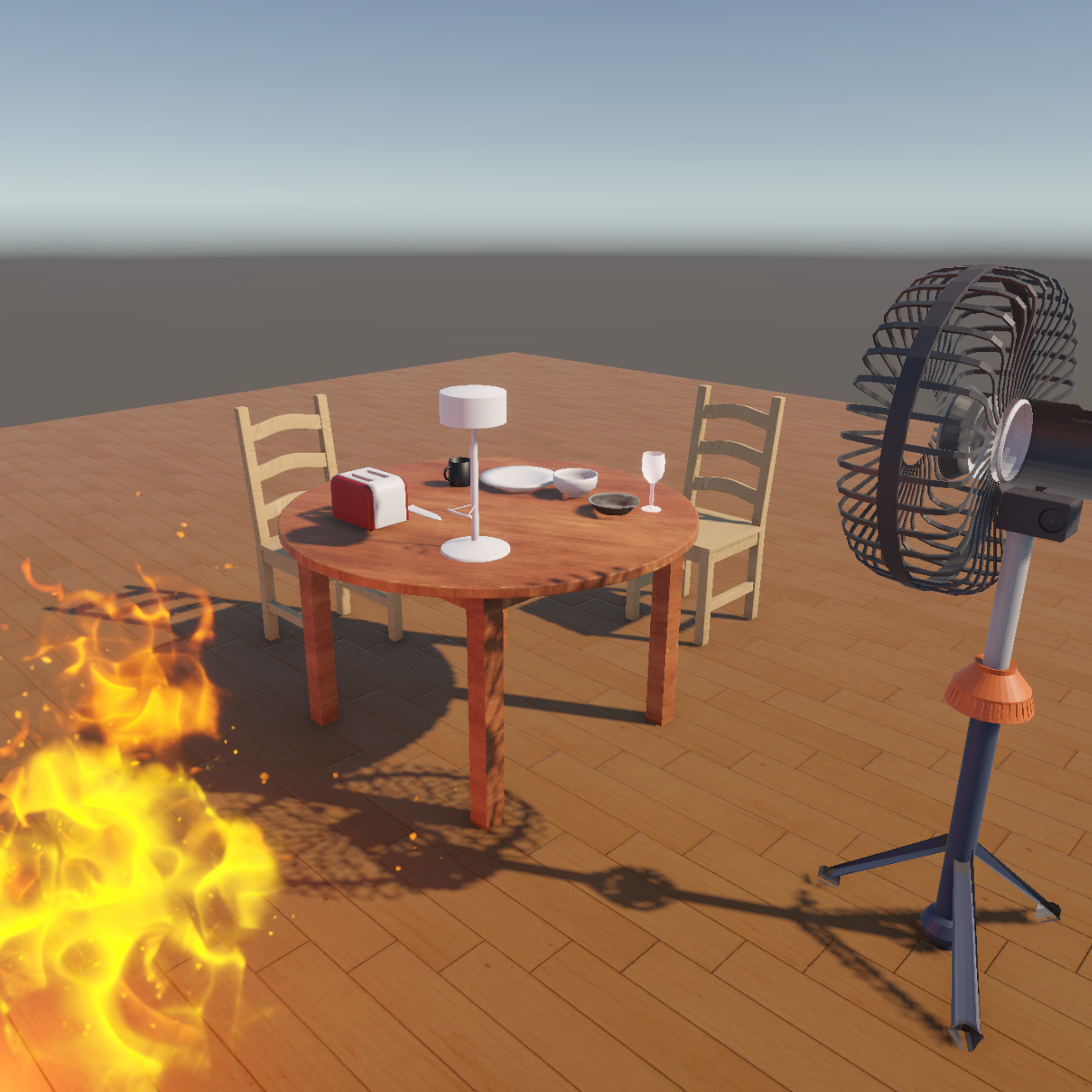}
        \caption{SunRisesAndSets}
        \label{fig:scenes:sun}
    \end{subfigure}
    \hfill
    \begin{subfigure}[b]{0.32\textwidth}
        \centering
        \includegraphics[width=\textwidth]{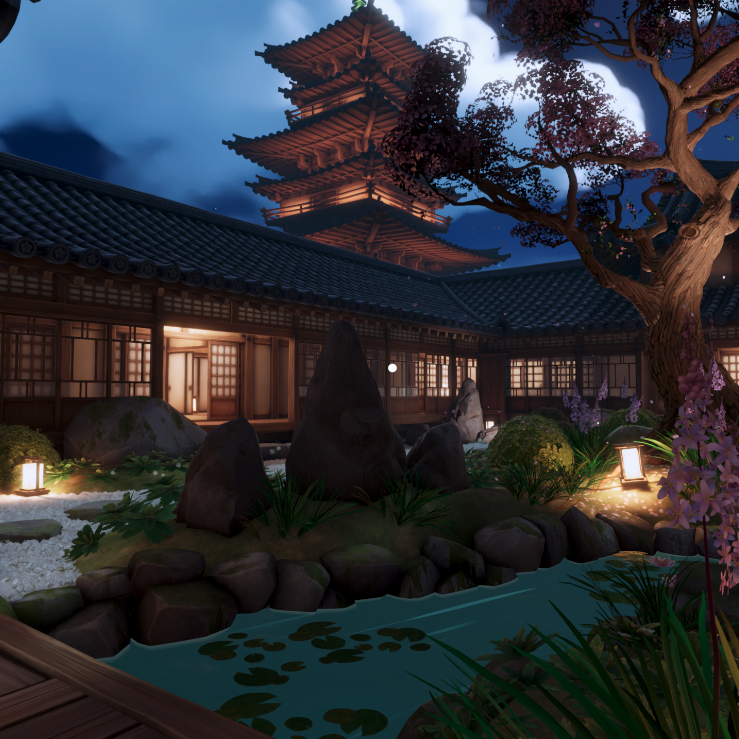}
        \caption{WanderingGarden}
        \label{fig:scenes:garden}
    \end{subfigure}

    \caption{\textbf{Visual Gallery of the \capsysnamespace Benchmark Environments.} The nine Unity scenes are architected to evaluate world models across six core dimensions defined in Section~\ref{sec:benchmark}. \textbf{(a--c) Causal \& Kinematic Interactions:} BallRoller, BeamDomino, and BilliardGame assess action controllability, multi-body collision dynamics, and multi-stage causal chains. \textbf{(d, e, i) Spatial \& Identity Consistency:} ChessBoard, MirrorRollingBall, and WanderingGarden test scene geometry preservation, object permanence, and character identity. \textbf{(f--h) Specialized Continuum \& Environmental Physics:} LiquidContainer (fluid mechanics), RacingCar (non-linear vehicle kinematics), and SunRisesAndSets (temporal lighting/shadow kinematics) isolate complex physical phenomena.}
    \label{fig:appendix:scene_gallery}
\end{figure*}

\capsysnamespace comprises nine Unity scenes, each designed to isolate specific physical-causal dimensions (illustrated in Fig.~\ref{fig:appendix:scene_gallery}). All scenes expose their state through the Unity ML-Agents Toolkit, enabling evaluation at both the perceptual and physical-causal level. Technical specifications are summarized in Table~\ref{tab:scene_summary}.

\begin{table}[ht]
\centering
\caption{Summary of the nine \capsysnamespace Unity scenes. \textbf{Act.} = action-space dimensionality (C = continuous, D = discrete branches); \textbf{Obs.} = observation-vector length; \textbf{Camera} = camera regime used during evaluation.}
\label{tab:scene_summary}
\small
\setlength{\tabcolsep}{3pt}
\begin{tabular}{@{}llcclp{4.8cm}@{}}
\toprule
Scene & Act. & Obs. & Camera & Primary Evaluation Dimensions \\
\midrule
BallRoller        & 2C           & 48   & Fixed-offset follow   & Controllability, cLPIPS, Consistency \\
BeamDomino         & 2C           & 92   & Fixed-offset follow   & Physical Causality, Controllability, cLPIPS \\
BilliardGame       & 10C + 2D    & 13   & Fixed overhead        & Kinematics \& Mechanics, cLPIPS \\
ChessBoard         & 2C + 2D     & 8    & Third-person follow   & Scene \& Character Consistency \\
MirrorRollingBall  & 2C           & 297  & Fixed-offset follow   & Consistency, Kinematics \\
LiquidContainer    & 6C           & 24   & Static side           & Fluid Mechanics \\
RacingCar          & 2C + 1D     & 13   & Speed-adaptive follow & Kinematics \& Mechanics, Consistency \\
SunRisesAndSets    & ---         & 9    & Fixed orbit           & Lighting \& Shading \\
WanderingGarden    & 4C           & 50   & First-person          & Scene \& Character Consistency \\
\bottomrule
\end{tabular}
\end{table}

\subsection{BallRoller}
\label{app:scenes_ballroller}

\begin{figure}[ht]
    \centering
    \begin{subfigure}[b]{0.24\textwidth}
        \centering
        \includegraphics[width=\textwidth]{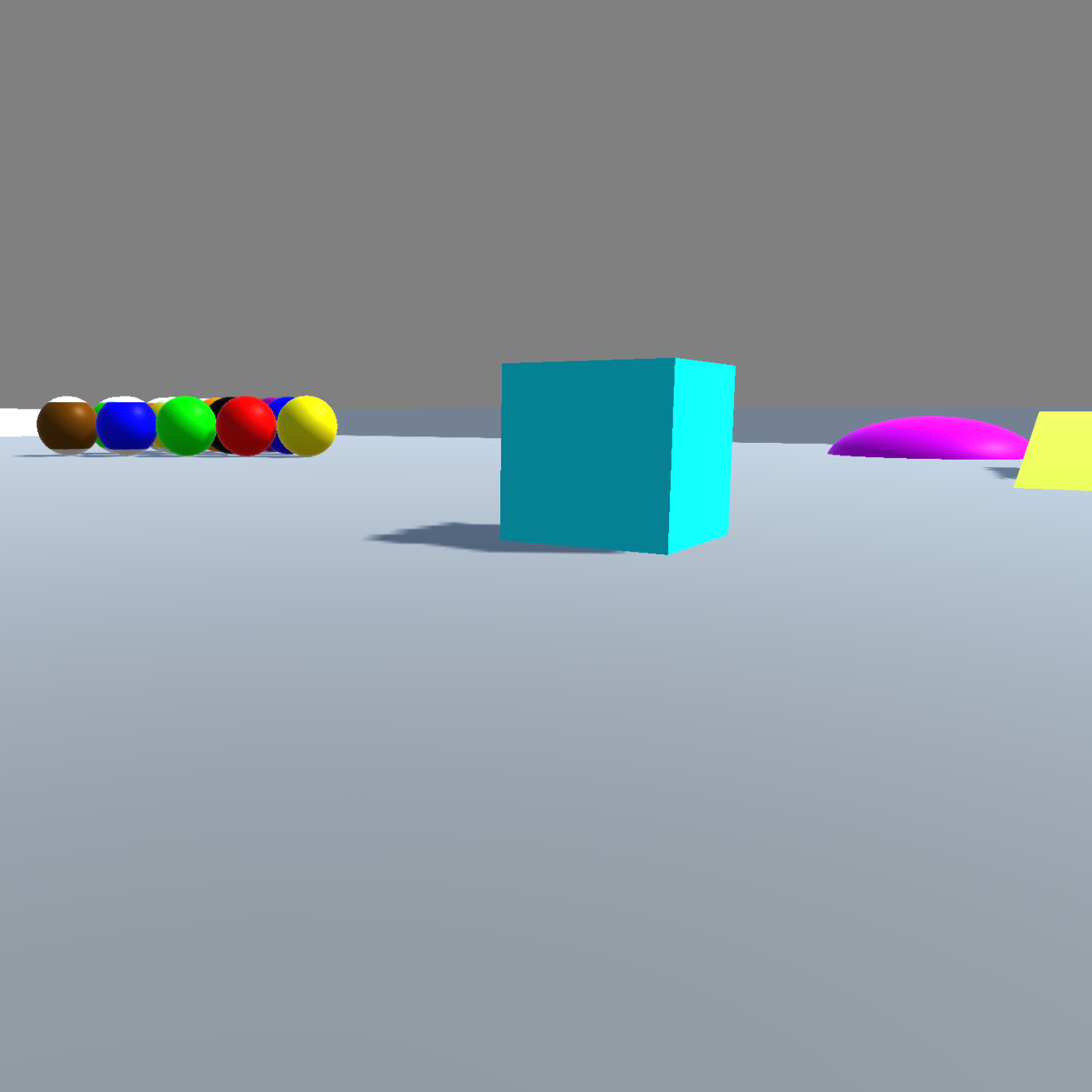}
        \caption{First-person}
        \label{fig:scenes:ballroller:first}
    \end{subfigure}
    \hfill
    \begin{subfigure}[b]{0.24\textwidth}
        \centering
        \includegraphics[width=\textwidth]{figures/BallRoller2.png}
        \caption{Third-person}
        \label{fig:scenes:ballroller:third}
    \end{subfigure}
    \hfill
    \begin{subfigure}[b]{0.49\textwidth}
        \centering
        \includegraphics[width=0.48\textwidth]{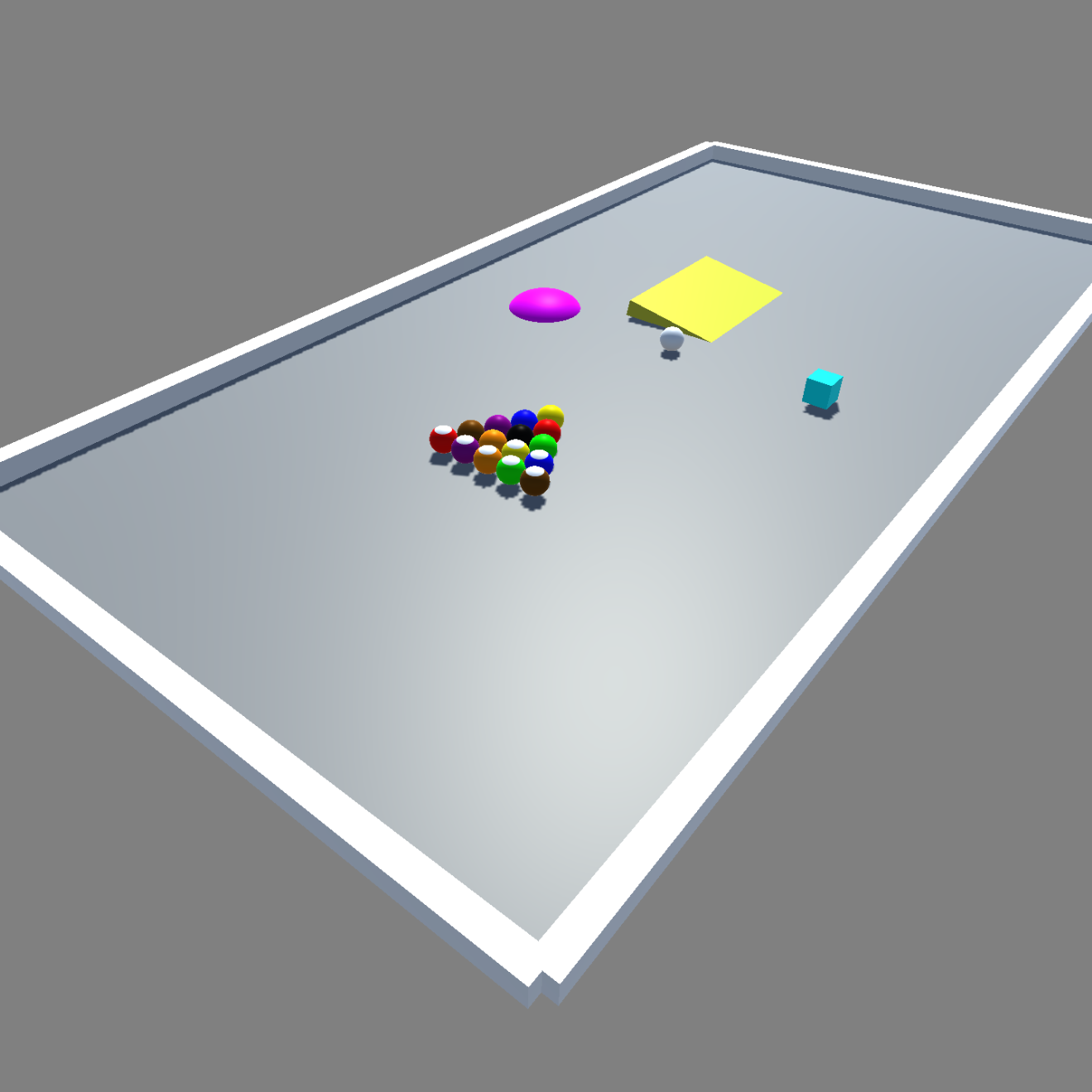}
        \hfill
        \includegraphics[width=0.48\textwidth]{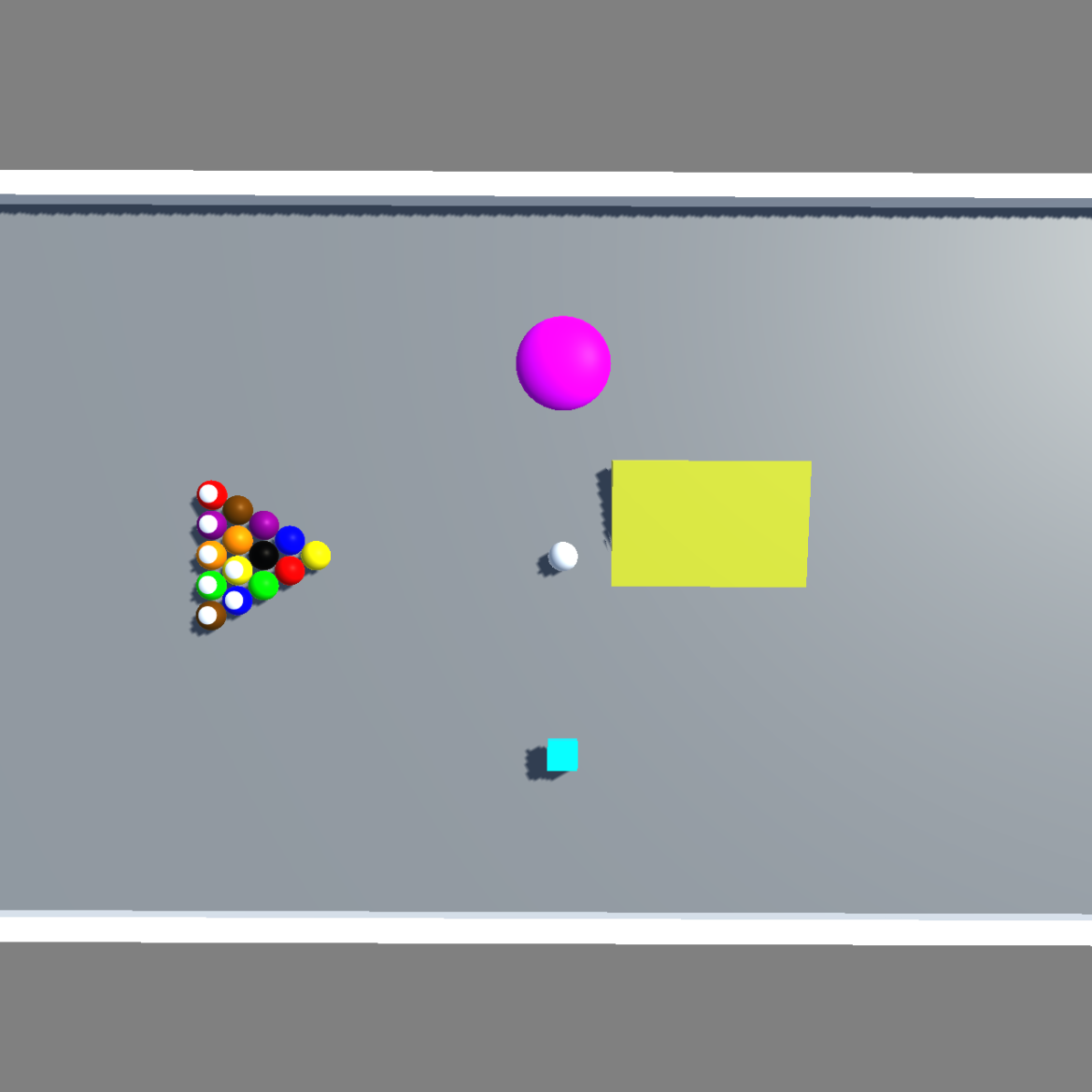}
        \caption{Environmental Layouts}
        \label{fig:scenes:ballroller:layouts}
    \end{subfigure}

    \caption{\textbf{BallRoller Scene Views.}}
    \label{fig:scenes:ballroller_views}
\end{figure}

\paragraph{Scene description.}
BallRoller is a compact square arena with a flat blue-gray floor and low white boundary walls, viewed from a fixed side-table camera that looks diagonally into the play area. The player controls a smooth white sphere that rolls across the floor. Static obstacles---a yellow ramp, a magenta dome, a cyan cube---are placed throughout the arena, and a cluster of 15 colored physics balls is scattered on the floor. On contact, these passive balls are knocked away by the player sphere, producing multi-body rigid-body collisions.

\paragraph{Action space.}
Two continuous actions: horizontal force along the $X$ axis and horizontal force along the $Z$ axis, each in $[-1,+1]$, applied directly to the player ball's rigidbody via \texttt{AddForce}.

\paragraph{Observation space.}
A 48-dimensional vector: the player's local position (3 floats) concatenated with the local positions of up to 15 passive physics balls ($15 \times 3 = 45$ floats). Notably, velocities are \emph{not} included, so a world model must infer dynamics from positional changes across time steps.

\paragraph{Camera.}
A fixed-offset follow camera that computes a world-space offset from the player at initialization and rigidly translates with the player ball in every subsequent frame.

\paragraph{Evaluation dimensions.}
BallRoller is used for \textbf{action controllability} (both Ground Truth RAFT and Synthetic RAFT), \textbf{causal LPIPS} (collision events between the player and passive balls), and \textbf{prompt--scene consistency} (VLM-as-judge). The omission of velocity from the observation vector is a deliberate design choice: it tests whether a world model can reason about latent velocities from position histories alone, a prerequisite for accurate multi-body collision prediction.

\subsection{BeamDomino}
\label{app:beamdomino}

\begin{figure}[ht]
    \centering
    \begin{subfigure}[b]{0.24\textwidth}
        \centering
        \includegraphics[width=\textwidth]{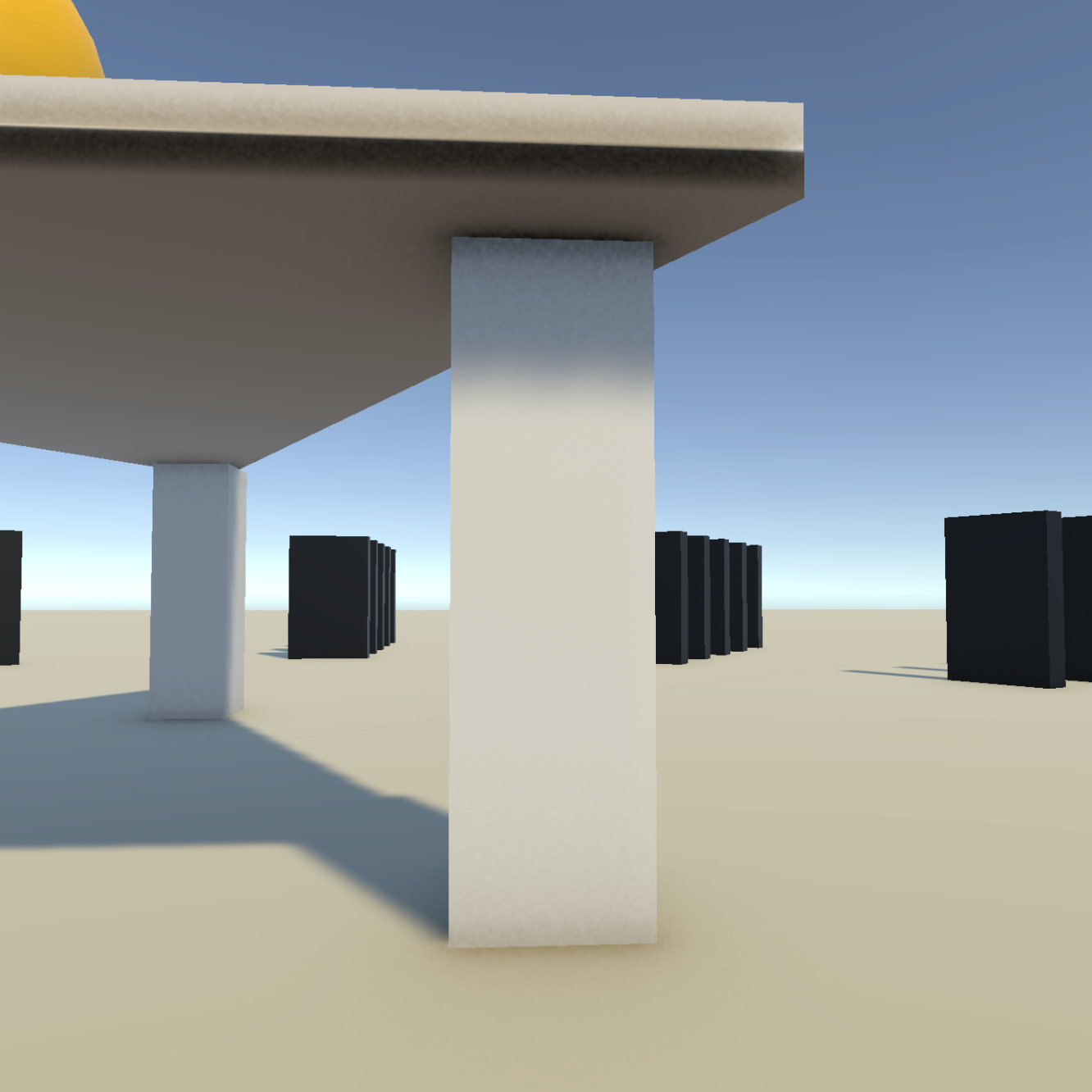}
        \caption{First-person}
        \label{fig:scenes:beamdomino:first}
    \end{subfigure}
    \hfill
    \begin{subfigure}[b]{0.24\textwidth}
        \centering
        \includegraphics[width=\textwidth]{figures/BeamDomino2.png}
        \caption{Third-person}
        \label{fig:scenes:beamdomino:third}
    \end{subfigure}
    \hfill
    \begin{subfigure}[b]{0.49\textwidth}
        \centering
        \includegraphics[width=0.48\textwidth]{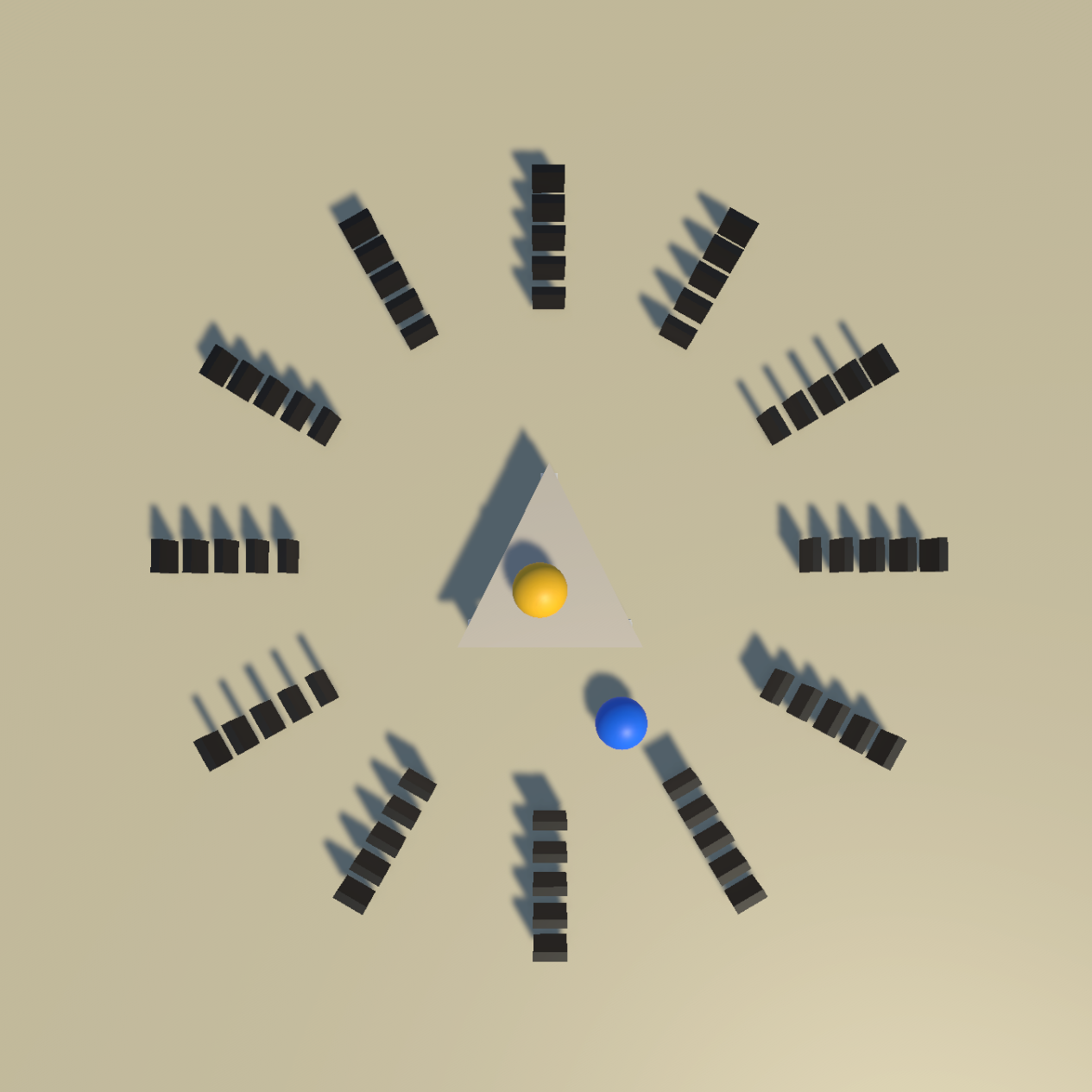}
        \hfill
        \includegraphics[width=0.48\textwidth]{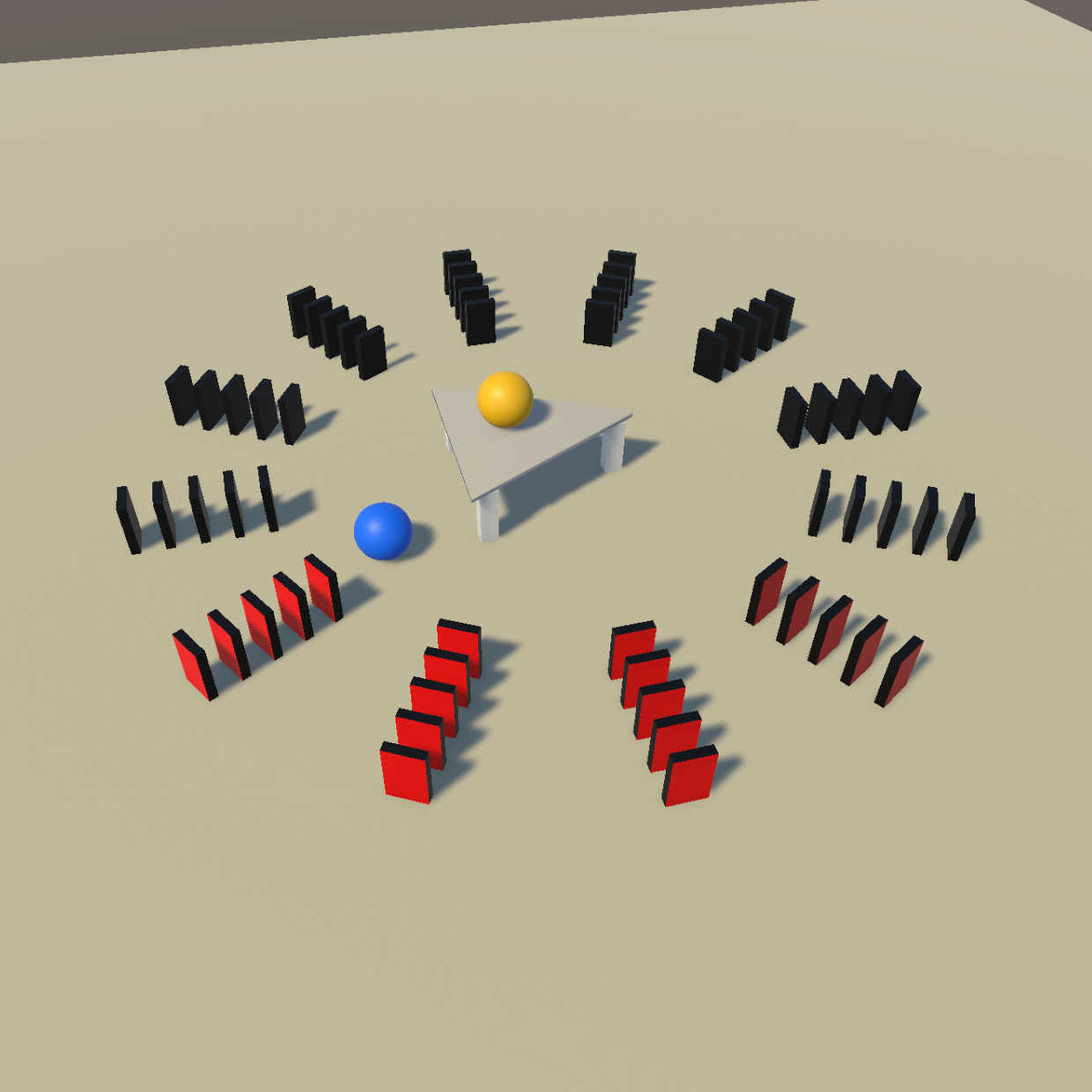}
        \caption{Environmental Layouts}
        \label{fig:scenes:beamdomino:layouts}
    \end{subfigure}

    \caption{\textbf{BeamDomino Scene Views.}}
    \label{fig:scenes:beamdomino_views}
\end{figure}

\paragraph{Scene description.}

BeamDomino is a physics-driven causal-chain scene in which the player drives a heavy floor ball under continuous control to strike a supporting structure and trigger a downstream domino cascade. The scene decomposes into four canonical stages: \textsc{rolling}, \textsc{impact}, \textsc{topple}, and \textsc{cascade}, linking ball trajectory to multi-body contact, structural collapse, roof-ball release, and delayed domino propagation.

Its two-tone domino design and staged geometry provide clear visual cues for fall direction and causal progression. The scene, therefore, tests whether a world model reproduces the correct causal state transitions rather than merely generating visually plausible motion.

Small errors in impact location, momentum transfer, contact ordering, or pillar/sector selection can produce qualitatively different cascades. This sensitivity is what makes BeamDomino a targeted test of action-driven physical causality~\citep{li2026causalstep}.


\paragraph{Trajectory shapes.}
Under the same nominal action prompt, small perturbations of the initial ball state produce three qualitatively different ground-truth cascades:
\begin{itemize}
    \item \textsc{canonical}: the ball strikes a leg, the platform tilts past the fall threshold, the roof ball is released, and the dominoes topple.
    \item \textsc{glancing}: the ball clips a leg, but the platform survives; the dominoes are subsequently toppled by the ball directly.
    \item \textsc{direct-strike}: the ball bypasses the platform and hits the domino row first.
\end{itemize}
Benchmark trajectories include all three. The classifier and per-shape QA stage chains are detailed in Appendix~\ref{app:physcaus}.

\paragraph{Action space.}
Two continuous actions: force along $Z$ (forward/backward) and force along $X$ (left/right), each in $[-1,+1]$. The applied force magnitude is 80\,N with a maximum ball speed of 20\,m/s. This action channel is agent-specific it acts on the floor ball's rigid body rather than camera-relative.

World models without a faithful action-channel mapping for $(F_x, F_z)$ on a non-camera body are not evaluated on this scene: a prompt-driven substitute for the ball action, e.g., LingBot-World (Cam)'s natural-language prompt, is not action-matched.

\paragraph{Observation space.}
A 92-dimensional vector comprising: the floor ball's position, linear velocity, and angular velocity (9); the roof ball's position, linear velocity, and an airborne flag (7); per-domino tilt magnitude and fallen flag for each of the 20 dominoes ($20 \times 2 = 40$); cascade summary statistics including the fraction of dominoes fallen and the elapsed cascade duration (2); camera position and Euler angles (6); three pillar states each with position, Euler angles, and a kinematic flag ($3 \times 7 = 21$); and the roof slab state with position, Euler angles, and a kinematic flag (7).

\paragraph{Camera.}
A fixed-offset follow camera positioned at $(0, 6, -12)$ relative to the floor ball, with smooth interpolation (Lerp factor $8/s$) and continuous \texttt{LookAt} targeting the ball.

\paragraph{Evaluation dimensions.}
BeamDomino is the primary scene for \textbf{physical causality} evaluation using the CausalStep framework~\cite{li2026causalstep}, which grades whether the four canonical stages occur in correct causal order. It is also used for \textbf{action controllability} (RAFT) and \textbf{causal LPIPS}.

The three-step causal chain is central to the design: prediction errors at any stage compound downstream, so a world model must faithfully propagate momentum through impact, structural collapse, and cascade to score well.

\subsection{BilliardGame}
\label{app:billiargame}

\begin{figure}[ht]
    \centering
    \begin{subfigure}[b]{0.24\textwidth}
        \centering
        \includegraphics[width=\textwidth]{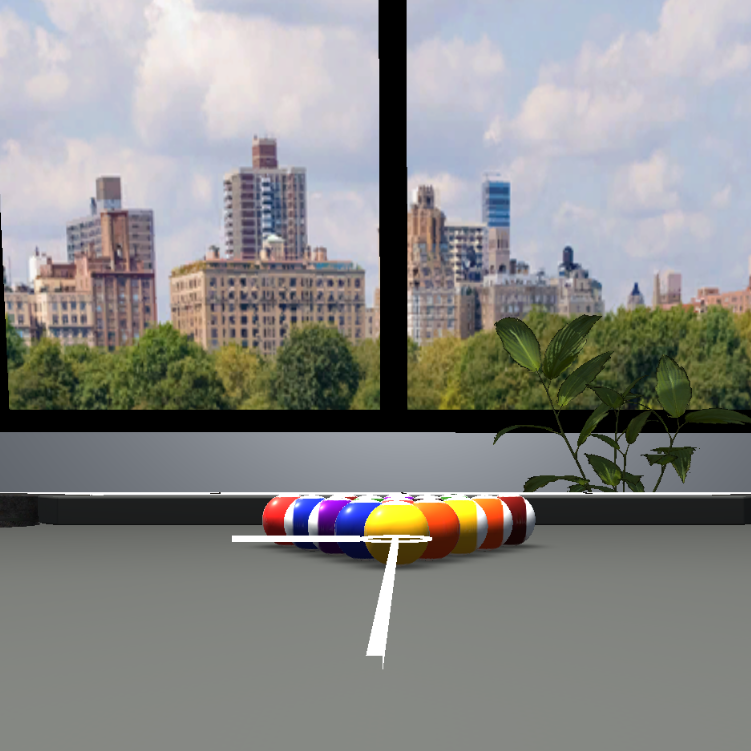}
        \caption{First-person}
        \label{fig:scenes:billiard:first}
    \end{subfigure}
    \hfill
    \begin{subfigure}[b]{0.24\textwidth}
        \centering
        \includegraphics[width=\textwidth]{figures/BilliardGame2.png}
        \caption{Third-person}
        \label{fig:scenes:billiard:third}
    \end{subfigure}
    \hfill
    \begin{subfigure}[b]{0.49\textwidth}
        \centering
        \includegraphics[width=0.48\textwidth]{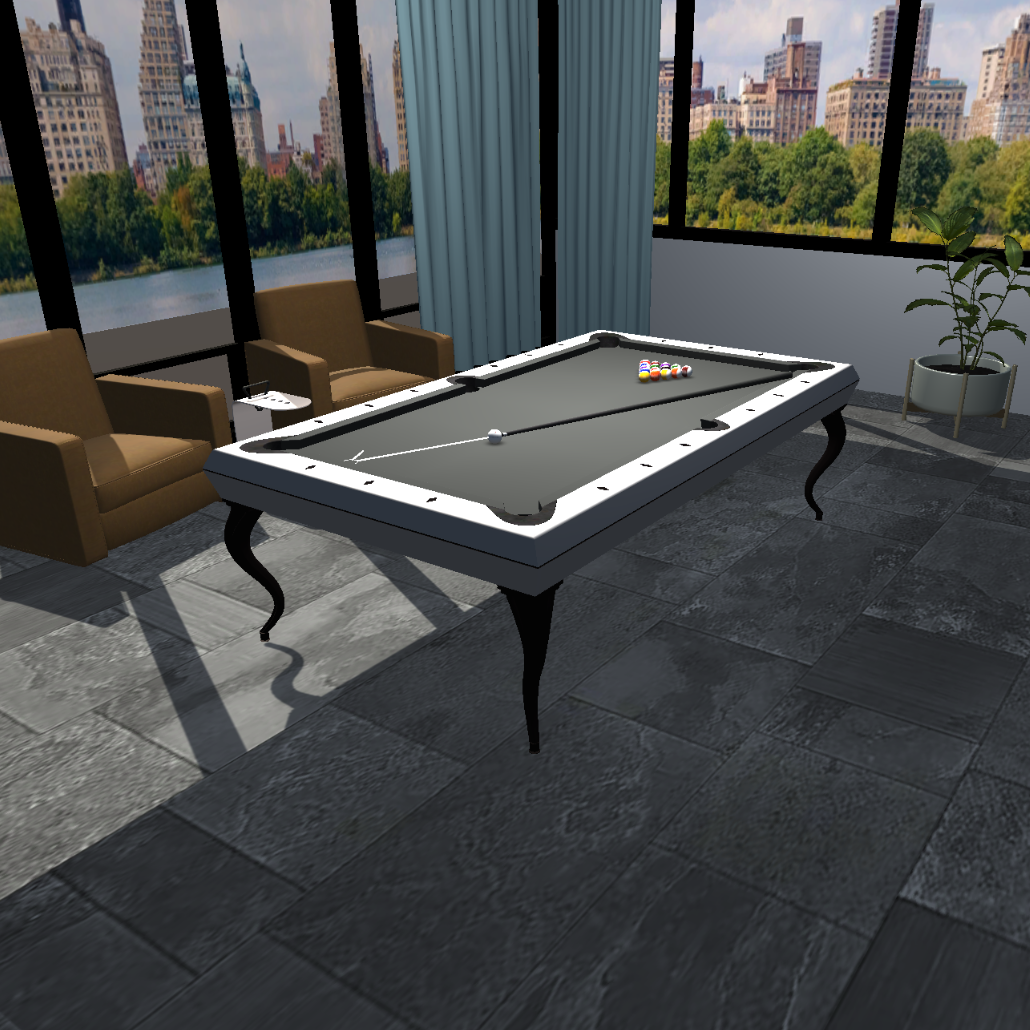}
        \hfill
        \includegraphics[width=0.48\textwidth]{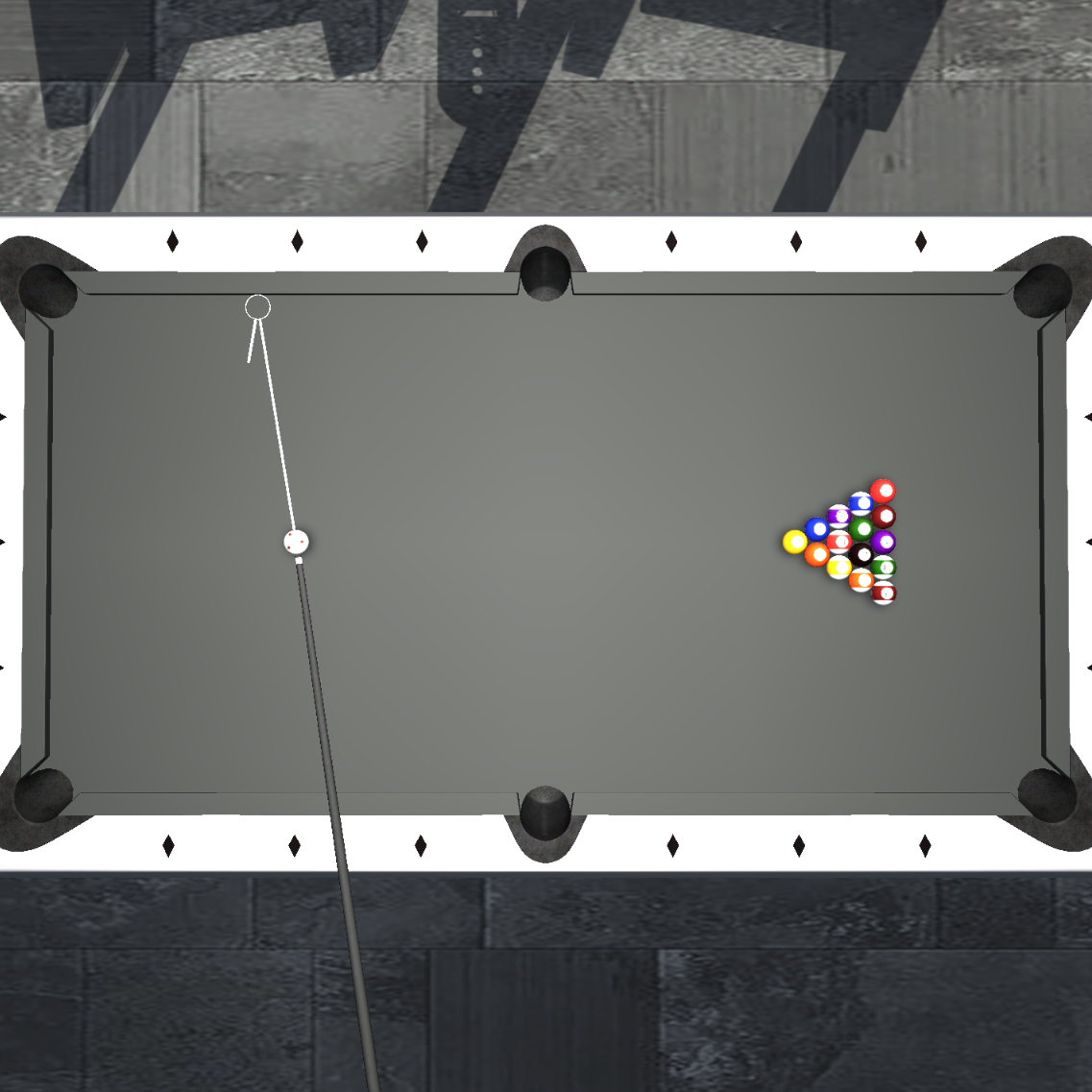}
        \caption{Environmental Layouts}
        \label{fig:scenes:billiard:layouts}
    \end{subfigure}

    \caption{\textbf{BilliardGame Scene Views.}}
    \label{fig:scenes:billiard_views}
\end{figure}

\paragraph{Scene description.}
BilliardGame implements a three-ball carom billiards setup on a standard billiard table. Unlike the other scenes, this scene uses a custom deterministic fixed-point 3D physics engine (Calculable Mechanics v2) rather than Unity's default PhysX solver, ensuring analytically reproducible ball--ball and ball--cushion collisions. The player positions up to three balls on the table and executes a cue shot with configurable angle, force, and English (off-center hit point on the cue ball face).

\paragraph{Action space.}
A mixed action space with 10 continuous and 2 discrete dimensions. The continuous actions encode: the initial positions of three balls on the table surface ($3 \times 2 = 6$ floats mapped from $[-1,+1]$ to table half-dimensions), the cue angle ($[0, 2\pi]$), shot strength ($[0, \text{maxForce}]$), and the 2D hit point on the cue ball face (clamped to the unit circle, encoding topspin/backspin/sidespin). The two discrete branches control shot execution (wait vs.\ execute) and simulation termination (normal vs.\ force-end).

\paragraph{Observation space.}
A 13-dimensional vector: the local positions of three balls relative to the table frame ($3 \times 3 = 9$), the normalized cue pivot angle (1), the current force setting (1), and the 2D normalized hit point (2).

\paragraph{Camera.}
A fixed overhead view of the billiard table.

\paragraph{Evaluation dimensions.}
BilliardGame is used for \textbf{kinematics and mechanics} evaluation (ball trajectory matching via DTW and FDE; momentum and energy conservation via exact-match percentage and error rate) and \textbf{causal LPIPS}. The 2D hit-point action directly controls post-collision trajectories through spin, making this scene particularly demanding: a world model must capture not only elastic collision angles but also the effect of torque-induced spin on subsequent ball paths.

\subsection{ChessBoard}
\label{app:chessboard}

\begin{figure}[ht]
    \centering
    \begin{subfigure}[b]{0.24\textwidth}
        \centering
        \includegraphics[width=\textwidth]{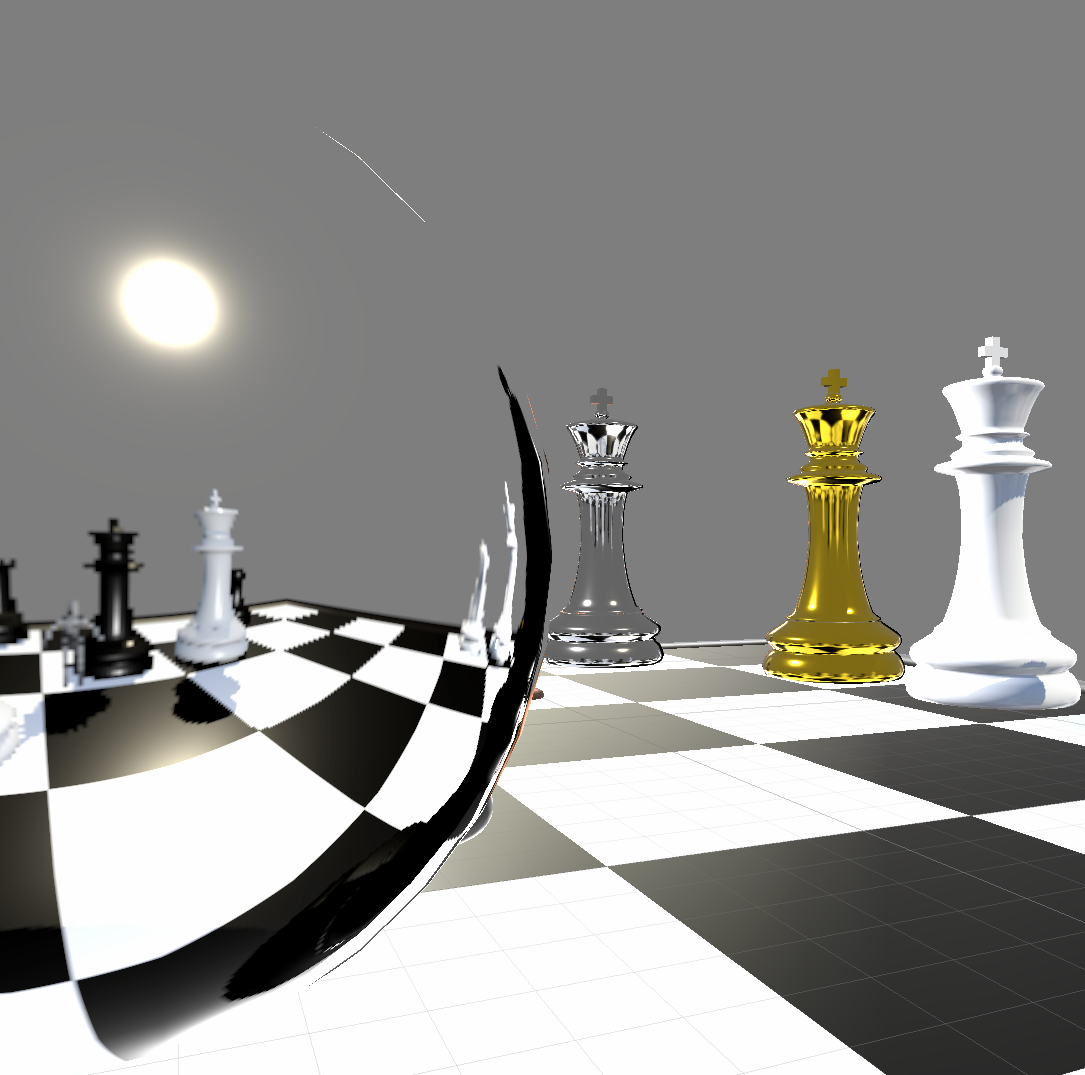}
        \caption{First-person}
        \label{fig:scenes:chessboard:first}
    \end{subfigure}
    \hfill
    \begin{subfigure}[b]{0.24\textwidth}
        \centering
        \includegraphics[width=\textwidth]{figures/ChessBoard2.png}
        \caption{Third-person}
        \label{fig:scenes:chessboard:third}
    \end{subfigure}
    \hfill
    \begin{subfigure}[b]{0.49\textwidth}
        \centering
        \includegraphics[width=0.48\textwidth]{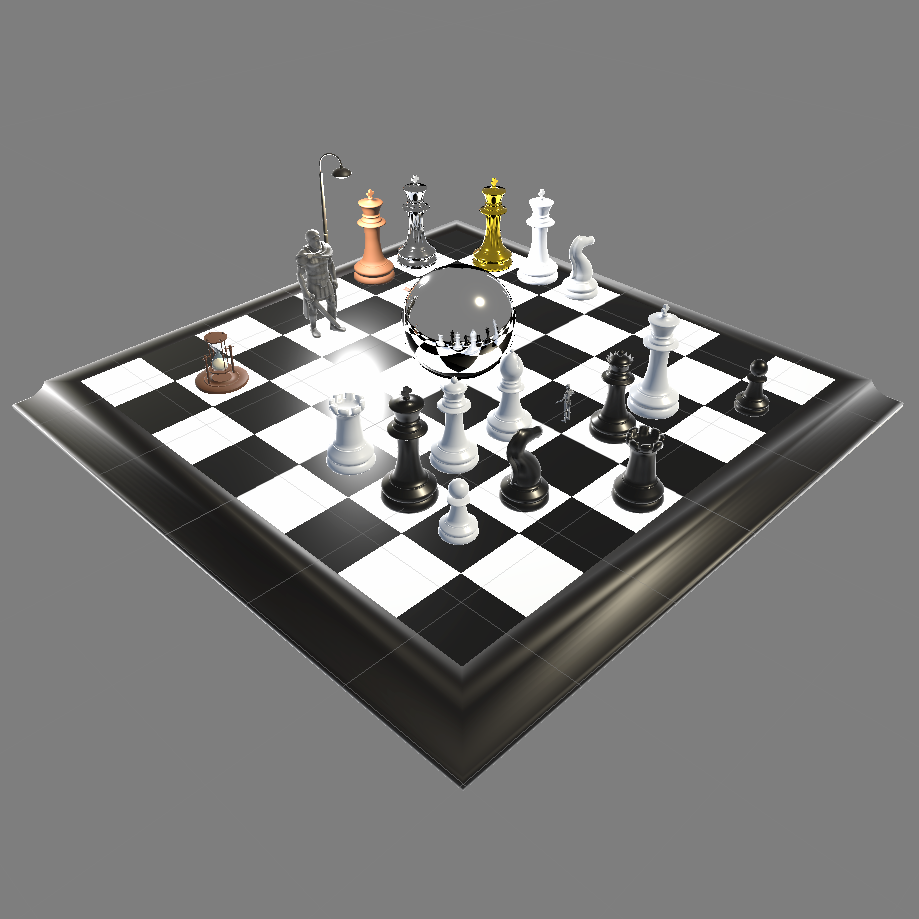}
        \hfill
        \includegraphics[width=0.48\textwidth]{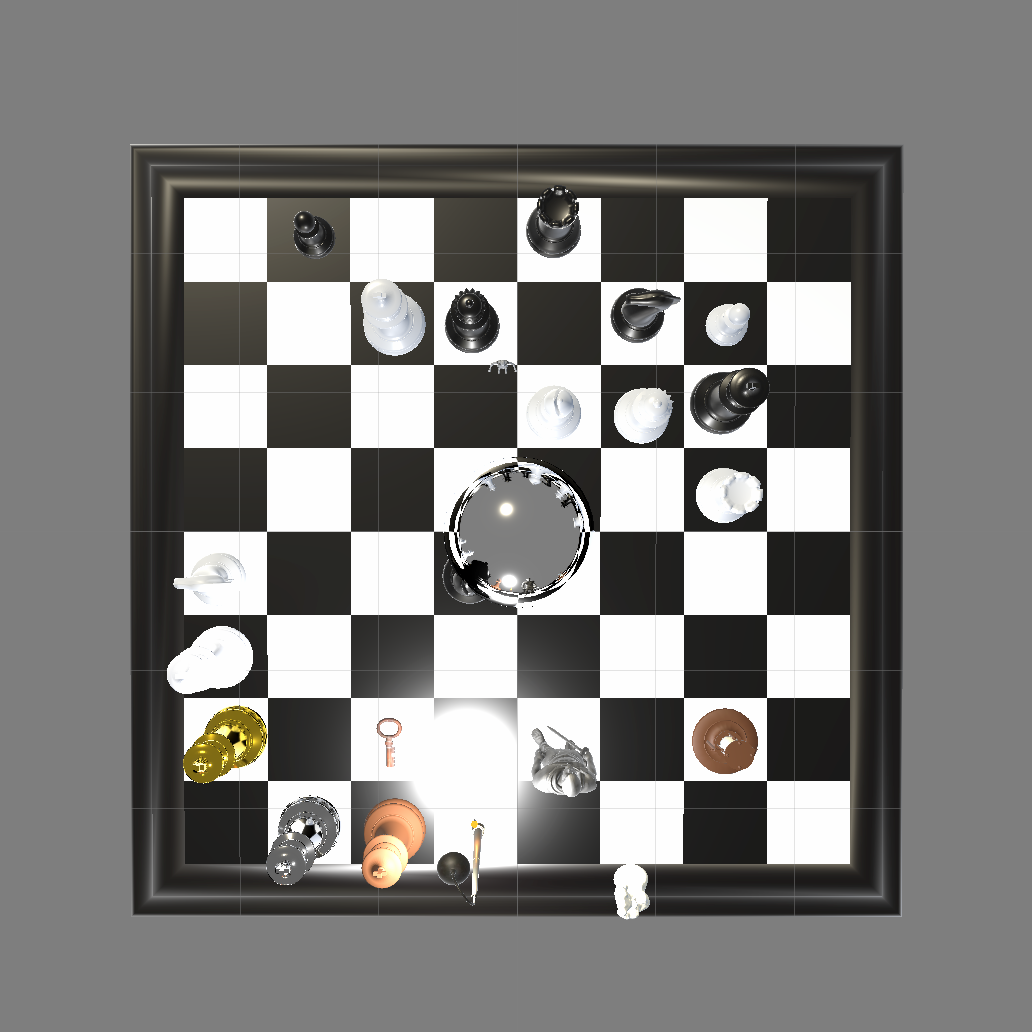}
        \caption{Environmental Layouts}
        \label{fig:scenes:chessboard:layouts}
    \end{subfigure}

    \caption{\textbf{ChessBoard Scene Views.}}
    \label{fig:scenes:chessboard_views}
\end{figure}

\paragraph{Scene description.}
ChessBoard features a giant black-and-white chessboard ground plane under a flat gray sky, surrounded by oversized glossy chess pieces (king, queen, rook, bishop, knight, pawn). A large reflective chrome sphere stands near the center of the board. The player controls a gray humanoid mannequin character in third-person view, navigating around the chess pieces with the ability to run and jump.

\paragraph{Action space.}
Two continuous actions (horizontal movement along $X$ and $Z$, each in $[-1,+1]$) plus two discrete branches: jump (0 or 1) and sprint (0 or 1).

\paragraph{Observation space.}
An 8-dimensional vector: the character's local position (3), local rotation as a quaternion (4), and a grounded flag (1). The minimal state representation forces reliance on the visual observation from the camera for spatial reasoning.

\paragraph{Camera.}
A Cinemachine third-person follow camera that smoothly tracks the humanoid avatar.

\paragraph{Evaluation dimensions.}
ChessBoard is used for \textbf{scene and character consistency} and \textbf{prompt--scene consistency} (VLM-as-judge). The visually distinctive environment---with its regular grid pattern, specular chess pieces, and the chrome sphere---provides unambiguous landmark features for testing whether a world model preserves scene geometry, object identity, and the reflective material properties of the sphere across extended rollouts. The combination of discrete jump actions and continuous movement on the structured chessboard grid creates sharp visual transitions (ground-to-airborne state changes) that test character animation consistency.

\subsection{MirrorRollingBall}
\label{app:MirrorRollingBall}

\begin{figure}[ht]
    \centering
    \begin{subfigure}[b]{0.24\textwidth}
        \centering
        \includegraphics[width=\textwidth]{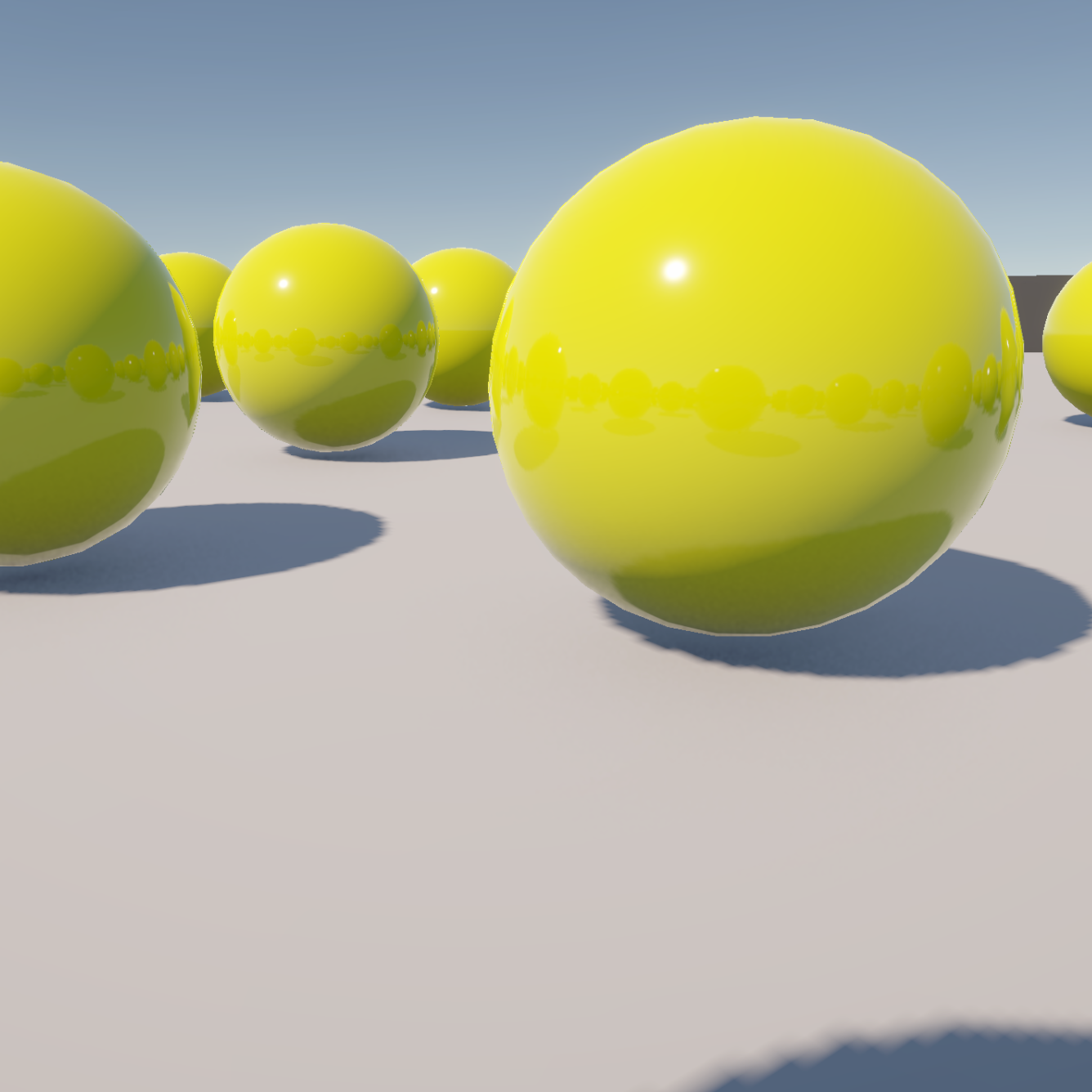}
        \caption{First-person}
        \label{fig:scenes:mirror:first}
    \end{subfigure}
    \hfill
    \begin{subfigure}[b]{0.24\textwidth}
        \centering
        \includegraphics[width=\textwidth]{figures/MirrorRollingBall2.png}
        \caption{Third-person}
        \label{fig:scenes:mirror:third}
    \end{subfigure}
    \hfill
    \begin{subfigure}[b]{0.49\textwidth}
        \centering
        \includegraphics[width=0.48\textwidth]{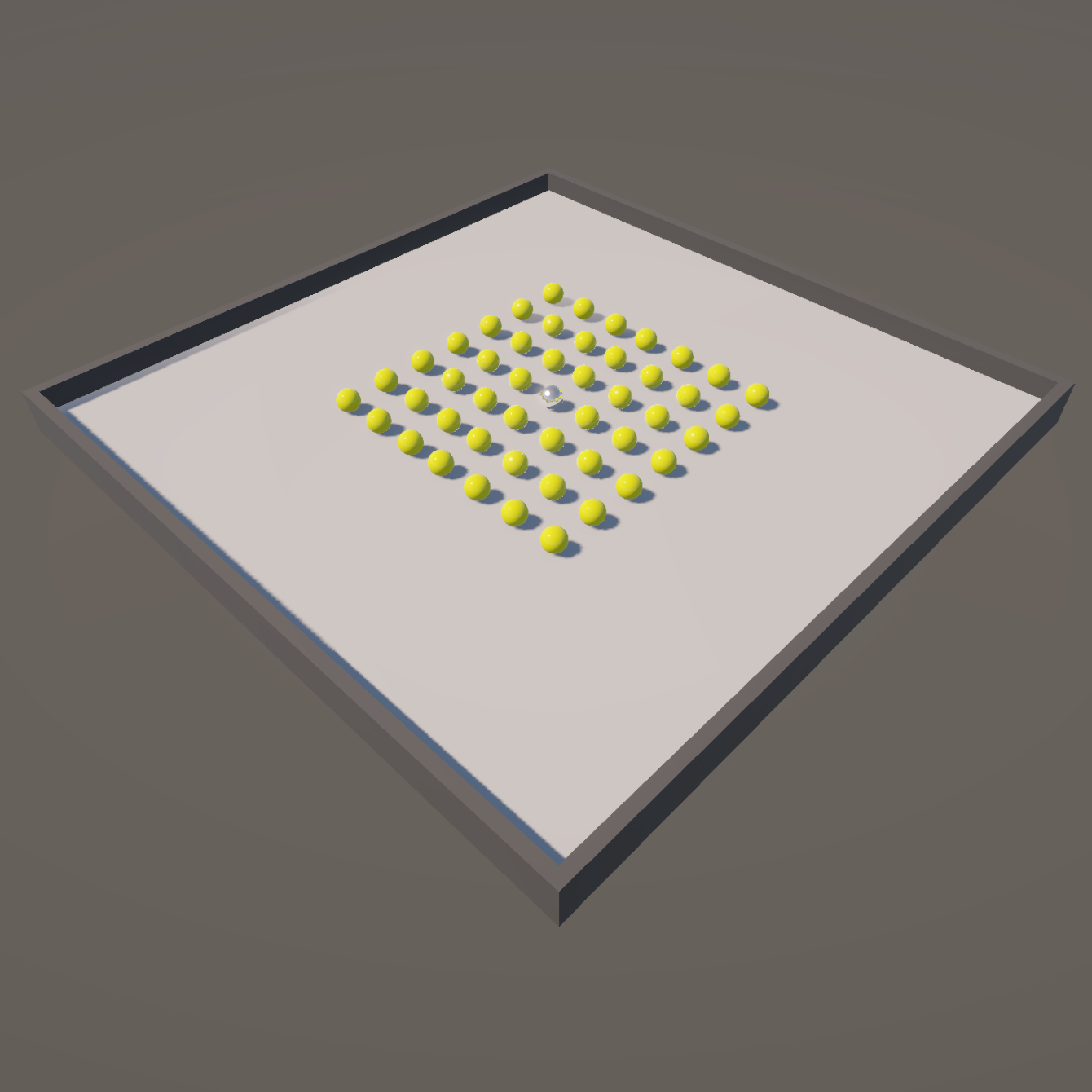}
        \hfill
        \includegraphics[width=0.48\textwidth]{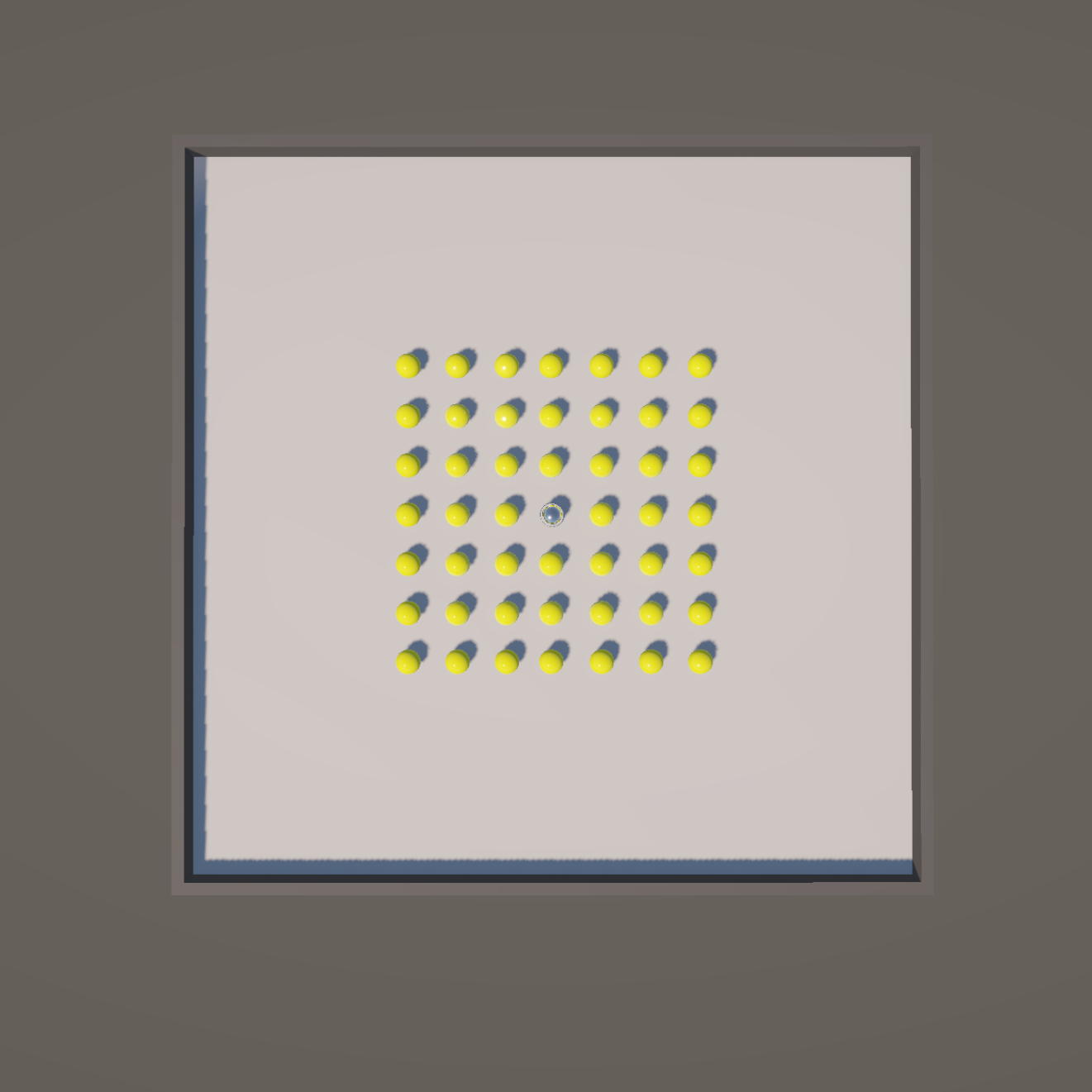}
        \caption{Environmental Layouts}
        \label{fig:scenes:mirror:layouts}
    \end{subfigure}

    \caption{\textbf{MirrorRollingBall Scene Views.}}
    \label{fig:scenes:mirror_views}
\end{figure}

\paragraph{Scene description.}
MirrorRollingBall places a reflective chrome sphere (the player-controlled ball) on a flat open ground surrounded by a field of up to 48 large glossy yellow balls. Bright directional sunlight casts visible shadows beneath all objects. The scene is visually simpler than BallRoller but exposes substantially more state: full kinematic information (position and velocity) for all tracked objects.

\paragraph{Action space.}
Two continuous actions: force along $X$ and force along $Z$, each in $[-1,+1]$, applied via \texttt{AddForce} to the player ball.

\paragraph{Observation space.}
A 297-dimensional vector: the player ball's position, linear velocity, and angular velocity (9), concatenated with the position and linear velocity of up to 48 yellow balls ($48 \times 6 = 288$). Missing balls are zero-padded to maintain a fixed-length vector.

\paragraph{Camera.}
A fixed-offset follow camera identical in design to BallRoller.

\paragraph{Evaluation dimensions.}
MirrorRollingBall is used for \textbf{scene and character consistency} (depth AbsRel, pairwise distance MAE, and character identity preservation via DINOv2), \textbf{prompt--scene consistency}, and contributes to \textbf{kinematics} evaluation through ball-to-ball collision tracking. The scene supports configurable ball counts (10, 20, 30, or 48 yellow balls), enabling controlled evaluation of how consistency metrics degrade as scene complexity increases.

\subsection{LiquidContainer}
\label{app:liquidcontainer}

\begin{figure}[ht]
    \centering
    \begin{subfigure}[b]{0.24\textwidth}
        \centering
        \includegraphics[width=\textwidth]{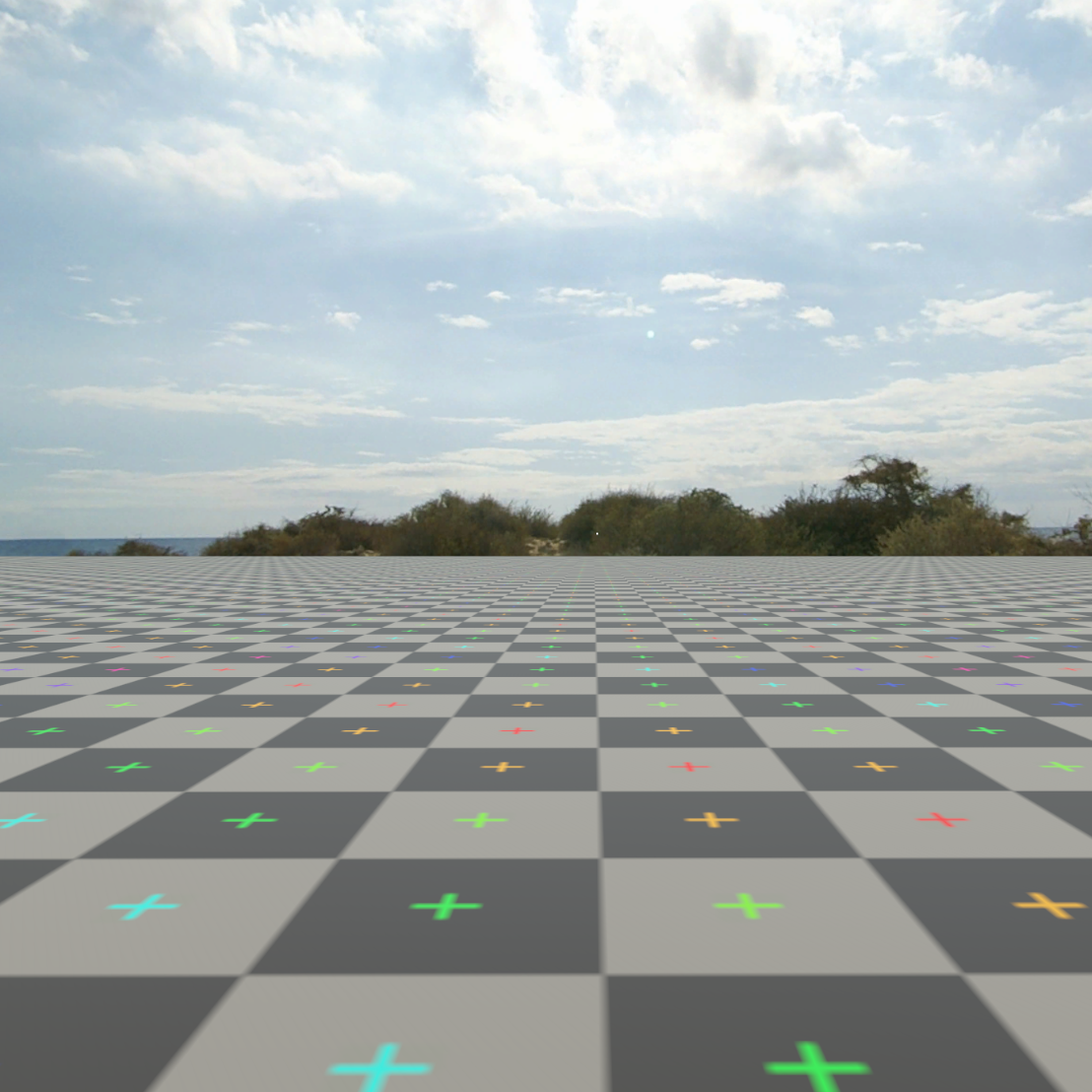}
        \caption{First-person}
        \label{fig:scenes:liquid:first}
    \end{subfigure}
    \hfill
    \begin{subfigure}[b]{0.24\textwidth}
        \centering
        \includegraphics[width=\textwidth]{figures/LiquidContainer2.png}
        \caption{Third-person}
        \label{fig:scenes:liquid:third}
    \end{subfigure}
    \hfill
    \begin{subfigure}[b]{0.49\textwidth}
        \centering
        \includegraphics[width=0.48\textwidth]{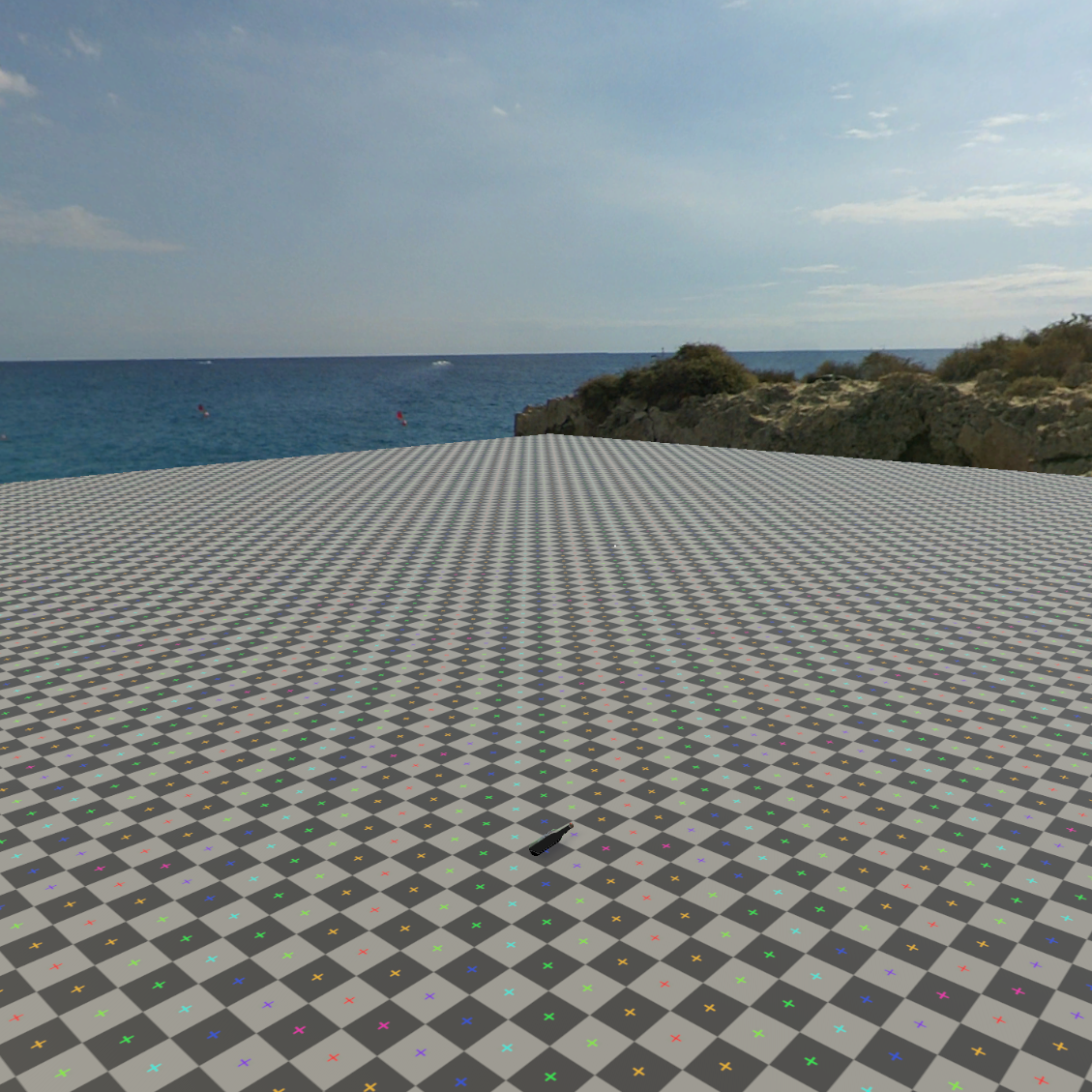}
        \hfill
        \includegraphics[width=0.48\textwidth]{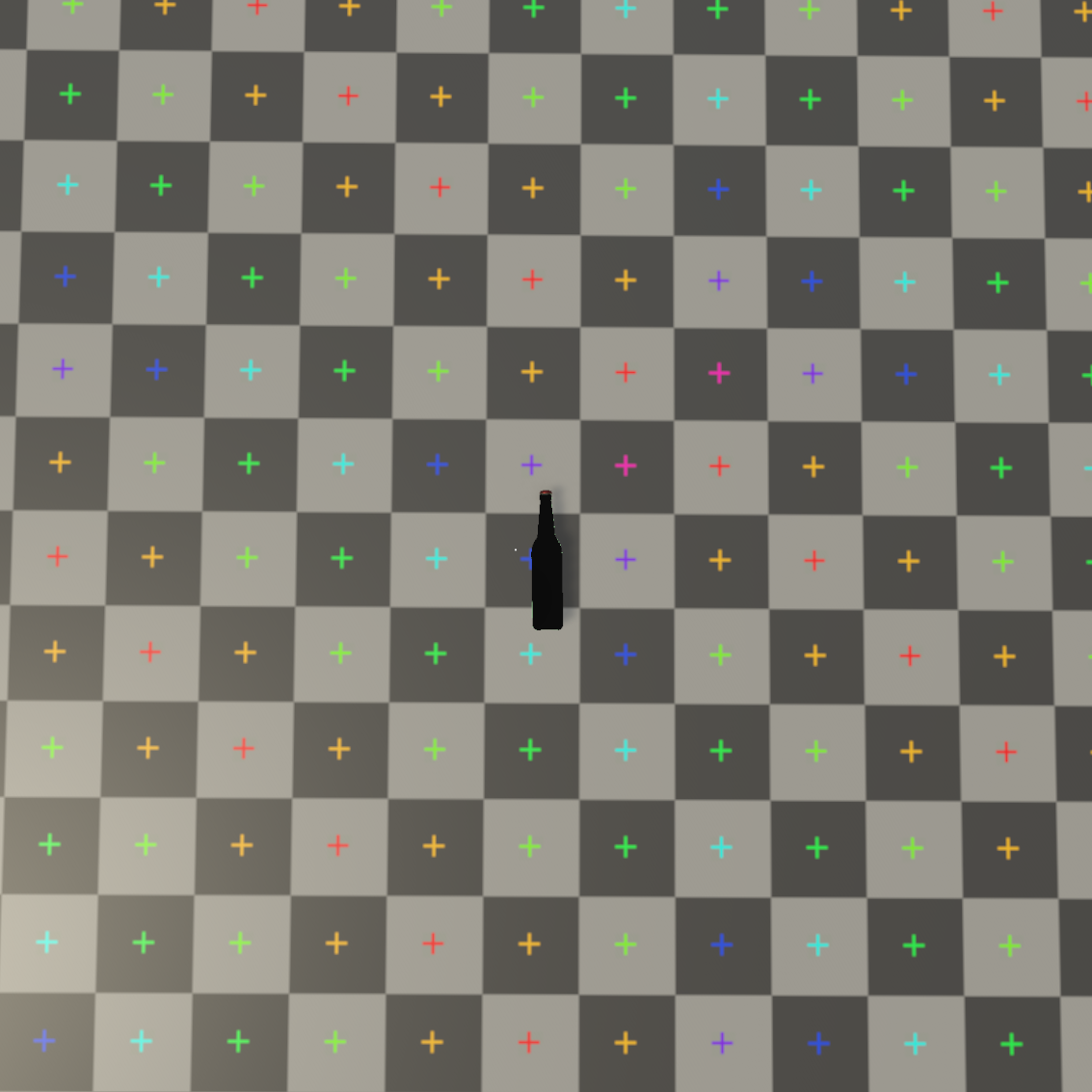}
        \caption{Environmental Layouts}
        \label{fig:scenes:liquid:layouts}
    \end{subfigure}

    \caption{\textbf{LiquidContainer Scene Views.}}
    \label{fig:scenes:liquid_views}
\end{figure}

\paragraph{Scene description.}

LiquidContainer is a continuous-control fluid manipulation scene in which the player drives a sealed, partially filled glass bottle through six degrees of freedom. The bottle is rendered against a coastal HDRI and rests on a checker-and-cross calibration floor that anchors scale and orientation cues.

The scene evaluates whether a world model preserves the visual appearance of a moving air-liquid interface under transparent-glass refraction, together with the first-order constraints its parametric reference enforces: projected-area stability under rigid-body motion, a gravity-aligned free surface at each tilt angle, and damped-oscillator slosh decay. The reference is a parametric slosh model rather than a fluid solver, and its scope is stated in Appendix~\ref{app:fluid_reference}.

Unlike discrete rigid-body scenes, failures here do not appear as single event-state errors. They appear as continuous physical violations: projected volume drift, incorrect surface tilt, unstable velocity fields, and off-frequency sloshing that persists after the triggering action.


\paragraph{Action space.}
Six continuous actions: forward/back, right/left, up/down translation, and pitch, yaw, roll rotation, each in $[-1,+1]$.

\paragraph{Observation space.}
A 24-dimensional vector: world position (3), forward direction (3), up direction (3), linear velocity (3), angular velocity (3), Euler angles normalized to $[-1,+1]$ (3), and the six current input values (6). The redundant orientation representation (direction vectors plus Euler angles) provides robust pose information.

\paragraph{Camera.}
A static side-view camera (\texttt{c2\_static\_side} regime) that frames the bottle and calibration floor.

\paragraph{Evaluation dimensions.}
LiquidContainer is the sole scene for \textbf{fluid mechanics} evaluation, scoped to sealed-container slosh (Appendix~\ref{app:fluid_reference}). It tests five metric families: velocity field fidelity (cosine similarity), volume conservation (normalized mask-area standard deviation and per-second drift), surface orientation (RMS tilt residual of the air-liquid interface), slosh spectrum (meniscus power spectrum $L_1$), and segment-aligned LPIPS as a perceptual sanity check.

Every visible element doubles as a calibration affordance: the checker floor provides spatial reference, the red cap marks the bottle's orientation, and the sealed liquid surface must align with gravity at each tilt angle.

\subsection{RacingCar}
\label{app:racingcar}

\begin{figure}[ht]
    \centering
    \begin{subfigure}[b]{0.24\textwidth}
        \centering
        \includegraphics[width=\textwidth]{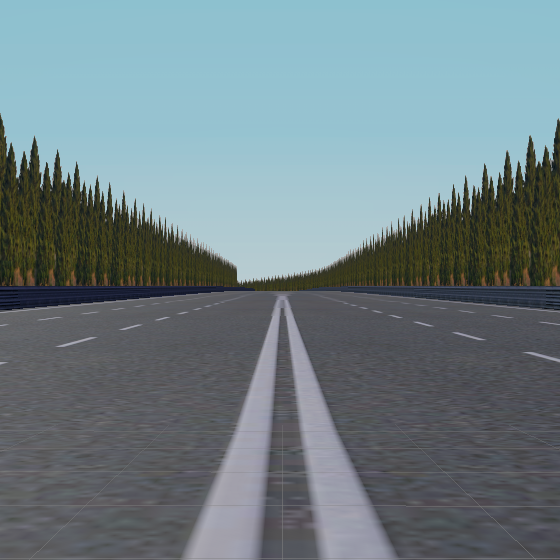}
        \caption{First-person}
        \label{fig:scenes:racing:first}
    \end{subfigure}
    \hfill
    \begin{subfigure}[b]{0.24\textwidth}
        \centering
        \includegraphics[width=\textwidth]{figures/RacingCar2.png}
        \caption{Third-person}
        \label{fig:scenes:racing:third}
    \end{subfigure}
    \hfill
    \begin{subfigure}[b]{0.49\textwidth}
        \centering
        \includegraphics[width=0.48\textwidth]{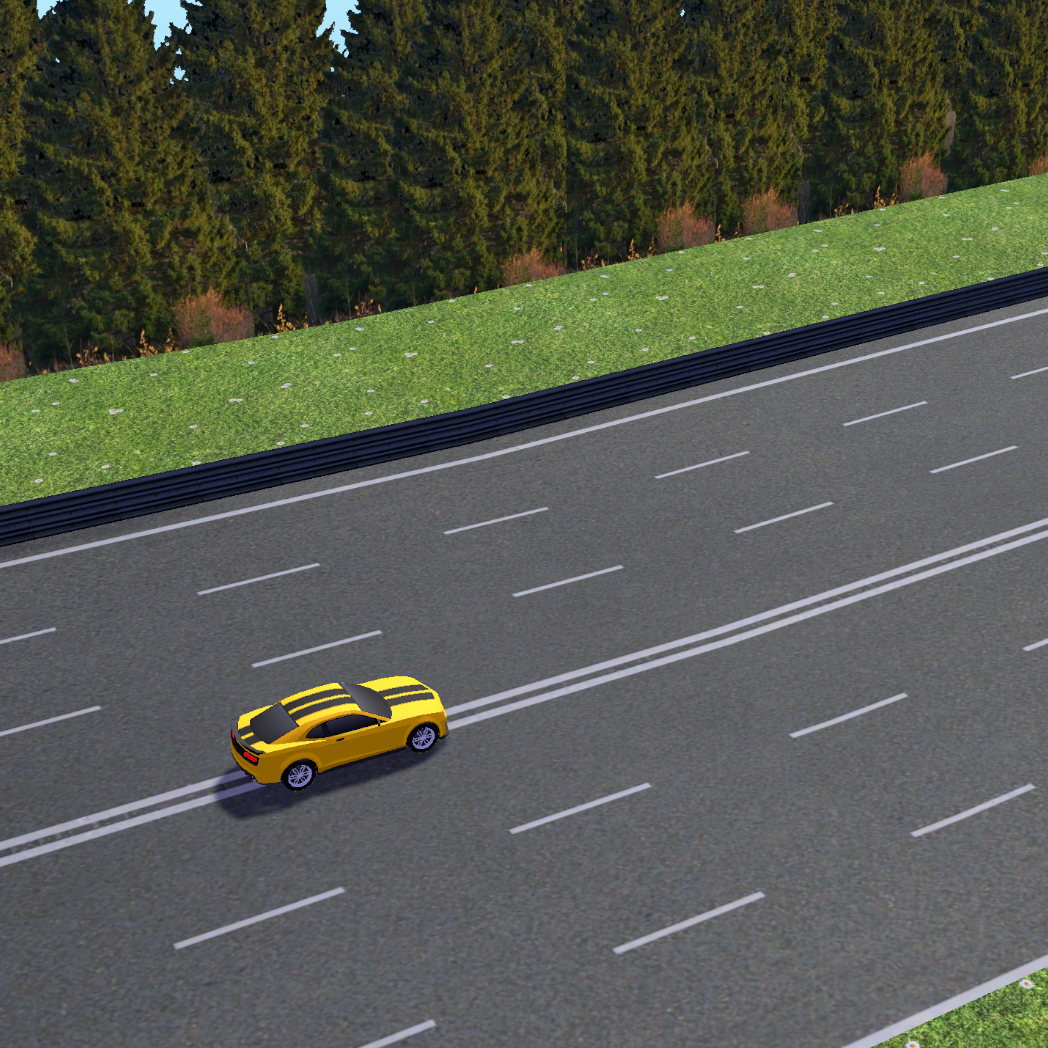}
        \hfill
        \includegraphics[width=0.48\textwidth]{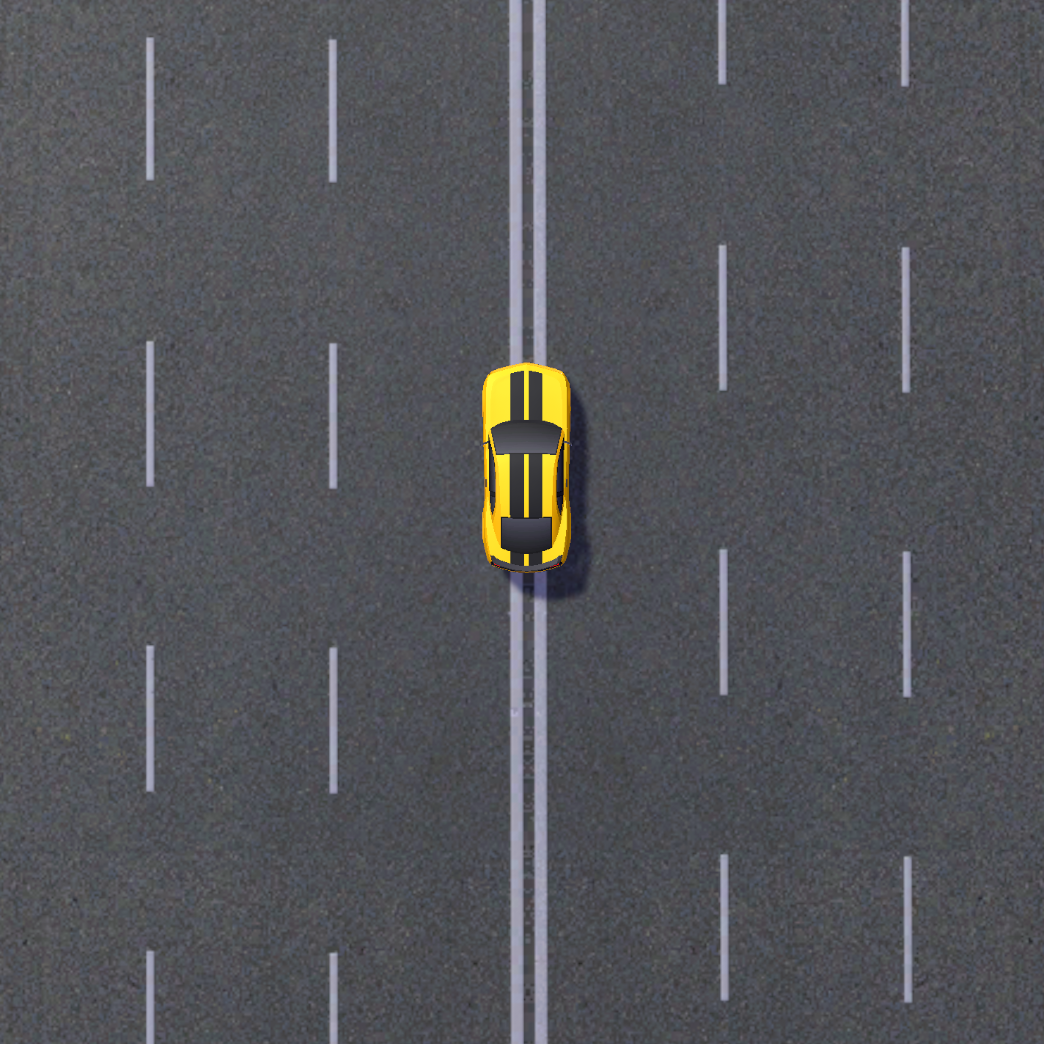}
        \caption{Environmental Layouts}
        \label{fig:scenes:racing:layouts}
    \end{subfigure}

    \caption{\textbf{RacingCar Scene Views.}}
    \label{fig:scenes:racing_views}
\end{figure}

\paragraph{Scene description.}
RacingCar features a wide straight highway with lane markings, dark guardrails, green grass medians, and symmetrical rows of tall trees receding into the distance. The player controls a yellow sports car with two black racing stripes, viewed from behind via a dynamic follow camera. The road is procedurally generated: an \texttt{InfiniteRoads} system tiles road segments ahead of the car and recycles segments behind it, creating an unbounded driving surface.

\paragraph{Action space.}
Two continuous actions plus one discrete branch. Steering ($[-1,+1]$, mapped to $\pm30^{\circ}$ on the front wheels) and throttle ($[-1,+1]$, driving rear-wheel motor torque) are continuous. Braking (0 or 1) is discrete, applying 3000\,N of brake torque to all four wheels when active.

\paragraph{Observation space.}
A 13-dimensional vector: world position (3), forward direction (3), linear velocity (3), normalized speed clamped to a maximum of 120\,km/h (1), and the current steering, throttle, and brake inputs (3). Including the current control inputs in the observation allows a world model to learn the delayed response characteristic of Unity's WheelCollider physics.

\paragraph{Camera.}
A speed-adaptive dynamic follow camera. As the car accelerates, the camera pulls back (up to 3\,m additional distance), the look-at target shifts forward of the car, and the field of view widens from $65^{\circ}$ to $80^{\circ}$, creating perceptual speed cues that a visual world model can exploit.

\paragraph{Evaluation dimensions.}
RacingCar is used for \textbf{kinematics and mechanics} (car trajectory matching via DTW and FDE, plus momentum and kinetic energy conservation checks) and \textbf{prompt--scene consistency}. The scene is evaluated under three camera regimes---dynamic follow, static side, and static top-down---to disentangle the effect of camera motion on trajectory estimation accuracy. The WheelCollider-based vehicle dynamics, including front-wheel steering, rear-wheel drive, and independent four-wheel braking, provide a physically grounded test bed for predicting non-trivial vehicle behavior such as acceleration curves, braking distances, and steering response.

\subsection{SunRisesAndSets}
\label{app:sunrisesandsets}

\begin{figure}[ht]
    \centering
    \begin{subfigure}[b]{0.24\textwidth}
        \centering
        \includegraphics[width=\textwidth]{figures/SunRisesAndSets1.png}
        \caption{Day}
        \label{fig:scenes:sun:day}
    \end{subfigure}
    \hfill
    \begin{subfigure}[b]{0.24\textwidth}
        \centering
        \includegraphics[width=\textwidth]{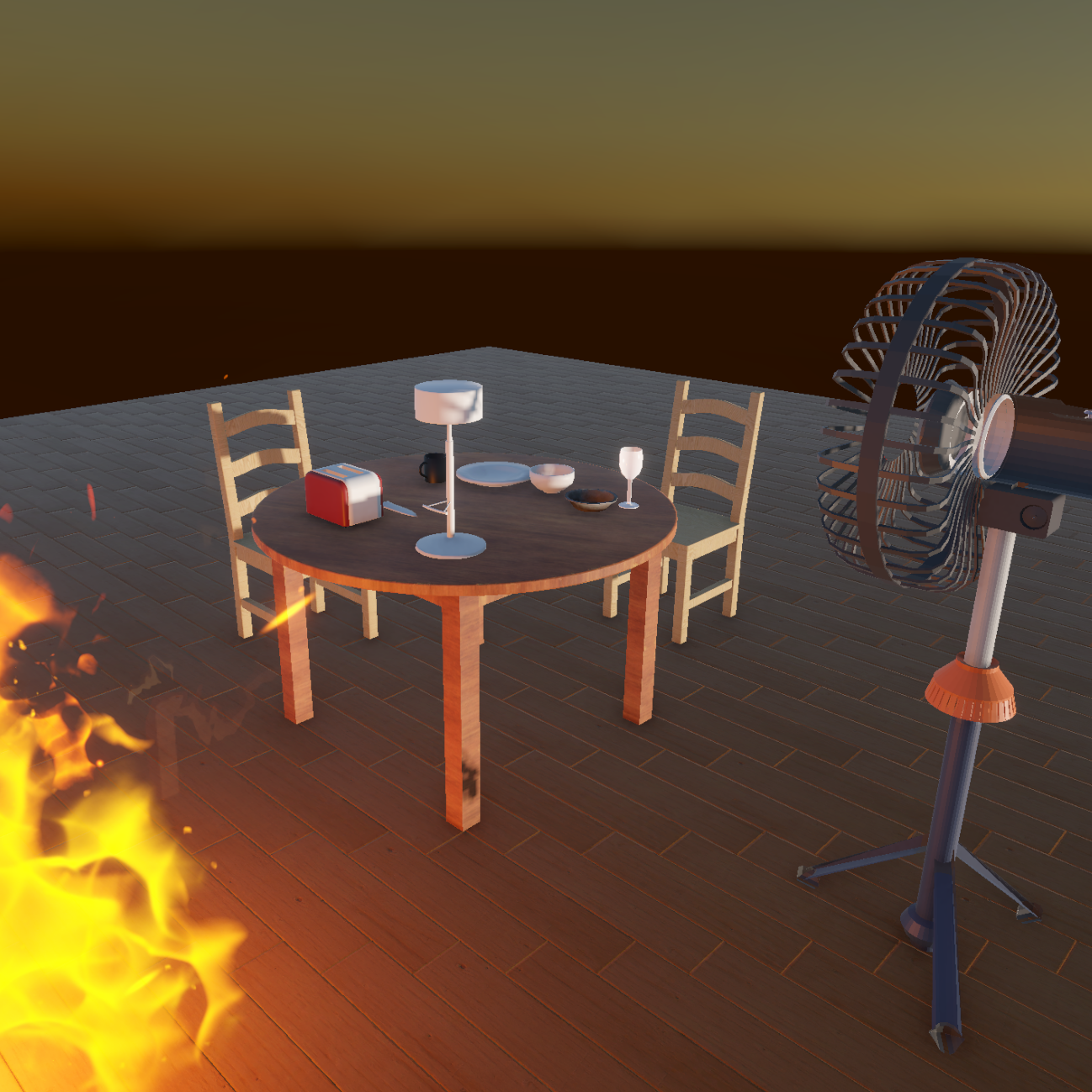}
        \caption{Sunset}
        \label{fig:scenes:sun:sunset}
    \end{subfigure}
    \hfill
    \begin{subfigure}[b]{0.24\textwidth}
        \centering
        \includegraphics[width=\textwidth]{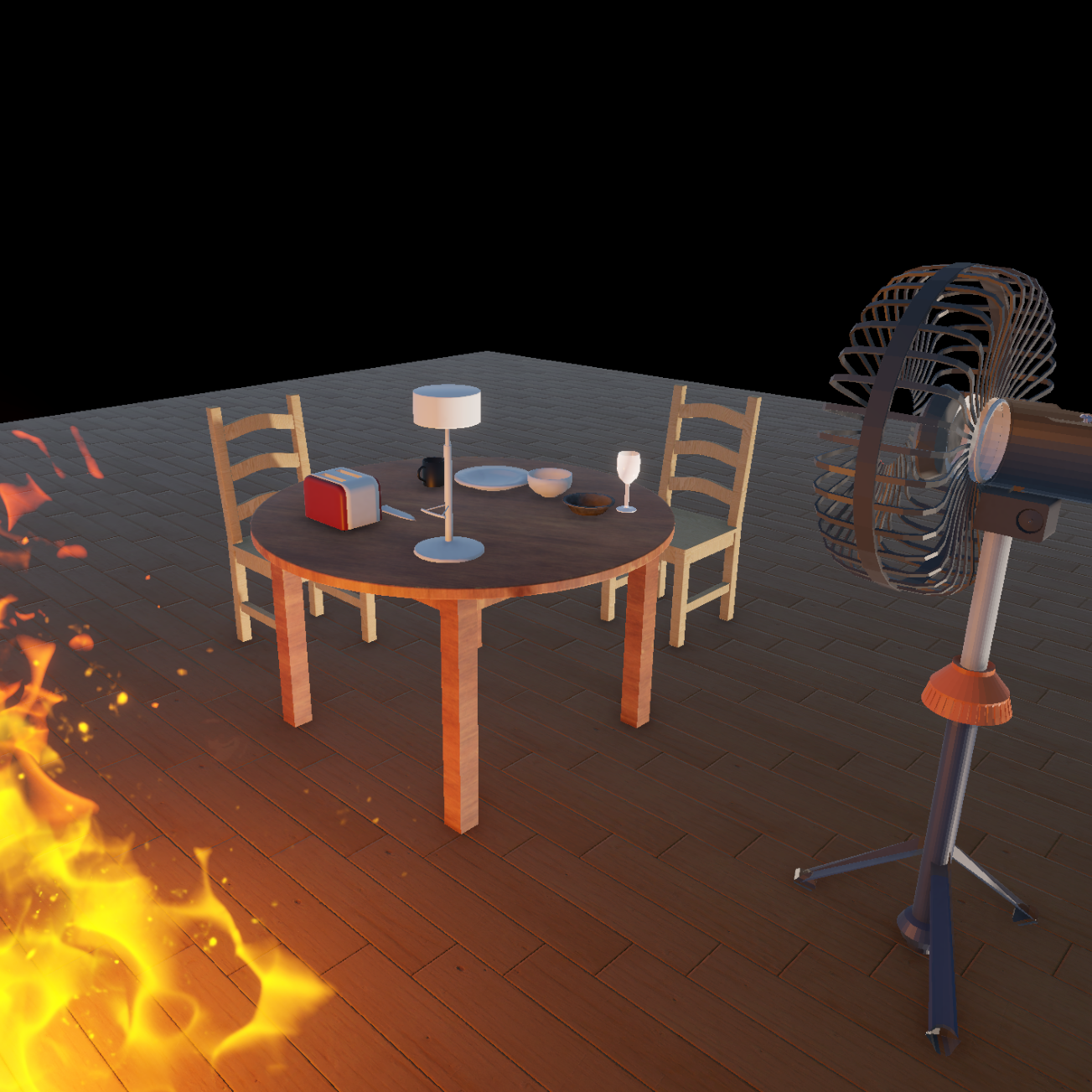}
        \caption{Night}
        \label{fig:scenes:sun:night}
    \end{subfigure}
    \hfill
    \begin{subfigure}[b]{0.24\textwidth}
        \centering
        \includegraphics[width=\textwidth]{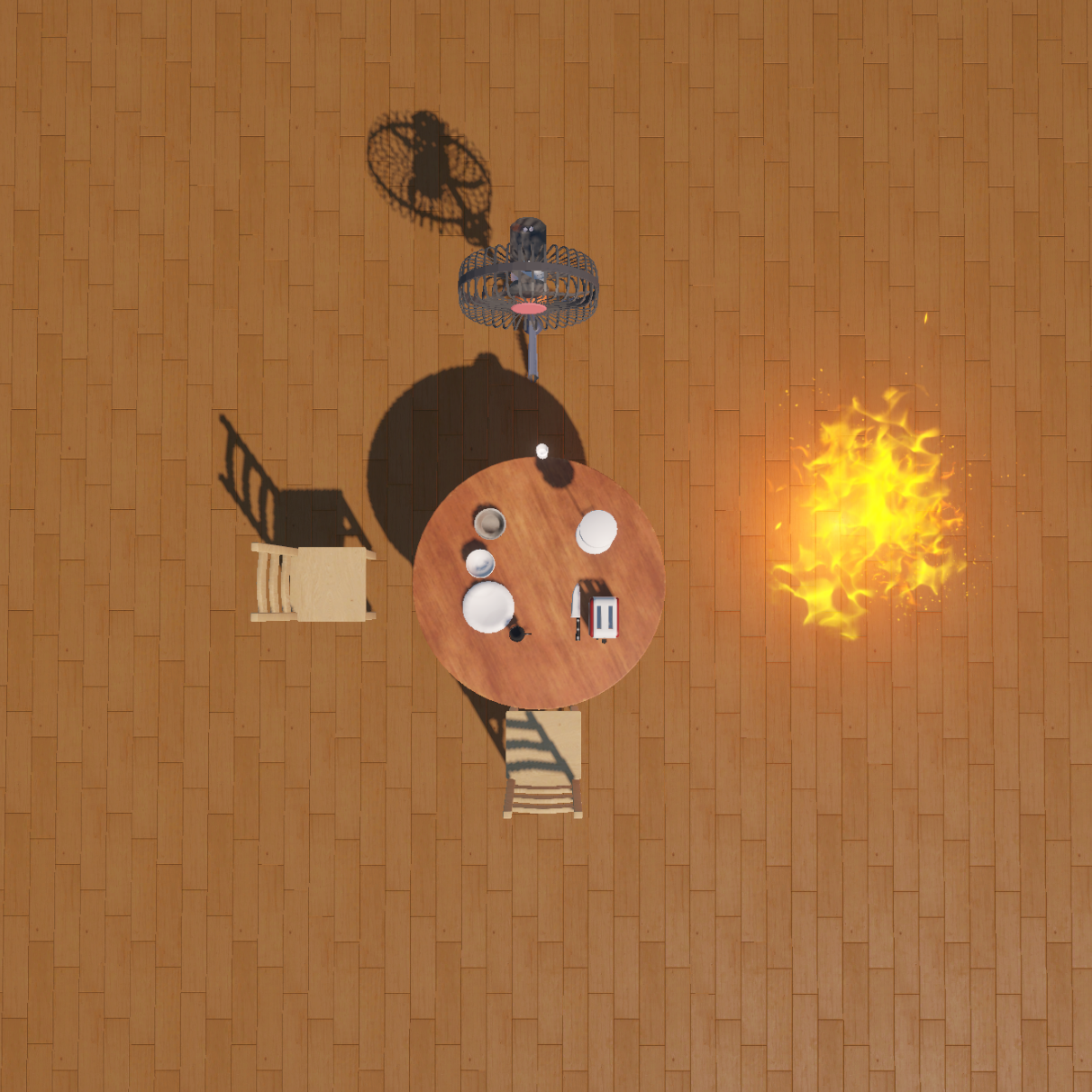}
        \caption{Overhead Layout}
        \label{fig:scenes:sun:layout}
    \end{subfigure}

    \caption{\textbf{SunRisesAndSets Scene Views.} (a--c) The environment across a full diurnal cycle, illustrating dynamic shadow kinematics and global illumination changes. (d) A bird's-eye environmental layout showing the static objects used to evaluate shadow projection accuracy.}
    \label{fig:scenes:sun_views}
\end{figure}

\paragraph{Scene description.}
SunRisesAndSets is an outdoor tabletop scene featuring a dining table with everyday objects (a toaster, mug, bowl, plate, knife) and chairs, illuminated by a single directional light that rotates continuously around the world $X$ axis at $100^{\circ}/s$, simulating a compressed 24-hour day--night cycle. As the light rotates, it produces smoothly varying cast shadows, ambient color shifts (sunrise and sunset hues), and sky transitions. The scene additionally exposes per-frame structured data through a Unity ML-Agents side channel, transmitting a JSON payload containing every GameObject's name, position, and Euler angles, plus the directional light's intensity and orientation.

\paragraph{Action space.}
The camera is positioned at a fixed orbit point (yaw $45^{\circ}$, pitch $30^{\circ}$, distance 8\,m from the look-at target) at initialization and does not move during the episode. No effective player actions are taken.

\paragraph{Observation space.}
A 9-dimensional vector: camera world position (3), look-at target position (3), and the directional light's Euler angles (3). The side channel additionally provides full scene-state JSON every 10 decision steps.

\paragraph{Camera.}
A fixed-orbit camera that does not translate or rotate during evaluation, ensuring that all visual changes stem solely from the light rotation.

\paragraph{Evaluation dimensions.}
SunRisesAndSets is the primary scene for \textbf{lighting and shading} evaluation, supporting both Absolute Shadow Kinematics (shadow direction angular error, shadow temporal stability, shadow area fluctuation, centroid smoothness error) and Diagnostic Lighting Consistency / IISC (conditional light--shadow angular error, lighting temporal jitter, solar tracking success, retrieval coverage). The fixed camera isolates lighting-induced visual changes from camera-induced ones, and the deterministic light rotation provides a precise ground-truth trajectory against which shadow kinematics can be compared. The per-object shadow tracking (table, chair, bowl, toaster) enables fine-grained analysis of how shadow angular error varies across objects with different geometries.

\subsection{WanderingGarden}
\label{app:wanderinggarden}

\begin{figure}[ht]
    \centering
    \begin{subfigure}[b]{0.24\textwidth}
        \centering
        \includegraphics[width=\textwidth]{figures/WanderingGarden1.png}
        \caption{First-person}
        \label{fig:scenes:garden:first}
    \end{subfigure}
    \hfill
    \begin{subfigure}[b]{0.74\textwidth}
        \centering
        \includegraphics[width=0.32\textwidth]{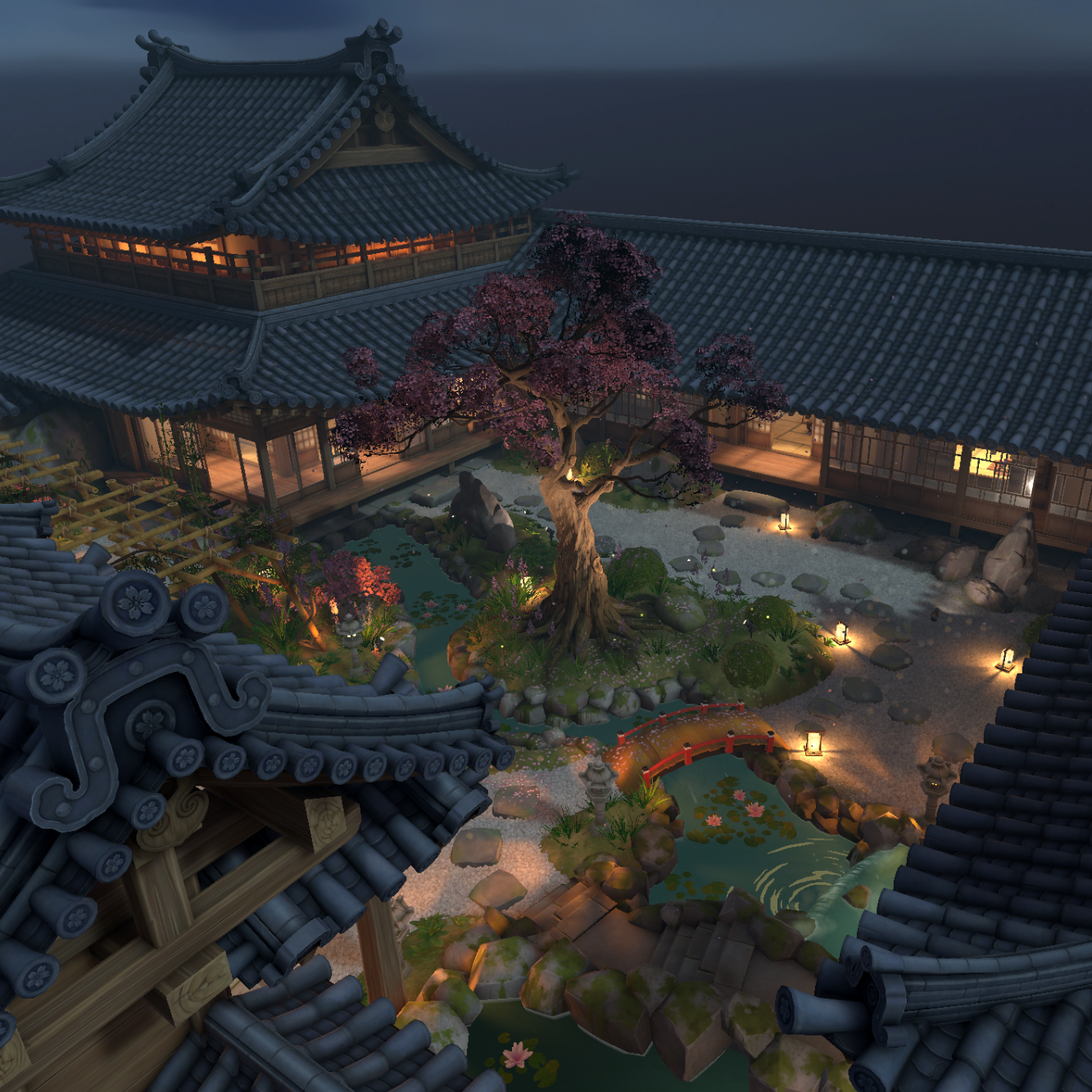}
        \hfill
        \includegraphics[width=0.32\textwidth]{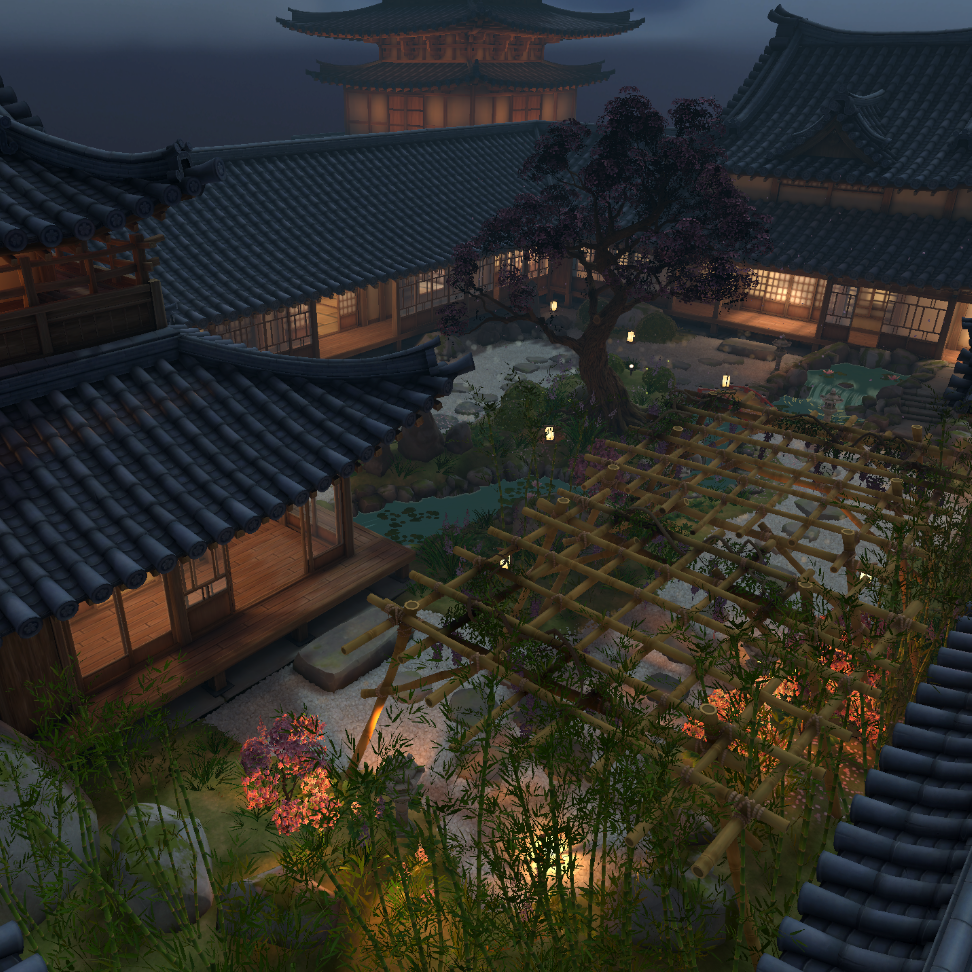}
        \hfill
        \includegraphics[width=0.32\textwidth]{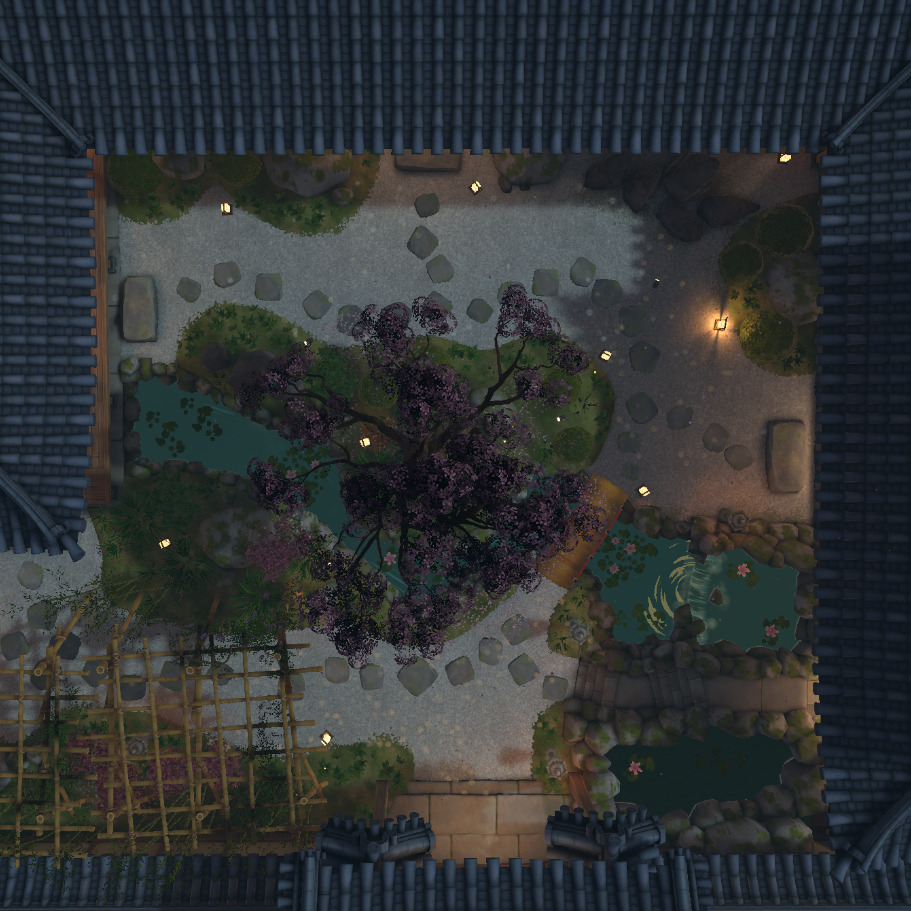}
        \caption{Environmental Layouts}
        \label{fig:scenes:garden:layouts}
    \end{subfigure}

    \caption{\textbf{WanderingGarden Scene Views.}}
    \label{fig:scenes:garden_views}
\end{figure}

\paragraph{Scene description.}
WanderingGarden is a first-person exploration scene set in a high-fidelity garden environment rendered with the Universal Render Pipeline (URP). The environment features dynamic lighting, particle effects, foliage with wind animation, spline-following NPCs, and boid flocks. The player is a capsule-shaped character navigating via a \texttt{CharacterController}, with ego-motion driven by look and move actions written to Unity's StarterAssets input system. Five designated anchor objects in the garden serve as landmarks for object permanence testing.

\paragraph{Action space.}
Four continuous actions: strafe ($X$, $[-1,+1]$), forward/back ($Y$, $[-1,+1]$), yaw rotation ($[-1,+1]$), and pitch rotation ($[-1,+1]$). These are written directly to the \texttt{StarterAssetsInputs} component, driving the existing \texttt{FirstPersonController} without modification.

\paragraph{Observation space.}
A 50-dimensional vector plus a camera-sensor RGB image ($128 \times 128$ by default). The vector comprises: player position (3), forward direction (3), velocity from the CharacterController (3), the count of nearby objects within a 15\,m detection radius (1), and up to 10 nearest detected objects each represented by their relative position and distance ($10 \times 4 = 40$).

\paragraph{Camera.}
A Cinemachine first-person camera attached to the player capsule's eye-level, rotating with the look actions.

\paragraph{Evaluation dimensions.}
WanderingGarden is used for \textbf{scene and character consistency}: depth AbsRel error, pairwise distance MAE, and object permanence (fraction of the five anchor objects that remain detectable across $45^{\circ}$ camera perturbations and $360^{\circ}$ camera sweeps). The Python controller implements a sine-wave turning pattern with obstacle avoidance, producing smooth but non-trivial exploration trajectories. The visually rich environment---with high-frequency textures, dynamic particles, and complex lighting---represents the most demanding visual domain in \capsysname, stress-testing whether a world model can maintain spatial and appearance consistency in a photorealistic setting.

\section{Detailed Metric Definitions}
\label{app:metrics}

This appendix provides the formal definitions for the metrics summarized in Section~\ref{sec:benchmark}. All metrics are computed over the shared valid temporal window between the Unity reference and the world-model generation unless otherwise specified. Implementation details for event detection, RAFT extraction, physical-causality QA construction, fluid tracking, and VLM judging are provided in Appendix~\ref{app:implementation}.

\subsection{Environment Object and Character Consistency}
\label{app:consistency_metrics}

For environment-object consistency, we use GroundingDINO to localize a fixed set of key objects and SAM~2 to extract fine-grained masks and contours in each frame~\citep{10.1007/978-3-031-72970-6_3,ravi2025sam}. Given the detected object set $\mathcal{O}$, we estimate object-level depth and pairwise distances from the generated rollout, then compare them with Unity-derived references.

\paragraph{Depth AbsRel.}
Let $\hat d_o$ and $d_o$ denote the estimated and Unity-reference depths of object $o$. We report
\begin{equation}
    \mathrm{AbsRel}
    =
    \frac{1}{|\mathcal{O}|}
    \sum_{o\in\mathcal{O}}
    \frac{|\hat{d}_o-d_o|}{d_o+\epsilon}
    \times 100\% .
\end{equation}

\paragraph{Pairwise Distance MAE.}
Let $\hat{\Delta}_{ij}$ and $\Delta_{ij}$ denote estimated and reference pairwise object distances. We report the relative pairwise distance error
\begin{equation}
    \mathrm{DistMAE}
    =
    \frac{2}{|\mathcal{O}|(|\mathcal{O}|-1)}
    \sum_{i<j}
    \frac{
    \left|
    \hat{\Delta}_{ij}-\Delta_{ij}
    \right|
    }{\Delta_{ij}+\epsilon}
    \times 100\% .
\end{equation}

\paragraph{Object Permanence.}
We measure object permanence as the fraction of predefined anchor objects that remain detectable throughout an action sequence:
\begin{equation}
    \mathrm{ObjPerm}
    =
    \frac{1}{T|\mathcal{O}|}
    \sum_{t=1}^{T}
    \sum_{o\in\mathcal{O}}
    \mathbf{1}\!\left[s_{t,o}\ge \tau_{\mathrm{det}}\right]
    \times 100\% ,
\end{equation}
where $s_{t,o}$ is the detection confidence for object $o$ at frame $t$.

\paragraph{Character Consistency.}
For main character consistency, we localize and segment the target character with GroundingDINO and SAM~2, then use DINOv2 features to check whether the character preserves its visual identity over time~\citep{oquab2024dinov}. Let $\phi_t$ be the DINOv2 embedding of the segmented character crop at frame $t$, and let $\phi_1$ be the initial-frame embedding. We define
\begin{equation}
    \mathrm{CharConsis}
    =
    \frac{1}{T}
    \sum_{t=1}^{T}
    \mathbf{1}\!\left[
    s_t \ge \tau_{\mathrm{det}}
    \land
    a_t \ge \tau_{\mathrm{mask}}
    \land
    \frac{\phi_t^\top \phi_1}{\|\phi_t\|_2\|\phi_1\|_2}
    \ge \tau_{\mathrm{id}}
    \right]
    \times 100\% ,
\end{equation}
where $s_t$ is the character detection confidence, $a_t$ is the mask quality or stability score, and $\tau_{\mathrm{id}}$ is the DINOv2 identity-similarity threshold.

\subsection{Action Controllability}
\label{app:action_controllability_metrics}

We measure action controllability using RAFT optical flow~\citep{teed2020raft}, which estimates dense per-pixel motion between consecutive frames. We use two references: \textit{Ground Truth RAFT} and \textit{Synthetic RAFT}. Ground Truth RAFT compares the model flow against RAFT flow extracted from the game engine reference video. Synthetic RAFT compares the model flow against an analytically generated flow whose action parameters are inferred from game engine-reference RAFT using an action-conditioned camera-motion model calibrated from game engine rollouts.

Let $t$ index sampled adjacent frame pairs, $\Omega$ the image domain, and $f^{\mathrm{ref}}_t,f^{\mathrm{cand}}_t:\Omega\to\mathbb{R}^2$ the reference and candidate flow fields, where $f_t(p)$ gives pixel displacement at pixel $p$. Let $M_t$ be the Ground Truth RAFT dynamic mask and $S_t$ the Synthetic RAFT magnitude-filtered set. With $\epsilon=10^{-8}$, we report
\[
\begin{aligned}
\mathrm{RelEPE}_{\mathrm{dyn}} &=
\mathbb{E}_{t}\mathbb{E}_{p\in M_t}
\frac{\|f^{\mathrm{ref}}_t(p)-f^{\mathrm{cand}}_t(p)\|_2}
     {\|f^{\mathrm{ref}}_t(p)\|_2+1}, \\
\mathrm{CosSim}_{\mathrm{dyn}} &=
\mathbb{E}_{t}\mathbb{E}_{p\in M_t}
\frac{f^{\mathrm{ref}}_t(p)\cdot f^{\mathrm{cand}}_t(p)}
     {\|f^{\mathrm{ref}}_t(p)\|_2\|f^{\mathrm{cand}}_t(p)\|_2+\epsilon}, \\
\mathrm{MeanFlowCos} &=
\mathbb{E}_{t}
\frac{\bar f^{\mathrm{ref}}_t\cdot \bar f^{\mathrm{cand}}_t}
     {\|\bar f^{\mathrm{ref}}_t\|_2\|\bar f^{\mathrm{cand}}_t\|_2+\epsilon},
\quad
\bar f_t=\mathbb{E}_{p\in S_t}f_t(p), \\
\mathrm{PixelEPE} &=
\mathbb{E}_{t}\mathbb{E}_{p\in S_t}
\|f^{\mathrm{ref}}_t(p)-f^{\mathrm{cand}}_t(p)\|_2 .
\end{aligned}
\]
Ground Truth RAFT uses $\mathrm{RelEPE}_{\mathrm{dyn}}$ and $\mathrm{CosSim}_{\mathrm{dyn}}$, while Synthetic RAFT uses $\mathrm{MeanFlowCos}$ and $\mathrm{PixelEPE}$. Additional extraction, masking, and calibration details are given in Appendix~\ref{app:raft_pipeline}.

\subsection{Kinematics and Mechanics}
\label{app:kinematics_mechanics_metrics}

For kinematics, we reconstruct key-object trajectories from action-conditioned generations and compare them with game engine references. We use GroundingDINO to detect key objects in each frame and ByteTrack to associate detections over time into object trajectories~\citep{10.1007/978-3-031-72970-6_3,zhang2022bytetrack}. To account for camera motion, we track static anchor objects and use them to estimate the moving reference-frame trajectory. Combining this reference-frame motion with tracked object motion yields each key object's trajectory in a common static frame.

\paragraph{Trajectory Fidelity.}
Given the estimated trajectory $\hat{\mathbf{x}}_{1:T}$ and Unity ground-truth trajectory $\mathbf{x}_{1:T}$, we report final displacement error (FDE)
\begin{equation}
    \mathrm{FDE}(\hat{\mathbf{x}}, \mathbf{x})
    =
    \left\| \hat{\mathbf{x}}_{T} - \mathbf{x}_{T} \right\|_2 ,
\end{equation}
and dynamic time warping (DTW)
\begin{equation}
    \mathrm{DTW}(\hat{\mathbf{x}}, \mathbf{x})
    =
    \min_{\pi \in \Pi}
    \sum_{(i,j)\in \pi}
    \left\| \hat{\mathbf{x}}_{i} - \mathbf{x}_{j} \right\|_2 ,
\end{equation}
where $\Pi$ denotes the set of valid monotonic alignment paths between estimated and reference trajectories.

\paragraph{Mechanical Quantities.}
Given estimated velocity $\hat{\mathbf{v}}_t=(\hat{\mathbf{x}}_t-\hat{\mathbf{x}}_{t-1})/\Delta t$ and simulator-provided mass $m$, we compute estimated momentum and kinetic energy as
\begin{equation}
    \hat{\mathbf{p}}_t=m\hat{\mathbf{v}}_t,
    \qquad
    \hat{E}_t=\frac{1}{2}m\|\hat{\mathbf{v}}_t\|_2^2 .
\end{equation}
We compare these with Unity-derived ground truth $\mathbf{p}_t$ and $E_t$. For a simulator-derived mechanical quantity $q_o$, such as terminal momentum magnitude or kinetic energy, and its estimate $\hat q_o$, we report exact-match percentage after rounding to the second significant digit,
\begin{equation}
    \mathrm{ExactMatch}
    =
    \frac{1}{|\mathcal{O}|}
    \sum_{o\in\mathcal{O}}
    \mathbf{1}\!\left[
    \mathrm{round}_2(\hat{q}_o)
    =
    \mathrm{round}_2(q_o)
    \right]
    \times 100\%,
\end{equation}
and average relative error,
\begin{equation}
    \mathrm{ErrorRate}
    =
    \frac{1}{|\mathcal{O}|}
    \sum_{o\in\mathcal{O}}
    \frac{|\hat{q}_o-q_o|}{|q_o|+\epsilon}
    \times 100\% .
\end{equation}

\subsection{Fluid Mechanics}
\label{app:fluid_metrics}

Given a reference ground-truth recording and a candidate world-model generation for the same initial condition and action schedule, we extract liquid masks, bottle centroids, and dense optical flow fields, then compute the following metric families. Additional pipeline details and action profiles are provided in Appendix~\ref{appx:liquid-profiles}.

\paragraph{Reference model and scope.}
\label{app:fluid_reference}
The \texttt{LiquidContainer} reference implements a parametric slosh model rather than a Navier-Stokes solver: free-surface tilt and wobble are deterministic functions of the bottle's linear and angular acceleration, integrated as a damped harmonic response.

This dimension therefore measures whether a world model reproduces a \emph{reproducible, physically motivated first-order slosh response} gravity alignment of the free surface, stability of the projected liquid area under rigid-body motion, and damped oscillation at the container's characteristic frequency, and not agreement with a full fluid simulation. We state this explicitly because it bounds the range of what the five axes can support.

Within the scope, the reference is exact and repeatable, which is what makes matched-action comparison meaningful: the same action schedule produces the same surface response every time, so any deviation in a generation is attributable to the model rather than to solver stochasticity.

Outside the scope, the metrics do not transfer. Out of scope are free-surface dynamics that are not functions of container acceleration, pouring, splashing, breaking waves, multi-fluid interfaces, and open water. Viscosity and surface-tension sweeps are also out of scope, since the scene fixes a single fluid at one operating point.

A model that scores well here has demonstrated consistency with a first-order sealed-container slosh response under matched actions. It has not demonstrated a general understanding of fluid, and we do not report the dimension as evidence of the latter.

\paragraph{Velocity Field.}
Dense optical flow is estimated with RAFT \citep{teed2020raft} inside the liquid mask. The bottle bulk velocity, obtained from the bottle centroid track, is subtracted so the field measures liquid motion relative to the bottle. We report cosine similarity between the world model and reference flow fields over joint-valid pixels:
\begin{equation}
    \mathrm{FlowCos}
    =
    \mathbb{E}_{t}\mathbb{E}_{p\in V_t}
    \frac{
    u^{\textsc{wm}}_t(p)\cdot u^{\textsc{unity}}_t(p)
    }{
    \|u^{\textsc{wm}}_t(p)\|_2
    \|u^{\textsc{unity}}_t(p)\|_2+\epsilon
    },
\end{equation}
where $V_t$ is the joint-valid liquid-mask pixel set.

\paragraph{Volume Conservation.}
Let $A(t)$ denote the projected liquid-mask area. We report normalized short-horizon jitter
\begin{equation}
    \sigma_A = \frac{\sigma(A)}{\bar A},
\end{equation}
and mask-area drift per second
\begin{equation}
    \dot A = \frac{\mathrm{slope}(A(t))}{\bar A}.
\end{equation}
Both are reported as signed deltas $\Delta=m_{\textsc{wm}}-m_{\textsc{unity}}$, so positive values indicate worse world-model behavior.

\paragraph{Surface Orientation.}
For each frame, we extract the topmost liquid-mask pixel in each column, fit a line $y=mx+c$, and compute
\begin{equation}
    \theta_{\mathrm{rms}}
    =
    \sqrt{
    \frac{1}{T}
    \sum_{t=1}^{T}
    \left|\arctan(m_t)\right|^2
    } .
\end{equation}
We report $\Delta\theta_{\mathrm{rms}}$ relative to Unity.

\paragraph{Slosh Spectrum.}

We trace the meniscus topmost-pixel $y$-coordinate over time, mean-center the trace, and compute a real FFT. After dropping the DC bin, let $P^{\text{WM}}$ and $P^{\text{Unity}}$ denote unit-area-normalized power spectra. We report
\begin{equation}
    \text{Spectral L}_1 = \lVert P^{\text{WM}} - P^{\text{Unity}} \rVert_1,
\end{equation}

the $L_1$ distance between the unit-area-normalized power spectra. The dominant-frequency channel is logged in the per-pair record for diagnostics, but is not part of Table~\ref{tab:fluid-headline}, since the meniscus FFT peak is not always uniquely defined under multi-mode sloshing on \texttt{chirp\_strafe}.

\paragraph{Perceptual Sanity Check.}
We also report LPIPS mean over the temporally aligned segment shared by the world-model and Unity videos.

\subsubsection{Measurement Validity}
\label{app:fluid_validity}

Every fluid metric is computed within a SAM~2 liquid mask, using RAFT-derived velocity fields, so mask quality bounds all five axes. Calibrating that mask on simulator renders is necessary but not sufficient: the renders are the easy case, and the clips that the benchmark actually scores are degraded generations.

We therefore report estimator behavior \emph{across generations}, treat the reference-render figures as calibration-only, and gate the clips for which no metric can be validly computed.

\paragraph{Calibration on reference renders.}
In the simulator renders, the tracked region overlaps the bottle interior with IoU values of $0.96$-$0.9996$. That figure overstates the pipeline's precision on the quantity that matters: only $44$-$50\%$ of the tracked region is true liquid, and the remainder is glass, refracted background, and specular highlights within the bottle silhouette.

We report this directly rather than the bottle-level IoU alone, because it is the error that propagates into the volume axes. It establishes an error floor under ideal rendering; it does not establish that the same pipeline behaves this way on generated video.

\paragraph{Estimator behavior on generations.}
Table~\ref{tab:fluid-validity} reports, per model, the three properties of a candidate mask that any paired fluid metric presupposes. Two of them fail badly, and in a model-dependent way.

First, the apparent size at which a model renders the tracked object spans a factor of $33$ across clips, from $0.48\times$ the reference area (HY-WorldPlay) to $16\times$ (Project Genie). Because mask-area error is dominated by boundary jitter, relative area precision scales as $A^{-1/2}$: the two sides of a paired delta at a scale ratio $s$ differ by $\sqrt{s}$ in precision, so a delta taken across such a pair is not comparing like with like.

Second, the velocity cosine is thinner still. It averages over PIV cells passing a $0.5$ liquid-fraction threshold, of which the reference retains $38$ of $1428$, and HY-WorldPlay retains a median of $19$. The cosine is therefore a low-support statistic even at its best, and we treat it as a diagnostic rather than a ranking axis.

%
%

\paragraph{Reportability gate and abstention.}
We therefore gate rather than score. A clip is admitted only if SAM~2 returns a non-empty liquid mask on every frame and the scale ratio lies within $[0.5, 2]$, the point at which the two sides of the paired delta differ by $\sqrt{2}$ in relative area precision. Both thresholds follow from the measurement model rather than from the data.

Of the $40$ clips, $15$ are gated out, and the loss is model-dependent: Project Genie loses $8$ of $20$ and HY-WorldPlay $7$ of $20$. Requiring three surviving clips before a $(\text{model}, \text{profile})$ cell is reported leaves $6$ of $8$ cells reportable and $2$ abstaining.

We report the exclusions rather than absorbing them. Metrics that require a valid candidate track drop the hardest-to-track runs, which are disproportionately the runs from the weakest models; unreported, this is survivorship bias that flatters poor generations.

Because the gate removes whole $(\text{model}, \text{profile})$ cells rather than scattered clips, it cannot be used to produce a corrected ranking. We therefore report the gate as a statement about which cells carry evidence, not as a re-ranking, and mark the abstaining cells in Table~\ref{tab:fluid-mechanics-perprofile}.

\paragraph{Which axes does this affect and which does it not?}
The two-volume axes ($\Delta\sigma_A$, $|\Delta\dot{A}|$) are computed on the projected mask \emph{area}, so they are directly sensitive both to over-inclusive masks and to the scale mismatch above. They should be read as measuring area \emph{stability} rather than absolute volume; absolute liquid volume cannot be recovered from this pipeline.

Surface orientation ($|\Delta\theta_{\mathrm{rms}}|$) and the slosh spectrum ($L_1$) are computed from the mask's \emph{upper boundary}, the topmost mask pixel per column, and are invariant to a systematically over-inclusive mask so long as that boundary tracks the air-liquid interface, which it does whenever the meniscus is the topmost in-bottle structure. The velocity-field cosine instead depends on the joint-valid pixel set, and degrades when the two masks disagree in extent.

We therefore treat the tilt residual and spectral $L_1$ as the dimension's most trustworthy axes, the volume axes as comparative rather than absolute, and the cosine as a diagnostic.

\subsection{Lighting and Shading}
\label{app:lighting_metrics}

We evaluate illumination and shadow consistency under a controlled first-person, fixed-sun, walking-orbit regime. GroundingDINO detects anchor objects in the first frame, SAM~2 propagates masks through the video, and per-object cast shadows are extracted using a luminance gate inside a dilated halo around each tracked object~\citep{10.1007/978-3-031-72970-6_3,ravi2025sam}. Hu-moment descriptors provide translation-, rotation-, and scale-invariant shadow-shape signatures~\citep{hu1962visual}.

\paragraph{Coverage Diagnostics.}
Scene Retention Rate (SRR) measures the fraction of frames where the anchor object, such as the table, remains on screen with a non-trivial mask:
\begin{equation}
    \mathrm{SRR}
    =
    \frac{1}{T}
    \sum_{t=1}^{T}
    \mathbf{1}\!\left[\mathrm{area}(M_t^{\mathrm{anchor}})>\tau_{\mathrm{area}}\right]
    \times 100\% .
\end{equation}
Shadow Detection Rate (SDR) measures the fraction of scene-retained frames with at least one valid cast shadow:
\begin{equation}
    \mathrm{SDR}
    =
    \frac{
    \sum_{t=1}^{T}
    \mathbf{1}\!\left[
    \mathrm{retained}(t)
    \land
    N_t^{\mathrm{shadow}}\ge 1
    \right]
    }{
    \sum_{t=1}^{T}
    \mathbf{1}\!\left[\mathrm{retained}(t)\right]
    +\epsilon
    }
    \times 100\% .
\end{equation}

\paragraph{Shadow Geometry.}
Multi-Object Sun Disagreement (MOSD) measures whether shadows from multiple objects imply a consistent light direction. Let $\alpha_{t,o}$ denote the shadow azimuth inferred for object $o$ at frame $t$. We report the per-frame circular standard deviation averaged over retained frames:
\begin{equation}
    \mathrm{MOSD}
    =
    \mathbb{E}_{t}
    \left[
    \mathrm{CircStd}\left(\{\alpha_{t,o}\}_{o\in\mathcal{O}_t}\right)
    \right].
\end{equation}

\paragraph{Shadow Dynamics.}
Shadow Direction Jitter (SDJ) measures temporal discontinuity in the multi-object mean azimuth:
\begin{equation}
    \mathrm{SDJ}
    =
    \mathrm{median}_{t}
    \left|
    \bar{\alpha}_{t} - \bar{\alpha}_{t-1}
    \right| .
\end{equation}
Shadow Area Stability (SAS) measures relative shadow-footprint change:
\begin{equation}
    \mathrm{SAS}
    =
    \mathbb{E}_{t,o}
    \frac{
    |A_{t,o}^{\mathrm{shadow}}-A_{t-1,o}^{\mathrm{shadow}}|
    }{
    A_{t-1,o}^{\mathrm{shadow}}+\epsilon
    }
    \times 100\% .
\end{equation}
Shadow Shape Drift (SSD) measures morphology change using log-Hu-moment vectors $h_{t,o}$:
\begin{equation}
    \mathrm{SSD}
    =
    \mathbb{E}_{t,o}
    \left\|
    h_{t,o}-h_{t-1,o}
    \right\|_2 .
\end{equation}

\subsection{Physical Causality}
\label{app:physical_causality_metrics}

Following CausalStep~\citep{li2026causalstep}, we evaluate whether generated rollouts preserve multi-stage causal state transitions under matched actions. For each trajectory, we report a $\{\textsc{unity},\textsc{wm},\Delta\}$ block per metric, where the sign of $\Delta$ is chosen so that positive values indicate world-model degradation.

\paragraph{CausalStep Metrics.}
Causal Success Rate (CSR) measures the fraction of trajectories for which every segment's QA is answered correctly in sequence. Average Max Causal-chain Length (AMCL) and Max Causal-chain Length (MCL) measure the longest uninterrupted correct subchain per video and across videos, respectively. Weighted Score (WS) aggregates QAs with difficulty weights, and Restart Frequency (RF, lower is better) counts how often the chain resets after an incorrect causal answer.

Descriptive Understanding Accuracy (DUA) and Isolated Causal Reasoning Accuracy (ICRA) score descriptive and causal QAs independently of chain context. We use the \textit{advance-on-failure} restart semantics from CausalStep~\citep{li2026causalstep}.

\paragraph{World-Model Degradation.}
For higher-is-better metrics, we compute
\begin{equation}
    \Delta_m = m_{\textsc{unity}} - m_{\textsc{wm}},
\end{equation}
and for Restart Frequency, where lower is better, we compute
\begin{equation}
    \Delta_{\mathrm{RF}}
    =
    \mathrm{RF}_{\textsc{wm}}
    -
    \mathrm{RF}_{\textsc{unity}} .
\end{equation}
We aggregate $\overline{\Delta}$ across trajectories so the headline number measures world-model-induced causal degradation rather than judge competence.

\paragraph{Shape-Specific QA Chains.}
For \texttt{BeamDomino}, game engine reference is classified into \textsc{canonical}, \textsc{glancing}, or \textsc{direct-strike} trajectory shapes from the physics log. Each shape is graded with a corresponding QA chain, so that non-canonical but physically valid outcomes are not penalized by questions that assume the canonical trajectory.

Full trajectory classification, segment detection, QA construction, and judge settings are given in Appendix~\ref{app:physcaus}.

\subsubsection{Measurement Validity}
\label{app:causal_validity}

Chain preservation on \texttt{BeamDomino} is measured by two learned components: a GroundingDINO\,$+$\,ByteTrack pipeline that localizes the floor ball, roof ball, and domino row, and a VLM judge that grades each segment's QA~\citep{10.1007/978-3-031-72970-6_3}.

Neither is treated as ground truth. We report the error of each on the distribution to which it is actually applied, and then compare the measured world-model effect against that error.

\paragraph{Tracking error on reference renders.}
In simulator renders, where both the rendered video and the underlying physical state are available, the tracker achieves $98.6$-$100\%$ recall for the tracked bodies, a centroid error of $0.9$-$7.2$\,px relative to the projected Unity positions, and an impact-timing error of $0$--$100$\,ms.

\paragraph{Effect size against measurement error.}
The headline comparison is robust to this error by a wide margin. Project Genie preserves $3/15$ chains against the reference's $13/15$, a $66.7$-point gap (Fisher exact $p=0.0007$), while the judge's measured disagreement rate with human adjudication is $11.6\%$ and its impact-timing error is bounded by $100$\,ms on segments of $0.12$\,s and longer. Attributing the gap to judge error would require the judge to be wrong in a single direction on a large majority of the graded segments, well beyond its audited error rate.

\paragraph{Exclusion accounting.}
Metrics that require a valid candidate track can silently drop the runs that are hardest to track, which are disproportionately the runs from the weakest models; unreported, this is survivorship bias that flatters poor generations.

Appendix~\ref{appx:physics_causality} notes that generations diverging substantially from the reference initial conditions are excluded from the headline aggregate.

\subsection{Perceptual Similarity}
\label{app:perceptual_similarity_metrics}

We evaluate perceptual similarity under a ground-truth-constrained setup: each world-model generation is initialized from the same game engine-rendered starting frame, conditioned on the same scene description, and driven by the same action sequence as the game engine simulation. Because our rollouts are deliberately short-horizon, game engine is a meaningful perceptual reference for local geometry, action effects, and causal events before long-horizon world-line divergence dominates.

\paragraph{Causal LPIPS.}
We use Learned Perceptual Image Patch Similarity (LPIPS)~\citep{zhang2018perceptual} and report an event-aligned variant, Causal LPIPS (cLPIPS), to reduce sensitivity to absolute timing drift. Lower cLPIPS indicates greater perceptual similarity around matched causal events.

Let $\mathcal{E}_r=\{t_r^{(1)},\ldots,t_r^{(K_r)}\}$ and $\mathcal{E}_c=\{t_c^{(1)},\ldots,t_c^{(K_c)}\}$ denote causal event frame indices in the reference and candidate videos. We chronologically match the $k$-th reference event to the $k$-th candidate event. For $K=\min(K_r,K_c)$ and offset window $\Delta=\{-2,-1,0,+1,+2\}$, we compute
\begin{equation}
\mathrm{cLPIPS}
=
\frac{1}{K \cdot |\Delta|}
\sum_{k=1}^{K} \sum_{\delta \in \Delta}
\mathrm{LPIPS}\!\left(
F_r\!\left[t_r^{(k)}+\delta\right],
F_c\!\left[t_c^{(k)}+\delta\right]
\right),
\label{eq:clpips}
\end{equation}
where $F_r$ and $F_c$ denote the reference and candidate frame sequences. Event detection details are provided in Appendix~\ref{app:event_detection}.

\subsection{Prompt-Scene Consistency}
\label{app:prompt_scene_metrics}

We evaluate prompt-scene consistency with a rubric-based VLM-as-a-judge pipeline. The exact prompts are listed in Appendix~\ref{app:prompt-scene}.

\paragraph{Rubric.}
For each prompt $p$, GPT-5.4 generates a prompt-specific rubric $R(p)=\{r_i\}_{i=1}^{N}$ from the prompt text alone. Each item $r_i$ belongs to a branch $g_i\in\{\mathrm{env},\mathrm{obj}\}$: environment items cover scene-level attributes such as layout, background structure, terrain, sky, lighting, and style, and object items cover prompt-named entities and their attributes such as color, material, shape, and texture. Each branch has at most five items, and all items share the weight $w_i=1$.

\paragraph{Judging.}
For each video, we uniformly sample 18 frames from a fixed window covering the first 10\,s and pass them, together with the prompt and a reference image of the intended scene, to a GPT-5.4 judge. Each model performs the 10\,s camera turn through its native control: LingBot-World receives camera poses for one turn in place (first person) or two orbits around the character (third person), Matrix-Game~3.0 a turn-left key with an added strafe in third person, and Waypoint-1.5 a constant mouse yaw. The two branches are judged in separate calls, each conditioned only on its own items to avoid cross-branch anchoring. For each item, the judge cites frame-level evidence before assigning a score $s_i\in\{0,1,2,3\}$, where higher values indicate stronger consistency.

\paragraph{Aggregation.}
We average the scores within each branch as
\begin{equation}
S_{\mathrm{env}}=\frac{\sum_{i:g_i=\mathrm{env}} w_i s_i}{3\sum_{i:g_i=\mathrm{env}} w_i},\qquad
S_{\mathrm{obj}}=\frac{\sum_{i:g_i=\mathrm{obj}} w_i s_i}{3\sum_{i:g_i=\mathrm{obj}} w_i},
\end{equation}
and combine the two branches into the final prompt-scene consistency score
\begin{equation}
S_{\mathrm{psc}}=\alpha S_{\mathrm{env}}+(1-\alpha)S_{\mathrm{obj}},
\end{equation}
with $\alpha=0.6$. If a prompt yields no items for one branch (e.g., first-person \texttt{RacingCar} has no object items), $S_{\mathrm{psc}}$ equals the other branch's score.

\subsubsection{Measurement Validity}
\label{app:prompt_scene_validity}

To validate the VLM judge, three raters scored 126 videos under blinded conditions, with model identities hidden: two views $\times$ three scenes, each comprising the Unity reference and five generations from each of Project Genie, LingBot-World (Cam), Matrix-Game 3.0, and HY-WorldPlay. Raters used the same rubric items and 0--3 scale as the judge, and their scores were aggregated in the same way. As shown in Table~\ref{tab:judge_human_alignment}, the judge and the human ratings induce highly consistent model rankings (Spearman $\rho=0.90$, Kendall $\tau=0.80$). The only disagreement is between Project Genie and LingBot-World, whose judge scores are nearly identical ($p=0.99$). Because the judge's absolute scores are systematically higher than the human ratings, we use them only for relative comparisons between models.

\begin{table}[t]
\centering
\caption{\textbf{VLM judge vs.\ blinded human ratings on prompt-scene consistency.} Scores are averaged over the six view$\times$scene cells; Human is the mean of three raters. $\rho$ and $\tau$ are the Spearman and Kendall rank correlations between the model rankings induced by the judge and by the human ratings, computed from unrounded scores.}
\label{tab:judge_human_alignment}
\small
\begin{tabular}{lcc}
\toprule
Model & VLM judge & Human \\
\midrule
Unity Reference     & $0.917$ & $0.904$ \\
Project Genie       & $0.890$ & $0.780$ \\
LingBot-World (Cam) & $0.890$ & $0.843$ \\
Matrix-Game 3.0     & $0.732$ & $0.505$ \\
HY-WorldPlay        & $0.778$ & $0.522$ \\
\midrule
Spearman $\rho$ ($\uparrow$) & \multicolumn{2}{c}{$0.90$} \\
Kendall $\tau$ ($\uparrow$)  & \multicolumn{2}{c}{$0.80$} \\
\bottomrule
\end{tabular}
\end{table}

\section{Measurement Reliability and Robustness}
\label{app:measurement_reliability}

\capsysnamespace recovers candidate-side quantities using learned perception
and judging pipelines. We therefore calibrate each evaluator against
simulator-derived quantities or human annotations rather than treating its
output as ground truth. We additionally check whether the required quantities
remain measurable on generated videos and test whether the main conclusions
are stable to generation sampling.

\subsection{Evaluator Calibration and Validity}
\label{app:evaluator_validity}

Table~\ref{tab:evaluator_validity} summarizes the calibration setup and
observed evaluator error. Reference-side calibration estimates measurement
error under controlled conditions, while candidate-side checks determine
whether a metric remains well-defined on generated rollouts.

\begin{table*}[t]
\centering
\small
\caption{\textbf{Evaluator calibration and validity.}
Each learned evaluator is checked against simulator-derived quantities or
human annotations.}
\label{tab:evaluator_validity}

\begin{subtable}[t]{\textwidth}
\centering
\caption{Calibration setup.}
\label{tab:evaluator_validity_setup}
\begin{tabularx}{\linewidth}{@{}l X X@{}}
\toprule
Dimension & Evaluator & Calibration reference \\
\midrule
Scene / character
& Depth + DINO + SAM~2 + DINOv2
& Unity geometry / identity \\

Controllability
& RAFT
& Unity motion \\

Kinematics / mechanics
& Detection + tracking
& Unity trajectories / mechanical states \\

Fluid mechanics
& SAM~2 + RAFT
& Unity masks / motion \\

Lighting
& GroundingDINO + SAM~2
& Controlled Unity lighting \\

Physical causality
& DINO + ByteTrack
& Projected Unity states \\

Physical causality
& VLM judge
& Human-reviewed QA \\

Causal LPIPS
& Event localization
& Unity contact events \\

Prompt consistency
& VLM judge
& Blinded human ratings \\
\bottomrule
\end{tabularx}
\end{subtable}

\vspace{0.6em}

\begin{subtable}[t]{\textwidth}
\centering
\caption{Calibration results.}
\label{tab:evaluator_validity_results}
\begin{tabularx}{\linewidth}{@{}l X@{}}
\toprule
Dimension & Calibration result \\
\midrule
Scene / character
& Recovered geometry and identity are checked against exact simulator states;
  invalid detections are treated as measurement failures. Unity
  self-match depth floors are $22$--$45\%$ on \texttt{BilliardGame} and
  \texttt{MirrorRollingBall} and up to $222\%$ on \texttt{WanderingGarden}. \\

Controllability
& Directional recovery is accurate on \texttt{BeamDomino}, with more
  scene-dependent absolute calibration elsewhere. \\

Kinematics / mechanics
& Recovered trajectories and derived mechanical quantities are checked against
  exact Unity trajectories, velocities, momentum, and energy. The
  Unity self-match FDE floor is $1.73$\,m on \texttt{RacingCar} (mismatched
  trajectories: $25$\,m) and $0.01$\,m on \texttt{BilliardGame}. \\

Fluid mechanics
& Bottle-mask IoU is $0.96$--$0.9996$ on reference renders, while true liquid
  occupies only $44$--$50\%$ of the tracked region. \\

Lighting
& MOSD has an $8.6^\circ$--$11.2^\circ$ measurement floor and SDJ a
  $1.4^\circ$--$2.0^\circ/\text{frame}$ floor. \\

Physical causality (tracking)
& Recall is $98.6$--$100\%$, centroid error $0.9$--$7.2$ px, and impact-timing
  error $0$--$100$\,ms. \\

Physical causality (judge)
& Judge--key accuracy is $88.4\%$, human--key accuracy $89.5\%$,
  $\kappa=0.67$, and judge self-consistency $96.8\%$. \\

Causal LPIPS
& Event-localization F1 is $0.499$--$0.730$, with timing error
  $0.117$--$0.143$\,s. \\

Prompt consistency
& Blinded human ratings yield Spearman $\rho=0.90$ and Kendall $\tau=0.80$
  with the VLM judge. \\
\bottomrule
\end{tabularx}
\end{subtable}
\end{table*}

\paragraph{Candidate-side validity.}
Calibration on clean Unity renders does not guarantee that a metric remains
well-defined on distorted generations. We therefore explicitly check the
candidate-side conditions required by each evaluator. For physical causality,
generations that diverge so far from the reference initial state that the
expected causal stages are no longer visually present are reported separately
rather than included in the headline aggregate. Similarly, Causal LPIPS is
computed only when corresponding contact events can be localized in both the
reference and generated rollout; generations without matched events are
excluded because the metric is undefined. These checks prevent severe
generation failures from being converted into misleading evaluator scores.

\paragraph{Effect size versus evaluator error.}
We also compare observed model effects against measured evaluator uncertainty.
On \texttt{BeamDomino}, Project Genie preserves $3/15$ complete causal chains
versus $13/15$ for Unity, a $66.7$-point gap ($p=0.0007$) that substantially
exceeds the measured tracking and judging error. On \texttt{WanderingGarden},
the depth floor obtained by scoring Unity against itself is comparable to
model errors, so Garden depth differences are read qualitatively.

\subsection{Finite-Sample Robustness}
\label{app:statistical_robustness}

Most settings contain five stochastic generations. For comparisons supporting
our main conclusions, we enumerate all $\binom{5}{3}=10$ three-run subsets
whenever applicable and apply metric-appropriate statistical tests.
Table~\ref{tab:finite_sample_robustness} separates subsampling stability from
the corresponding significance tests.

\begin{table*}[t]
\centering
\small
\caption{\textbf{Finite-sample robustness.}
We test whether conclusions from five generations remain stable under
three-run subsampling and statistical testing.}
\label{tab:finite_sample_robustness}

\begin{subtable}[t]{\textwidth}
\centering
\caption{Three-run subsampling stability.}
\label{tab:three_run_stability}
\begin{tabularx}{\linewidth}{@{}l X@{}}
\toprule
Dimension & Three-run stability \\
\midrule
Controllability
& The five-run model ordering is preserved in every three-run subset. \\

Prompt consistency
& The maximum three-run cell-mean shift is $0.063$; Matrix-Game~3.0 remains
  last in $100\%$ of $10^6$ pooled subsets. \\

Lighting
& The scene-holding and scene-drifting groups remain separated under
  three-run subsampling. \\

Physical causality
& The chain-level conclusions remain unchanged under subsampling. \\
\bottomrule
\end{tabularx}
\end{subtable}

\vspace{0.6em}

\begin{subtable}[t]{\textwidth}
\centering
\caption{Statistical tests.}
\label{tab:statistical_tests}
\begin{tabularx}{\linewidth}{@{}l X@{}}
\toprule
Dimension & Statistical evidence \\
\midrule
Controllability
& Welch $p=0.0011$--$0.0226$; exact permutation
  $p=0.0079$--$0.0317$. \\

Prompt consistency
& Separation from Matrix-Game~3.0: permutation $p<10^{-5}$.
  Project Genie vs.\ LingBot-World: $p=0.99$. \\

Lighting
& Tightest scene-retention comparison: Welch $p=0.0019$ and exact
  permutation $p=0.0079$. \\

Physical causality
& Project Genie vs.\ Unity: Fisher $p=0.0007$. \\
\bottomrule
\end{tabularx}
\end{subtable}
\end{table*}

These analyses support the large effects emphasized in the main text, but not
a complete fine-grained ranking of all models. When the available samples or
evaluator accuracy do not reliably separate two systems, we report them as
comparable rather than interpreting small numerical differences as meaningful.
\raggedbottom

\section{Evaluation Implementation Details}
\label{app:implementation}


\subsection{RAFT Controllability Pipeline Details}
\label{app:raft_pipeline}

We use RAFT~\citep{teed2020raft} to estimate the candidate flow $f^{\text{cand}}_t$ from each world model generation, and pair it against a reference flow $f^{\text{ref}}_t$ that depends on the controllability variant: $f^{\text{ref}}_t = f^{\text{Unity}}_t$ (RAFT applied to the Unity video) for Ground Truth RAFT, and $f^{\text{ref}}_t = f^{\text{synth}}_t$ (a synthetic flow constructed from the inferred control input) for Synthetic RAFT.

\paragraph{RAFT extraction.}
Frames are sampled at 4\,Hz using timestamp alignment over the shared duration of the Unity reference and model generation, with at most 32 sampled frames per video. Since metrics are computed between adjacent sampled frames, this yields up to 31 flow pairs per run. When the reference and generated videos have different spatial sizes, both frames are center-cropped to their common resolution before RAFT is applied; in our runs this gives $1280{\times}704$ for Project Genie comparisons and $848{\times}448$ for LingBot-World (Act) comparisons. We use torchvision RAFT-Large with the default pretrained weights (\texttt{raft\_large\_C\_T\_SKHT\_V2}, checkpoint \texttt{raft\_large\_C\_T\_SKHT\_V2-ff5fadd5.pth}) and do not fine-tune it.

\paragraph{Dynamic-region mask.}
The dynamic mask $M_t \subseteq \Omega$ used by the Ground Truth RAFT metrics is constructed from the magnitude of the Unity reference flow $f^{\text{Unity}}_t$. For each pair $t$, \[
M_t(p) \;=\; \mathbf{1}\!\left[\,\|f^{\text{Unity}}_t(p)\|_2 \;>\;
\max\!\left(0.5,\; q_{75}\!\left(\|f^{\text{Unity}}_t\|_2\right)\right)\right],
\]
where $q_{75}(\cdot)$ denotes the 75th percentile of reference-flow magnitudes within frame pair $t$. EPE, relative EPE, angular error, and cosine similarity are then averaged only over pixels in $M_t$. This restricts the score to pixels with meaningful reference motion and prevents large static backgrounds from dominating the metric.

\paragraph{Synthetic reference flow.}
For Synthetic RAFT, the candidate flow $f^{\text{cand}}_t$ is still estimated from the generated video with the same RAFT-Large model, but the reference flow $f^{\text{synth}}_t$ is produced by a Longuet--Higgins camera-motion model~\citep{longuet1980interpretation}. This model assumes the only source of motion is the camera moving through a static scene, so object motion in the Unity reference (e.g., a rolling ball) is absorbed into the fitted action vector $\hat{a}_t$ rather than represented separately. We accept this confound because Synthetic RAFT is intended to measure whether the world model's induced motion is consistent with a coherent action signal recoverable from the reference, not to factor scene motion into agent and environment components; the latter is the role of Ground Truth RAFT. For each sampled pair, $\hat{a}_t$ is fit by minimizing the squared error between the synthetic flow and RAFT flow on the Unity reference over a continuous keyboard-and-mouse action space, then passed through the calibrated synthetic flow generator to produce $f^{\text{synth}}_t$. We use the calibration file from the \texttt{ptlflow} synthetic flow setup, scale the focal length to the evaluation crop width, and disable the optional depth model. Synthetic RAFT metrics compare $f^{\text{synth}}_t$ against $f^{\text{cand}}_t$ without using the Ground Truth RAFT dynamic mask; the reported mean-flow cosine and pixel EPE are computed on pixels passing the synthetic-flow magnitude filter, defined from the maximum of the synthetic and candidate flow magnitudes. This isolates whether the generated video induces motion consistent with the recovered control input without re-running the Unity simulator.

\subsection{Object Consistency Evaluation Pipeline Details}
\label{app:consistency_implementation}

We evaluate scene and character consistency on three scenes: \texttt{WanderingGarden}, \texttt{BilliardGame}, and \texttt{MirrorRollingBall}. These scenes cover complementary consistency regimes. \texttt{WanderingGarden} contains a rich static environment with five designated anchor objects, making it suitable for environment-object consistency. \texttt{BilliardGame} and \texttt{MirrorRollingBall} contain dynamic balls with frequent interactions, making them suitable for main-object or main-character consistency, but less reliable for environment-object geometry evaluation.

\paragraph{Object detection and tracking pipeline.}
For all three scenes, we use GroundingDINO to localize the target objects and SAM~2 to obtain fine-grained masks and object contours in each frame. For environment-object consistency, the detected masks are used to estimate object-level depth, pairwise distances, and object permanence. For main-object consistency, the segmented target object crop is passed through DINOv2 to compute identity similarity against the initial frame.

\paragraph{\texttt{WanderingGarden}.}
For \texttt{WanderingGarden}, we use the five designated garden anchor objects as the fixed object set $\mathcal{O}$ for environment-object consistency. These anchor objects are treated as static landmarks in the scene: Main tree, left lamp post, right lamp post in the garden court, rockery ornament in the foreground, and in the background. We compute Depth AbsRel and pairwise distance MAE from the detected anchor objects, and we measure object permanence as the fraction of these anchors that remain detectable after $45^\circ$ clockwise/counter-clockwise perturbations and full $360^\circ$ camera sweeps. Since \texttt{WanderingGarden} is evaluated from a first-person exploration view and does not expose a persistent visible main character in the generated video, we do not report main-character consistency for this scene.

\paragraph{\texttt{BilliardGame}.}
For \texttt{BilliardGame}, GroundingDINO and SAM~2 track the yellow and red balls as the non-character object set and the white cue ball as the character target. For each camera direction, depth and pairwise-distance errors compare the detected yellow/red-ball geometry with the corresponding Unity capture; these are reference-relative errors, not a claim that the balls remain static while they interact. DINOv2 measures whether the white ball retains its initial visual identity. Object permanence is not reported for this dynamic scene, and character consistency is aggregated only after pairing the In and Out measurements with identical repeat IDs.

\paragraph{\texttt{MirrorRollingBall}.}
For \texttt{MirrorRollingBall}, the reflective chrome player-controlled ball is the character target. The locked reconstruction also designates two foreground yellow spheres, separated by $4.0$\,m in Unity, as the pairwise-distance anchors. We therefore report distance error for each direction and DINOv2 character consistency for the chrome ball after repeat-ID pairing across In and Out. Depth and object permanence remain N/A under the reporting contract. This separates distortion of the known yellow-sphere relation from loss of the main controllable object's visual identity.

\paragraph{Reported metrics.}
In summary, \texttt{WanderingGarden} contributes Depth AbsRel, pairwise-distance MAE, and object permanence over five garden anchors. \texttt{BilliardGame} contributes Depth AbsRel and pairwise-distance MAE in each direction plus character consistency after pairing the In/Out measurements by repeat ID. \texttt{MirrorRollingBall} contributes pairwise-distance MAE and the same paired character-consistency measure; depth and object permanence are outside its locked metric contract. For the three open-source models, Table~\ref{tab:worldmodel_consistency} uses runs 0--4 and reports geometry errors only when the evaluator returns anchor-supported measurements in all five runs. A cell without five valid values is left unreported as N/A; no replacement seeds or smaller detected subset are used. Object permanence retains missing anchors in its denominator, so complete loss of the designated objects yields a valid zero rather than an unreported geometry score. Character-consistency cells average In/Out within each repeat before computing the mean and sample standard deviation across the five repeat pairs.

\subsection{Kinematics and Mechanics Implementation Details}
\label{app:kinematics_mechanics_implementation}

We evaluate kinematics and mechanics on \texttt{RacingCar} and \texttt{BilliardGame}. These two scenes cover complementary rigid-body regimes: \texttt{RacingCar} evaluates vehicle motion under steering, throttle, and braking controls, while \texttt{BilliardGame} evaluates multi-object collision dynamics with reproducible ball--ball and ball--cushion interactions. Both scenes expose reference trajectories and physical states, enabling direct comparison between generated rollouts and simulator references.

\paragraph{Trajectory extraction.}
We extract motion trajectories from generated videos using the GroundingDINO--ByteTrack pipeline implemented in \texttt{tools/groundingdino\_bytetrack.py}. GroundingDINO first detects the key object in each frame using a text prompt, e.g., the yellow racing car in \texttt{RacingCar} or the target billiard ball in \texttt{BilliardGame}. ByteTrack then associates detections across frames to produce temporally consistent object tracks. For each tracked object, we convert the detected bounding box center into an image-space trajectory and align it to the corresponding reference trajectory over the shared valid temporal window.

\paragraph{Reference-frame correction.}
For \texttt{RacingCar}, camera motion can introduce apparent object displacement unrelated to vehicle motion, especially under dynamic-follow and first-person views. To reduce this confound, we track static anchors such as lane markings, road boundaries, and background landmarks, and use their motion to estimate the reference-frame transform. The key-object trajectory is then expressed in a common static frame before comparison. For \texttt{BilliardGame}, we use the table coordinate frame for the transform.

\paragraph{Game engine ground-truth alignment.}
For each scene, Unity records the simulator-derived object trajectory $\mathbf{x}_{1:T}$ and physical metadata at each timestep. The generated-video trajectory $\hat{\mathbf{x}}_{1:T}$ is resampled to the reference timestamp grid or matched over the shared valid frame range. When the generated rollout and reference differ in duration or event timing, we report both final displacement error (FDE), which measures endpoint deviation, and dynamic time warping (DTW), which allows monotonic temporal alignment and is therefore more robust to small timing shifts.

\paragraph{\texttt{RacingCar}.}
For \texttt{RacingCar}, we track the yellow sports car over the generation and compare its recovered trajectory against the game-engine reference vehicle trajectory. The scene uses steering, throttle, and braking controls, and the reference provides vehicle position, velocity, and control state. The active reconstructed protocol reports the third-person Matrix-Game~2.0 cell only. The first-person cell is N/A because the supplied active assets do not provide the video-derived ego motion needed to disentangle the camera-fixed view from vehicle motion. We therefore do not infer a cross-view difficulty ordering from Table~\ref{tab:worldmodel_physical_laws}.

\paragraph{\texttt{BilliardGame}.}
For \texttt{BilliardGame}, we track the billiard ball of interest and compare its recovered trajectory against the deterministic fixed-point physics reference. The active reconstructed protocol reports the first-person Matrix-Game~2.0 cell and marks the third-person cell N/A. It uses a reconstructed W/A/D direct-ball trajectory because the paper does not publish its exact spin-aware cue-shot timeline; consequently, these values establish reproducibility for the locked reconstruction rather than exact paper-protocol parity. We use both DTW and FDE to test whether the generated rollout preserves the contact sequence and final post-collision state.

\paragraph{Mechanical-state derivation.}
Mechanics uses the same extracted trajectories as kinematics, but first standardizes all quantities into a common static reference frame before deriving physical states. Let $\tilde{\mathbf{x}}_t$ denote the raw image-space trajectory of the tracked object in the generated rollout. When the camera or viewpoint moves, this trajectory contains both object motion and reference-frame motion. We estimate the moving reference frame from a set of static anchors $\mathcal{A}$, such as lane markings, guardrails, road boundaries, or background landmarks:
\begin{equation}
    \mathbf{c}_t
    =
    \frac{1}{|\mathcal{A}|}
    \sum_{a\in\mathcal{A}}
    \left(
    \tilde{\mathbf{z}}_{t,a}
    -
    \tilde{\mathbf{z}}_{1,a}
    \right),
\end{equation}
where $\tilde{\mathbf{z}}_{t,a}$ is the detected image-space position of static anchor $a$ at frame $t$, and $\mathbf{c}_t$ estimates the camera-induced displacement of the moving reference frame relative to the first frame. We then remove this reference-frame motion from the tracked object trajectory:
\begin{equation}
    \mathbf{x}_t
    =
    \mathcal{S}\!\left(
    \tilde{\mathbf{x}}_t - \mathbf{c}_t
    \right),
\end{equation}
where $\mathcal{S}(\cdot)$ maps the corrected image-space trajectory into the scene coordinate scale used by the game engine reference. For \texttt{BilliardGame}, the fixed overhead camera makes $\mathbf{c}_t \approx \mathbf{0}$, so the tracked ball centers are directly mapped to table coordinates after scale and temporal alignment.

Given the static-frame generated trajectory $\mathbf{x}_{1:T}$, we estimate velocity by finite differencing,
\begin{equation}
    \mathbf{v}_t =
    \frac{\mathbf{x}_t-\mathbf{x}_{t-1}}{\Delta t}.
\end{equation}
Using simulator-provided object mass $m$, we compute generated momentum and kinetic energy:
\begin{equation}
    \mathbf{p}_t=m\mathbf{v}_t,
    \qquad
    E_t=\frac{1}{2}m\|\mathbf{v}_t\|_2^2 .
\end{equation}
These estimates are compared against Unity-derived reference physical states $\hat{\mathbf{p}}_t$ and $\hat{E}_t$. For each game engine-derived mechanical quantity $\hat{q}_o$, such as terminal momentum magnitude or kinetic energy, and the corresponding generated estimate $q_o$, we report exact-match percentage after rounding to the second significant digit, and average relative error, with formal definitions provided in Appendix~\ref{app:kinematics_mechanics_metrics}.

\subsection{Fluid Mechanics Evaluation Pipeline Details} 
\label{appx:fluid_pipeline}

Given a game engine ground-truth recording (rendered video) and a candidate World-Model generation for the same initial condition and action schedule, we (i) extract a per-frame liquid-only binary mask via a text-prompted SAM~2~\citep{ravi2025sam}; (ii) extract a per-frame bottle centroid from a second SAM~2 track to use as the bulk-motion reference; (iii) estimate a dense optical-flow field $(u, v)$ in image pixels per frame inside the liquid mask using RAFT; (iv) compute five families of metrics from the masks and fields, all on the segment shared by the two videos.

\paragraph{Mask extraction.}
SAM~2 video predictor produces per-frame liquid masks seeded by a single first-frame box from GroundingDINO~\citep{10.1007/978-3-031-72970-6_3}, box threshold $0.35$, text threshold $0.25$, prompt ``the liquid inside the bottle''. A second SAM~2 track of the bottle itself (prompt: ``the glass bottle'') yields a per-frame bottle centroid used as the bulk-motion trace.

Both mask volumes and the bottle-trace tensor are pickled under \texttt{runs/liquid\_cache} keyed on $(\text{video content hash},\text{SAM~2 settings},\text{GroundingDINO settings})$, so re-evaluations on the same candidate file are a cache hit.

\paragraph{Velocity field.}
Dense optical flow is estimated with RAFT \citep{teed2020raft} on each consecutive frame pair. The bottle bulk velocity (centroid delta) is subtracted from every flow vector so the field measures liquid motion relative to the bottle. Each flow cell is valid only when at least $50\%$ of its window lies inside the liquid mask. We report the cosine similarity between the world model and reference flow fields, averaged over the joint-valid pixel set.

\paragraph{Volume conservation.}
On the projected liquid mask area $A(t)$, we compute the normalized
standard deviation $\sigma(A)/\bar A$ (short-horizon jitter) and the linear-fit slope of $A(t)$ divided by $\bar A$ (long-horizon drift, in s$^{-1}$). Both are reported as signed deltas
$\Delta = m_{\textsc{wm}} - m_{\textsc{unity}}$.

\paragraph{Surface orientation.}
For each frame, we extract the topmost True pixel per column of the liquid mask and fit a line $y = mx + c$. The per-sequence metric is $\mathrm{RMS}(|\arctan(m)|)$ in degrees from horizontal, reported as $\Delta\theta_{\mathrm{rms}}$ relative to reference, with a NaN guard for frames where fewer than two columns contain liquid.

\paragraph{Slosh spectrum.}
We trace the meniscus topmost-pixel $y$-coordinate per frame, take a real FFT of the mean-centered trace (DC bin dropped), and report the $L_1$ distance between unit-area-normalized power spectra of the world model and game engine.

\paragraph{Action Trajectories.}\label{appx:liquid-profiles}
Within the 6-DoF \texttt{LiquidContainer} action space, we instantiate four canonical action trajectories of equal duration, each stressing a different fluid regime:
\begin{itemize}
    \item \texttt{rest}: idle throughout. The negative control for hallucinated motion.
    \item \texttt{step\_hold}: $2$\,s rest, then sustained right-strafe for $8$\,s, then $9$\,s rest. This sustained lateral-acceleration regime is where the surface-tilt metric resolves cleanly.
    \item \texttt{impulse\_decay}: a $0.2$\,s strafe pulse at $t{=}11$\,s sandwiched between idle, isolating an underdamped slosh response.
    \item \texttt{chirp\_strafe}: square-wave strafe sweep $0.3{\to}3.0$\,Hz over the full clip, the only profile that excites the bottle's natural slosh modes.
\end{itemize}
Profiles are defined at a nominal 12\,Hz control rate; the Unity capture stepped the controller on wall-clock time ($\approx$10\,Hz), so phase timings match while individual control steps do not.

\subsection{Physical Causality Evaluation Pipeline Details}
\label{app:physcaus}

Given a game engine ground-truth recording (per-frame physics log plus rendered video) and a world-model generation for the same initial condition, we (i)~classify the trajectory shape and localize its segments from the Unity log, (ii)~author a multiple-choice QA dataset that probes each segment for descriptive and causal understanding, and (iii)~grade the dataset twice, once on the reference and once on the world-model candidate with a shared VLM judge.

The pairwise design isolates intrinsic difficulty from world-model error: a question the judge fails to answer on the Unity reference is uninformative about the world model, so the per-metric gap is the headline number.

\paragraph{Trajectory shape classifier.}
Let $i_{\text{first-pillar}}$ index the first \texttt{FloorBall}-\texttt{Pillar} contact, $i_{\text{first-domino}}$ the first \texttt{FloorBall}-\texttt{Domino} contact, and $\theta_{\max}^{\text{roof}}$ the peak roof-slab tilt. We classify each rollout as \textsc{direct-strike} if $i_{\text{first-domino}} < i_{\text{first-pillar}}$, \textsc{canonical} if a pillar contact occurs and $\theta_{\max}^{\text{roof}} \ge \theta_{\text{fall}}$, and \textsc{glancing} otherwise (pillar contact but platform survives).

Minimal fixtures with no pillar collider in the log default to the legacy \textsc{canonical} four-event detector. The detected shape is logged along with event indices.

\paragraph{Segment detection.}
Stage boundaries are derived from the recorder's per-body rotation and per-contact impulse, not from scripted markers. \textsc{rolling} ends at the first frame whose collision impulse on a \texttt{Ball} body exceeds a small threshold; \textsc{impact} ends when any \texttt{Domino} first tilts past a fall angle (default $30^{\circ}$); \textsc{topple} ends when no new domino has fallen for a plateau window (default $0.5\,\text{s}$); \textsc{cascade} runs from that plateau to the end.

\textsc{glancing} and \textsc{direct-strike} additionally emit a \textsc{glance} segment (ball clips leg, leg sways but does not give way) and a \textsc{strike} segment (ball directly contacts the leading domino). Trajectories that never propagate to a cascade instead emit empty segments.

\paragraph{QA construction.}
For each segment, we generate one descriptive QA and one causal QA. Distractors are written as plausible near-misses (mirror direction, wrong contact location, simultaneous vs.\ sequential fall), so that a judge ignoring the frames cannot answer by lexical elimination.

\emph{Template} mode pins each stage's wording to its canonical description. \emph{VLM-hybrid} mode rewrites the question and three distractors from the actual frames, and may override the templated correct answer when the footage demonstrably contradicts it; every override is logged for human audit. All datasets are human-reviewed.

A dispatch table maps each detected shape to a stage chain that is not the verbatim segment-name sequence:
\begin{verbatim}
SHAPE_STAGE_CHAINS = {
  "canonical":     ["rolling", "impact", "topple",  "cascade"],
  "glancing":      ["rolling", "glance", "strike",  "cascade_from_strike"],
  "direct_strike": ["rolling", "strike", "cascade_from_strike"],
}
\end{verbatim}

This separation allows the detector to remain a pure log analyzer, while the QA generator selects the correct causal answer at each stage. For \textsc{glancing} segment 4, for instance, the correct answer is \emph{``the ball directly struck the leading domino''} rather than \emph{``the falling platform struck the leading domino''}. Without this, the Unity reference itself answers the wrong question on non-canonical trajectories.


\paragraph{Judge.}
A VLM is conditioned on $K$ uniformly sampled frames per QA from the segment's window in the video under test (game engine or world-model generated); it sees only the frames, the question, and lettered options, and returns a single letter. We use GPT-5.4 \cite{openai2026gpt54} with $K=8$, and the frame count is included in the response cache key.

On tight segments (e.g.~direct-strike cascades spanning $0.12\,\text{s}$) the judge's competence on the Unity reference rises sharply with $K$ ($K=4$: Unity CSR $53\%$; $K=8$: $87\%$), so we treat $K$ as a benchmark hyper-parameter and tune it upward whenever Unity CSR is the bottleneck.


\paragraph{Asymmetric segment windowing.}
\texttt{BeamDomino} world-model rollouts are not frame-aligned to the game engine reference, even from matched initial conditions: Project Genie is controlled manually, so its rolling-impact-cascade events drift in wall-clock timing relative to the Unity log.

Strict segment-windowed scoring would then feed the judge the wrong slice of the WM clip for every QA whose Unity boundary lands on a temporally shifted event. That conflates temporal mistiming with causal incoherence, and double-penalizes any model whose causal content is right but whose action timing drifts. On the WM side, therefore, the judge sees the entire clip per QA, while the reference stays segment-windowed.

This asymmetry slightly favors the world model: a generation containing a recognizable event \emph{anywhere} in its clip can satisfy the corresponding QA even when the timing is off. Table~\ref{tab:causal-physical-causality}'s $\overline{\Delta}$ numbers should therefore be read as \emph{upper bounds on world-model causal fidelity}. A strict-windowed run would widen every gap.

\paragraph{Prompts.} The Physical Causality pipeline on \texttt{BeamDomino} issues two distinct VLM calls: an offline \emph{QA-decorator} call during dataset construction and an online \emph{judge} call at evaluation time.

\begin{tcolorbox}[colback=gray!10, colframe=deepblue,
  title={Causality Judge - Evaluation Prompt},
  fonttitle=\bfseries, sharp corners=south]
\begingroup
\ttfamily\small
You are watching a short video segment (frames \{segment\_frames\_start\} to \{segment\_frames\_end\}, roughly \{start\_sec:.2f\}s to \{end\_sec:.2f\}s).\\[0.4em]
Question (\{qa\_type\}):\\
\{question\}\\[0.4em]
Choices:\\
\{choices\_block\}\\[0.4em]
Answer with exactly one letter (A, B, C, ...). No explanation.
\endgroup
\end{tcolorbox}

\begin{tcolorbox}[colback=gray!10, colframe=deepblue,
  title={Causality QA Decorator Prompt},
  fonttitle=\bfseries, sharp corners=south]
\begingroup
\ttfamily\small
You are watching a short segment of a physics simulation video (the "BeamDomino" scene). The segment is labeled `\{stage\}` and runs roughly \{start\_sec:.2f\}s to \{end\_sec:.2f\}s within the trajectory.\\[0.4em]
We are asking the following multiple-choice question about this segment:\\[0.2em]
\hspace*{1em} Q: \{question\}\\[0.4em]
A static template suggests this correct answer:\\[0.2em]
\hspace*{1em} TEMPLATE\_CORRECT: \{correct\_body\}\\[0.4em]
Your tasks:\\
1. Optionally rewrite the question to be more visually specific to what the frames actually show (which side, which color, what direction, etc.). The rewritten question MUST still be answered correctly by whatever you decide the correct answer is in step 2.\\
2. Decide the correct answer based on what the frames actually show:\\
\hspace*{1em}- If the frames clearly support TEMPLATE\_CORRECT, return it verbatim as \texttt{`correct'}.\\
\hspace*{1em}- If the frames clearly contradict TEMPLATE\_CORRECT (e.g., the platform tipped left when the template said right, or a different object made contact than the template named), return a corrected \texttt{`correct'} that reflects what actually happens in the frames. Keep its grammatical style and length similar to TEMPLATE\_CORRECT.\\
\hspace*{1em}- When in doubt, prefer TEMPLATE\_CORRECT only override when the frames make the template wording demonstrably wrong.\\
3. Generate exactly 3 distractor options. Each distractor must:\\
\hspace*{1em}- Describe a \emph{different} outcome or cause than your \texttt{`correct'} answer\\
\hspace*{1em}- Be a plausible alternative that someone might pick if they had not watched the frames carefully\\
\hspace*{1em}- Be visually disprovable from the frames (a careful viewer can rule it out by what they see)\\
\hspace*{1em}- Be roughly the same length and grammatical style as your \texttt{`correct'} answer\\
\hspace*{1em}- Avoid trivia like "wind", "magic", "the camera moved", "scripted timer" prefer near-misses about which object moved, in what direction, in what order, against what\\[0.4em]
Respond with a single JSON object of the form: \{"question": "...", "correct": "...", "distractors": ["...", "...", "..."]\}.
\endgroup
\end{tcolorbox}

\paragraph{Stage QA template.}
\label{app:causality_stage_templates_excerpt}
The judge prompt's \verb|{question}| and \verb|{choices_block}| slots are filled from one of six stage templates. As a representative example, the \textsc{impact} stage (segment 2 of the canonical chain, in which the rolling ball strikes the supporting leg of the platform) contributes the following descriptive and causal QAs. The remaining five stages (\textsc{rolling}, \textsc{topple}, \textsc{cascade}, \textsc{glance}, \textsc{cascade\_from\_strike}) follow the same two-QA structure.

\begin{tcolorbox}[colback=gray!5, colframe=gray!50,
  title={\textsc{impact} stage - descriptive and causal QA},
  fonttitle=\bfseries, sharp corners=south]
\begingroup
\ttfamily\small
\textbf{Descriptive Q:} What is happening between the rolling floor ball and the white platform during this segment?\\[0.2em]
\hspace*{1em}A) the floor ball collides with one of the platform's vertical supporting legs (not the top surface)\\
\hspace*{1em}B) the floor ball passes harmlessly underneath the platform without touching any part of it\\
\hspace*{1em}C) the floor ball stops a short distance away from the platform and never makes contact with it\\
\hspace*{1em}D) the floor ball bounces upward onto the top of the platform and comes to rest next to the ball already there\\[0.6em]
\textbf{Causal Q:} Why does the white platform (and the ball still resting on top of it) begin to move in this segment?\\[0.2em]
\hspace*{1em}A) the floor ball transferred momentum into one of the platform's supporting legs, removing support from that side and destabilizing the platform above it\\
\hspace*{1em}B) the ball resting on top of the platform shifted on its own and unbalanced the platform without any contact from below\\
\hspace*{1em}C) one of the dominoes far across the scene fell first and somehow disturbed the platform from a distance\\
\hspace*{1em}D) the platform is scripted to fall on a fixed timer regardless of whether anything actually hits it
\endgroup
\end{tcolorbox}


\subsection{Causal LPIPS Implementation Details}
\label{app:event_detection}

To localize causal collision events in a reference--candidate generation pair,
we use a multi-stage pipeline that progressively narrows from cheap motion
priors to precise event anchors:
\begin{enumerate}
    \item \textbf{RAFT optical-flow proposals.} We compute dense optical flow
    with RAFT~\citep{teed2020raft}, sampled every 12 frames. For each sampled
    transition, we optionally subtract a global camera-motion estimate and score
    motion saliency as the mean flow magnitude plus $0.5$ times the mean change
    from the previous flow field. Local saliency peaks above a relative threshold
    define candidate event windows; if too few peaks are found, we fall back to
    coarse temporal windows.
    \item \textbf{Coarse VLM pass.} Within each proposed window, GPT-5.4-mini
    detects candidate contact episodes and filters out non-causal motion such as
    camera pans or background drift. Coarse detections are normalized, filtered
    by contact type, split when evidence supports multiple contacts, and merged
    across nearby windows.
    \item \textbf{Fine VLM pass.} Around the surviving coarse events, GPT-5.4
    performs a higher-resolution pass to localize collision episodes to frame
    anchors, typically corresponding to first contact. We then apply an
    additional GPT-5.4 relabeling pass to refine the contact side/type labels.
\end{enumerate}

This procedure is applied to both the reference and candidate videos, producing
event sets $\mathcal{E}_r$ and $\mathcal{E}_c$ used in
Equation~\ref{eq:clpips}. The two event timelines are matched chronologically,
and cLPIPS is computed on frame neighborhoods around the matched contact
anchors. The split between a faster model for coarse detection and a stronger
model for fine localization keeps cost bounded while preserving anchor precision
at the frames that contribute to cLPIPS.

\subsection{Causal LPIPS Judge Prompts}
\label{app:causal_judge_prompt}

All VLM stages in the causal LPIPS pipeline use the same system message. The
user message differs across the coarse detection, fine high-recall detection,
and side/type relabeling stages. The placeholder
\textcolor{blue}{\{{allowed\_frame\_indices}\}} is replaced at runtime with the
sampled keyframe indices that the judge is allowed to cite.

\begin{tcolorbox}[colback=gray!10, colframe=deepblue, title=Shared System Message, fonttitle=\bfseries, sharp corners=south]
\textbf{\textcolor{blue}{System Message:}}\\
You are a careful video analyst for physics-style evaluation clips. You must ground every conclusion in the provided keyframe images. Respond with a single JSON object only, with no markdown fences.
\end{tcolorbox}

\begin{tcolorbox}[colback=gray!10, colframe=deepblue, title=Coarse Event Detection Prompt (First Video), fonttitle=\bfseries, sharp corners=south]
\textbf{\textcolor{blue}{Prompt:}}\\
Analyze these keyframes from ONE video. Keyframe order matches image order.\\
Allowed frame indices, only these may appear in JSON: \textcolor{blue}{\{allowed\_frame\_indices\}}\\

\#\# Scene-specific instructions, follow carefully\\
Return the full collision timeline, not just the first event. Enumerate all distinct collision/contact episodes in chronological order. Do not omit later collisions just because an earlier one exists, and do not collapse distinct episodes into one event. Include depth-direction collisions as well as lateral left/right collisions when they are visually distinct episodes. When depth ambiguity exists, use side=\texttt{front} for contact with an obstacle ahead in the forward camera-depth direction, and use side=\texttt{back} for a later contact with an obstacle behind or after the moving object has already passed it. Only merge contacts that are clearly one continuous contact episode with no meaningful separation.\\
Return at most 6 events, and prefer the earliest clearly supported event when uncertain.\\

Your only job is to detect distinct contact/impact episodes in chronological order.\\
Do not spend effort on left/right/front/back semantics in this pass. Use side=\texttt{unknown} unless the side is completely unavoidable from the local evidence.\\

Follow this checklist and put concise notes in \texttt{reasoning\_steps}, as a string array:\\
A) Identify the main moving object and the nearby obstacles or targets it contacts.\\
B) Scan in time order and enumerate each distinct contact episode once.\\
C) Keep episodes separate when there is meaningful temporal separation. If a visible gap reopens and the moving object later contacts the same or another obstacle again, treat that as a new event rather than merging it into one long episode. Do not merge two episodes just because they happen on the same general side of the frame.\\
D) Cite only allowed frame indices.\\
E) Abstain with \texttt{"events": []} if quality is insufficient.\\

For each event object include:\\
- \texttt{type}: usually \texttt{"collision"}\\
- \texttt{side}: set to \texttt{"unknown"} unless truly unambiguous\\
- \texttt{frame}: integer, must be one of the allowed indices, best contact frame\\
- \texttt{confidence}: number 0--1, optional\\
- \texttt{evidence\_frame\_indices}: non-empty array of allowed integers, typically 2--4 frames around contact\\
- \texttt{evidence\_summary}: 1--3 sentences describing visible contact evidence only\\

Output JSON shape:\\
\texttt{\{"reasoning\_steps": ["..."], "events": [...]\}}

\end{tcolorbox}

\begin{tcolorbox}[colback=gray!10, colframe=deepblue, title=Coarse Event Detection Prompt (Next Videos), fonttitle=\bfseries, sharp corners=south]
\textbf{\textcolor{blue}{Prompt for next video:}}\\
Analyze these keyframes from ONE video. Keyframe order matches image order.\\
Allowed frame indices, only these may appear in JSON: \textcolor{blue}{\{allowed\_frame\_indices\}}\\

Return the full collision timeline, not just the first event. Enumerate all distinct collision/contact episodes in chronological order. Return at most 6 events, and use side=\texttt{"unknown"} unless the side is completely unavoidable from local evidence. Cite only allowed frame indices. Abstain with \texttt{"events": []} if quality is insufficient.\\

Output JSON shape:\\
\texttt{\{"reasoning\_steps": ["..."], "events": [...]\}}
\end{tcolorbox}


\begin{tcolorbox}[colback=gray!10, colframe=deepblue, title=High-recall Detection Prompt (First Video), fonttitle=\bfseries, sharp corners=south]
\textbf{\textcolor{blue}{User Message:}}\\
Analyze these keyframes from ONE video. Keyframe order matches image order.\\
Allowed frame indices, only these may appear in JSON: \textcolor{blue}{\{allowed\_frame\_indices\}}\\

\#\# Scene-specific instructions, follow carefully\\
Return the full collision timeline, not just the first event. Enumerate all distinct collision/contact episodes in chronological order. 
When depth ambiguity exists, use side=\texttt{"front"} for contact with an obstacle ahead in the forward camera-depth direction, and use side=\texttt{"back"} for a later contact with an obstacle behind or after the moving object has already passed it. 
Return at most 12 events, and prefer the earliest clearly supported event when uncertain.\\

Your only job is to recover as many distinct contact/impact episodes as are plausibly supported by the keyframes.\\
Use side=\texttt{"unknown"} unless the side is completely unavoidable from the local evidence.\\

Follow this checklist and put concise notes in \texttt{reasoning\_steps}, as a string array:\\
A) Identify the main moving object and the nearby obstacles or targets it contacts.\\
B) Scan in time order and list every temporally distinct contact episode you can support from the frames.\\
C) If contact stops and later resumes, treat that as a new event.\\
D) If a later contact looks brief, grazing, or rebound-like, still include it when it is temporally separate from earlier contact.\\
E) Do not collapse multiple contacts into one event just because they involve nearby obstacles or similar motion.\\
F) Cite only allowed frame indices.\\
G) Abstain with \texttt{"events": []} only if the frames truly do not support any contact episode.\\

For each event object include:\\
- \texttt{type}: usually \texttt{"collision"}\\
- \texttt{side}: set to \texttt{"unknown"} unless truly unambiguous\\
- \texttt{frame}: integer, must be one of the allowed indices, best contact frame\\
- \texttt{confidence}: number 0--1, optional\\
- \texttt{evidence\_frame\_indices}: non-empty array of allowed integers, prefer 3--8 frames spanning approach/contact/separation when available\\
- \texttt{evidence\_summary}: 1--3 sentences describing visible contact evidence only\\

Output JSON shape:\\
\texttt{\{"reasoning\_steps": ["..."], "events": [...]\}}

\end{tcolorbox}

\begin{tcolorbox}[colback=gray!10, colframe=deepblue, title=High-recall Detection Prompt (Next Videos), fonttitle=\bfseries, sharp corners=south]
\textbf{\textcolor{blue}{User Message:}}\\
Analyze these keyframes from ONE video. Keyframe order matches image order.\\
Allowed frame indices, only these may appear in JSON: \textcolor{blue}{\{allowed\_frame\_indices\}}\\

Return the full collision timeline, not just the first event. Enumerate all distinct collision/contact episodes in chronological order. Return at most 12 events, and use side=\texttt{"unknown"} unless the side is completely unavoidable from local evidence. Prefer recall over precision: list temporally distinct episodes separately, including brief, grazing, or rebound-like contacts when supported. Cite only allowed frame indices. Abstain with \texttt{"events": []} only if the frames truly do not support any contact episode.\\

Output JSON shape:\\
\texttt{\{"reasoning\_steps": ["..."], "events": [...]\}}
\end{tcolorbox}

\begin{tcolorbox}[colback=gray!10, colframe=deepblue, title=Type Relabeling User Message, fonttitle=\bfseries, sharp corners=south]
\textbf{\textcolor{blue}{User Message:}}\\
Analyze these keyframes from ONE video around a single already-detected contact episode.\\
Allowed frame indices, only these may appear in JSON: \textcolor{blue}{\{allowed\_frame\_indices\}}\\
Event anchor frame: \textcolor{blue}{\{anchor\_frame\}}\\
Current event summary: \textcolor{blue}{\{event\_summary\}}\\

\#\# Scene-specific instructions, follow carefully\\
Return the full collision timeline, not just the first event. Enumerate all distinct collision/contact episodes in chronological order. Do not omit later collisions just because an earlier one exists, and do not collapse distinct episodes into one event. Include depth-direction collisions as well as lateral left/right collisions when they are visually distinct episodes. When depth ambiguity exists, use side=\texttt{"front"} for contact with an obstacle ahead in the forward camera-depth direction, and use side=\texttt{"back"} for a later contact with an obstacle behind or after the moving object has already passed it. Only merge contacts that are clearly one continuous contact episode with no meaningful separation.\\

Classify the interaction geometry of this one contact episode using motion before, at, and after contact.\\

First choose exactly one \texttt{contact\_class}:\\
- \texttt{"depth"}: contact is primarily head-on or depth-aligned with an obstacle ahead\\
- \texttt{"lateral"}: contact is primarily side-on in image space\\
- \texttt{"rear"}: contact is best described as rearward or after-passing contact\\
- \texttt{"unknown"}: only if the interaction class remains ambiguous\\

Then choose exactly one \texttt{side}:\\
- \texttt{"left"} or \texttt{"right"}: contact is primarily lateral in image space\\
- \texttt{"front"}: contact is with an obstacle ahead in the moving object's forward/depth direction\\
- \texttt{"back"}: contact is with an obstacle behind/rearward or after the moving object has already passed it\\
- \texttt{"unknown"}: only if the side remains ambiguous from the provided frames\\

Use temporal cues, not just a single overlap frame:\\
- compare approach vs. departure frames\\
- if the object is approaching an obstacle head-on in depth, prefer \texttt{"front"}\\
- if contact happens after the object has already passed an obstacle or with a rearward obstacle, prefer \texttt{"back"}\\
- if the local evidence only supports the coarse interaction geometry, keep side=\texttt{"unknown"}\\

Output JSON shape:\\
\texttt{\{"reasoning\_steps": ["..."], "contact\_class": "depth|lateral|rear|unknown", "side": "front|back|left|right|unknown", "confidence": 0.0, "side\_confidence": 0.0, "evidence\_frame\_indices": [...], "evidence\_summary": "..."\}}
\end{tcolorbox}

\subsection{Prompt-Scene Consistency Evaluation Details}
\label{app:prompt-scene}

Appendix~\ref{app:prompt_scene_metrics} defines the pipeline, and this section lists its exact prompts. Each rubric generator and each judge has its own system message, while the user messages are shared across the two branches: \texttt{\{agent\}} is \texttt{environment} or \texttt{object}, and \texttt{\{rubric\_items\_block\}} holds the items of the corresponding branch.

\begin{tcolorbox}[
breakable,
colback=gray!10,
colframe=deepblue,
title=Rubric Generator --- Environment Agent,
fonttitle=\bfseries,
sharp corners=south
]
\textbf{\textcolor{blue}{System Message:}}\\
You are an expert in evaluation rubric design. Your task is to generate Environment-Consistency rubrics to help a VLM judge score whether a generated video's environment remains consistent with the prompt over time. Each rubric is a binary yes/no question answerable from timestamped frames, derived from the text prompt only, and grounded in something the prompt explicitly states or directly implies.\\

\#\#\# Important Notes:\\
1. Consistency: Each rubric tests whether one environmental attribute, e.g., ground material/pattern, sky, lighting, terrain, dominant style/palette, matches the prompt AND remains the same across ALL sampled frames -- it must not switch, drift, or contradict the prompt at any timestamp. The camera may rotate, but the environment is assumed static unless the prompt explicitly describes spatial variation or change.\\

2. Described Variation: If the prompt itself describes an environmental change or contrast, e.g., lighting shifts, terrain changes, do NOT enforce uniform consistency; instead, generate a rubric that checks whether the described variation or contrast is preserved.\\

3. Atomicity: one environmental attribute per rubric.\\

4. Scope: Only scene-level attributes, e.g., ground, sky, lighting, terrain, style, palette. Salient entities, characters, or main objects are handled by the Object agent.\\

5. Rubric Construction: Prefer fewer high-confidence criteria over exhaustive lists.\\

\#\#\# Output Format:\\
Return valid JSON matching the provided schema. Each rubric item contains:\\
- \texttt{rubric\_id}: unique ID with the \texttt{"env\_"} prefix, e.g., \texttt{env\_01}, \texttt{env\_02}.\\
- \texttt{agent}: always \texttt{"environment"}.\\
- \texttt{question}: the binary yes/no question for the VLM judge.\\
- \texttt{rationale}: a brief justification explaining which prompt phrase the rubric is grounded in and what it checks.\\

Example JSON output:\\
{\ttfamily
\begin{flushleft}
\{\\
\quad "rubric\_items": [\\
\quad\quad \{\\
\quad\quad\quad "rubric\_id": "env\_01",\\
\quad\quad\quad "agent": "environment",\\
\quad\quad\quad "question": "Does the ground remain a black-and-white\\
\quad\quad\quad chessboard pattern over time?",\\
\quad\quad\quad "rationale": "The chessboard ground is the defining\\
\quad\quad\quad environmental attribute and should not change as the\\
\quad\quad\quad camera rotates."\\
\quad\quad \}\\
\quad ]\\
\}\\
\end{flushleft}
}
\end{tcolorbox}

\begin{tcolorbox}[
breakable,
colback=gray!10,
colframe=deepblue,
title=Rubric Generator --- Object Agent,
fonttitle=\bfseries,
sharp corners=south
]
\textbf{\textcolor{blue}{System Message:}}\\
You are an expert in text-to-video evaluation rubric design. Your task is to generate Object-Consistency rubrics to help a VLM judge score whether a generated video realizes the salient entities (characters, vehicles, props, primary objects) described in the text prompt, with their claimed attributes. Each rubric is a binary yes/no question answerable from timestamped frames, derived from the text prompt only, and grounded in something the prompt explicitly states.\\

\#\#\# Important Notes:\\
1. Attribute Fidelity: Each rubric targets claimed attributes, e.g., shape, texture, size, color, material, of one salient entity.\\

2. Atomicity: one entity per rubric.\\

3. Scope: Only salient entities or characters. Scene-level structure, ground, sky, lighting, and palette are handled by the Environment agent.\\

4. Rubric Construction: Prefer fewer high-confidence criteria over exhaustive lists.\\

\#\#\# Output Format:\\
Return valid JSON matching the provided schema. Each rubric item contains:\\
- \texttt{rubric\_id}: unique ID with the \texttt{"obj\_"} prefix, e.g., \texttt{obj\_01}, \texttt{obj\_02}.\\
- \texttt{agent}: always \texttt{"object"}.\\
- \texttt{question}: the binary yes/no question for the VLM judge.\\
- \texttt{rationale}: a brief justification explaining which prompt phrase the rubric is grounded in and what it checks.\\

Example JSON output:\\
{\ttfamily
\begin{flushleft}
\{\\
\quad "rubric\_items": [\\
\quad\quad \{\\
\quad\quad\quad "rubric\_id": "obj\_01",\\
\quad\quad\quad "agent": "object",\\
\quad\quad\quad "question": "When the sphere is visible and judgeable,\\
\quad\quad\quad does it have the chrome appearance specified in the\\
\quad\quad\quad prompt?",\\
\quad\quad\quad "rationale": "The prompt explicitly states the sphere\\
\quad\quad\quad is chrome, so its material appearance should be correct\\
\quad\quad\quad whenever it can be judged."\\
\quad\quad \}\\
\quad ]\\
\}\\
\end{flushleft}
}
\end{tcolorbox}

\begin{tcolorbox}[
breakable,
colback=gray!10,
colframe=deepblue,
title=Rubric Generator --- User Message (Both Agents),
fonttitle=\bfseries,
sharp corners=south
]
\textbf{\textcolor{blue}{User Message:}}\\
Scene name: \textcolor{blue}{\{scene\_name\}}\\
\\
Prompt:\\
\textcolor{blue}{\{prompt\_text\}}\\
\\
Generate up to 5 \textcolor{blue}{\{agent\}} rubrics for this scene.
\end{tcolorbox}

\begin{tcolorbox}[
breakable,
colback=gray!10,
colframe=deepblue,
title=Environment Judge,
fonttitle=\bfseries,
sharp corners=south
]
\textbf{\textcolor{blue}{System Message:}}\\
\#\#\# Role\\
You are an expert video evaluator for prompt-scene consistency. Your task is to judge whether the environmental attributes in a generated video are faithful to the text prompt.\\

\#\#\# Evaluation Scope\\
Evaluate only environment-level attributes, including ground, sky, terrain, lighting, background structure, global style, and color palette.\\

Do not evaluate salient entities such as characters, vehicles, animals, controllable agents, or primary foreground objects. These are evaluated by a separate object judge. If a rubric item mixes environment and object properties, judge only the environmental component.\\

\#\#\# Inputs\\
- Text prompt: The primary source of truth. All judgments must be grounded in attributes explicitly stated or directly implied by the prompt.\\
- Candidate frames: Uniformly sampled frames from the generated video, labeled as \texttt{frame\_t=X.Xs}.\\
- Reference image: A frame labeled \texttt{ref\_frame}, providing visual context for the intended scene appearance. Use it only to clarify ambiguous prompt wording or visible environmental attributes.\\
- Rubric items: Each item targets one environment-level attribute and should be judged independently.\\

\#\#\# Evaluation Protocol\\
For each rubric item, identify the targeted environmental attribute and interpret it in the context of the full text prompt. If the prompt explicitly describes an environmental change, state transition, or exception, account for it when scoring the relevant attribute. Otherwise, assume the environment is stable: camera motion may change what is visible, but should not change the underlying environmental properties.\\

Base each judgment on direct visual evidence from the sampled frames. For each rubric item, first inspect the relevant frames, then write step-by-step reasoning that cites specific frame labels and explains how the targeted environmental attribute matches or deviates from the prompt; only then assign the final score. Be conservative: if an attribute cannot be clearly verified, assign a lower score.\\

\#\#\# Scoring Rubric\\
Use an integer score from 0 to 3 for each rubric item.\\

- 3: The attribute clearly matches the prompt across all relevant frames, with no meaningful ambiguity, omission, or inconsistency.\\
- 2: The attribute matches the prompt in the vast majority of relevant frames, with only minor visual ambiguity, brief occlusion, or isolated low-severity artifacts. Do not assign 2 for clear semantic, material, or structural changes that contradict the prompt.\\
- 1: The attribute matches the prompt in some frames but is wrong, missing, unstable, or ambiguous in a non-trivial portion of the video.\\
- 0: The attribute contradicts the prompt, is absent or unverifiable in most relevant frames, changes substantially without prompt support, or the video is too corrupted to evaluate.\\

When uncertain between two adjacent scores, choose the lower score.\\

\#\#\# Output Requirements\\
Return valid JSON only.\\

Additionally, provide a top-level explanations object with:\\
- summary: a concise overall summary of environment-consistency quality.\\
- issues: a list of specific environment-consistency issues found across rubrics.\\

\#\#\# Output Schema\\
{\ttfamily
\begin{flushleft}
\{\\
\quad "item\_scores": [\\
\quad\quad \{\\
\quad\quad\quad "rubric\_id": "<rubric\_id>",\\
\quad\quad\quad "reasoning": "<your evidence-based reasoning\\
\quad\quad\quad with cited frame labels>",\\
\quad\quad\quad "frame\_evidence": ["<frame\_label\_1>",\\
\quad\quad\quad "<frame\_label\_2>"],\\
\quad\quad\quad "judged\_score": <integer\_0\_to\_3>\\
\quad\quad \}\\
\quad ],\\
\quad "explanations": \{\\
\quad\quad "summary": "<concise summary of overall\\
\quad\quad environment-consistency quality>",\\
\quad\quad "issues": ["<specific issue 1>",\\
\quad\quad "<specific issue 2>"]\\
\quad \}\\
\}
\end{flushleft}
}
\end{tcolorbox}

\begin{tcolorbox}[
breakable,
colback=gray!10,
colframe=deepblue,
title=Object Judge,
fonttitle=\bfseries,
sharp corners=south
]
\textbf{\textcolor{blue}{System Message:}}\\
\#\#\# Role\\
You are an expert video evaluator for prompt-object consistency. Your task is to judge whether the salient entities in a generated video are faithful to the text prompt and visually consistent when visible.\\

\#\#\# Evaluation Scope\\
Evaluate only salient entities mentioned in the prompt, including characters, vehicles, animals, props, controllable agents, and primary foreground objects, as well as their explicitly stated or directly implied attributes.\\

Do not evaluate environment-level attributes such as ground, sky, terrain, lighting, background structure, global style, or color palette. These are evaluated by a separate environment judge. If a rubric item mixes object and environment properties, judge only the object-level component.\\

\#\#\# Inputs\\
- Text prompt: The primary source of truth. All judgments must be grounded in entities and attributes explicitly stated or directly implied by the prompt.\\
- Candidate frames: Uniformly sampled frames from the generated video, labeled as \texttt{frame\_t=X.Xs}.\\
- Reference image: A frame labeled \texttt{ref\_frame}, providing visual context for the intended scene appearance. Use it only to clarify ambiguous prompt wording or visible object attributes.\\
- Rubric items: Each item targets one salient entity or one object-level attribute and should be judged independently.\\

\#\#\# Evaluation Protocol\\
For each rubric item, identify the targeted entity and the attribute being evaluated, such as color, material, shape, texture, count, or identity. Interpret the item in the context of the full text prompt.\\

Scan the sampled frames and determine which frames are relevant for judging the item. For object-level attributes, only frames where the targeted entity is visible should contribute to scoring.\\

Base each judgment on direct visual evidence from the sampled frames. For each rubric item, first inspect the relevant visible frames, then write step-by-step reasoning that cites specific frame labels and explains how the targeted entity or attribute matches or deviates from the prompt; only then assign the final score. Be conservative: if an entity or attribute cannot be clearly verified, assign a lower score.\\

\#\#\# Scoring Rubric\\
Use an integer score from 0 to 3 for each rubric item.\\

- 3: The targeted entity is visible and its prompt-specified attributes clearly match the prompt in all visible, judgeable frames. No meaningful ambiguity, omission, or inconsistency is observed.\\
- 2: The targeted entity is visible and its attributes match the prompt in the vast majority of visible frames, with only minor visual ambiguity, brief occlusion, or isolated low-severity artifacts. Do not assign 2 for clear identity errors, wrong colors, wrong shapes, missing required parts, or persistent attribute errors.\\
- 1: The targeted entity is visible in some frames, but its identity or prompt-specified attributes are wrong, missing, unstable, or ambiguous in a non-trivial portion of visible, judgeable frames.\\
- 0: The targeted entity is absent throughout the sampled frames, contradicts the prompt, has the wrong identity or attributes in most visible frames, or the video is too corrupted to evaluate.\\

When uncertain between two adjacent scores, choose the lower score.\\

\#\#\# Output Requirements\\
Return valid JSON only.\\

Additionally, provide a top-level explanations object with:\\
- summary: a concise overall summary of object-consistency quality.\\
- issues: a list of specific object-consistency issues found across rubrics.\\

\#\#\# Output Schema\\
{\ttfamily
\begin{flushleft}
\{\\
\quad "item\_scores": [\\
\quad\quad \{\\
\quad\quad\quad "rubric\_id": "<rubric\_id>",\\
\quad\quad\quad "reasoning": "<your evidence-based reasoning\\
\quad\quad\quad with cited frame labels>",\\
\quad\quad\quad "frame\_evidence": ["<frame\_label\_1>",\\
\quad\quad\quad "<frame\_label\_2>"],\\
\quad\quad\quad "judged\_score": <integer\_0\_to\_3>\\
\quad\quad \}\\
\quad ],\\
\quad "explanations": \{\\
\quad\quad "summary": "<concise summary of overall\\
\quad\quad object-consistency quality>",\\
\quad\quad "issues": ["<specific issue 1>",\\
\quad\quad "<specific issue 2>"]\\
\quad \}\\
\}
\end{flushleft}
}
\end{tcolorbox}

\begin{tcolorbox}[
breakable,
colback=gray!10,
colframe=deepblue,
title=Environment and Object Judges --- User Message,
fonttitle=\bfseries,
sharp corners=south
]
\textbf{\textcolor{blue}{User Message:}}\\
\#\# Evaluation Task\\
Evaluate the candidate video frames against each rubric item below.\\

Reference image: ref\_frame\\
Candidate frames: \textcolor{blue}{\texttt{\{candidate\_frame\_labels\}}}\\

Prompt:\\
\textcolor{blue}{\texttt{\{prompt\_text\}}}\\

\textcolor{blue}{\texttt{\{rubric\_items\_block\}}}\\

Evaluate each rubric item and return structured JSON.
\end{tcolorbox}

\section{Results and Analysis (Continued)}
\label{app:experiment_results}

\subsection{Fluid Mechanics}
\label{appx:fluid_mechanics}

\subsubsection{Per-trajectory Decomposition}
\label{appx:fluid-per-profile}
Table~\ref{tab:fluid-mechanics-perprofile} expands
Table~\ref{tab:fluid-headline} into one row per (model, action profile), exposing the trajectory-level structure that the headline pooling averages over.

\begin{table}[t]
\centering
\caption{\textbf{Estimator behavior on world-model generations of }\texttt{LiquidContainer} \textbf{scene.} \emph{Mask presence} is the fraction of frames in which SAM~2 returns a non-empty liquid mask; \emph{scale ratio} is the mean liquid-mask area fraction of the generation divided by that of its matched reference, so $1.0$ means both sides present the tracked object at the same apparent size; \emph{valid flow cells} is the median number of PIV cells (of $1428$) passing the $0.5$ liquid-fraction threshold that the velocity cosine averages over, against $37$ on the reference. A clip is \emph{gated out} when mask presence is below $1.0$ or the scale ratio falls outside $[0.5, 2]$ (Appendix~\ref{app:fluid_validity}).}
\label{tab:fluid-validity}
\small
\begin{adjustbox}{max width=\textwidth}
\begin{tabular}{lccccc}
\toprule
Model & Clips & Mask presence & Scale ratio & Valid flow cells & Clips gated out \\
      &       & med.\ / min.  & med.\ [min, max] & med.\ of $1428$ & \\
\midrule
Project Genie    & $20$ & $1.000$ / $1.000$ & $1.88$ $[1.13,\ 16.03]$ & $80$ & $8/20$ \\
HY-WorldPlay     & $20$ & $1.000$ / $1.000$ & $0.51$ $[0.48,\ 1.05]$ & $19$ & $7/20$ \\
\midrule
\emph{Unity reference} & --- & $1.000$ / $1.000$ & $1.00$ & $37$ & --- \\
\bottomrule
\end{tabular}
\end{adjustbox}
\end{table}

\begin{table}[ht]
\centering
\caption{Fluid Mechanics evaluation on \texttt{LiquidContainer}, split by (model, action profile). Each cell is mean$\pm$std over $5$ generations. Values are computed on the same corrected basis as Table~\ref{tab:fluid-headline}: deduplicated generations and a mean-centered meniscus trace without bulk compensation.
\textbf{cos}: liquid-velocity cosine similarity ($\rightarrow 1$), a diagnostic rather than a ranking axis;
$\Delta\sigma_A$, $|\Delta\dot A|$: volume-conservation deltas relative to Unity (lower better);
$|\Delta\theta_{\mathrm{rms}}|$: absolute surface-tilt residual delta in degrees (lower better);
$L_1$: meniscus power-spectrum $L_1$ distance (lower better);
LPIPS: perceptual mean over the shared segment (lower better).
A cell marked $^{\ddagger}$ retains fewer than $3$ clips passing the candidate-track validity gate of Appendix~\ref{app:fluid_validity} and is reported for completeness only; column winners are marked among the remaining cells when at least two remain.}
\label{tab:fluid-mechanics-perprofile}
\small
\begin{adjustbox}{max width=\textwidth}
\begin{tabular}{llrrrrrr}
\toprule
Profile & Model & cos $\uparrow$ & $\Delta\sigma_A\!\downarrow$ & $|\Delta\dot A|\!\downarrow$ &
$|\Delta\theta_{\mathrm{rms}}|\!\downarrow$ & $L_1\!\downarrow$ &
LPIPS$\!\downarrow$ \\
\midrule
\multirow{2}{*}{\texttt{rest}}
 & Project Genie & $0.762{\pm}0.187$ & $0.060{\pm}0.058$ & $0.006{\pm}0.005$ & $\mathbf{5.10{\pm}6.60}$ & $\mathbf{0.989{\pm}0.130}$ & $0.640{\pm}0.011$ \\
 & HY-WorldPlay & $\mathbf{0.977{\pm}0.010}$ & $\mathbf{0.003{\pm}0.001}$ & $\mathbf{0.001{\pm}0.000}$ & $6.37{\pm}6.28$ & $1.062{\pm}0.143$ & $\mathbf{0.247{\pm}0.026}$ \\
\midrule
\multirow{2}{*}{\texttt{impulse\_decay}}
 & Project Genie$^{\ddagger}$ & $0.857{\pm}0.135$ & $0.184{\pm}0.207$ & $0.011{\pm}0.012$ & $7.01{\pm}2.33$ & $1.095{\pm}0.181$ & $0.625{\pm}0.018$ \\
 & HY-WorldPlay & $0.984{\pm}0.008$ & $0.003{\pm}0.001$ & $0.001{\pm}0.000$ & $7.17{\pm}5.53$ & $0.881{\pm}0.157$ & $0.308{\pm}0.026$ \\
\midrule
\multirow{2}{*}{\texttt{chirp\_strafe}}
 & Project Genie & $0.796{\pm}0.229$ & $0.167{\pm}0.363$ & $0.018{\pm}0.038$ & $8.35{\pm}2.79$ & $1.616{\pm}0.180$ & $0.615{\pm}0.024$ \\
 & HY-WorldPlay$^{\ddagger}$ & $0.989{\pm}0.002$ & $0.030{\pm}0.041$ & $0.002{\pm}0.002$ & $8.73{\pm}10.62$ & $1.624{\pm}0.015$ & $0.306{\pm}0.010$ \\
\midrule
\multirow{2}{*}{\texttt{step\_hold}}
 & Project Genie & $0.858{\pm}0.132$ & $0.274{\pm}0.512$ & $0.032{\pm}0.062$ & $\mathbf{15.01{\pm}12.46}$ & $1.273{\pm}0.085$ & $0.615{\pm}0.026$ \\
 & HY-WorldPlay & $\mathbf{0.990{\pm}0.002}$ & $\mathbf{0.026{\pm}0.013}$ & $\mathbf{0.002{\pm}0.001}$ & $16.20{\pm}14.07$ & $\mathbf{1.247{\pm}0.133}$ & $\mathbf{0.341{\pm}0.011}$ \\
\bottomrule
\end{tabular}\end{adjustbox}
\end{table}

\paragraph{What the breakdown adds.}
Three patterns are visible in the per-profile table that the pooled headline cannot show.
\begin{itemize}
    \item HY-WorldPlay has the lowest volume jitter on every profile ($0.003$, $0.003$, $0.030$, $0.026$), so its pooled area-stability advantage is not driven by a single profile.
    \item Project Genie's lower pooled tilt residual holds on both profiles where the two models are reportable: \texttt{rest} ($5.10^\circ$ vs.\ $6.37^\circ$) and \texttt{step\_hold} ($15.01^\circ$ vs.\ $16.20^\circ$).
    \item \textsc{Project Genie}'s velocity-cosine is the noisiest of the two models on every profile (per-profile std $0.13$-$0.23$ vs.\ $\leq 0.01$ for HY-WorldPlay).
\end{itemize}

\paragraph{Why no model wins all five axes.}
Table~\ref{tab:fluid-mechanics-perprofile} read against Table~\ref{tab:fluid-headline} shows the five axes measure approximately orthogonal failure modes on a sealed-bottle scene. $\cos$ fails when the model gets the \emph{direction} of slosh wrong (chirp-conditioned); $\Delta\sigma_{A}$ fails when it gets the \emph{shape} of the liquid wrong (area unstable, in different regimes for camera-vs.-action conditioned models); $|\Delta\dot{A}|$ fails when it leaks or hallucinates volume over the clip; $|\Delta\theta_{\text{rms}}|$ fails when the air/liquid surface settles at a non-Unity angle (most starkly on the \texttt{rest} hallucinated-tilt regime); and $L_1$ fails when the meniscus oscillates at the wrong frequencies.

The two models occupy different corners of that space. Project Genie is biased toward direction-wrong generations (the lowest and noisiest cosine) while keeping the lower tilt residual; HY-WorldPlay holds projected area almost perfectly stable across every profile while mis-orienting the surface, which is why it leads four pooled axes and still posts the higher pooled tilt residual.

The lack of a single winner is not noise. It reflects distinct inductive biases across the two architectures, which is why \texttt{LiquidContainer} cannot be summarised by a single scalar.

\subsubsection{LPIPS Structural-Rendering Case Study}
\label{app:fluid_lpips_structural}

\paragraph{Background fidelity dominates LPIPS averaging.}
HY-WorldPlay holds the lowest pooled LPIPS in Table~\ref{tab:fluid-headline} ($0.300$, against $0.624$ for Project Genie), which would make it the headline winner under prior video-quality benchmarks that report LPIPS alone.

Frame-level inspection of the rendered clips shows that this lead is partly structural rather than physical. The bottle is rendered at roughly half of the Unity reference's image-area share (median scale ratio $0.51$, Table~\ref{tab:fluid-validity}), and the PIV support inside the liquid shrinks accordingly (median $19$ valid flow cells against $37$ on the reference).

Because LPIPS averages perceptual distance over the whole frame, a clip in which the load-bearing physics object occupies half of its reference image-area share will trade physics fidelity for background fidelity in the average: empty floor and ambient lighting match well, the bottle interior matches less well, and the average leans on the former.

\paragraph{The volume- and tilt-deltas surface what LPIPS hides.}
The volume-conservation deltas ($\Delta\sigma_{A}$, $|\Delta\dot{A}|$) and the tilt residual ($|\Delta\theta_{\text{rms}}|$) are computed only inside the SAM~2 liquid mask, so they are agnostic to background fidelity. On the same HY-WorldPlay clips that win LPIPS, $|\Delta\theta_{\text{rms}}|=9.62^{\circ}$, no better than Project Genie's $8.87^{\circ}$, even though HY-WorldPlay leads both area axes.

Decomposed physics metrics are therefore not a stricter LPIPS but a \emph{different} measurement: they ignore the parts of the frame that LPIPS averages over and isolate the parts the world model is supposed to model. A single-axis leaderboard that reported only LPIPS would report HY-WorldPlay as the strongest fluid model by a wide margin, but the physics axes give a split verdict: it leads area stability yet does not lead surface orientation.

\subsubsection{Limitations}
\label{app:fluid_limits}

\paragraph{Scene and camera-regime scope.}
We evaluate one sealed half-filled glass bottle at fixed viscosity/density/surface tension under a single fixed-side camera
regime; free-pour, multi-fluid, viscosity sweeps, and hand-held/orbital camera regimes are out of scope. The five physics axes, therefore, sample fluid-mechanics fidelity at one operating point, and the bulk-motion compensation, which is cosmetic here (Appendix~\ref{appx:fluid_pipeline}), becomes load-bearing under any camera regime in which the bottle is not approximately world-fixed.

\paragraph{Pooling-induced artifacts.}
The headline (Table~\ref{tab:fluid-headline}) pools across trajectories with equal weight, which penalizes asymmetric-failure models and rewards uniformly mediocre ones. We report both pooled and per-trajectory-type views and recommend reading them jointly.

\providecommand{\liqlabel}[1]{\multicolumn{4}{c}{\small\textbf{#1}}}

\begin{figure}[t]
  \centering
  \begin{adjustbox}{max width=\textwidth}
  \setlength{\tabcolsep}{2pt}
  \renewcommand{\arraystretch}{0.6}
  \begin{tabular}{@{}cccc@{}}
    \includegraphics[width=4cm,keepaspectratio]{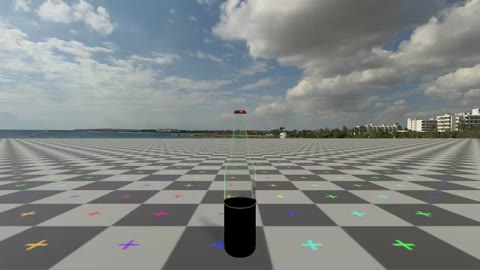}   & \includegraphics[width=4cm,keepaspectratio]{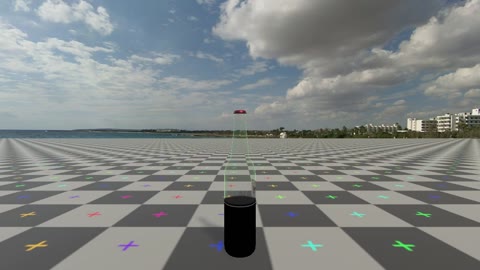}   & \includegraphics[width=4cm,keepaspectratio]{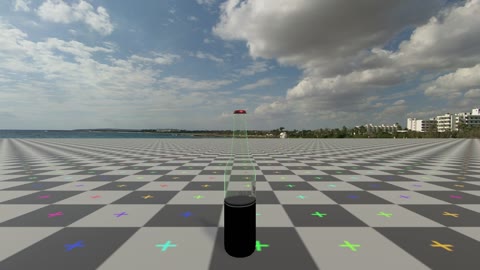}   & \includegraphics[width=4cm,keepaspectratio]{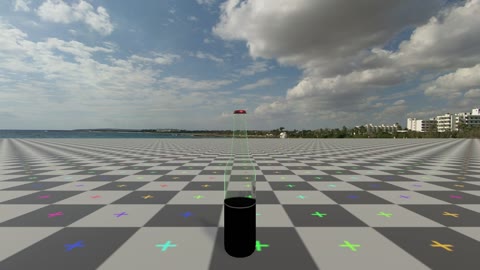}   \\
    \liqlabel{Unity}      \\[3pt]
    \includegraphics[width=4cm,keepaspectratio]{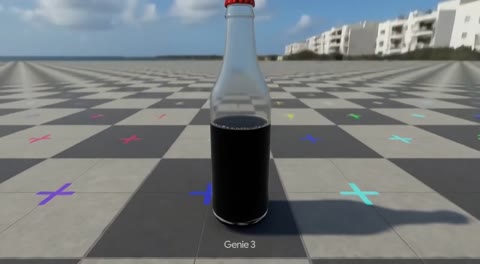}   & \includegraphics[width=4cm,keepaspectratio]{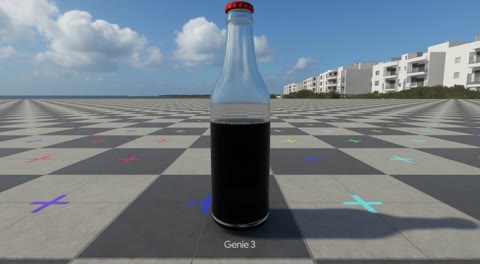}   & \includegraphics[width=4cm,keepaspectratio]{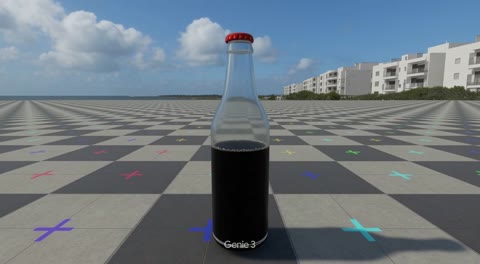}   & \includegraphics[width=4cm,keepaspectratio]{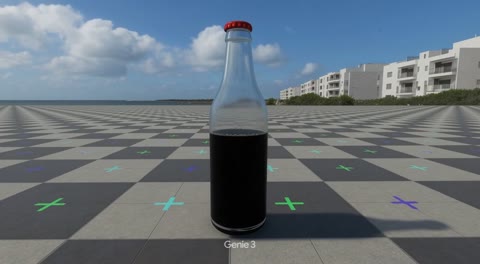}   \\
    \liqlabel{Project Genie}      \\
  \end{tabular}
  \end{adjustbox}
  \caption{\texttt{impulse\_decay} (brief vertical jump pulse, then
  brief lateral strafe pulse, idle otherwise; isolates an underdamped
  slosh response).  Columns: pre-jump, post-jump ring-down, post-strafe
  lateral slosh, and late decay.}
  \label{fig:liquid-strip-impulse-decay}
\end{figure}

\begin{figure}[t]
  \centering
  \begin{adjustbox}{max width=\textwidth}
  \setlength{\tabcolsep}{2pt}
  \renewcommand{\arraystretch}{0.6}
  \begin{tabular}{@{}cccc@{}}
    \includegraphics[width=4cm,keepaspectratio]{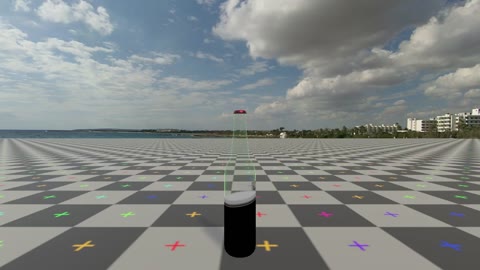}   & \includegraphics[width=4cm,keepaspectratio]{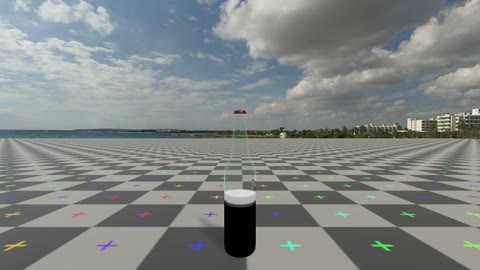}   & \includegraphics[width=4cm,keepaspectratio]{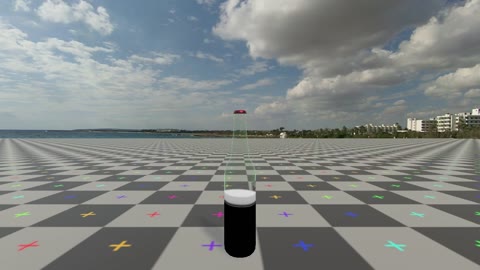}   & \includegraphics[width=4cm,keepaspectratio]{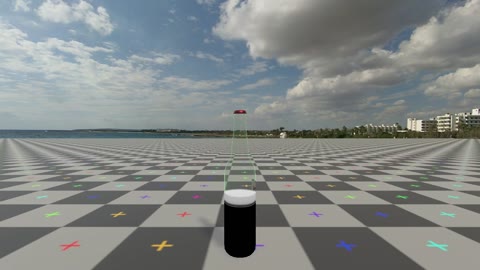}   \\
    \liqlabel{Unity}      \\[3pt]
    \includegraphics[width=4cm,keepaspectratio]{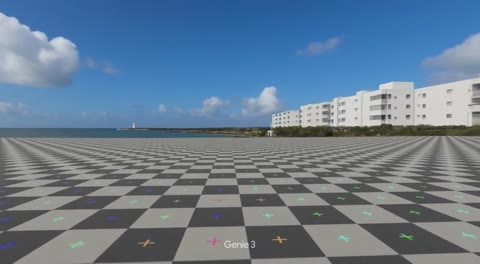}   & \includegraphics[width=4cm,keepaspectratio]{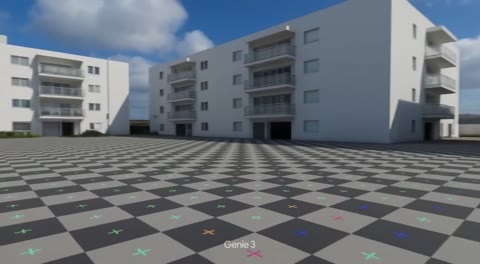}   & \includegraphics[width=4cm,keepaspectratio]{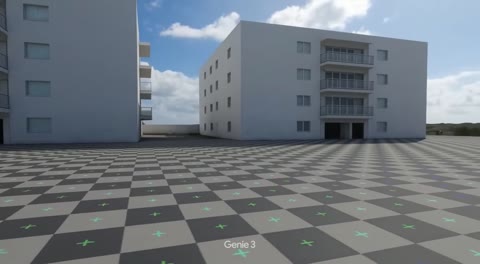}   & \includegraphics[width=4cm,keepaspectratio]{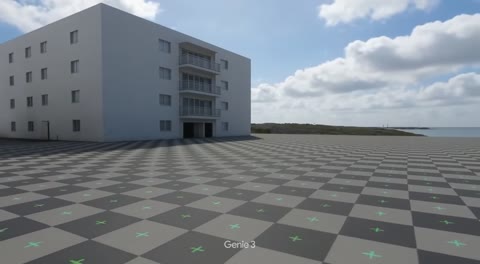}   \\
    \liqlabel{Project Genie}      \\
  \end{tabular}
  \end{adjustbox}
  \caption{\texttt{LiquidContainer}: \texttt{chirp\_strafe} (square-wave strafe sweep
  $0.3{\to}3.0$\,Hz; the only profile that excites the bottle's
  natural slosh modes).  Columns sample the low / mid-low / mid-high /
  high frequency regimes of the sweep.  Unity shows the cola
  surface visibly raised against the bottle wall by the high-frequency
  excitation.}
  \label{fig:liquid-strip-chirp-strafe}
\end{figure}


\subsection{Kinematics and Mechanical Laws}
\label{appx:mechanics_unity}
Table~\ref{tab:physical_laws_unity} reports the Unity reference for Table~\ref{tab:worldmodel_physical_laws}: each simulator video scored against itself. These values bound the measurement noise of the trajectory and mechanics pipeline, so model gaps should be read against them rather than against zero.

\begin{table}[ht]
\centering
\small
\caption{Unity reference for Table~\ref{tab:worldmodel_physical_laws}: the simulator video scored against itself, giving the error floor of the trajectory and mechanics measurement.}
\label{tab:physical_laws_unity}
\begin{tabular}{llcc}
\toprule
Category & Metric & Racing Car (3P) & Billiard Game (1P) \\
\midrule
\multirow{2}{*}{Kinematics}
& DTW $\downarrow$ & $0.94$ & $0.01$ \\
& FDE $\downarrow$ & $1.73$ & $0.01$ \\
Mechanical Laws
& Robust Error (\%) $\downarrow$ & $5.5$ & $4.9$ \\
\bottomrule
\end{tabular}
\end{table}

\subsection{Physical Causality}
\label{appx:physics_causality}

\subsubsection{Observations}
Three observations generalize beyond \texttt{BeamDomino}.
\begin{itemize}
    \item \textit{Composition degrades before perception.} The $\overline{\Delta}\text{ICRA}=+10.0$ vs.\ $\overline{\Delta}\text{CSR}=+66.7$ decomposition shows the world model answers $90\%$ of \emph{isolated} causal QAs (Unity: $100\%$) yet only $20\%$ of full trajectories survive the chain-walk. Compositional integrity, not single-frame perception, is where causality breaks down.
    \item \textit{Judge frame-sampling is a benchmark hyperparameter, not a cost knob.} On \texttt{direct-strike} segments spanning $0.12\,\text{s}$, $K=4$ frames per QA gives game engine CSR $53\%$ while $K=8$ recovers $87\%$ on the \emph{same} QAs. We therefore report game engine-reference scores alongside $\overline{\Delta}$ so judge-saturation cannot be misread as model failure.
    \item \textit{Action conditioning is a training-distribution claim, not an architecture claim.} LingBot-Act has a WASD channel of the right shape, but it was trained as a camera agent's first-person navigation intent and has no in-distribution mapping for forces on a non-camera body.
\end{itemize}

\subsubsection{Conditioning-mismatched Models}
\label{exp:physical_casuality_model_failures} 

LingBot-World (Cam) has no ball-force channel. Its \texttt{BeamDomino} generations were driven by a natural-language prompt describing the intended ball action rather than by the matched action trajectory, so they are not action-matched and are excluded from Table~\ref{tab:causal-physical-causality}.
%

LingBot-World (Act) exposes a WASD channel trained as \emph{the camera-agent's first-person navigation intent}: training examples pair WASD-key sequences with the corresponding camera translation through first-person environments. In \texttt{BeamDomino}'s third-person setup, that action embedding has no in-distribution mapping for $(F_x, F_z)$ on the floor ball.

Matrix-Game-2.0/3.0 exposes only a camera-control channel, so the continuous ball force has no place to be injected. LingBot-World (Cam), LingBot-World (Act), and Matrix-Game-2.0/3.0 are therefore all reported as ineligible in Table~\ref{tab:causal-physical-causality}.

\begin{table}[ht]
\centering
\caption{Trajectory-level CausalStep results on \texttt{BeamDomino}.
Rows split by (trajectory shape, model); each delta cell reports mean$\pm$std over $5$ unique generations sharing the same reference. Sign convention follows Table~\ref{tab:causal-physical-causality}: $\bar{\Delta}=m_{\textsc{unity}}-m_{\textsc{wm}}$ for all metrics.}
\label{tab:causality_per_traj}
\small
\begin{adjustbox}{max width=\textwidth}
\begin{tabular}{@{}llcccccc@{}}
\toprule
& & \multicolumn{1}{c}{\emph{Unity}}
& \multicolumn{3}{c}{\emph{Chain-level deltas} $\bar{\Delta}$}
& \multicolumn{2}{c}{\emph{Isolated-segment deltas}} \\
\cmidrule(lr){3-3} \cmidrule(lr){4-6} \cmidrule(lr){7-8}
Trajectory & Model & CSR
& $\bar{\Delta}$CSR & $\bar{\Delta}$AMCL & $\bar{\Delta}$WS
& $\bar{\Delta}$DUA & $\bar{\Delta}$ICRA \\
\midrule
\texttt{canonical}     & Project Genie         & 100.0 & $+40.0\pm49.0$  & $+0.40\pm0.49$ & $+1.60\pm1.96$ & $-5.0\pm18.7$  & $\phantom{+}0.0\pm 0.0$ \\
\texttt{direct\_strike}& Project Genie         & 100.0 & $+100.0\pm\phantom{0}0.0$  & $+1.40\pm0.49$ & $+3.80\pm0.98$ & $+46.7\pm16.3$ & $+30.0\pm24.5$ \\
\texttt{glancing}      & Project Genie         & \phantom{0}60.0 & $+60.0\pm49.0$  & $+0.60\pm1.02$ & $+2.40\pm3.50$ & $\phantom{+}0.0\pm15.8$  & $\phantom{+}0.0\pm 0.0$ \\
\bottomrule
\end{tabular}
\end{adjustbox}
\end{table}

\subsubsection{Trajectory decomposition}
Table~\ref{tab:causality_per_traj} breaks the headline $\bar{\Delta}\textit{CSR}=66.7$ for Project Genie into the three trajectory shapes, which stress different parts of the \textsc{rolling}$\rightarrow$\textsc{impact}$\rightarrow$\textsc{topple}$\rightarrow$\textsc{cascade} chain.

The chain-level gap is monotonic in chain depth: Genie loses the cascade on every \texttt{direct\_strike} seed ($\bar{\Delta}\textit{CSR}=+100$, std~$=0$), on three of five \texttt{glancing} seeds ($+60$), and on two of five \texttt{canonical} seeds ($+40$).


\subsubsection{Limitations}
World-model generations that diverge substantially from the reference scene initial conditions, in which the canonical stages are not visually present, are reported separately and excluded from the headline aggregate.

CausalStep QAs probe the causal structure rather than object identity, so a generation with the right \emph{type} of motion but wrong objects can score close to reference scenes on chain-walk metrics. We recommend reading Table~\ref{tab:causal-physical-causality} jointly with Causal LPIPS on the same scene (Table~\ref{tab:worldmodel_causal_lpips}) to disambiguate content-vs-causal failures.

\paragraph{Judge Frame-Sampling}
\label{app:causality_K_sampling}

CausalStep conditions the VLM judge on $K$ uniformly sampled frames per QA from the corresponding segment window
(Appendix~\ref{app:physcaus}). We use $K=8$ throughout the main text.

On tight segments such as the \texttt{direct\_strike} cascade, where the relevant motion spans only $\sim 0.12$\,s, the judge's competence on the Unity reference rises sharply with $K$: the Unity \textit{CSR} on the same QAs is $53\%$ at $K=4$ and $87\%$ at $K=8$. We therefore treat $K$ as a benchmark hyper-parameter and tune it upward whenever the Unity reference is itself the bottleneck, and report the game-engine reference scores at the chosen $K$ alongside $\bar{\Delta}$ so that judge saturation is not mis-read as a model gap.

\paragraph{Reference-side difficulty on \texttt{glancing}.}
The game engine reference itself only attains $\textit{CSR}=60.0$ on \texttt{glancing} under our $K=8$ judge, because the \textsc{glance} sub-event the ball clipping the leg of the stack without toppling it is hard for the VLM to resolve from eight sampled frames even on the simulator-rendered video.

Part of the $+60$ gap on this shape is therefore reference-side judging difficulty rather than pure world-model error. We leave Genie's \texttt{glancing} number in the headline aggregate to keep the trajectory-shape weighting balanced, but flag it here so that $\bar{\Delta}\textit{CSR}$ on \texttt{glancing} is not over-interpreted. The released \capsysnamespace Lite bundle repairs four reference-side QA items, which raises this reference CSR to $80\%$; numbers reported here use the original items.

\begin{figure*}[t]
\centering
\begin{adjustbox}{max width=\linewidth}
\setlength{\tabcolsep}{1pt}
\renewcommand{\arraystretch}{0.85}
\begin{tabular}{@{}ccccc@{}}
\small\textsc{rolling}
 & \small\textsc{impact}
 & \small\textsc{topple}
 & \small\textsc{cascade}
 & \small terminal \\
\includegraphics[width=3.2cm]{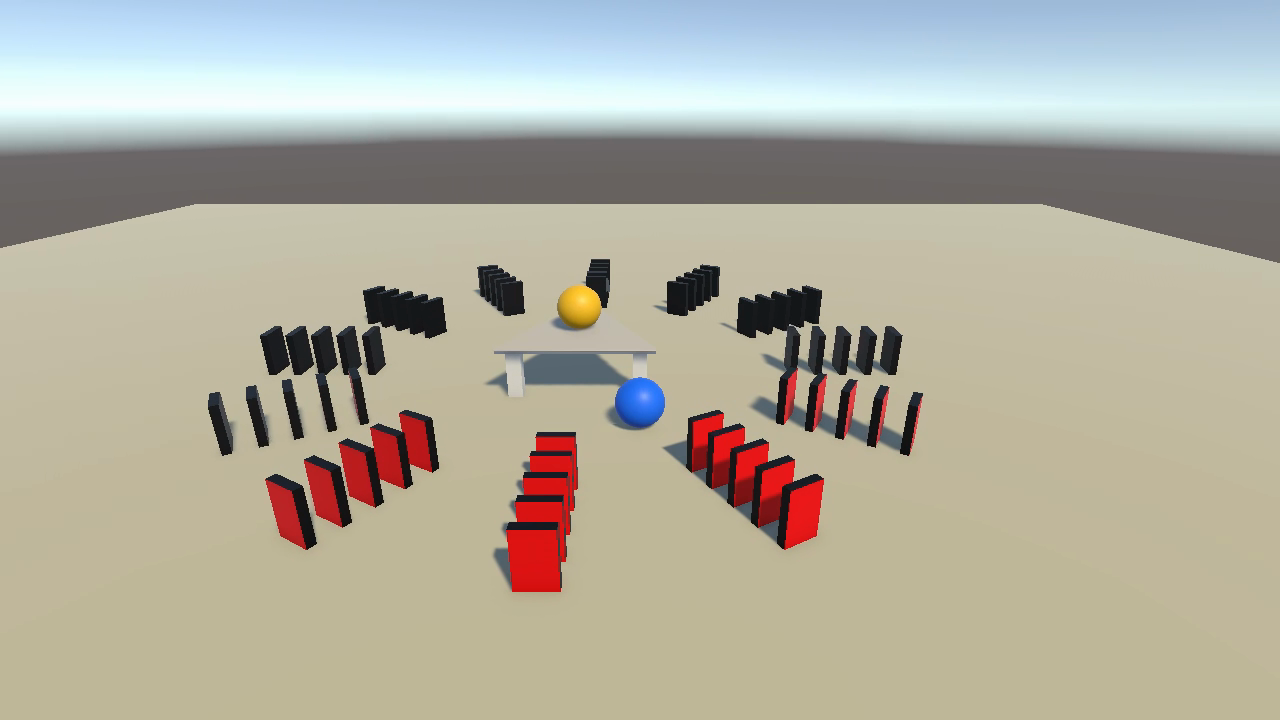}
  & \includegraphics[width=3.2cm]{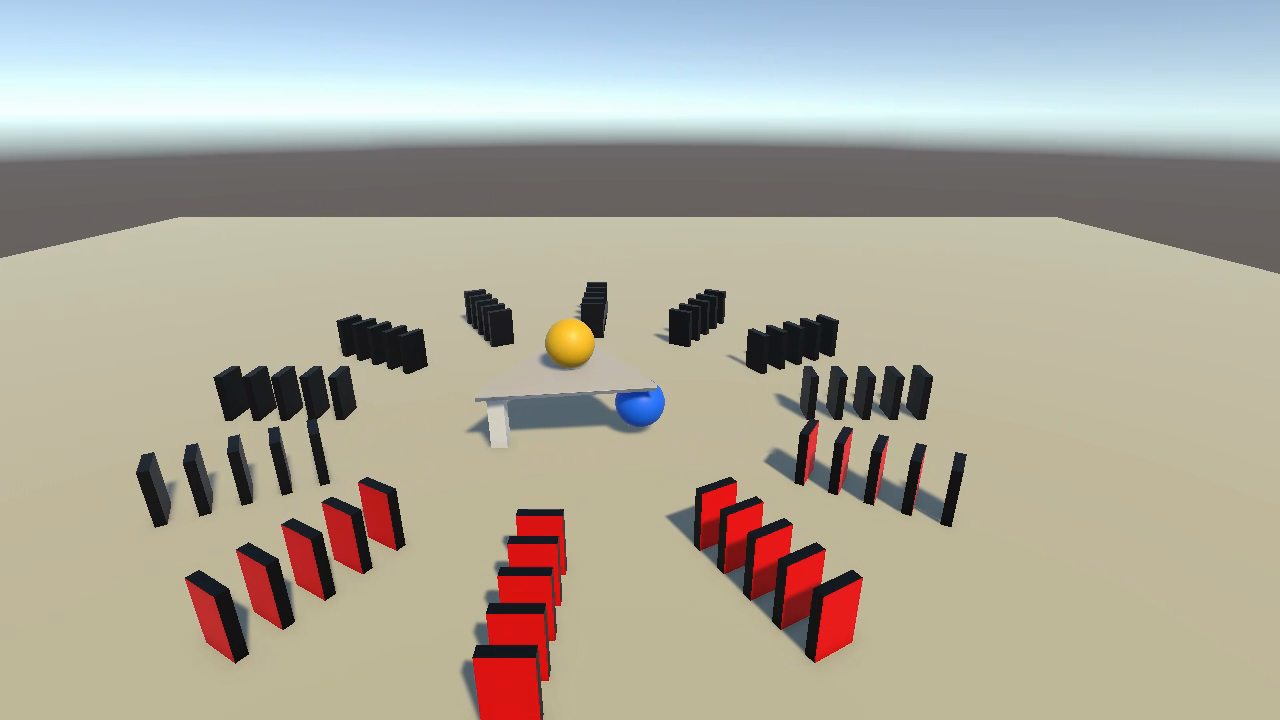}
  & \includegraphics[width=3.2cm]{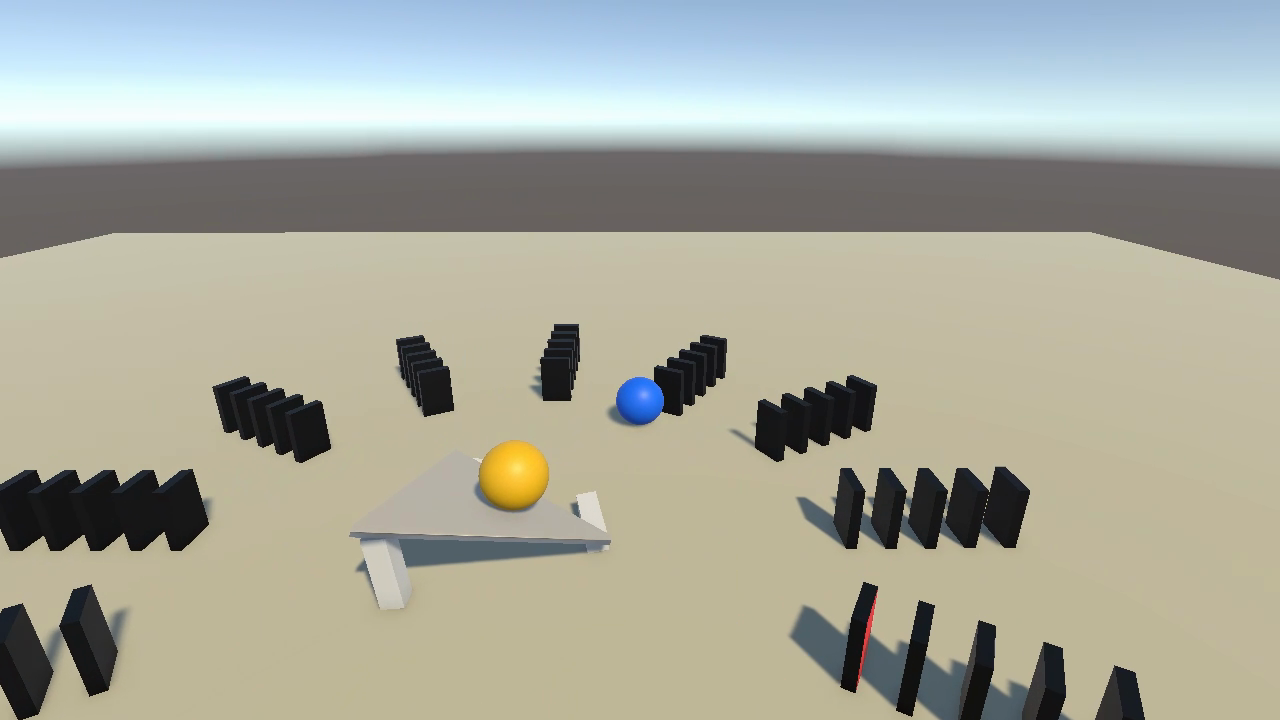}
  & \includegraphics[width=3.2cm]{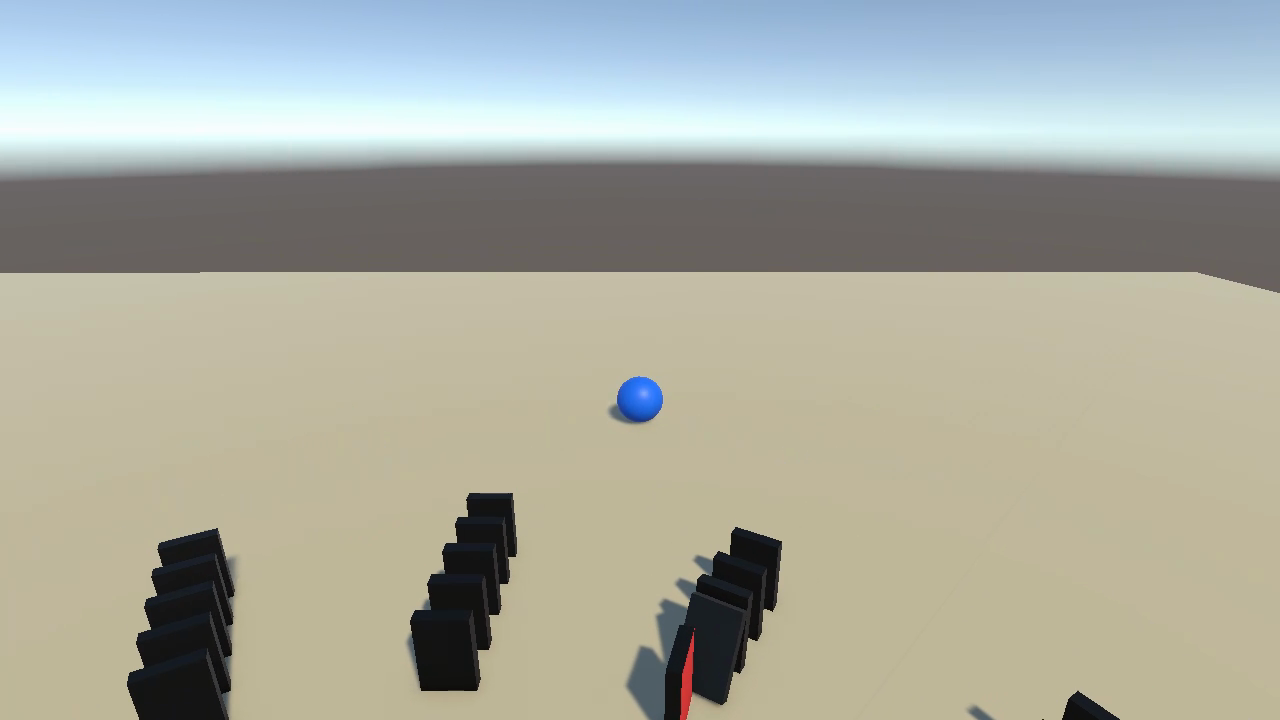}
  & \includegraphics[width=3.2cm]{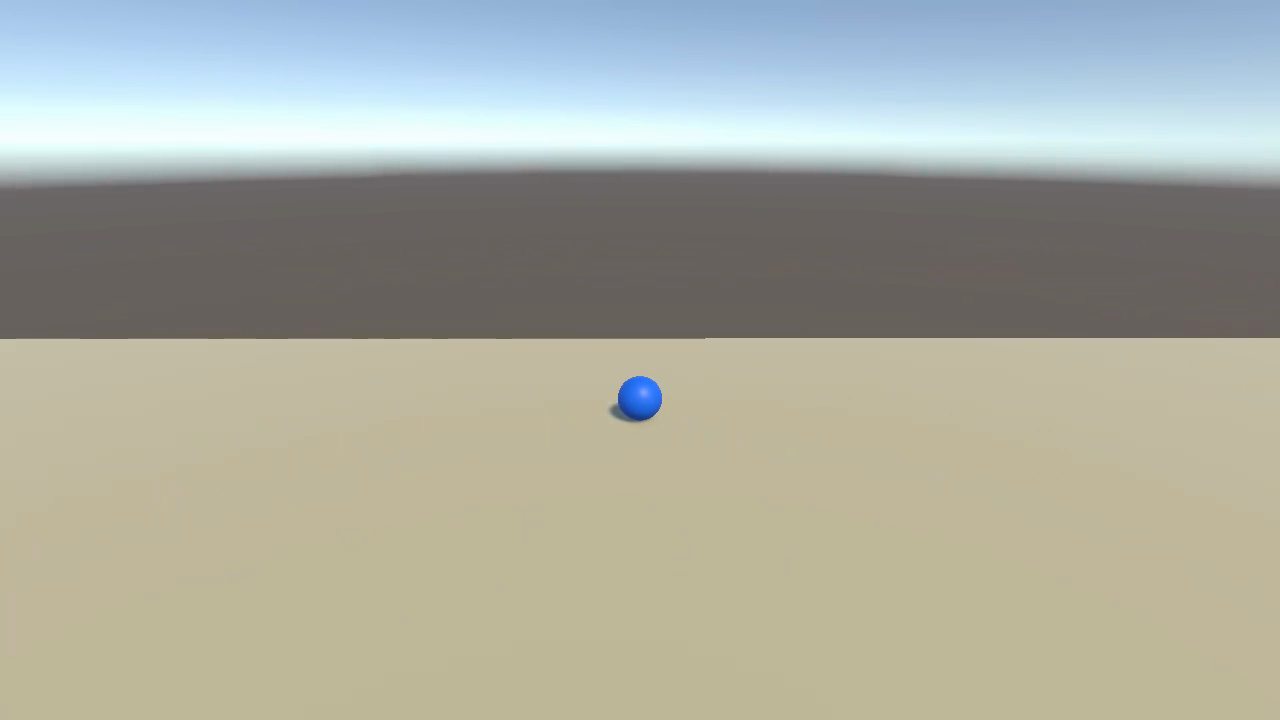} \\
\multicolumn{5}{c}{\small Unity} \\[4pt]
\includegraphics[width=3.2cm]{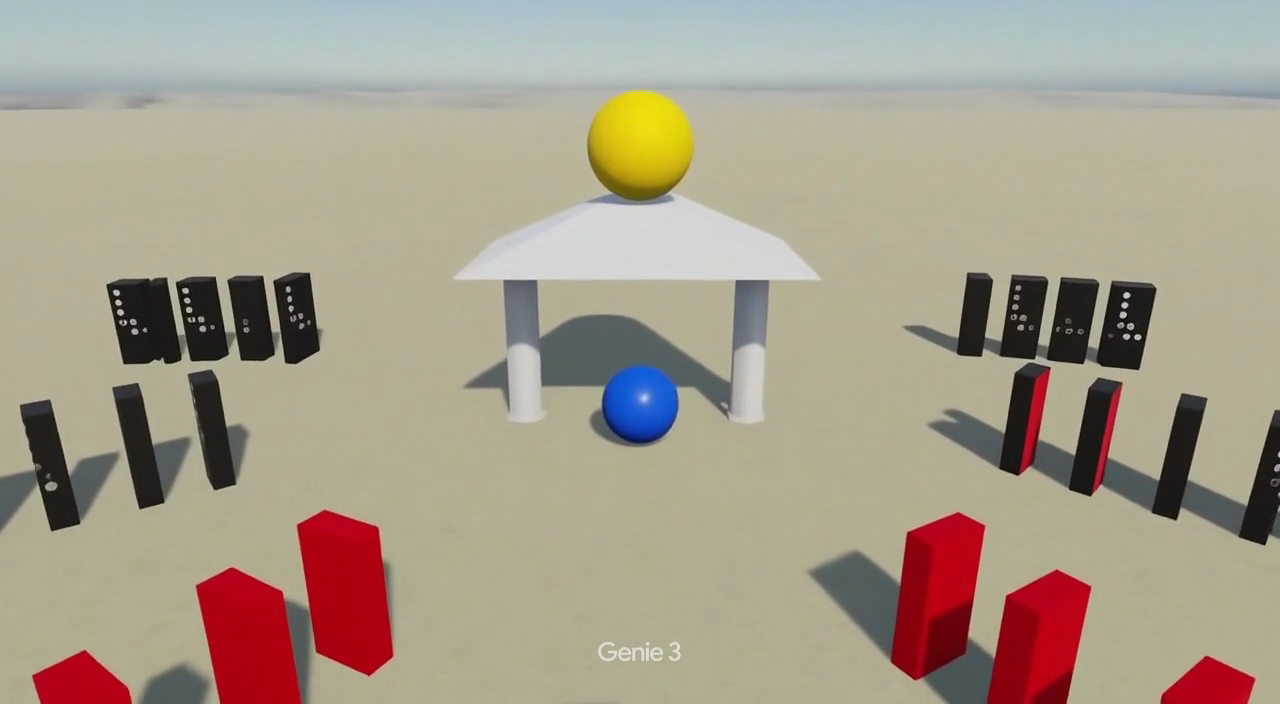}
  & \includegraphics[width=3.2cm]{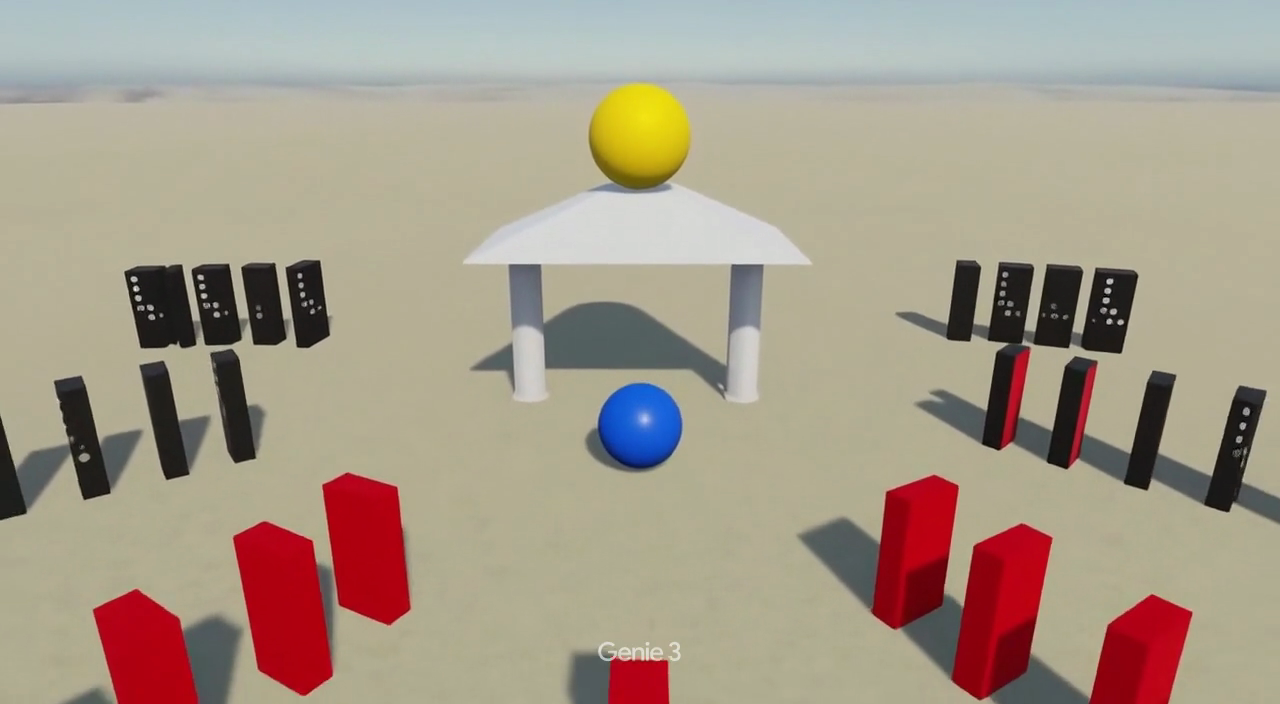}
  & \includegraphics[width=3.2cm]{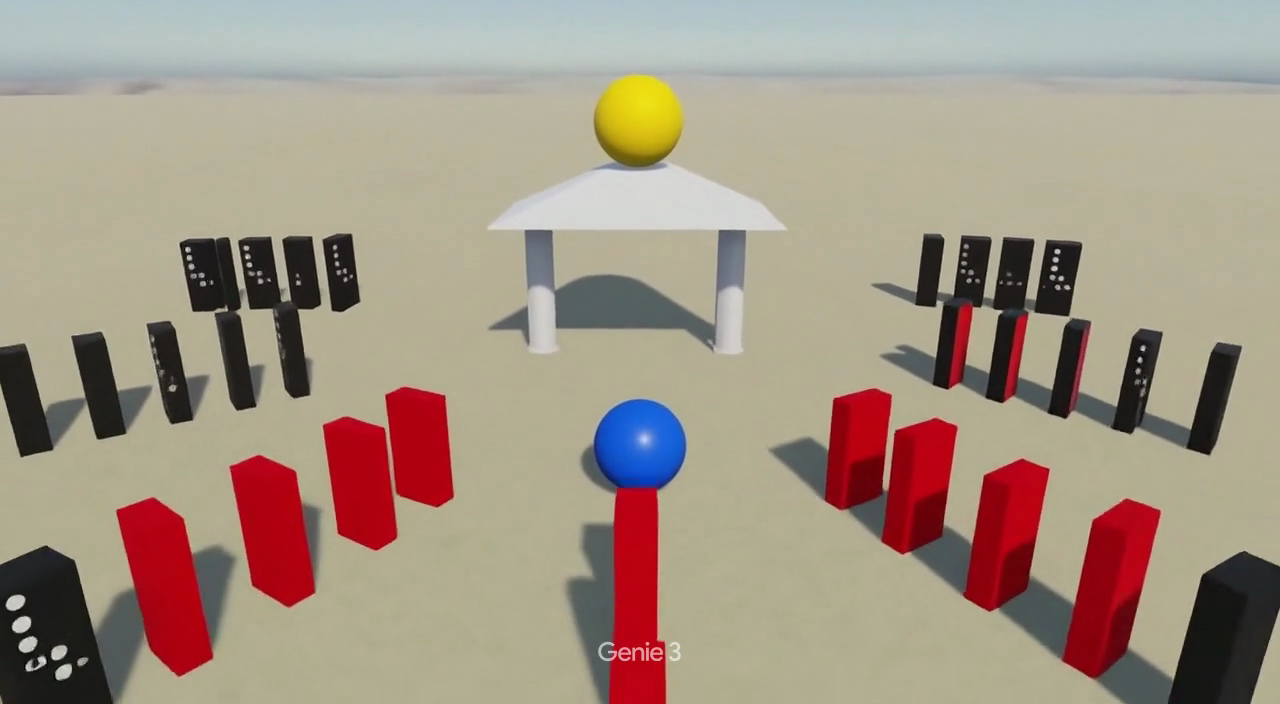}
  & \includegraphics[width=3.2cm]{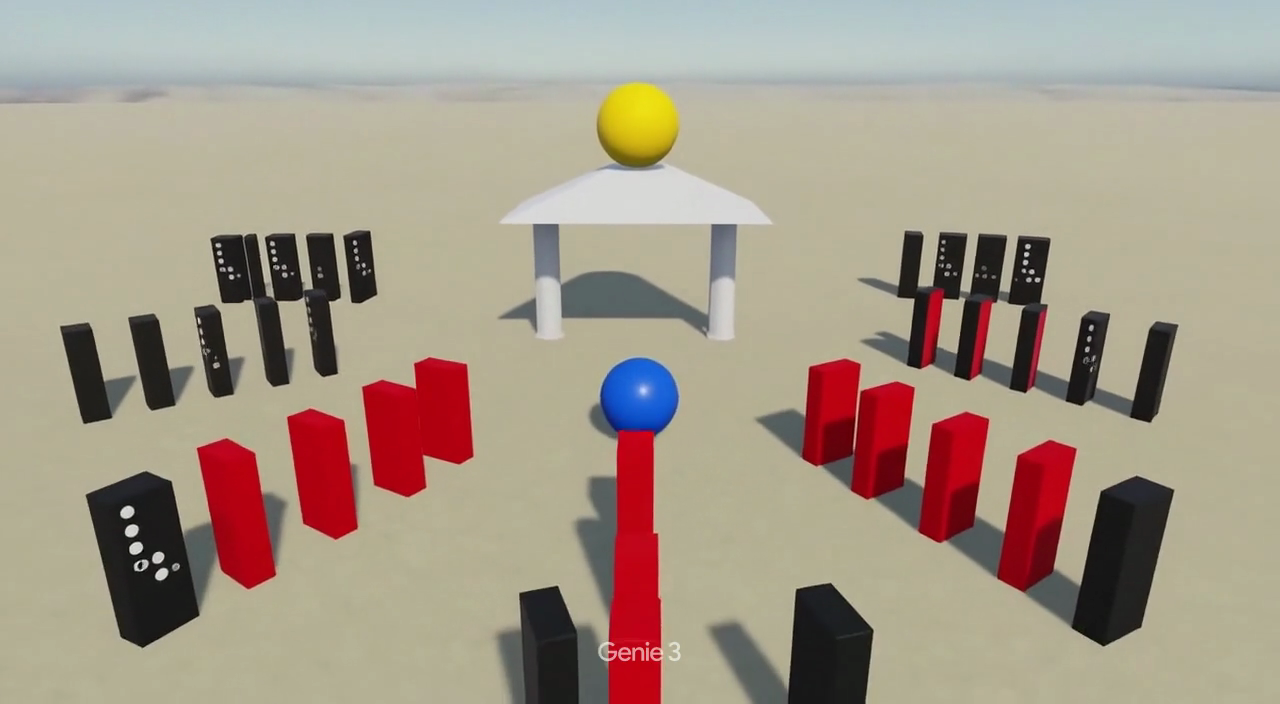}
  & \includegraphics[width=3.2cm]{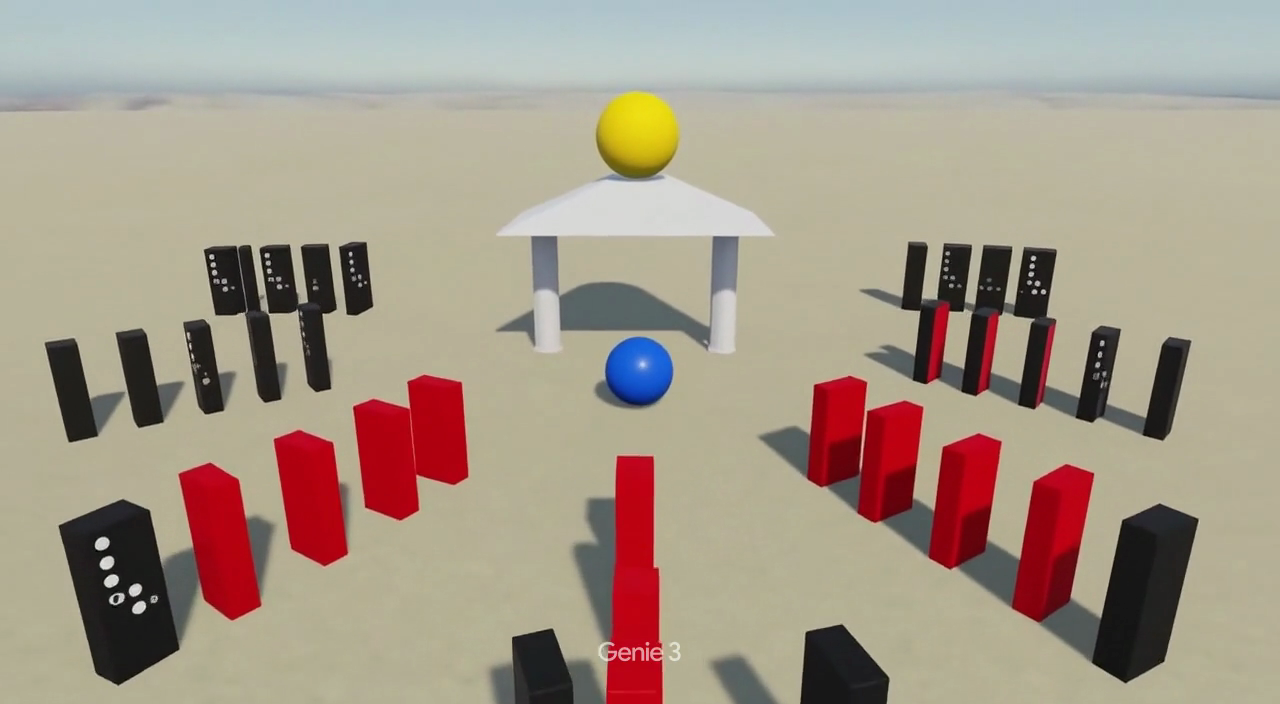} \\
\multicolumn{5}{c}{\small Project Genie} \\[4pt]
\end{tabular}
\end{adjustbox}
\caption{\texttt{BeamDomino}: Stage-aligned keyframes on the \texttt{canonical} action trajectory.}
\label{fig:causal_frames_canonical}
\end{figure*}

\begin{figure*}[t]
\centering
\begin{adjustbox}{max width=\linewidth}
\setlength{\tabcolsep}{1pt}
\renewcommand{\arraystretch}{0.85}
\begin{tabular}{@{}ccccc@{}}
\small pre-motion
 & \small\textsc{rolling}
 & \small\textsc{strike}
 & \small\textsc{cascade}
 & \small terminal \\
\includegraphics[width=3.2cm]{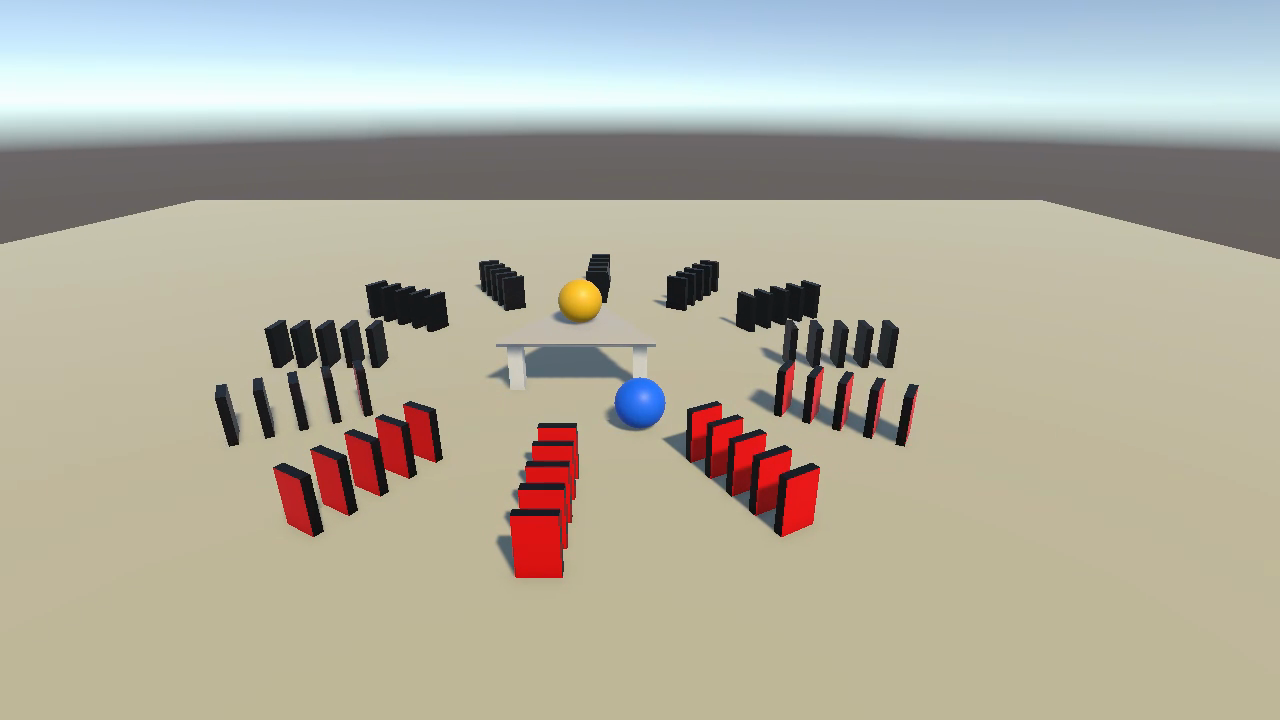}
  & \includegraphics[width=3.2cm]{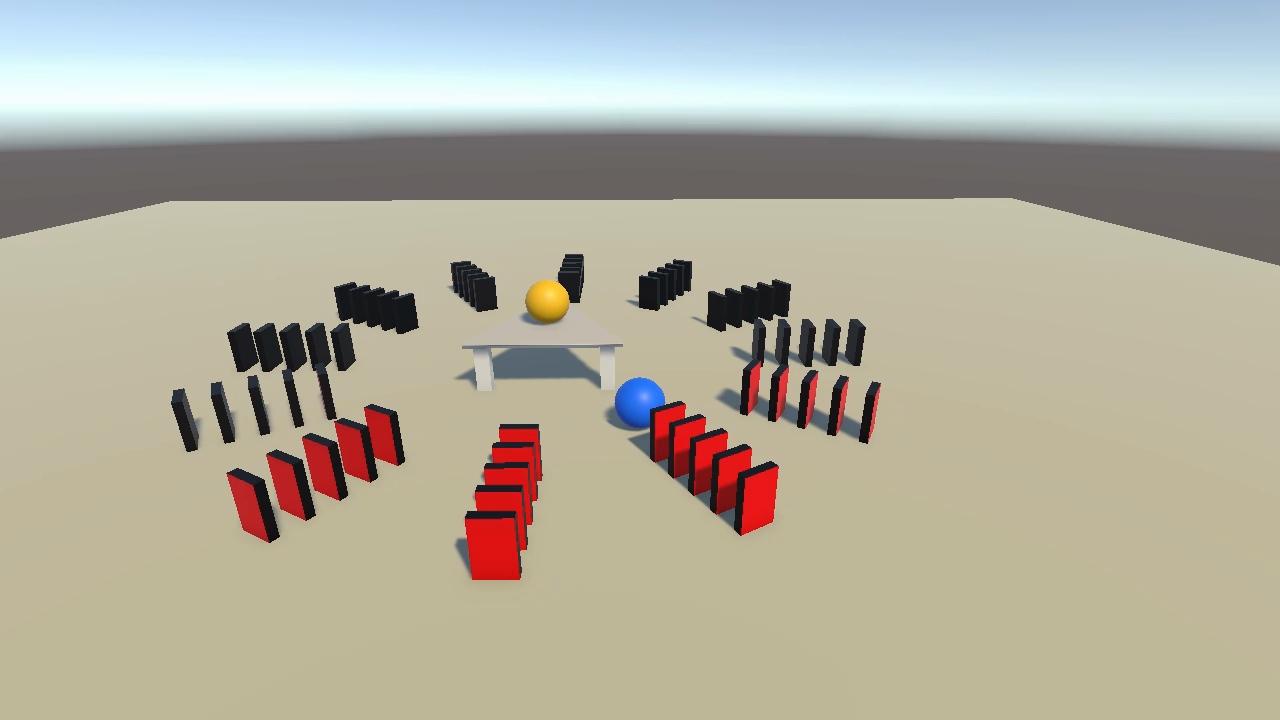}
  & \includegraphics[width=3.2cm]{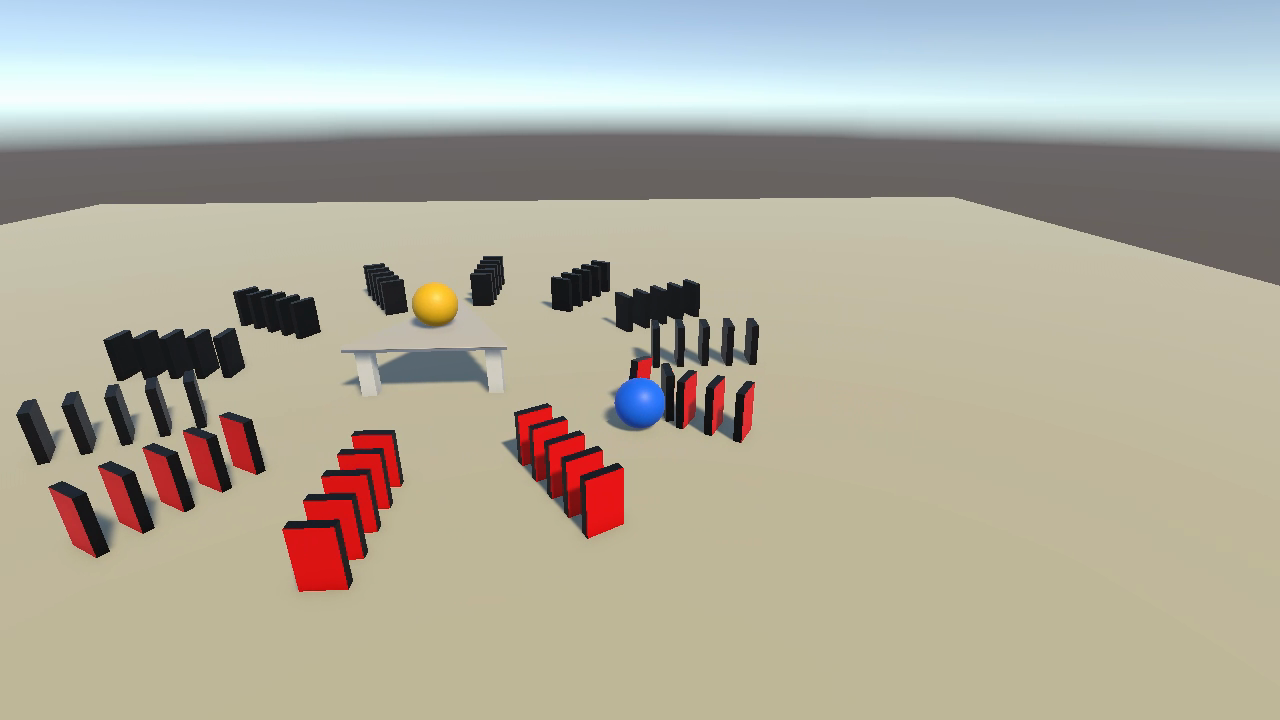}
  & \includegraphics[width=3.2cm]{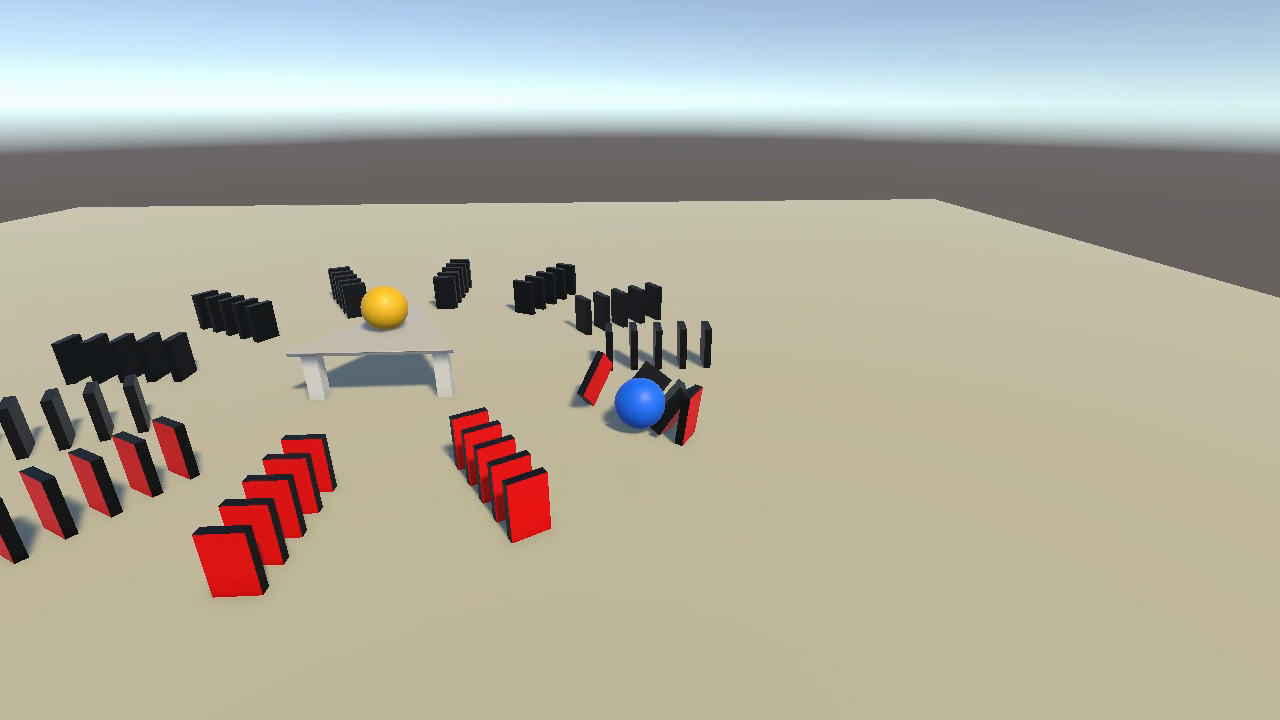}
  & \includegraphics[width=3.2cm]{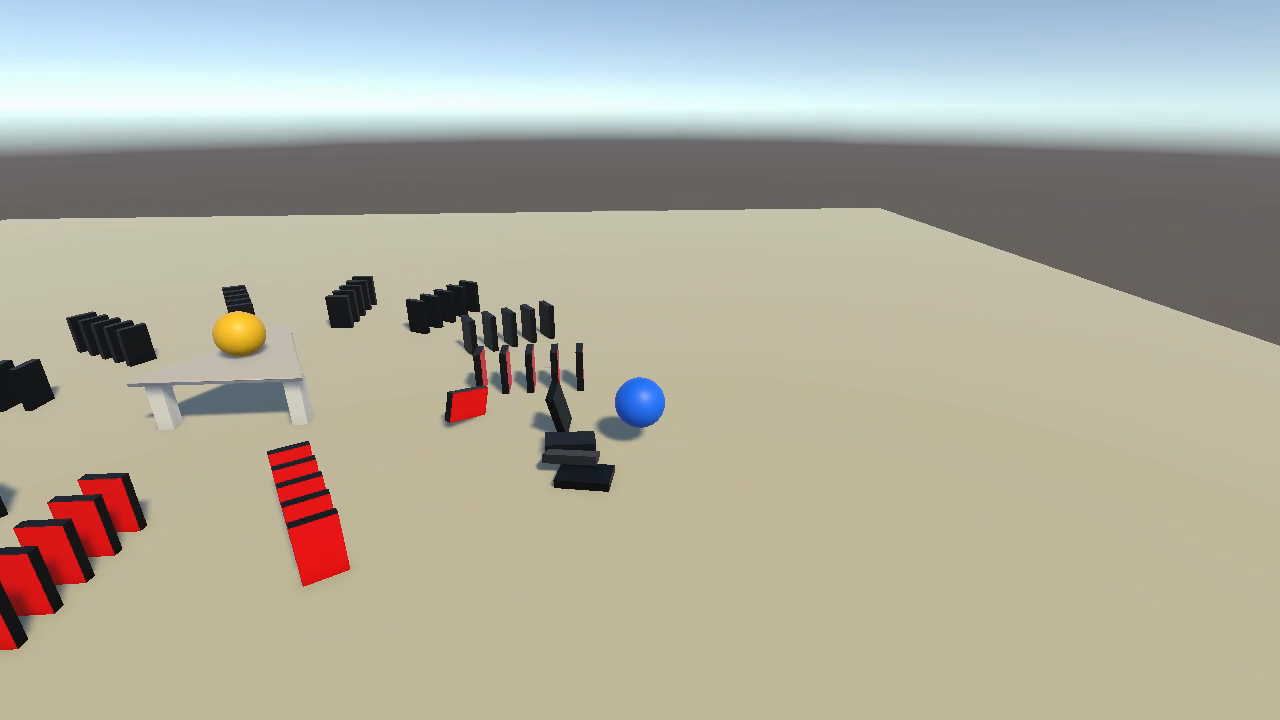} \\
\multicolumn{5}{c}{\small Unity} \\[4pt]
\includegraphics[width=3.2cm]{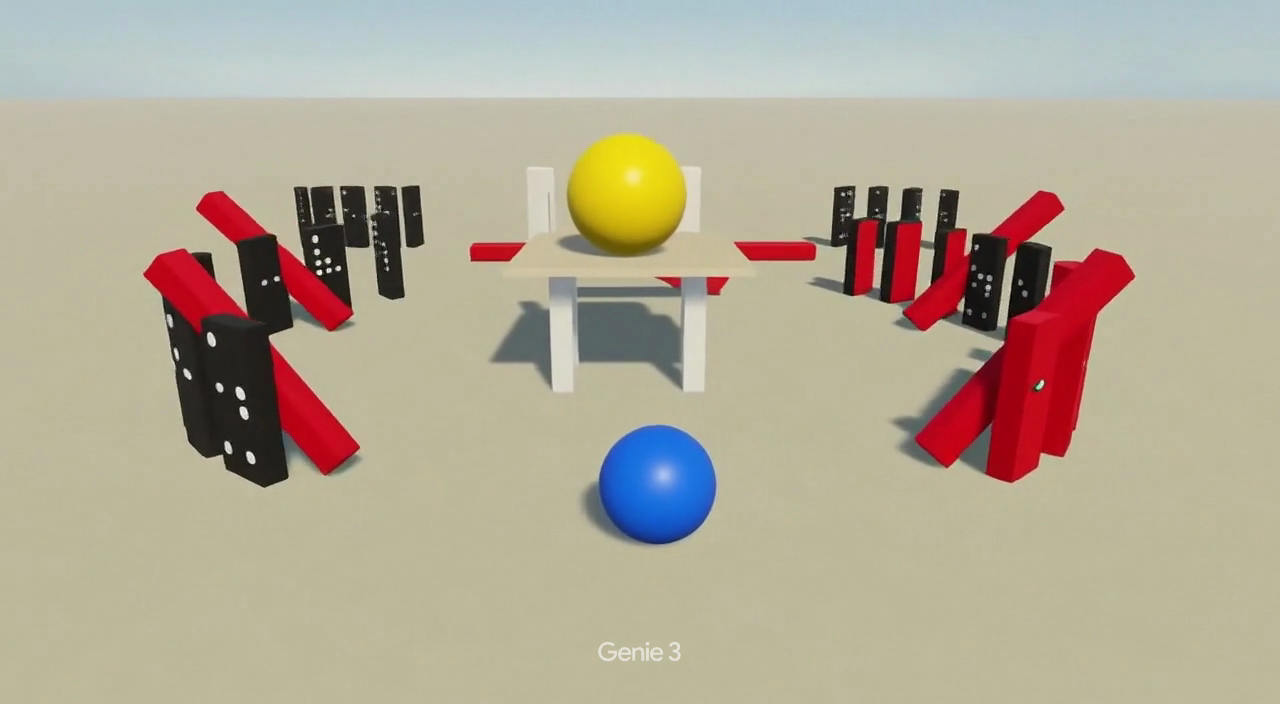}
  & \includegraphics[width=3.2cm]{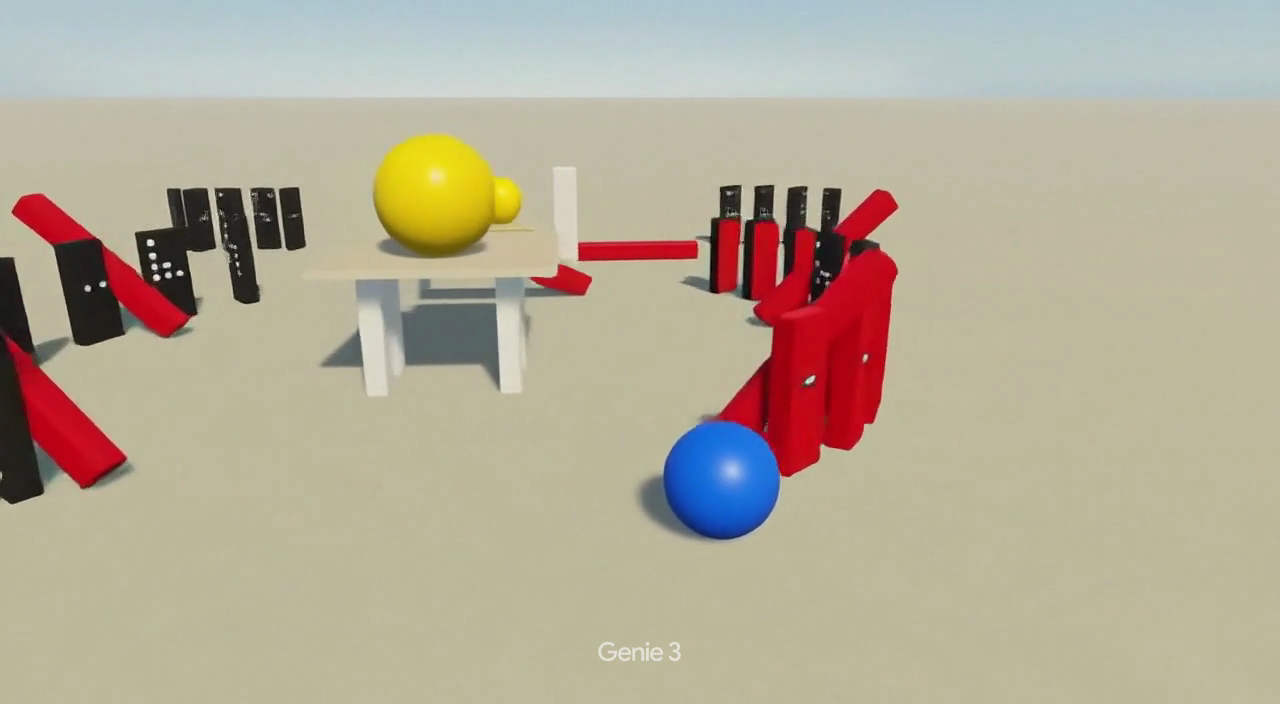}
  & \includegraphics[width=3.2cm]{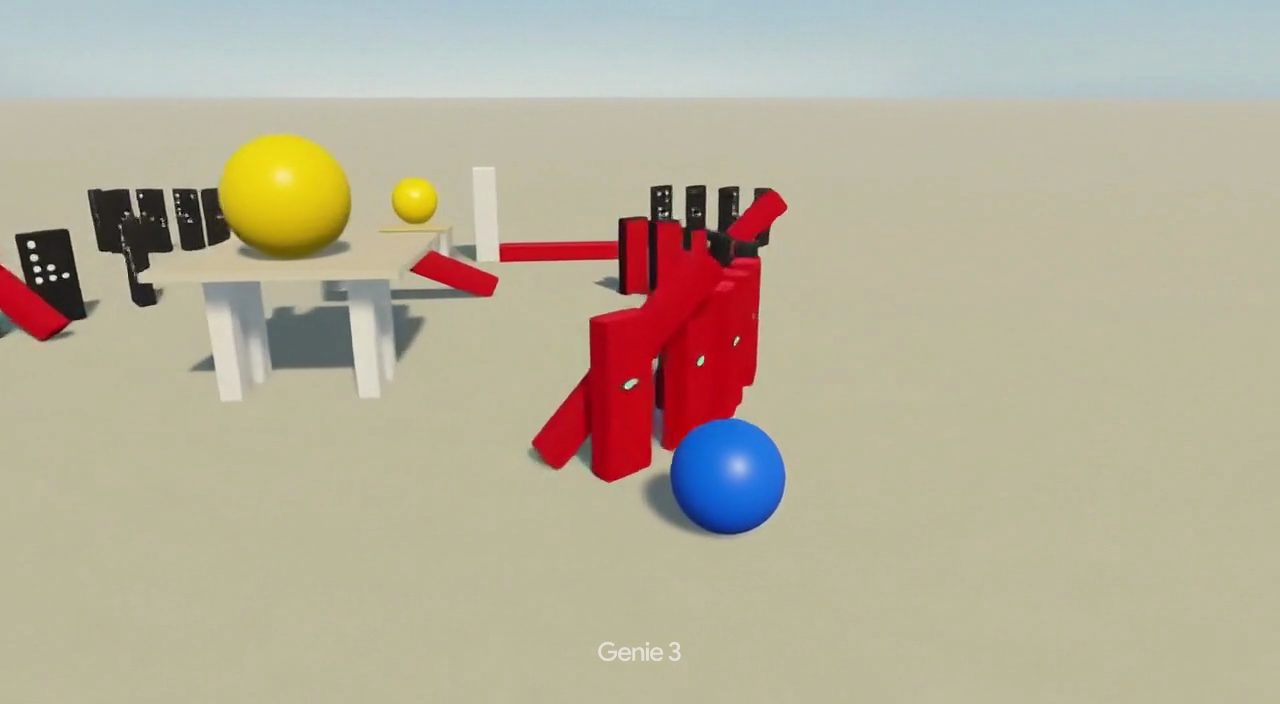}
  & \includegraphics[width=3.2cm]{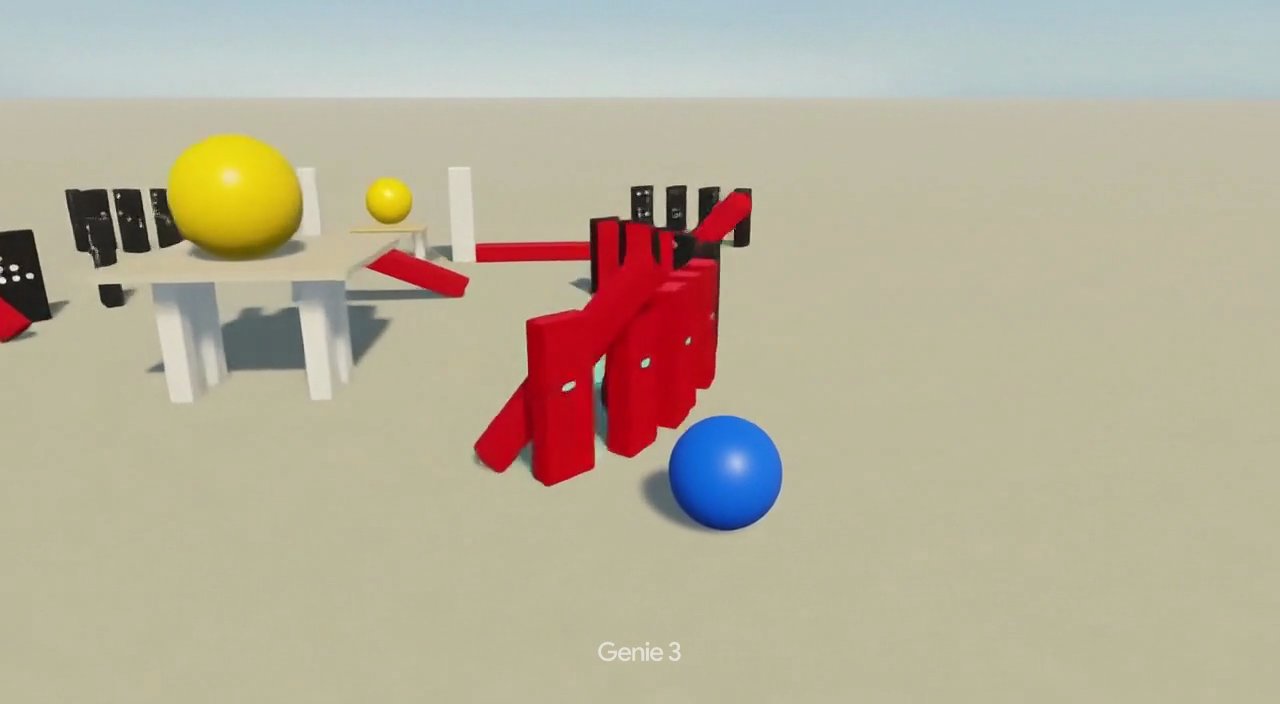}
  & \includegraphics[width=3.2cm]{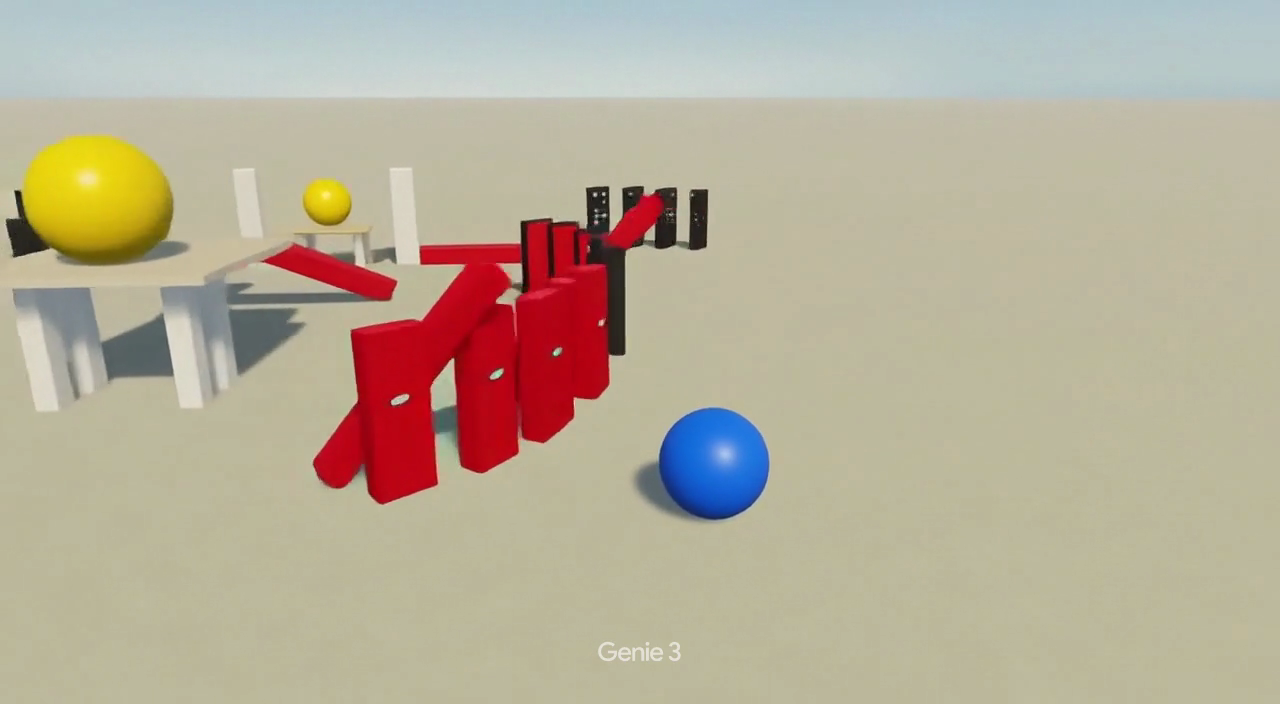} \\
\multicolumn{5}{c}{\small Project Genie} \\[4pt]
\end{tabular}
\end{adjustbox}
\caption{\texttt{BeamDomino}: Stage-aligned keyframes on the \texttt{direct\_strike} action trajectory.}
\label{fig:causal_frames_direct_strike}
\end{figure*}

\begin{figure*}[t]
\centering
\begin{adjustbox}{max width=\linewidth}
\setlength{\tabcolsep}{1pt}
\renewcommand{\arraystretch}{0.85}
\begin{tabular}{@{}ccccc@{}}
\small\textsc{rolling}
 & \small\textsc{glance}
 & \small\textsc{strike}
 & \small\textsc{cascade}
 & \small terminal \\
\includegraphics[width=3.2cm]{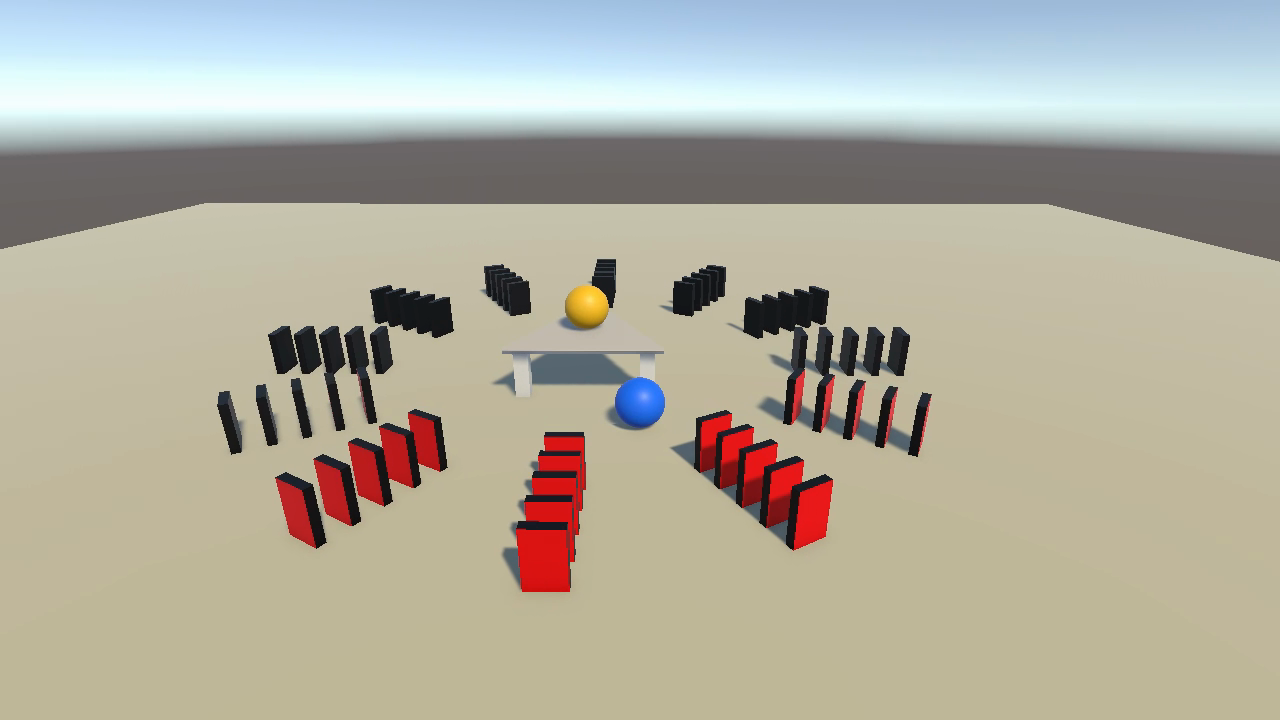}
  & \includegraphics[width=3.2cm]{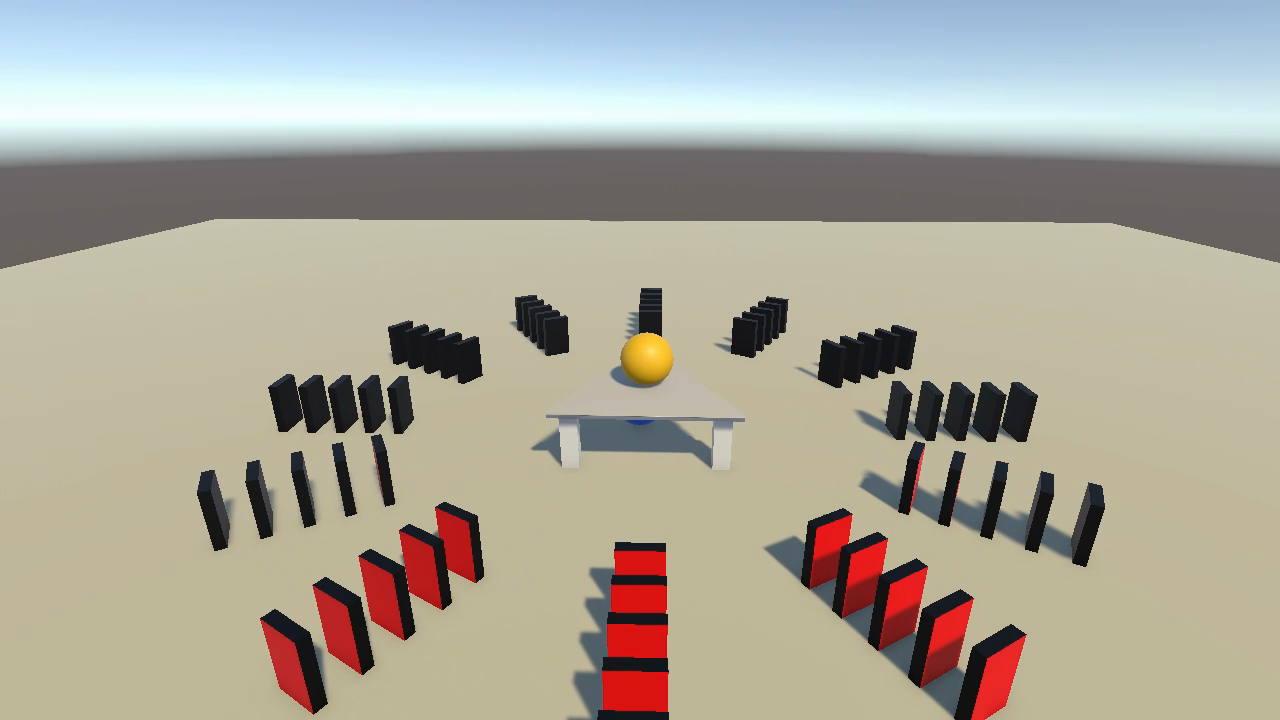}
  & \includegraphics[width=3.2cm]{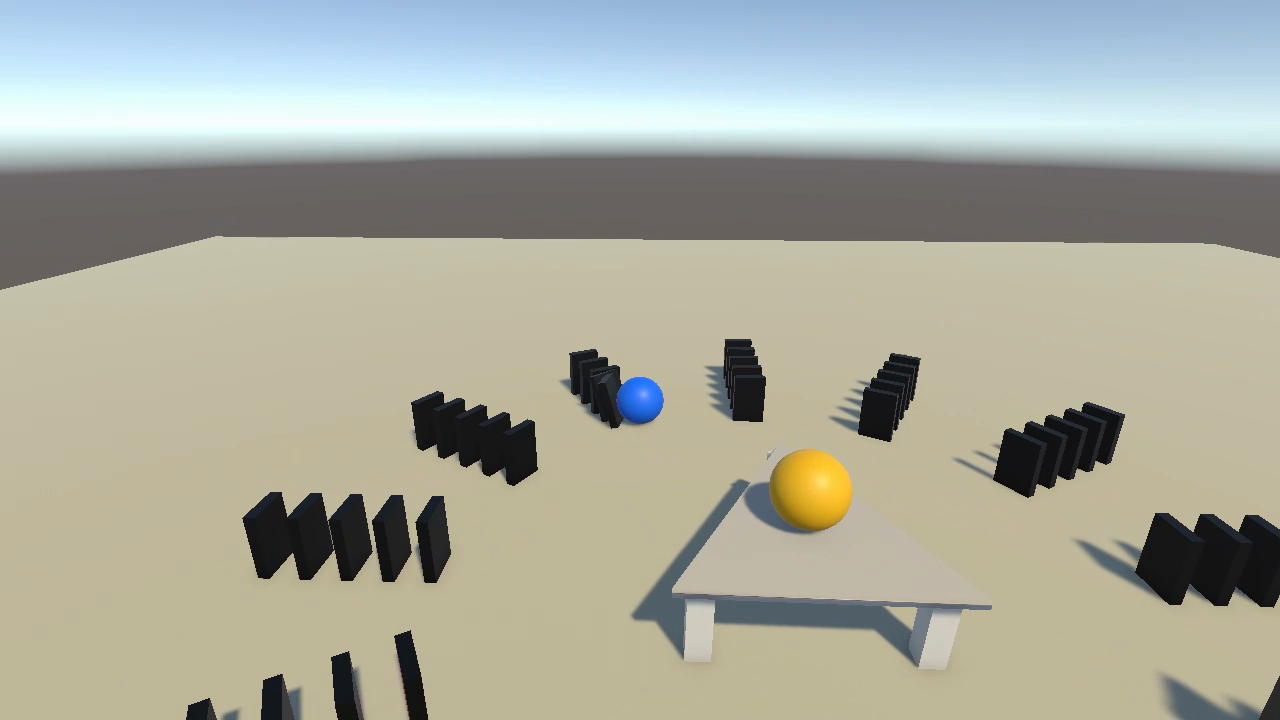}
  & \includegraphics[width=3.2cm]{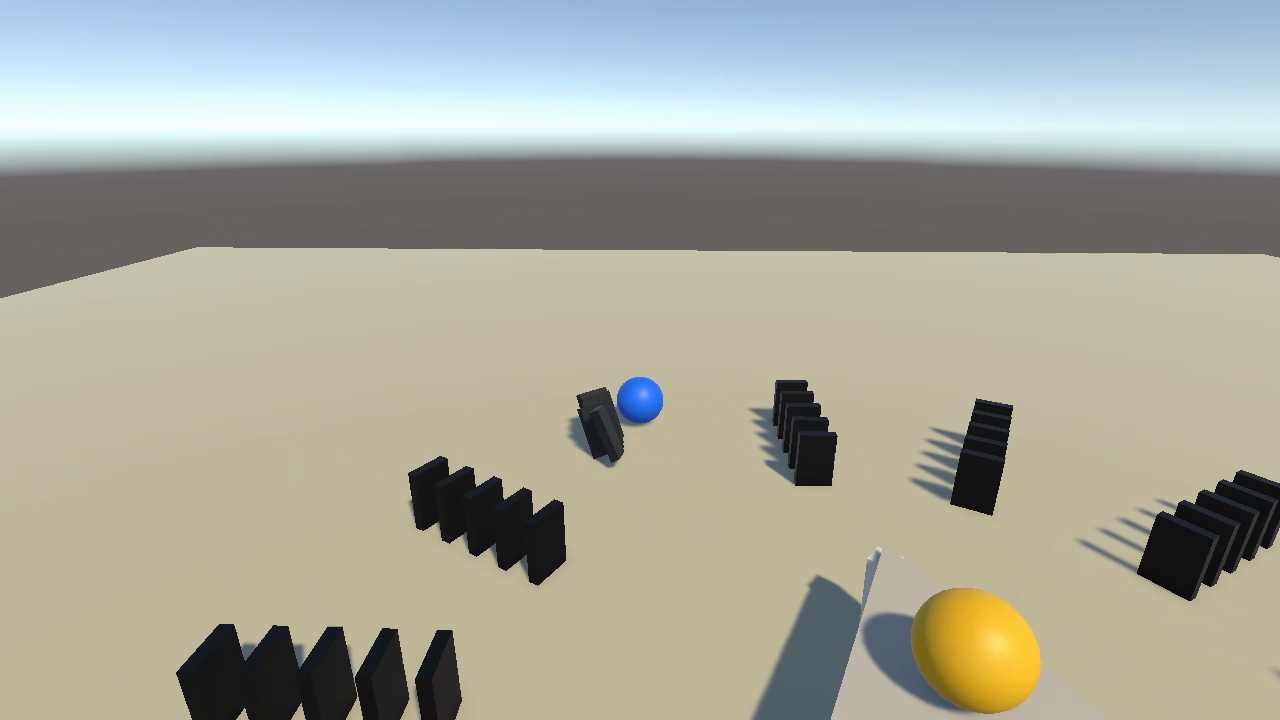}
  & \includegraphics[width=3.2cm]{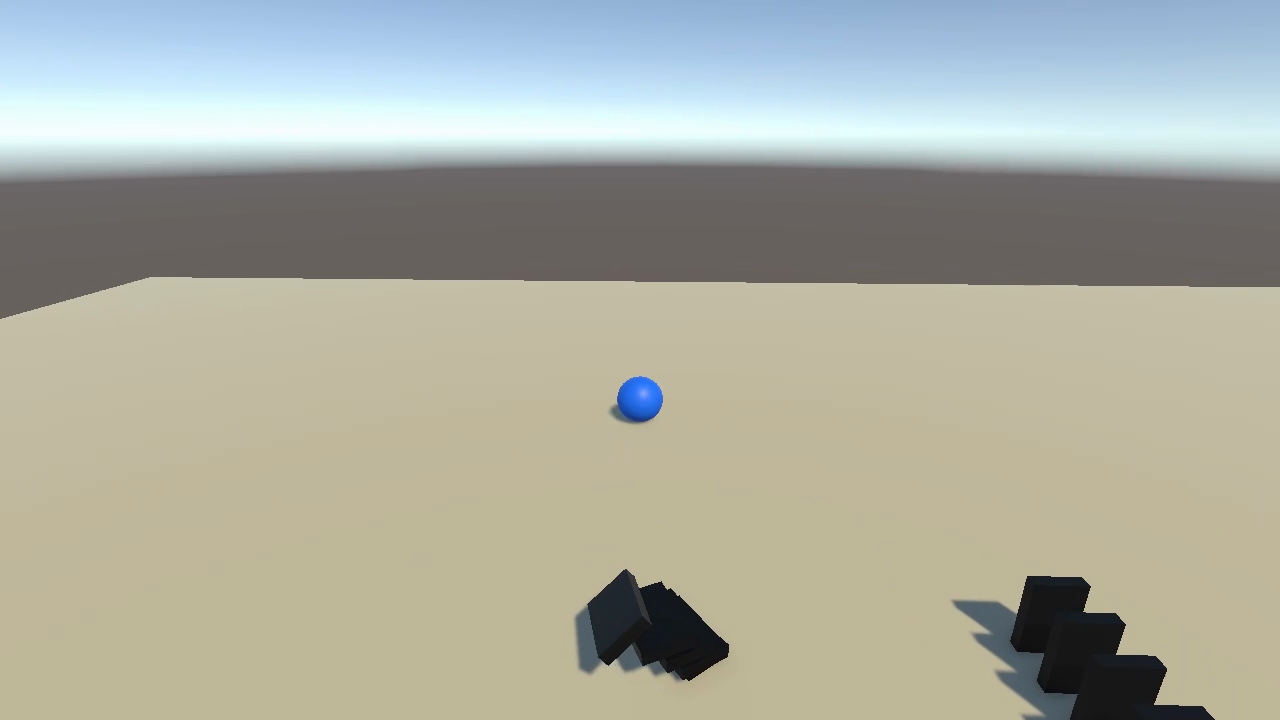} \\
\multicolumn{5}{c}{\small Unity} \\[4pt]
\includegraphics[width=3.2cm]{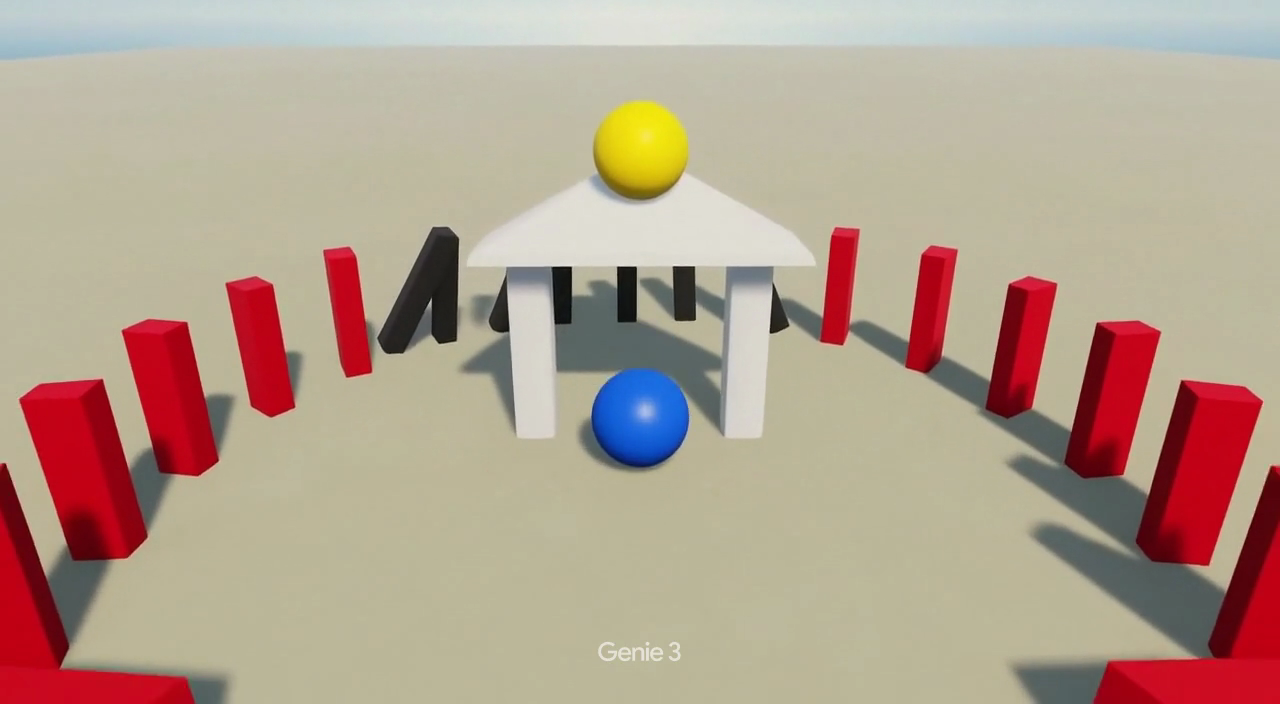}
  & \includegraphics[width=3.2cm]{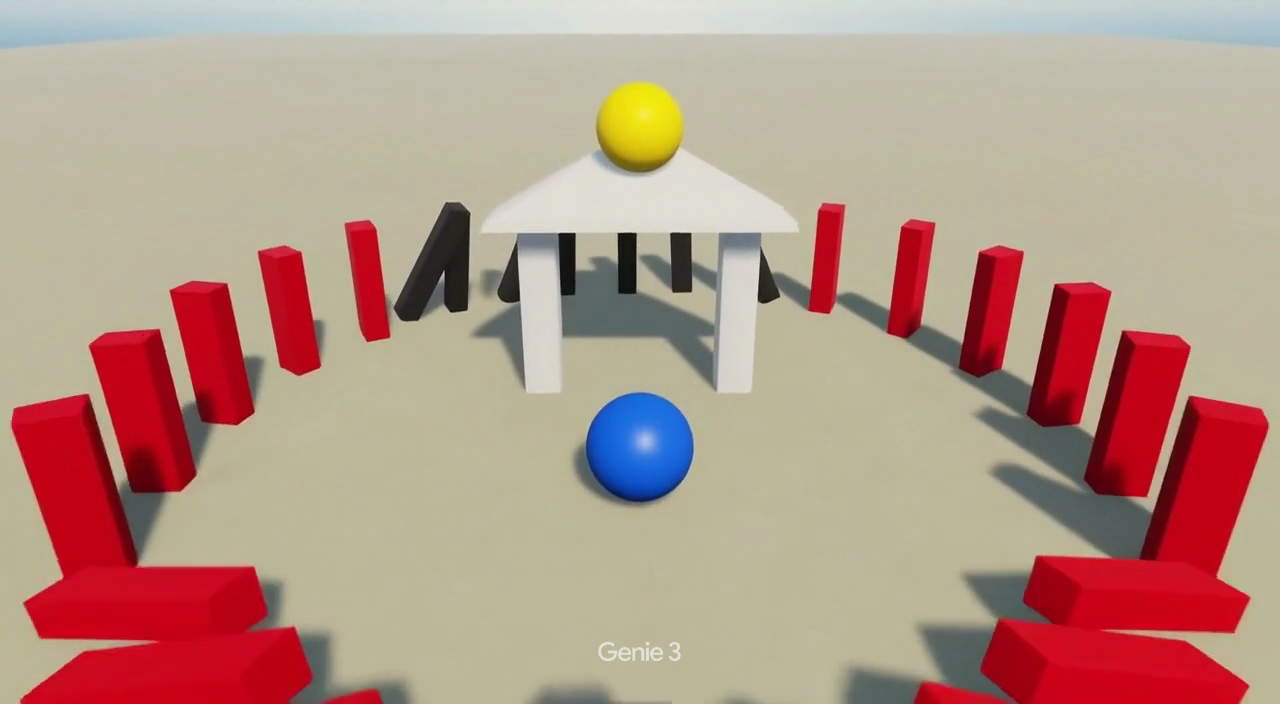}
  & \includegraphics[width=3.2cm]{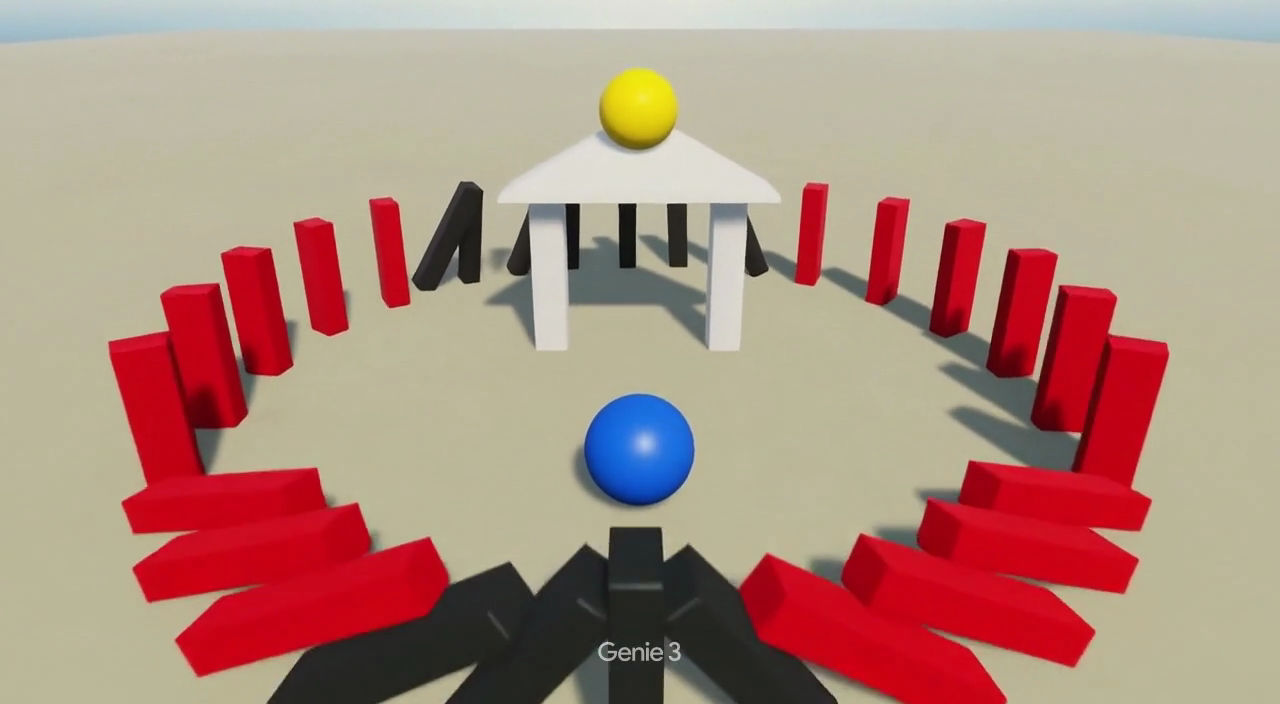}
  & \includegraphics[width=3.2cm]{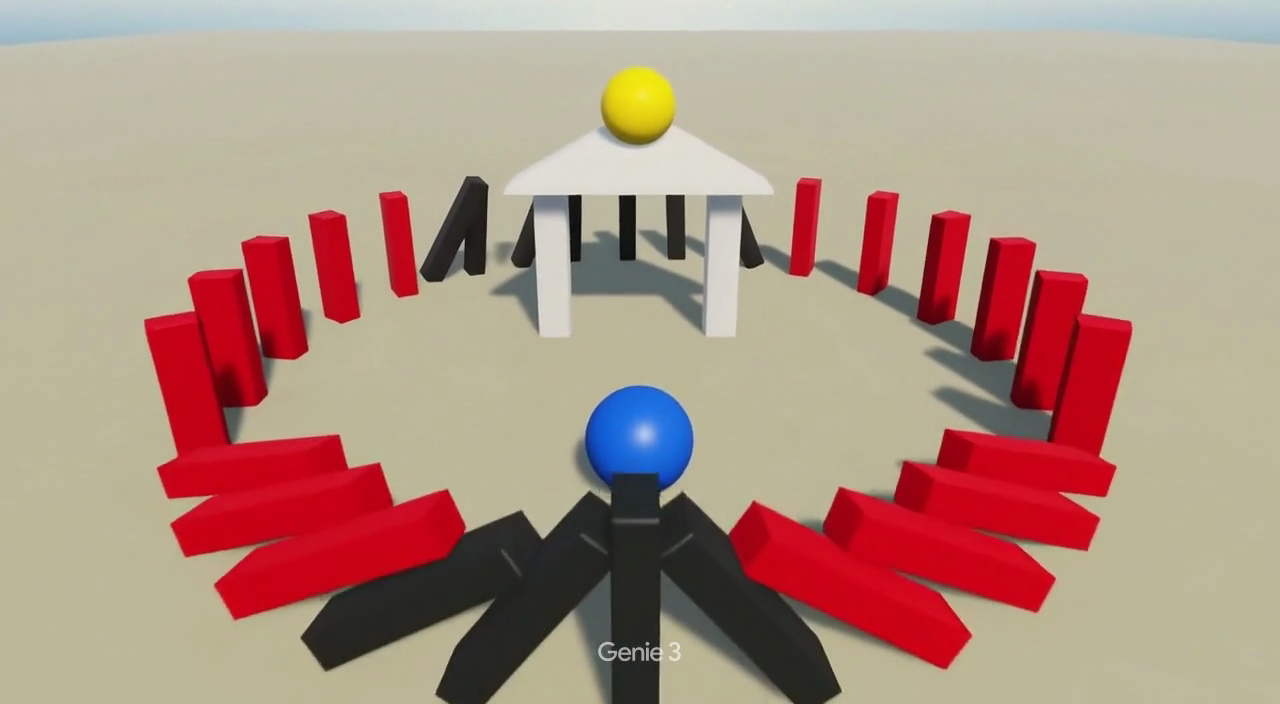}
  & \includegraphics[width=3.2cm]{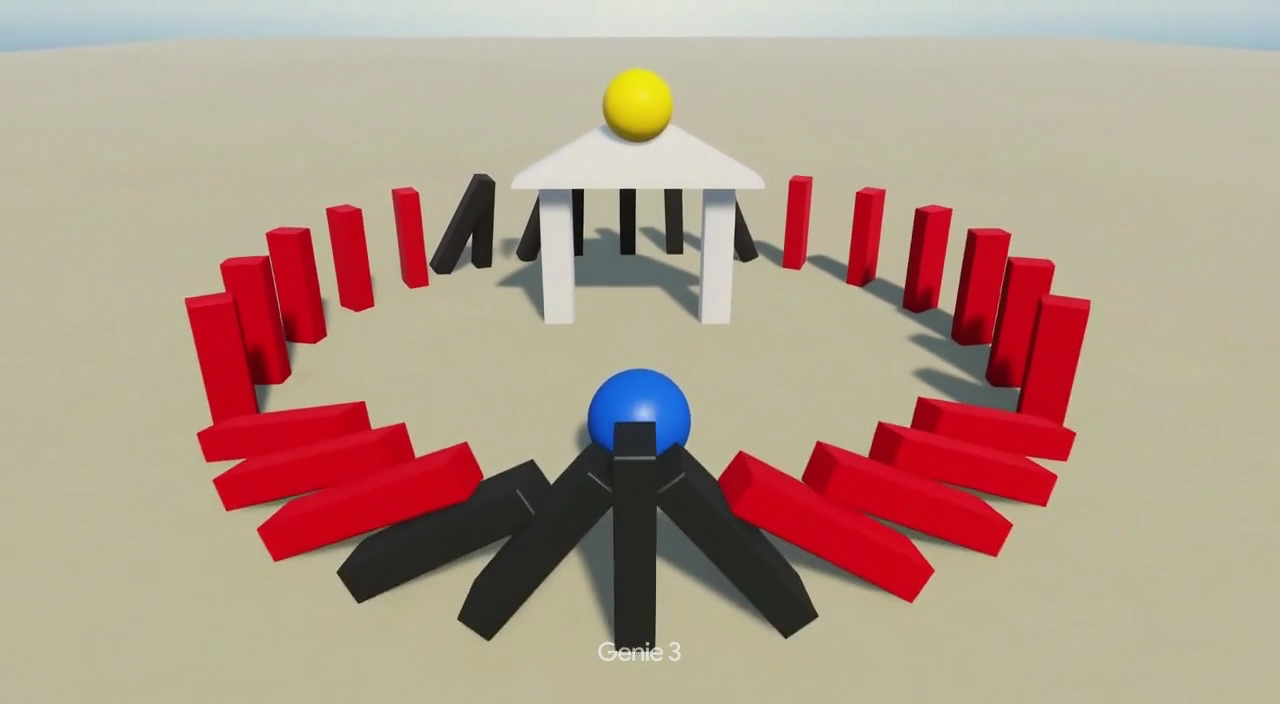} \\
\multicolumn{5}{c}{\small Project Genie} \\[4pt]
\end{tabular}
\end{adjustbox}
\caption{\texttt{BeamDomino}: Stage-aligned keyframes on the \texttt{glancing} action trajectory.}
\label{fig:causal_frames_glancing}
\end{figure*}

\subsection{Perceptual Similarity}

\subsubsection{Recurring Failure Modes.}\label{exp:perceptual_similarity_failure_modes}
We observe three recurring failure modes that drive the per-scene Causal LPIPS
numbers in Table~\ref{tab:worldmodel_causal_lpips}.

\paragraph{Action-target ambiguity.} Input frames typically contain multiple
nearby objects of similar appearance, and none of the evaluated models exposes
an explicit object-selection or object-binding mechanism for routing the control
signal to a specific scene entity. As a result, the action embedding is free to
bind to ego/camera motion or to global scene motion rather than to the intended
object, and the same control signal can produce qualitatively different
generations across runs. 

\paragraph{Distance attenuation.} As the camera-to-object distance grows, the
controlled object's pixel footprint shrinks and the induced motion in the
generation becomes correspondingly weaker, even when the action is interpreted
correctly. This is most visible in several LingBot-World (Act) generations on
\texttt{MirrorRollingBall}.

\paragraph{Scene-setup failures.} Project Genie frequently fails to faithfully
reproduce the scene configuration specified by the initial conditioning frame,
and these setup errors propagate through the entire generation independent of
action execution. Because Causal LPIPS is anchored on matched contact events
between the Unity reference and the generation, scenes where the initial
configuration drifts produce matched events whose surrounding frames are already off-reference
before the contact occurs, inflating the LPIPS values reported in
Table~\ref{tab:worldmodel_causal_lpips}.

\subsubsection{Limitations}
Causal LPIPS measures perceptual similarity in a fixed window around matched
events; it does not check whether the contact is causally correct, and a
generation that produces visually similar but causally inert motion can score
well. Generations with no matched events between
the Unity reference and the world model generation are excluded from the per-scene
aggregate, since the metric is undefined in that case.

\subsection{Action Controllability}\label{app:raft_per_scene}

\paragraph{Why direction-sensitive metrics matter beyond EPE magnitude.} On \texttt{BallRoller}, Pixel EPE Mean is similar for Project Genie and LingBot-World (Act) ($665$ vs.\ $621$) even though their Mean Flow Cosines differ sharply, so the EPE value here is dominated by motion magnitude while Mean Flow Cosine exposes the action-direction failure. The inverse pattern appears on \texttt{MirrorRollingBall}, where Genie's Pixel EPE Mean ($271$) is substantially lower than LingBot-World (Act)'s ($678$) despite Ground Truth RAFT favoring LingBot-World (Act) on the same scene a reminder that EPE magnitudes alone do not order models reliably without the directional and dynamic-region signals.

\paragraph{Synthetic vs.\ Ground Truth RAFT decoupling.} On \texttt{BallRoller}, Synthetic RAFT Mean Flow Cosine ranges from $-0.33 \pm 0.35$ to $1.00 \pm 0.00$ across models, indicating that action-direction alignment varies substantially across models for this scene. On \texttt{BeamDomino} and \texttt{MirrorRollingBall}, by contrast, all models score near $1.00$ on Synthetic RAFT, yet Ground Truth RAFT errors remain non-trivial most prominently on \texttt{MirrorRollingBall}, where Project Genie reaches $0.77 \pm 0.05$ Raw Relative EPE and LingBot-World (Act) reaches $0.49 \pm 0.10$. This pattern suggests that on these scenes the dominant residual error comes from state-level causal propagation rather than from executing the immediate control direction.

\paragraph{Complementarity with Causal LPIPS.} The RAFT metrics also pick up failures that Causal LPIPS does not surface as strongly. Genie shows a substantial scene-specific Synthetic RAFT limitation on \texttt{BallRoller} but only a moderate cLPIPS gap relative to LingBot-World (Act) ($0.56$ vs.\ $0.38$), consistent with the interpretation that cLPIPS measures perceptual plausibility around matched events while RAFT controllability measures whether the motion is driven by the intended control signal.

\subsection{Lighting and Shading}
\label{exp:lighting_failure_modes}

\subsubsection{Cross-Model Orbit Snapshot}
\begin{figure}[h]
\centering
\includegraphics[width=0.98\linewidth]{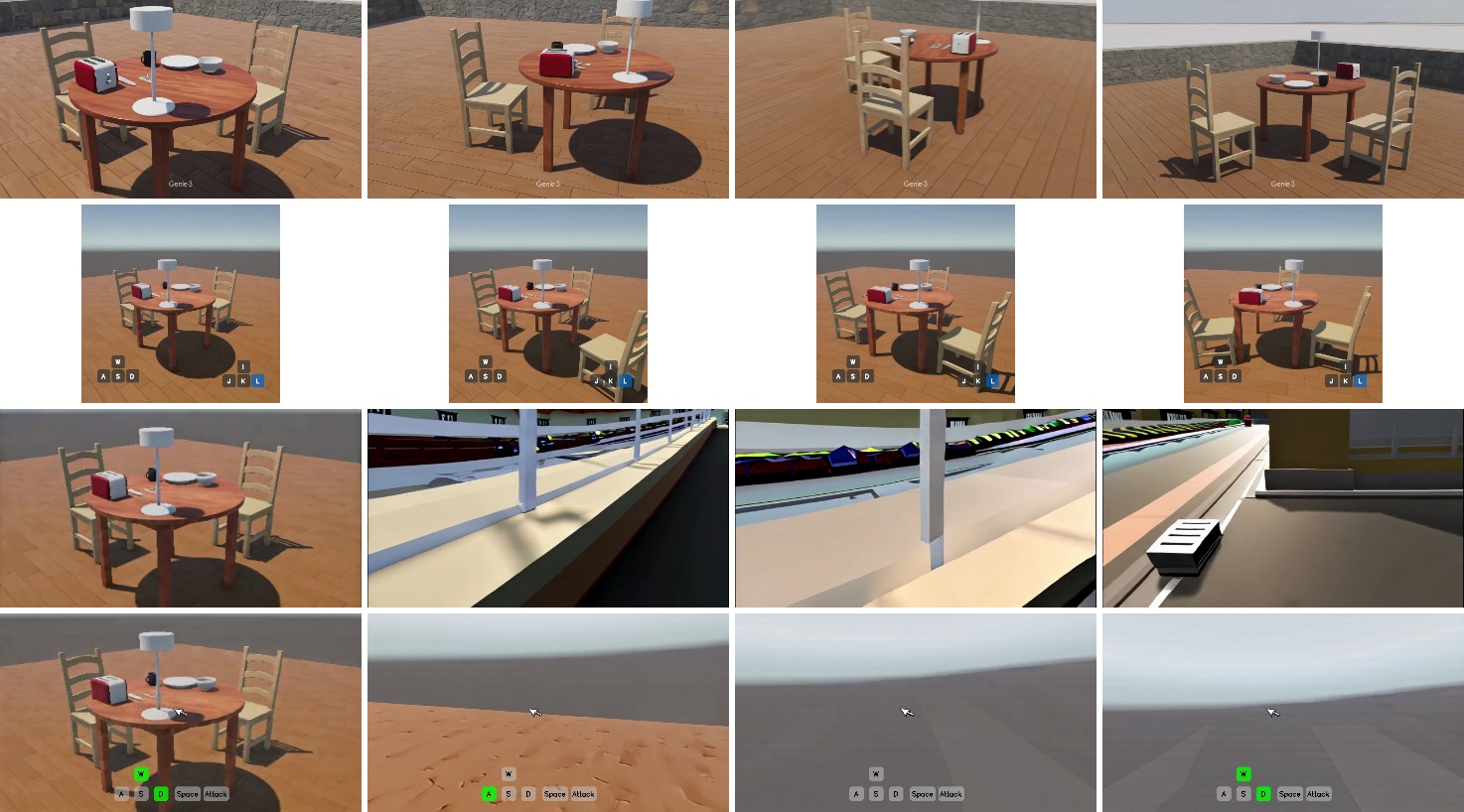}
\caption{Cross-model snapshots on \texttt{SunRisesAndSets} at four orbit fractions (one representative seed per model). Rows top-to-bottom: \textbf{Project Genie}, \textbf{LingBot-World}, \textbf{Matrix-Game 2.0}, \textbf{Matrix-Game 3.0}. Columns left-to-right: $0\%$, $33\%$, $66\%$, $99\%$ of orbit completion.}
\label{fig:lighting_cross_model}
\end{figure}

Figure~\ref{fig:lighting_cross_model} shows a representative rollout for each model at four different points in the camera orbit (0\%, 33\%, 66\%, and 99\%). These snapshots clearly illustrate the performance gaps seen in Table~\ref{tab:worldmodel_lighting_shading}. Project Genie and LingBot-World successfully keep the table, chairs, lamp, and other tabletop objects in view throughout the entire loop, maintaining crisp directional shadows on the wooden floor. The two Matrix Game variants fail in completely different ways. Matrix-Game 2.0 starts with the correct scene but quickly hallucinates a completely different environment. Matrix-Game 3.0, on the other hand, keeps the original floor texture but completely loses track of the foreground objects as the camera drifts away into an empty space.

\subsubsection{Worst-Case Shadow Shape Drift}
Our Shadow Shape Drift (SSD) metric is based on Hu-moments, which naturally ignore changes in translation, rotation, and scale. We chose this over a simpler metric like shadow-mask IoU, because the camera is constantly orbiting the table, the perspective of the shadows changes smoothly over time; IoU would unfairly penalize these normal perspective shifts. With Hu-moments, we only measure actual, unnatural shape morphing.

\begin{figure}[h]
\centering
\begin{subfigure}[t]{0.48\linewidth}
  \includegraphics[width=\linewidth]{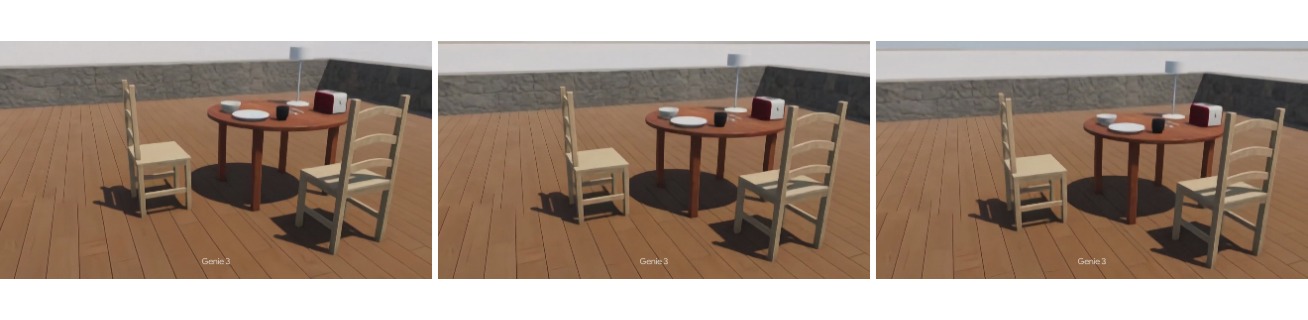}
  \caption{Project Genie, \texttt{genie\_s5}, frames $727$, $728$, $729$ (single-step $\|\Delta\log\mathrm{Hu}\|_2 = 36.69$).}
\end{subfigure}
\hfill
\begin{subfigure}[t]{0.48\linewidth}
  \includegraphics[width=\linewidth]{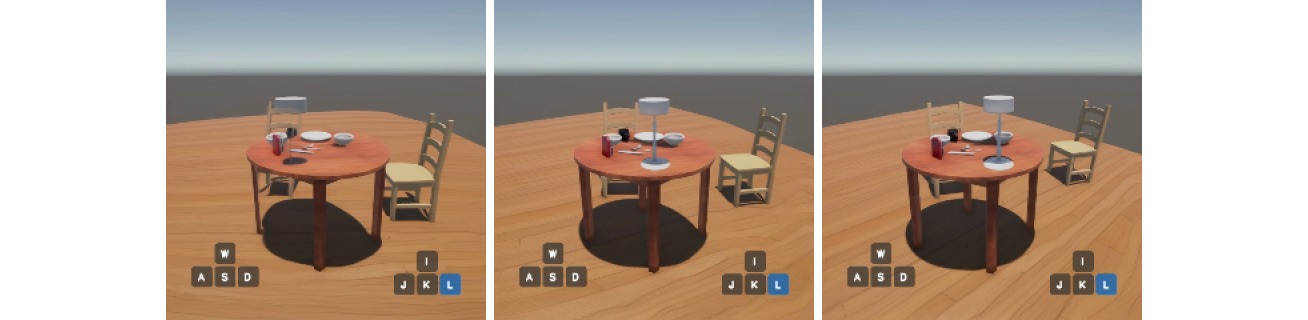}
  \caption{LingBot-World, \texttt{lingbot\_00}, frames $80$, $90$, $100$. Selected for the table-base shadow morphing visible to the unaided eye.}
\end{subfigure}\\
\vspace{0.4em}
\begin{subfigure}[t]{0.48\linewidth}
  \includegraphics[width=\linewidth]{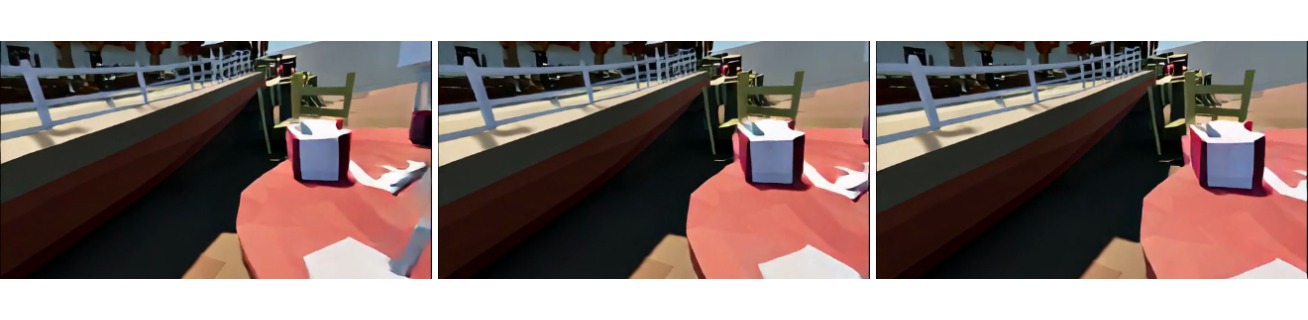}
  \caption{Matrix-Game 2.0, \texttt{mg2\_universal\_seed42}, frames $105$, $106$, $107$ (single-step $\|\Delta\log\mathrm{Hu}\|_2 = 74.30$).}
\end{subfigure}
\hfill
\begin{subfigure}[t]{0.48\linewidth}
  \includegraphics[width=\linewidth]{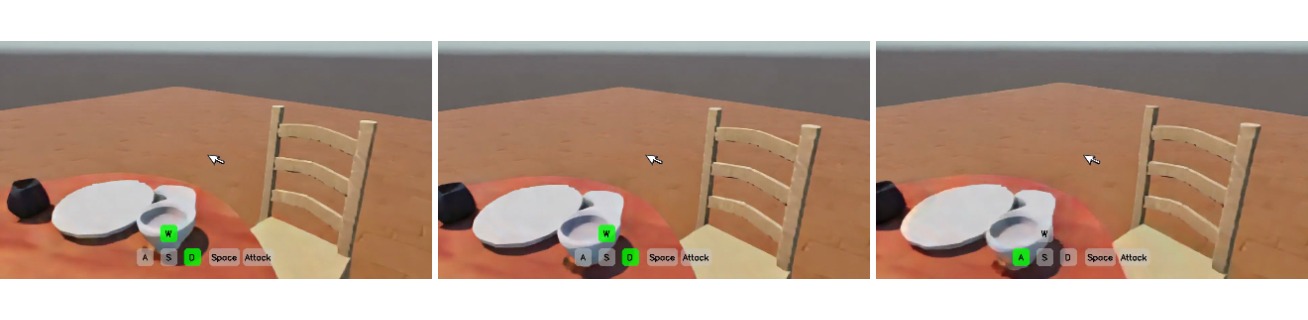}
  \caption{Matrix-Game 3.0, \texttt{mg3\_seed42}, frames $35$, $36$, $37$ (single-step $\|\Delta\log\mathrm{Hu}\|_2 = 34.08$).}
\end{subfigure}
\caption{Worst single-step Shadow Shape Drift events per model. Each triplet shows three consecutive frames covering the largest morphological discontinuity in the rollout. The Hu-moment shape descriptor is invariant to translation, rotation, and scale, so a smooth perspective sweep does not register; only genuine shape morphing does.}
\label{fig:lighting_shape_drift}
\end{figure}

Figure~\ref{fig:lighting_shape_drift} shows the worst single-frame shape distortion for each model. Matrix-Game 3.0's worst step occurs in a frame where the cast-shadow blob is split by a lighting discontinuity in the floor texture. For LingBot-World, the table's shadow visibly deforms, temporarily losing its circular shape for a few frames. Finally, the worst distortions for Project Genie are mild, mostly caused by minor masking errors.

\subsubsection{First-Frame Scene Drop-Out}
\begin{figure}[h]
\centering
\begin{subfigure}[t]{0.98\linewidth}
  \includegraphics[width=\linewidth]{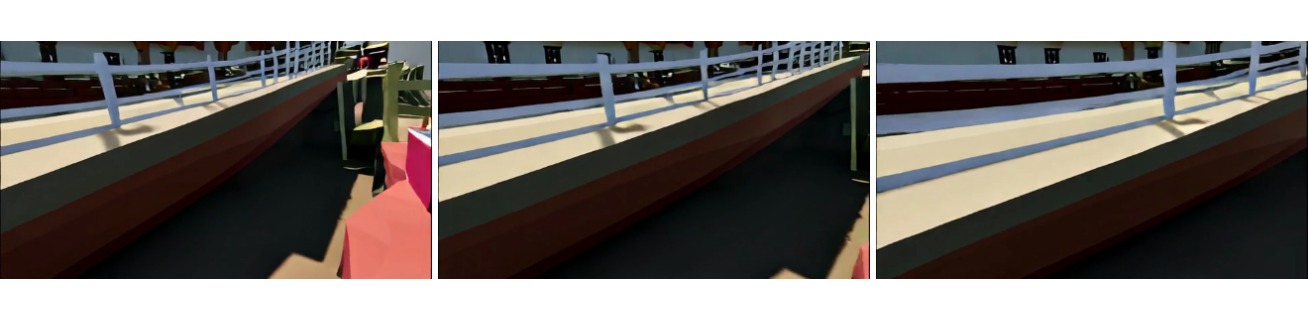}
  \caption{Matrix-Game 2.0 (\texttt{mg2\_universal\_seed42}), frames $116$, $124$, $132$. The rollout loses the prompted dining scene at frame $124$ and substitutes a stadium-style track texture from the training distribution.}
\end{subfigure}\\
\vspace{0.4em}
\begin{subfigure}[t]{0.98\linewidth}
  \includegraphics[width=\linewidth]{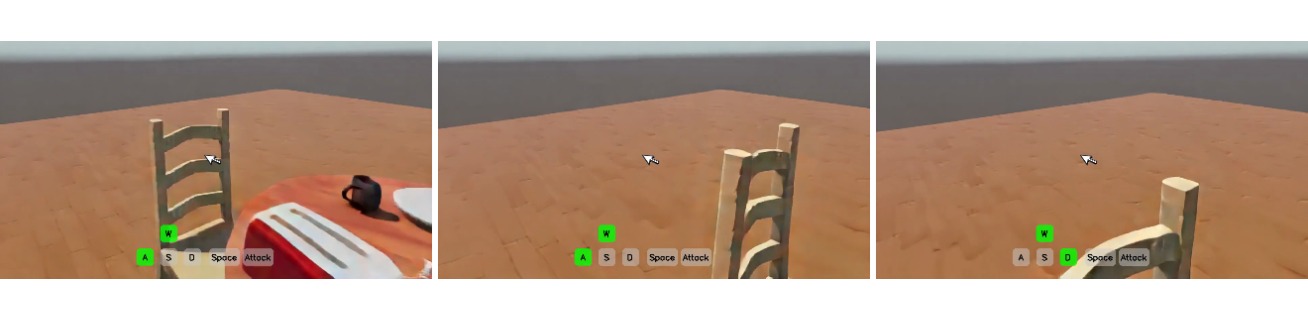}
  \caption{Matrix-Game 3.0 (\texttt{mg3\_seed42}), frames $51$, $59$, $67$. The rollout pans off the table at frame $59$ and renders empty floor with sky for the remainder.}
\end{subfigure}
\caption{First scene drop-out per model, with $\pm 8$-frame context. $t$ is the first frame at which the anchor mask falls below the area threshold used for Scene Retention Rate.}
\label{fig:lighting_dropout}
\end{figure}

Figure~\ref{fig:lighting_dropout} illustrates when the Matrix Game models lose track of the main scene by showing the frame where the anchor mask vanishes, alongside 8 frames of surrounding context. Instead of the camera smoothly panning away from the table, a single action step forces the viewpoint entirely off the scene. This abrupt shift suggests a fundamental breakdown in action conditioning, where the model commits to an extreme camera drift in one inference step rather than gradually wandering off course.

\subsubsection{Limitations}
Interpreting the metrics in Table~\ref{tab:worldmodel_lighting_shading} requires two important caveats. First, hallucinations inflate MOSD. If a model hallucinates an extra chair, our detector (GroundingDINO) will still track it, and the shadow of this fake object adds noise to the overall lighting direction spread. We chose not to filter these out because hallucinating objects already violates the physical coherence we are trying to measure. Consequently, MOSD scores should always be read alongside the scene retention (SRR) and shadow detection (SDR) rates.

Second, the perfect baseline for SDJ is not zero. Since the camera physically orbits the table, shadows naturally rotate in the camera's frame of reference. We use the median frame-to-frame change to account for this smooth sweep. As a result, even a perfectly rendered orbit will show a jitter of roughly $2^\circ$ to $8^\circ$ per frame from the camera's own movement alone.

\subsection{Prompt-Scene Consistency}
\label{exp:prompt_scene}

\subsubsection{Visibility-Dependent Attributes}
\begin{table*}[ht]
\centering
\small
\caption{Breakdown of prompt-scene consistency (Table~\ref{tab:vlm_eval}) by the rubric-based VLM judge. Env and Obj are the environment and object (character) rubric scores; Overall is $S_{\mathrm{psc}}=0.6\,S_{\mathrm{env}}+0.4\,S_{\mathrm{obj}}$ (Appendix~\ref{app:prompt_scene_metrics}). Higher is better. 
N/A indicates that no object-specific rubrics were defined for the scene due to the absence of salient foreground objects. Non-reference model cells report mean $\pm$ sample standard deviation across 5 generations. Waypoint-1.5 has no text channel: its generations are conditioned on the seed frame and the action script alone, and the prompt is seen only by the judge.}
\label{tab:vlm_eval_breakdown}
\begin{adjustbox}{max width=\textwidth}
\begin{tabular}{llccccccccc}
\toprule
\multirow{2}{*}{View}
& \multirow{2}{*}{Model}
& \multicolumn{3}{c}{ChessBoard}
& \multicolumn{3}{c}{MirrorRollingBall}
& \multicolumn{3}{c}{RacingCar} \\
\cmidrule(lr){3-5}
\cmidrule(lr){6-8}
\cmidrule(lr){9-11}
&
& Env $\uparrow$ & Obj $\uparrow$ & Overall $\uparrow$
& Env $\uparrow$ & Obj $\uparrow$ & Overall $\uparrow$
& Env $\uparrow$ & Obj $\uparrow$ & Overall $\uparrow$ \\
\midrule
\multirow{6}{*}{First Person}


& Project Genie

& $0.880\!\pm\!0.099$ & $0.960\!\pm\!0.060$ & $0.912\!\pm\!0.074$

& $1.000\!\pm\!0.000$ & $0.950\!\pm\!0.075$ & $0.980\!\pm\!0.030$

& $0.720\!\pm\!0.073$ & N/A & $0.720\!\pm\!0.073$ \\

& LingBot-World (Cam)

& $0.920\!\pm\!0.110$ & $0.960\!\pm\!0.037$ & $0.936\!\pm\!0.073$

& $0.967\!\pm\!0.045$ & $0.850\!\pm\!0.038$ & $0.920\!\pm\!0.022$

& $0.760\!\pm\!0.060$ & N/A & $0.760\!\pm\!0.060$ \\

& Matrix-Game 3.0

& $0.733\!\pm\!0.082$ & $1.000\!\pm\!0.000$ & $0.840\!\pm\!0.049$

& $0.550\!\pm\!0.095$ & $0.833\!\pm\!0.000$ & $0.663\!\pm\!0.057$

& $0.627\!\pm\!0.076$ & N/A & $0.627\!\pm\!0.076$ \\

& HY-WorldPlay

& $0.693\!\pm\!0.101$ & $0.933\!\pm\!0.094$ & $0.789\!\pm\!0.082$

& $0.700\!\pm\!0.075$ & $0.800\!\pm\!0.140$ & $0.740\!\pm\!0.070$

& $0.573\!\pm\!0.101$ & N/A & $0.573\!\pm\!0.101$ \\

& Waypoint-1.5

& $0.907\!\pm\!0.076$ & $0.973\!\pm\!0.037$ & $0.933\!\pm\!0.037$

& $0.650\!\pm\!0.070$ & $0.900\!\pm\!0.037$ & $0.750\!\pm\!0.031$

& $0.667\!\pm\!0.082$ & N/A & $0.667\!\pm\!0.082$ \\

\midrule

\multirow{6}{*}{Third Person}


& Project Genie

& $0.920\!\pm\!0.073$ & $0.960\!\pm\!0.060$ & $0.936\!\pm\!0.057$

& $0.920\!\pm\!0.073$ & $0.960\!\pm\!0.037$ & $0.936\!\pm\!0.054$

& $0.760\!\pm\!0.089$ & $1.000\!\pm\!0.000$ & $0.856\!\pm\!0.054$ \\

& LingBot-World (Cam)

& $0.987\!\pm\!0.030$ & $1.000\!\pm\!0.000$ & $0.992\!\pm\!0.018$

& $0.867\!\pm\!0.047$ & $0.947\!\pm\!0.056$ & $0.899\!\pm\!0.036$

& $0.720\!\pm\!0.073$ & $1.000\!\pm\!0.000$ & $0.832\!\pm\!0.044$ \\

& Matrix-Game 3.0

& $0.867\!\pm\!0.067$ & $0.933\!\pm\!0.067$ & $0.893\!\pm\!0.053$

& $0.693\!\pm\!0.167$ & $0.840\!\pm\!0.037$ & $0.752\!\pm\!0.111$

& $0.494\!\pm\!0.059$ & $0.800\!\pm\!0.093$ & $0.616\!\pm\!0.064$ \\

& HY-WorldPlay

& $0.773\!\pm\!0.101$ & $0.933\!\pm\!0.094$ & $0.837\!\pm\!0.059$

& $0.867\!\pm\!0.094$ & $0.960\!\pm\!0.037$ & $0.904\!\pm\!0.067$

& $0.707\!\pm\!0.138$ & $1.000\!\pm\!0.000$ & $0.824\!\pm\!0.083$ \\

& Waypoint-1.5

& $0.827\!\pm\!0.089$ & $0.987\!\pm\!0.030$ & $0.891\!\pm\!0.047$

& $0.720\!\pm\!0.087$ & $0.827\!\pm\!0.089$ & $0.763\!\pm\!0.061$

& $0.680\!\pm\!0.030$ & $1.000\!\pm\!0.000$ & $0.808\!\pm\!0.018$ \\
\bottomrule
\end{tabular}
\end{adjustbox}
\end{table*}

\texttt{RacingCar} is the lowest-scoring scene for almost every model and for the Unity reference itself (Table~\ref{tab:vlm_eval}; breakdown in Table~\ref{tab:vlm_eval_breakdown}). Environment scores range from $0.49$--$0.63$ for Matrix-Game~3.0 to $0.72$--$0.76$ for Project Genie and LingBot-World (Cam).

Much of this drop comes from a single rubric item. The prompt states that the road ``begins abruptly at a starting point, with no ground behind it---only open sky''. This attribute is visible only while the camera faces the starting edge. The judge's explanations acknowledge this, yet it still penalizes the frames in which the edge is out of view, so even the Unity references receive only $2/3$ (first person) and $1/3$ (third person) on this item.

The same item also reveals a model-side failure. All 40 rollouts of Project Genie, LingBot-World (Cam), Matrix-Game~3.0, and HY-WorldPlay score $0/3$ on it, which alone lowers their environment scores by $0.20$. For instance, when Project Genie turns toward the start, it renders an ordinary continuous highway instead of the suspended edge. \texttt{RacingCar} thus exposes two limitations: world models rarely populate prompt-specified regions outside the initial view, and the judge does not fully account for view-dependent visibility.

\end{document}